\documentclass[sn-apa,iicol]{sn-jnl}% Default with double column layout

\usepackage{graphicx}
\usepackage{subfigure}
\usepackage{multirow}
\usepackage{multicol}
\usepackage{xcolor}

\jyear{2021}%

\theoremstyle{thmstyleone}%
\theoremstyle{thmstyletwo}%

\theoremstyle{thmstylethree}%

\def\A{{\bf A}}
\def\E{{\bf E}}
\def\F{{\bf F}}
\def\G{{\bf G}}
\def\I{{\bf I}}
\def\L{{\bf L}}
\def\R{{\bf R}}
\def\Z{{\bf Z}}
\def\d{{\bf d}}
\def\D{{\bf D}}
\def\B{{\bf B}}
\def\R{{\bf R}}
\def\g{{\bf g}}
\def\i{{\bf i}}
\def\l{{\bf l}}
\def\r{{\bf r}}
\def\x{{\bf x}}

\begin{document}
\title[RAUNA]{Low-light Image Enhancement by Retinex Based Algorithm Unrolling and Adjustment}


\author{\fnm{Xinyi} \sur{Liu}}\email{xlyqnhqs98@stu.xjtu.edu.cn}

\author{\fnm{Qi} \sur{Xie}}\email{xie.qi@xjtu.edu.cn}

\author{\fnm{Qian} \sur{Zhao}}\email{timmy.zhaoqian@xjtu.edu.cn}

\author{\fnm{Hong} \sur{Wang}}\email{hongwang01@stu.xjtu.edu.cn}

\author{\fnm{Deyu} \sur{Meng}}\email{dymeng@xjtu.edu.cn}

\affil{\orgdiv{School of Mathematics and Statistics}, \orgname{Xi'an Jiaotong University}, \orgaddress{\street{28 West Xianning Road}, \city{Xi'an}, \postcode{710049}, \state{Shaanxi}, \country{P.R. China}}}

%\author[1]{\fnm{Xinyi} \sur{Liu}}\email{xlyqnhqs98@stu.xjtu.edu.cn}
%
%\author[1]{\fnm{Qi} \sur{Xie}}\email{xq.liwu@stu.xjtu.edu.cn}
%%\equalcont{These authors contributed equally to this work.}
%
%\author[1]{\fnm{Qian} \sur{Zhao}}\email{timmy.zhaoqian@xjtu.edu.cn}
%%\equalcont{These authors contributed equally to this work.}
%
%\author[1]{\fnm{Hong} \sur{Wang}}\email{hongwang01@stu.xjtu.edu.cn}
%
%\author[1]{\fnm{Deyu} \sur{Meng}}\email{dymeng@xjtu.edu.cn}
%
%\affil[1]{\orgdiv{School of Mathematics and Statistics}, \orgname{Xi'an Jiaotong University}, \orgaddress{\street{28 West Xianning Road}, \city{Xi'an}, \postcode{710049}, \state{Shaanxi}, \country{P.R. China}}}

%\affil[2]{\orgdiv{Department}, \orgname{Organization}, \orgaddress{\street{Street}, \city{City}, \postcode{10587}, \state{State}, \country{Country}}}
%
%\affil[3]{\orgdiv{Department}, \orgname{Organization}, \orgaddress{\street{Street}, \city{City}, \postcode{610101}, \state{State}, \country{Country}}}


\abstract{Motivated by their recent advances, deep learning techniques have been widely applied to low-light image enhancement (LIE) problem. Among which, Retinex theory based ones, mostly following a decomposition-adjustment pipeline, have taken an important place due to its physical interpretation and promising performance. However, current investigations on Retinex based deep learning are still not sufficient, ignoring many useful experiences from traditional methods. Besides, the adjustment step is either performed with simple image processing techniques, or by complicated networks, both of which are unsatisfactory in practice. To address these issues, we propose a new deep learning framework for the LIE problem. The proposed framework contains a decomposition network inspired by algorithm unrolling, and adjustment networks considering both global brightness and local brightness sensitivity. By virtue of algorithm unrolling, both implicit priors learned from data and explicit priors borrowed from traditional methods can be embedded in the network, facilitate to better decomposition. Meanwhile, the consideration of global and local brightness can guide designing simple yet effective network modules for adjustment. Besides, to avoid manually parameter tuning, we also propose a self-supervised fine-tuning strategy, which can always guarantee a promising performance. Experiments on a series of typical LIE datasets demonstrated the effectiveness of the proposed method, both quantitatively and visually, as compared with existing methods.}


\maketitle

\section{Introduction}
Due to the lack of sufficient amount of light, photos taken under low-light environment suffer from various of degradations, such as low visibility, color biases and noise. These unexpected degradations not only reduce the user experience of consumer photography, but also degrade the performance of downstream vision tasks \citep{LIE_review} in industrial purpose, including object detection \citep{zhao2019object} and tracking \citep{tracking}. Therefore, it is indispensable to enhance the images captured under low-light conditions for further utilization. In the last decades, there are various methods have been developed for this low-light image enhancement (LIE) problem, including both traditional methods and deep learning based ones.

%\IEEEPARstart{A}{lthough} current photography equipment is advanced, photos taken under low-light environment still suffer from low visibility, color biases, noise and so on. The unexpected degradation has bringing extensive inconvenience to people's daily life and scientific researches, including object detection \citep{zhao2019object}, object recognition \citep{bansal20212d}, image restoration \citep{rani2016brief} and so on. Therefore, it's indispensable to do image enhancement for further utilization. 

In the early years, LIE was achieved by traditional image processing techniques, specifically value mapping methods, such as histogram equalization (HE) \citep{pizer1987adaptive, CLAHE, kim1997contrast, pisano1998contrast, reza2004realization, acharya2005image, sun2005dynamic} and gamma correction (GC) \citep{guan2009image, huang2012efficient, gamma_correction1, gamma_correction2}. The main idea behind these methods is applying certain transformation to the pixel values of the observed low-light image, in order to stretch the dynamic range of it. However, these methods intrinsically try to enhance the contrast of the images, rather than directly adjust the illumination, and also neglect other degaradations, such as noise, leading to inferior performance.

Later, model based methods become more popular, which assume that the low-light image is obtained following a physical observation model, and then enhance the image based on it. Most powerful methods along this line motivated their observation models from the well-known Retinex theory \citep{land1977retinex}, where the basic assumption is that the observed image can be decomposed into reflectance and illumination layers, while both the low-light image and the corresponding normal-light one share the same refletance layer. Therefore, one can first do such a decomposition to the low-light image (or only estimate the illumination layer), and then directly recover the reflectance layer as the final result \citep{jobson1997properties, rahman1996multi}, or further adjust the illumination layer to enhance the image \citep{6512558}. The key to such methods is designing proper priors for images or their decompositions, and various effective priors have been considered, including dark channel prior \citep{guo2016lime}, sparse gradient prior \citep{srie}, structure-revealing prior \citep{robust_LIE} and low-rank prior \citep{LR3M}. In addition to Retinex theory, there are also other observation models having been considered. For example, in some studies, the authors treated LIE as the inverted dehazing problem \citep{LIE_dehaze, li2015low}, and thus constructed algorithms based on the atmospheric scattering model \citep{atmosphere}. These model based methods, however, highly rely on the hand-crafted priors for images or their decompositions, which largely limits their performance.

Inspired by recent advances of deep neural networks (DNNs) in low-level computer vision \citep{pami2015_dong_cnn_sr, tip2017_zhang_dncnn}, deep learning based methods have also been developed for LIE. Among them, there are mainly two methodologies in applying deep learning to LIE. The first is to learn a direct mapping, parameterized by a DNN, from the low-light image to the corresponding normal-light one \citep{LLNet, Lv2018MBLLEN, xu2020learning, ma2021learning}, regardless of the physical model underlying the low-light observation. The second is introducing Retinex theory to the network designing \citep{shen2017msr, progressive_retinex, yang2020fidelity, liu2021retinex}, in which the DNN firstly tries to decompose the low-light image into two components, i.e., reflectance and illumination, and then doing postprocessing (possibly achieved also by a DNN) \citep{zhang2021beyond} or directly using the reflectance layer as the final result \citep{li2018lightennet, wang2019underexposed}, inheriting the similar idea as that of traditional Retinex based methods. Comparatively, Retinex inspired methods have attracted more attention, since they generally incorporate more prior knowledge by the physical observation model, and thus are expected to be more effective.

Though progress has been continuously made, the performance of current DNNs for LIE is still not significant, compared with their applications to other image restoration or enhancement tasks, such as image denoising and super-resolution, due to relatively more complicated degradations involved in low-light images. After comprehensively investigating current deep learning based LIE studies, we find one major issue possibly leading to such lower-than-expected performance of DNNs, that more attention is paid to fancy and powerful modules in designing the network architectures, while many valuable experiences from traditional methods are ignored, especially from the ones inspired by Retinex theory. This deficiency not only leads to less interpretability of DNNs, but could also limit their performance. We have noticed that there are studies concerning this issue \citep{liu2021retinex}, but current investigations are still not sufficient and have large room for improvement. Besides, as a post-processing step in using Retinex model, adjusting one or both of the decomposed reflectance and illumination is important. However, current studies either resort to traditional image processing techniques, such as Gamma Correction, or design relatively complicated network, for this goal, both of which are not very satisfactory in practice.

To address the aforementioned issues, in this paper we propose an effective and interpretable DNN for LIE, by fully considering useful experiences from traditional Retinex methods, and carefully designing the adjusting modules for the decomposed image layers. Specifically, following the pipeline that has been shown to be effective \citep{zhang2021beyond}, we propose a DNN that firstly decompose the image into reflectance and illumination layers, and then adjusting both of them to obtain the final enhancement results:
\begin{itemize}
	\item For the decomposition, we design the network architecture inspired by unrolling the optimization procedure for solving an Retinex based image decomposition model. Different from the previous study that also considered the algorithm unrolling of an Retiniex based optimization \citep{liu2021retinex}, while constructed the network with neural architecture search, we pay more attention to the algorithm inspired operations with more sophisticated prior borrowed from traditional methods. To be specific, in addition to implicit priors that can be learned by network as many other unrolling type DNN, we introduce the structure-revealing prior \citep{robust_LIE, LR3M} specifically designed for the LIE problem to the network. This strategy on the one hand takes the superiority of data-driven methods in flexibly prior learning, and on the other hand also naturally inherits the advantageous experience from traditional methods.
	\item For the adjustment, we design simple yet effective networks, which integrate global brightness and local brightness sensitivity, for illumination and reflectance, respectively. The physical mechanism of the proposed adjustment networks is also relatively easier to explain, as compared with existing methods.
	\item Besides, as a supplement to user-controlled enhancement level, we also propose a self-supervised strategy to fine-tune the adjustment networks at test time. This self-supervised fine-tuning strategy can generally produce a satisfactory result, without user intervention.
\end{itemize}
All of these advantages have been substantiated by comprehensive experiments on typical LIE datasets.

The paper is organized as follows. In Section \ref{sec:rw}, we review related work on the studied problem. Section \ref{sec:approach} presents the details of the proposed framework for LIE. In Section \ref{sec:exp}, extensive experimental results on typical LIE datasets are introduced to demonstrate the superiority of the proposed method. The conclusion is finally made in Section \ref{sec:conclusion}.

\section{Related Work} \label{sec:rw}
In this section, we review related work on the LIE problem. Our main focus is on Retinex theory based methods, both traditional and deep learning ones, while interested readers could refer to \citep{LIE_review} for a more comprehensive review.

\subsection{Traditional Methods}
\subsubsection{Value Mapping Methods}
The early attempts for the LIE problem are mainly focused on value mapping in traditional image processing techniques, such as Histogram Equalization (HE) \citep{HUMMEL1977184} and Gamma Correction (GC). The main idea of HE type methods \citep{pizer1987adaptive, CLAHE, kim1997contrast, pisano1998contrast, reza2004realization, acharya2005image, sun2005dynamic} is to enhance images by mapping the histogram distribution of an image to approximately a uniform one. Different from HE, GC type methods \citep{guan2009image, huang2012efficient, gamma_correction1, gamma_correction2} realize image enhancement by nonlinear editing of its gamma curve, which results in change to the tone of the image. There are also studies that used different value mappings. For example, Yuan and Sun (\citeyear{yuan2012automatic}) considered region-level illumination and applied a detail-preserved S-curve adjustment to enhance exposure. Generally, these methods are too specific with less flexibility to real scenarios, and tend to amplify noise existing in low-light images. 

\subsubsection{Retinex Theory Based Methods}
Retinex theory \citep{land1977retinex} plays an important role in traditional model-based LIE methods. The key step of applying Retinex theory is to decompose the image into reflectance and the illumination layers, representing the reflected objects and the light shining on the surface of objects, respectively. Such a Retinex decomposition can be achieved by traditional image processing techniques, such as filtering \citep{jobson1997properties, rahman1996multi, wang2013naturalness}. However, such methods did not take sophisticated image priors into considerations, and thus the performance is not satisfactory. A more preferable way in using Retinex theory is to formulate the Retinex decomposition as an optimization problem. However, since the solution to Retinex decomposition is mathematically non-unique, it is crucial to design proper priors to make the optimization well-defined. Along this line, multiple priors have been adopted or newly designed. For example, Guo (\citeyear{guo2016lime}) used the dark channel prior for images and sparse gradient prior to illumination; Fu et al. (\citeyear{srie}) proposed to consider sparse gradient prior to reflectance while smoothness prior to illumination; Li et al. (\citeyear{robust_LIE}) designed a structure-revealing prior for reflectance while also considered the robustness issue; Ren et al. (\citeyear{LR3M}) improved the model in \citep{robust_LIE} by introducing low-rank prior to similar patches extracted from reflectance. These methods, however, highly rely on the hand-crafted priors for images or their decompositions, which tends to limit their performance.

%Additionally, BIMEF \citep{ying2017bio} and CRF \citep{crf} introduced camera response model for better preservation of details.

\subsection{Deep Learning Based Methods}
\subsubsection{Retinex Inspired Deep Learning Methods}
Inspired by its successful applications in traditional LIE methods, Retinex theory has also been adopted in deep learning era, and there are mainly two methodologies in applying it. The first is to directly treat the estimated reflectance layer (can be achieved by first estimating the illumination and then doing element-wise division) as the enhanced image. For example, Shen et al. (\citeyear{shen2017msr}) simulated multi-scale Retinex with a convolution neural network; Li et al. (\citeyear{li2018lightennet}) proposed an improved network framework for correcting the illumination; Wang et al. (\citeyear{wang2019progressive}) designed a network to do Retinex decomposition in a progressive way; Wang et al. (\citeyear{wang2019underexposed}) addressed this problem with the help of a down-sample image; Lv et al. (\citeyear{lv2020fast}) built a noise compression network for reflectance denoising; Liu et al. (\citeyear{liu2021retinex}) proposed a Retinex inspired unrolling scheme with network architecture search (NAS). However, as analyzed in \citep{lu2020tbefn}, directly treating the reflectance might not be very reasonable, and therefore the second methodology, that additionally takes illumination adjustment into consideration, has also been adopted in many studies. For example, Chen et al. (\citeyear{Chen2018Retinex}) introduced additional feature extraction process for illumination mapping; Zhang et al. (\citeyear{zhang2019kindling}) constructed an LIE processing pipeline, which first decomposes the image and then adjusts both the reflectance and illumination layers, and later improved it \citep{zhang2021beyond}; Fan et al. (\citeyear{fan2020integrating}) considered the semantic information for better decomposition; Yang et al. (\citeyear{yang2021sparse}) introduced the sparse gradient regularization for the balance between details preservation and noise removal; Ma et al. (\citeyear{ma2021learning}) designed a context-sensitive decomposition module and used spatially varing illumination guidance for estimation.
 
Above methods were all built in a fully supervised fashion, and there are also some recent studies on unsupervised learning. For example, Zhao et al. (\citeyear{zhao2021retinexdip}) built Retinex-DIP model by combining the Retinex model and deep image prior \citep{dip}; Zhu et al. (\citeyear{rrdnet}) proposed RRDNet by training the network in a zero-shot way with specifically designed loss functions.

The current Retinex inspired deep learning studies, however, mostly try to build their inference architecture with fancy modules, while ignore advantageous experiences from traditional ones, and thus still has room to improve. It should be mentioned that, there has been an attempt to this issue \citep{liu2021retinex}, but it still did not fully make use of the useful priors in the optimization based Retinex decomposition models.

\subsubsection{Other Deep Learning Methods}
There are deep learning based methods that did not make use of Retinex theory. For example, Lore et al. (\citeyear{LLNet}) designed an autoencoder structure to learn a direct mapping from low-light image to the corresponding normal-light one; Lv et al.(\citeyear{Lv2018MBLLEN}) built a multi-branch network for this task; Lim and Kim (\citeyear{lim2020dslr}) introduced Laplacian pyramid to a multi-scale structure for better feature extraction; Zheng et al. (\citeyear{zheng2021adaptive}) presented an algorithm unrolling scheme mainly focusing on denoising.

Unsupervised and semi-supervised methodologies have also been considered. For example, Jiang et al. (\citeyear{jiang2021enlightengan}) proposed a GAN-based framework for LIE; Guo et al. (\citeyear{guo2020zero}) put forward a zero-shot method, inspired by curve adjusting for light enhancement, which was later extended \citep{Zero-DCE++}; Yang et al. (\citeyear{yang2020fidelity}) introduced unpaired data in addition to paired ones in a semi-supervised fashion.

These non-Retinex methods can also achieve promising results due to the power of DNNs, but are generally with less interpretability compared with those Retinex inspired ones.

\begin{figure*}[!t]
	\centering
	\includegraphics[width=\textwidth]{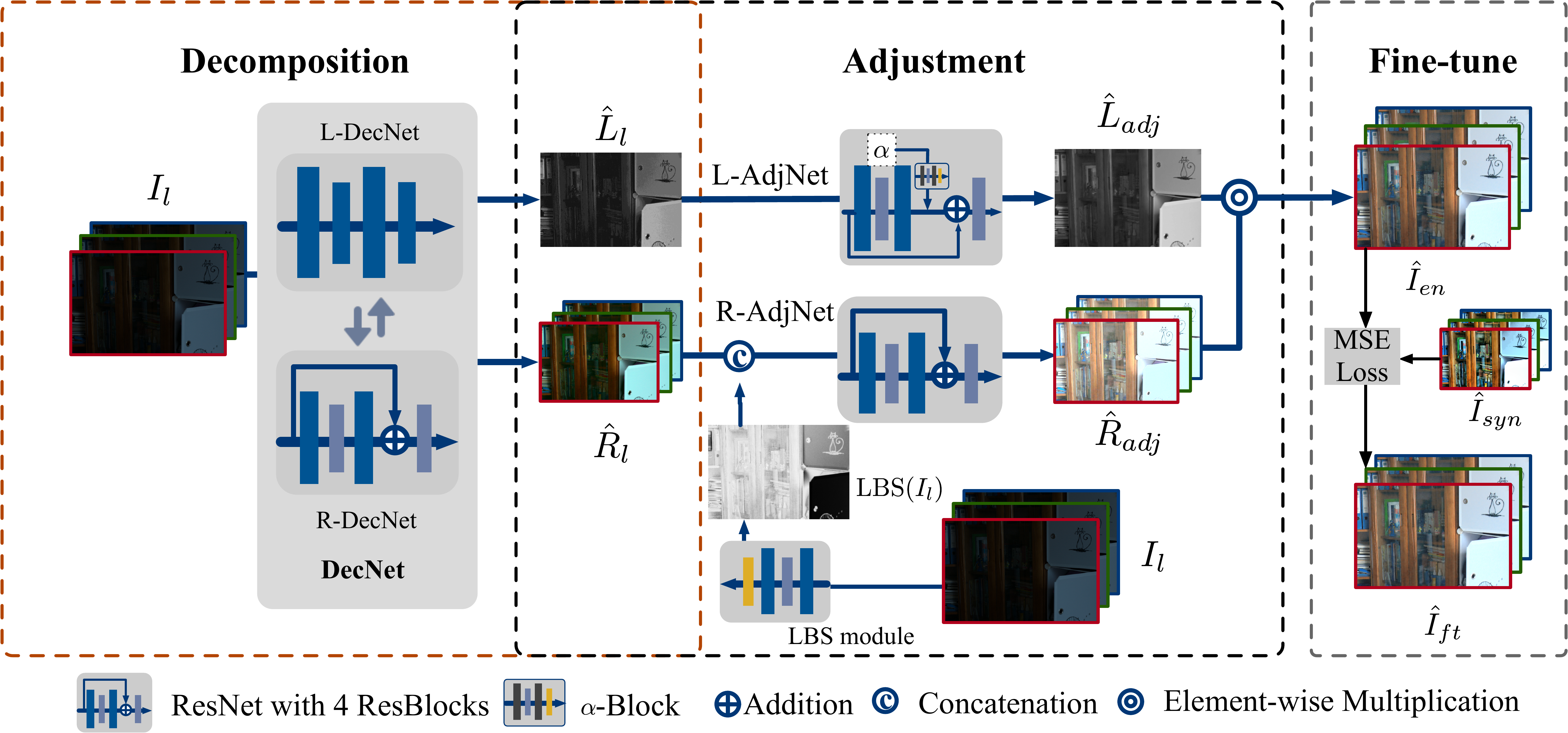}
	\caption{Illustration of the proposed LIE framework.}
	\label{fig:overview}
\end{figure*}

\section{Proposed Approach}\label{sec:approach}
%In this section, we present our DNN framework for the LIE problem. We first describe the designing details of the two main components of our framework, i.e., decomposition and adjustment networks. Then we discuss the loss functions and other details for training the overall DNN. At last, we propose our self-supervised fine-tuning strategy, together with discussions of practical usage.

\subsection{Overview of Proposed Framework}
\subsubsection{Retinex Theory and LIE}
For an observed image, denoted as $\I\in \mathbb{R}^{H\times W}$, it can be decomposed into two components based on Retinex theory \citep{land1977retinex} as
\begin{equation}
\label{eeq:1}
\I = \R\circ\L,
\end{equation}
where $\R\in\mathbb{R}^{H\times W}$ denotes the reflectance layer, $\I\in \mathbb{R}^{H\times W}$ represents the illumination layer, and $\circ$ means the element-wise product\footnote{Note that for color image, both $\I$ and $\R$ have three channels, and \eqref{eeq:1} should be more rigorously written as $\I^c = \R^c\circ\L,c\in\{r,g,b\}$. In the remaining of this paper, we mainly discuss with one-channel image for notation simplicity, if not emphasized. Nevertheless, the discussions can be straightforwardly extended to color image, on which our proposed framework indeed operates.}. The key assumption of Retinex theory is that, ideally, images taken under the same scene, with different light conditions, can have different illumination layers, while share the same reflectance layer. 

According to this theory, for the LIE problem, if such a decomposition for the low-light image can be accurately estimated, its enhancement can be obtained correspondingly. Specifically, suppose low-light image $\I_l$ can be decomposed into relfectance $\R$ and illumination $\L_l$, and then there are two possible ways to obtain the enhanced image $\hat{\I}_{\mathrm{en}}$:
\begin{itemize}
	\item Let $\hat{\I}_{\mathrm{en}}=\R$, or equivalently $\hat{\I}_{\mathrm{en}}=\I_l\oslash\L_l$, where $\oslash$ denotes the element-wise divide.
	\item Adjust $\L_l$ to $\hat{\L}_h$ (as an approximation to groundtruth illumination $\L_h$ of normal-light image), and then let $\hat{\I}_{\mathrm{en}}=\R\circ\hat{\L}_h$.
\end{itemize}
As discussed in \citep{guo2016lime, lu2020tbefn}, the first approach, though works, is less physically reasonable, since illumination is also an important factor to human perception \citep{lu2020tbefn}. Therefore, we focus on the second one, and the following discussions are also based on it, if not emphasized.

%As mentioned before, there are mainly two ways to implementing enhancement after Retinex decomposition
%
%in a straightforward way. Specifically, in the ideal case, if low-light image $\I_l$ can be decomposed into relfectance $\R$ and illumination $\L_l$, one can obtain the enhanced image $\hat{\I}_h$, as an approximation to groundtruth normal-light image $\I_h$, in two ways: 1) letting $\hat{\I}_h=\R$, or equivalently $\hat{\I}_h^c=\I_l^c\oslash\L_l$ for $c\in\{r,g,b\}$, where $\oslash$ denotes the element-wise divide; 2) adjusting $\L_l$ to $\hat{\L}_h$ (as an approximation to groundtruth illumination $\L_h$ of normal-light image), and letting $\hat{\I}_h^c=\R^c\circ\hat{\L}_h$ for $c\in\{r,g,b\}$. As aforementioned, in this work, we focus on the second way, and the following discussions are also based on it, if not emphasized.

In real scenarios, however, the estimated reflectance $\hat{\R}_l$ of the low-light image may contain unexpected degradations \citep{fan2020integrating, zhang2021beyond}, such as noise and contrast biases, and thus not necessarily equals to the ideal reflectance $\R$. Therefore, adjustment to $\hat{\R}_l$ is also important. Consequently, Retinex theory based LIE can be summarized as the following steps:
\begin{itemize}
	\item Decompose the low-light image to reflectence and illumination layers, such that
	\begin{equation}
	\I_l\approx\hat{\R}_l\circ\hat{\L}_l.
	\end{equation}
	\item Adjust both the reflectence and illumination layers, which can be expressed as
	\begin{equation}
	\left\{
	\begin{split}
	\hat{\R}_{\mathrm{adj}}&=f_R(\hat{\R}_l),\\
	\hat{\L}_{\mathrm{adj}}&=f_L(\hat{\L}_l).
	\end{split}
	\right.
	\end{equation}
	where, $f_R(\cdot)$ and $f_L(\cdot)$ denote the adjustment operations for reflectance and illumination layers, respectively.
	\item Recompose the reflectence and illumination to obtain the enhanced image by
	\begin{equation}
	\hat{\I}_{\mathrm{en}}=\hat{\R}_{\mathrm{adj}}\circ\hat{\L}_{\mathrm{adj}}.
	\end{equation}
\end{itemize}

\subsubsection{Proposed Solution to Retinex Based LIE}
%As mentioned before, there are mainly two pipelines in applying Retinex theory for the LIE problem. One is to first decompose the low-light image into reflectance layer and illumination layer, and then adjust both layers and recompose them together to get the enhanced image \citep{lu2020tbefn, yang2021sparse, zhang2021beyond}; the other is to directly treat the reflectance layer as the enhanced image after Retinex decomposition and possibly refletance adjustment \citep{wang2019underexposed, liu2021retinex}. As discussed in \citep{guo2016lime, lu2020tbefn}, the latter, though also works, is less physically reasonable, since illumination is also an important factor to human perception \citep{lu2020tbefn}.
Following the Retinex based LIE pipeline discussed above, the key is how to decompose the low-light image and adjust the decomposition results, which turns to designing effective network architectures for doing decomposition and adjustment, from deep learning perspective. We briefly introduce our solution to this as follows.

For decomposition, considering that most of existing deep learning methods did not fully make use of the experiences from traditional methods, we design a network inspired by algorithm unrolling of a Retinex based optimization model. This unrolling scheme can encode both the explicit prior, which has shown to be effective in traditional methods, and implicit prior, which can be learned by DNNs in a data-driven way. For adjustment, we design relatively simple and intuitively interpretable, yet effective modules to adjust the reflectance and illumination layers, respectively. In addition, we also introduce a self-supervised fine-tuning strategy as an optional post-processing step when doing inference. By this fine-tuning, a promising performance can generally be guaranteed, without much user intervention. 

The overall framework of our method is illustrated in Fig. \ref{fig:overview}, and in the following subsections, we will discuss the details of each component, together with training strategies. 

\subsection{Retinex-based Decomposition Network}
Recently, the algorithm unrolling methodology has attracted much attention, and been successfully applied to many low-level vision tasks \citep{deconv_unroll, ADMM-Net, dehaze_unroll, MHF-Net, RCD-Net}. This network designing methodology on the one hand makes the DNN more explainable, since each stage can be related to one iteration in solving an optimization problem; and on the other hand can often enhance the performance of a specific task, since the inference goal can be done in a progressive way by the network, and implicit prior can be learned in a data-driven manner. Therefore, we also design our network inspired by algorithm unrolling, aiming at a better image decomposition.

\subsubsection{Retinex-based Optimization Model for Image Decomposition}
To design a DNN by algorithm unrolling, we first need to construct the optimization model for the concerned problem. For LIE, the general Retinex based optimization can be formulated as
\begin{equation}
\label{model_basic}
\min_{\R,\L}{\frac{1}{2}}\Vert \I-\R\circ \L\Vert_F^2+g_1(\L) + g_2(\R),
\end{equation}
where $g_1(\cdot)$ and $g_2(\cdot)$ are regularizations to characterize the prior knowledge for illumination $\L$ and reflectance $\R$, respectively, such that the Retinex decomposition is well-defined (note that decomposition $\I=\R\circ \L$ is not unique). Then one can unroll the iterative algorithm for solving this optimization, and implicitly embed each of the priors $g_1(\cdot)$ and $g_2(\cdot)$ to a network module, respectively. Such a strategy has been adopted in earlier work \citep{liu2021retinex} that also considered algorithm unrolling based DNN for LIE.

In this work, we take a step further by adding an explicit prior term to the objective function in \eqref{model_basic}:
\begin{equation}
\label{model_used}
\begin{split}
\min_{\R,\L}{\frac{1}{2}}\Vert \I-\R\circ \L\Vert_F^2&+g_1(\L) + g_2(\R)\\
&+\frac{\gamma}{4}\!\sum_{i=x,y}\!\Vert\d_i\otimes\R-\G_i\Vert_F^2,
\end{split}
\end{equation}
where $\d_x=[1,0,-1]^\mathsf{T}$ and $\d_y=[1,0,-1]$ are difference operators on vertical and horizontal directions, respectively, to approximate the gradient of image, $\otimes$ denotes convolution (specifically 1D convolution here), $\gamma$ is a balance parameter, and $\G_i=\left(1+\lambda e^{-\frac{\vert \d_i\otimes \I\vert}{\sigma}}\right)\circ \left(\d_i\otimes \I\right)$ is approximately the amplified gradient of $\I$. As discussed in \citep{li2018structure}, the additional prior term $\sum_{i=x,y}\Vert\d_i\otimes\R-\G_i\Vert_F^2$ can reveal more structure details in the enhanced result. Since simple yet effective, we explicitly adopt such a structure revealing prior in optimization \eqref{model_used}, which is also easy to be embedded in network, by virtue of algorithm unrolling, as shown in the following.

\subsubsection{Solving Algorithm for Decomposition}
Now we need to construct the algorithm for solving optimization problem \eqref{model_used}. Jointly optimizing $\R$ and $\L$ is generally difficult, while if we only optimize one of them with the other fixed, the sub-problem can be relatively easier. Therefore, we adopt this alternative search strategy.

\begin{algorithm*}[tb]
	\caption{Algorithm for solving optimization \eqref{model_used} (differences in network realization are shown in {\color{blue}BLUE})}
	\label{algo1}
	\begin{algorithmic}[1]
		\Statex {\bf For} $k=1,2,\cdots,K$ {\bf Do}
		\Statex {\bf Update} $\L${\bf :}
		\State $d_{\L}^{(k)}=(\R^{(k-1)}\circ (\R^{(k-1)}\circ\L^{(k-1)}-\I))\oslash(\R^{(k-1)}\circ\R^{(k-1)})$
		\State $\hat{\L}^{(k)}=\L^{(k-1)}-\eta_1^{(k)}d_{\L}^{(k)}$ {\color{blue}$~~\Rightarrow~~\hat{\L}^{(k)}=\mathrm{concat}\left(\L^{(k-1)},\R^{(k-1)},d_{\L}^{(k)}\right)$}
		\State $\L^{(k)}=\mathrm{prox}_{\eta_1 g_1(\cdot)}\left(\hat{\L}^{(k)}\right)$ {\color{blue}$~~\Rightarrow~~\{\L^{(k)},\Z_L^{(k)}\}=\mathrm{proxNet}_{\theta_l^{(k)}}\left(\mathrm{concat}\left(\hat{\L}^{(k)},\Z_L^{(k-1)}\right)\right)$}
		\Statex {\bf Update} $\R${\bf :}
		\State $\tilde{\A}_i^{(k)}=\d_i\otimes\R^{(k-1)}-\G_i,~~i=x,y$
		\State $\hat{\A}_i^{(k)}=\d_i\otimes^{\mathsf{T}}\tilde{\A}_i^{(k)},~~i=x,y$
		\State $d_{\R}^{(k)}=\Big(\left(\R^{(k-1)}\circ\L^{(k)}-\I\right)\circ\L^{(k)}+\frac{\gamma}{2}\sum_{i=x,y}\hat{\A}_i^{(k)}\Big)\oslash\left(\L^{(k)}\circ\L^{(k)}+4\gamma\E\right)$
		\State $\hat{\R}^{(k)}=\R^{(k-1)}-\eta_2^{(k)}d_{\R}^{(k)}$
		\State $\R^{(k)}=\mathrm{prox}_{\eta_2 g_2(\cdot)}\left(\hat{\R}^{(k)}\right)$ {\color{blue}$~~\Rightarrow~~\{\R^{(k)},\Z_R^{(k)}\}=\mathrm{proxNet}_{\theta_r^{(k)}}\left(\mathrm{concat}\left(\hat{\R}^{(k)},\Z_R^{(k)}\right)\right)$}
		\Statex {\bf End For}
		\normalsize
	\end{algorithmic}
\end{algorithm*}

{\bf Updating $\L$:} With reflectance $\R$ being fixed, illumination $\L$ can be updated by solving the following optimization:
\begin{equation}
\label{sub_L}
\min_{\L}\frac{1}{2}\Vert\I-\R\circ \L\Vert_{F}^{2} + g_1(\L).
\end{equation}
This problem can be solved by proximal gradient descent (PGD) \citep{prox_alg} type method with the following form of iterations:
\begin{equation}\label{pgd_L}
	\L:=\mathrm{prox}_{\eta_1 g_1(\cdot)}\left(\L-\eta_{1}d_{\L}\right),
\end{equation}
where $\eta_1$ is the step size, $\mathrm{prox}_{f(\cdot)}(\cdot)$ is the proximal operator depending on function $f(\cdot)$, and $d_{\L}$ is the descent direction related to $\L$. Though commonly chosen as gradient, $d_{\L}$ can indeed be any decent direction with respect to function $f_1(\L)=\frac{1}{2}\Vert\I-\R\circ \L\Vert_{F}^{2}$. Therefore, inspired by the fact that Newton method \citep{newton} has better convergence than gradient descent, we adopt the following Newton descent direction (detailed calculations are provided in Appendix):
\begin{equation}\label{direction_L}
\begin{split}
d_{\L}&=\left(\nabla^2f_1(\L)\right)^{-1}\nabla f_1(\L)\\
&=(\R\circ (\R\circ\L-\I))\oslash(\R\circ \R),
\end{split}
\end{equation}
where $\nabla f_1(\L)$ and $\nabla^2f_1(\L)$ denote the gradient and Hessian of $f_1(\L)$, respectively. Note that only element-wise operations are involved in Eq. \eqref{direction_L}, and thus its computational cost is with the same order as that of computing gradient. 

{\bf Updating $\R$:} The optimization for reflectance $\R$, by fixing illumination $\L$, can be written as
\begin{equation}
	\min_{\R}{\frac{1}{2}}\Vert \I-\R\circ \L\Vert_F^2+ g_2(\R)+\frac{\gamma}{4}\!\sum_{i=x,y}\!\Vert\d_i\otimes\R-\G_i\Vert_F^2,
\end{equation}
which can also be solved by PGD type iterations with the following form:
\begin{equation}\label{pgd_R}
	\R:=\mathrm{prox}_{\eta_2 g_2(\cdot)}\left(\R-\eta_{2}d_{\R}\right).
\end{equation}
Following the similar idea for updating $\L$, we try to calculate the Newton direction for descending:
\small
\begin{equation}\label{direction_R}
\begin{split}
	d_{\R}=&\left(\nabla^2f_2(\R)\right)^{-1}\nabla f_2(\R)\\
	\approx&\Big(\!\left(\R\!\circ\!\L\!-\!\I\right)\!\circ\!\L\!+\!\frac{\gamma}{2}\!\sum_{i=x,y}\!\d_i\!\otimes^{\mathsf{T}}\!\!\left(\d_i\!\otimes\!\R\!-\!\G_i\right)\!\Big)\\
	&\oslash\left(\L\!\circ\!\L\!+\!4\gamma\E\right),                 
\end{split}
\end{equation}
\normalsize
where $f_2(\R)={\frac{1}{2}}\Vert \I-\R\circ \L\Vert_F^2+ g_2(\R)+\frac{\gamma}{4}\sum_{i=x,y}\Vert\d_i\otimes\R-\G_i\Vert_F^2$, $\E$ is the matrix with all elements being one, $\otimes^{\mathsf{T}}$ refers to the transposed convolution, and we used the diagonal approximation to the Hessian matrix (Details are provided in Appendix).

{\bf Overall Algorithm:} Putting Eqs. \eqref{pgd_L} \eqref{direction_L} \eqref{pgd_R} \eqref{direction_R} together, %at the $k$th iteration, 
the overall updating procedure can be summarized as Algorithm \ref{algo1}. This algorithm can then be used to guide the design of our decomposition network.

\subsubsection{Network Architecture Inspired by Algorithm Unrolling}
\begin{figure*}[!t]
	\centering
	\subfigure[The entire decomposition network.]{
	\includegraphics[width=\textwidth]{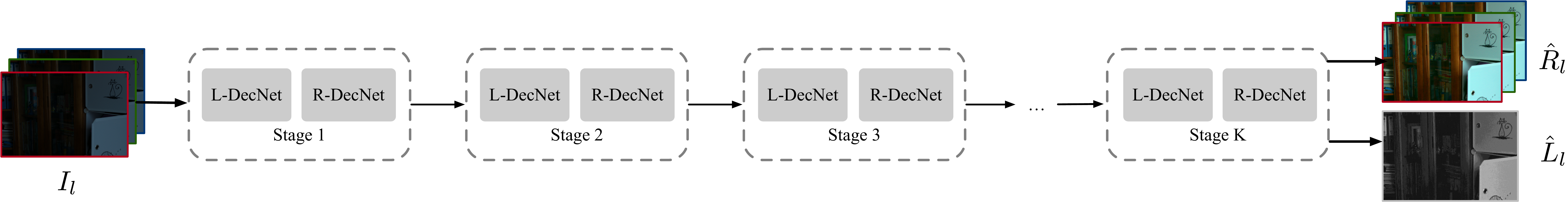}
	}\\
	\subfigure[Design of a single stage in the decomposition network.]{
	\includegraphics[width=\textwidth]{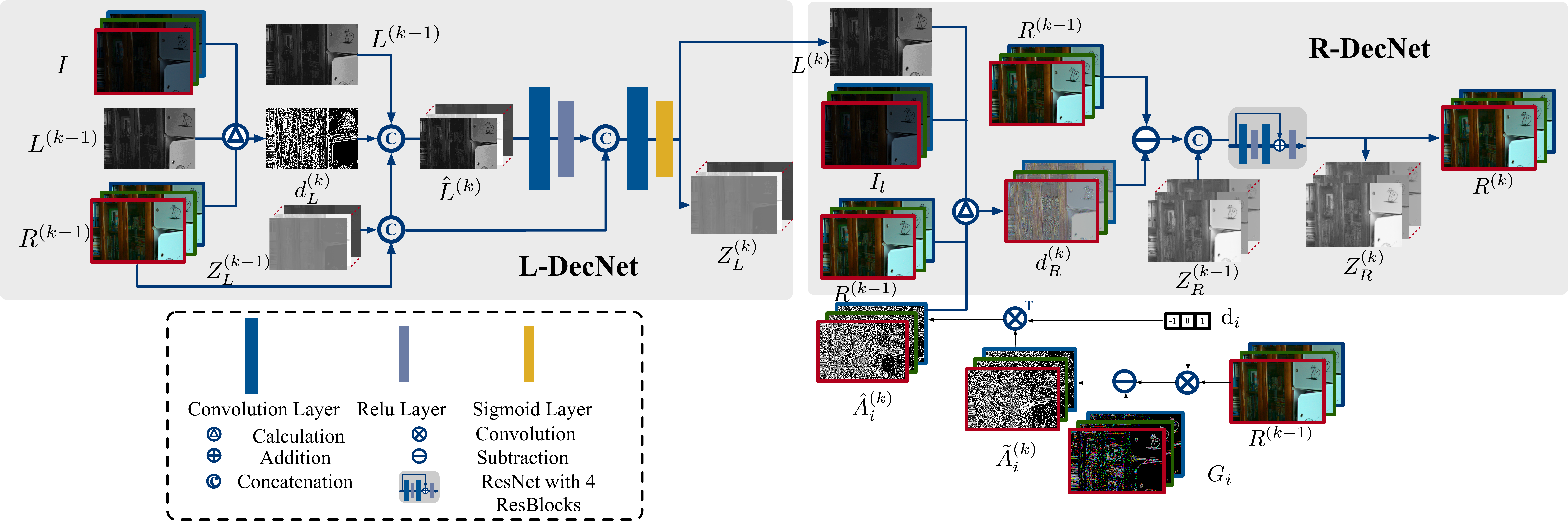}}
%	\caption{(a) presents the proposed DecNet with K stages. The network takes the low-light image $I_l$ as input and outputs the corresponding decomposed reflectance layer $R_l$ and illumination layer $L_l$. (b) presents the illustration of the network architecture at the $k_{th}$ stage. Each stage consists of L-DecNet and R-DecNet to accomplish the update of decomposing reflectance $R$ and illumination $L$.}
	\caption{Illustration of the proposed algorithm unrolling inspired network for Retinex decomposition.}
	\label{fig:dec}
\end{figure*}

Now we can present our network architecture for Retinex based image decomposition inspired by algorithm unrolling. It can be seen that, most of the operations are explicit, and thus can be directly transferred to network structures. The only operations not specified are the two proximal operators, which depend on the choices of $g_1(\cdot)$ and $g_2(\cdot)$. Following the idea in previous work \citep{MHF-Net, RCD-Net}, we can parameterize the proximal operators by network modules, in order to fully make use of the learning capacity of DNNs. In specific, we replace Step 3 and Step 8 in Algorithm \ref{algo1} by
\begin{equation}
	\L^{(k)}=\mathrm{proxNet}_{\theta_l^{(k)}}\left(\hat{\L}^{(k)}\right),
\end{equation}
and
\begin{equation}
	\R^{(k)}=\mathrm{proxNet}_{\theta_r^{(k)}}\left(\hat{\R}^{(k)}\right),
\end{equation}
respectively, where $\mathrm{proxNet}_{\theta_l^{(k)}}(\cdot)$ refers to simply two convolution layers with parameters $\theta_l^{(k)}$, and $\mathrm{proxNet}_{\theta_r^{(k)}}(\cdot)$ is composed of four ResBlocks \citep{he2016deep} parameterized by $\theta_r^{(k)}$ without batch normalization. Then we can obtain two sub-networks for estimating $\L$ and $\R$ each stage, respectively, denoted as \emph{L-DecNet} and \emph{R-DecNet}, which correspond to the updating steps for $\L$ and $\R$ in Algorithm \ref{algo1}. Sequentially stacking stages containing L-DecNet and R-DecNet, the whole network for decomposition can then be constructed, which is illustrated in Fig. \ref{fig:dec}. It can be seen that, the overall decomposition network is constructed with very simple building blocks, such as convolution layers and ResBlocks.

As can be observed from Fig. \ref{fig:dec}, also shown in blue in Algorithm \ref{algo1}, there are two architecture designs inconsistent with the solving algorithm for optimization \eqref{model_used}. First, instead of computing $\hat{\L}^{(k)}$ by $\hat{\L}^{(k)}=\L^{(k-1)}-\eta_1^{(k)}d_{\L}^{(k)}$, we formulate $\hat{\L}^{(k)}=\mathrm{concat}\left(\L^{(k-1)},\R^{(k-1)},d_{\L}^{(k)}\right)$, where $\mathrm{concat}(\cdot)$ is the concatenation operator. Intuitively, this can provide more flexibility for $\mathrm{proxNet}_{\theta_l^{(k)}}(\cdot)$ to integrate the information provided by last estimate $\L^{(k-1)}$ and descent direction $d_{\L}^{(k)}$ (note that $\R^{(k-1)}$ also appears in the calculation of $d_{\L}^{(k)}$), than explicitly doing so with fixed form in Line 3 of Algorithm \ref{algo1}. Second, auxiliary features $\Z_L$ and $\Z_R$ are used as input and output of $\mathrm{proxNet}$ for $\L$ and $\R$, respectively. This is because both $\L$ and $\R$ are with only few channels (1 and 3, respectively), and if we exactly follow the algorithm, they will be firstly expanded to multiple channels and then fused again in each stage of the network, which means that two adjacent stages can only interact with each other through only few fused feature channels, and thus obviously lead to loss of useful information. This strategy has also been used in \citep{RCD-Net}, though the authors did not emphasize it.

\begin{figure}[!t]
	\centering
	\includegraphics[width=\linewidth]{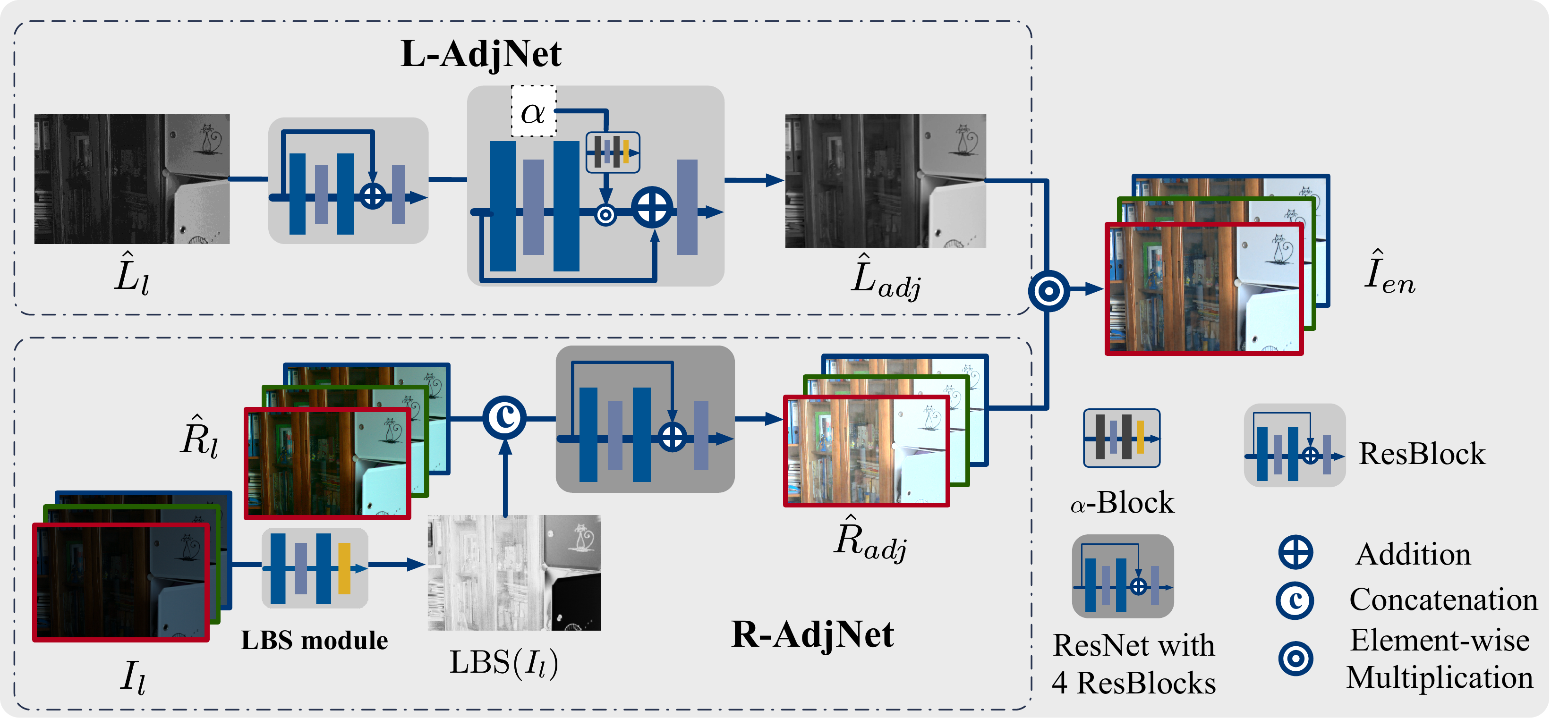}
	\caption{Illustration of the proposed adjustment network.}
	\label{fig:adj}
\end{figure}
 
\begin{figure*} 
	\centering
	\subfigure{
		\includegraphics[width=0.16\linewidth]{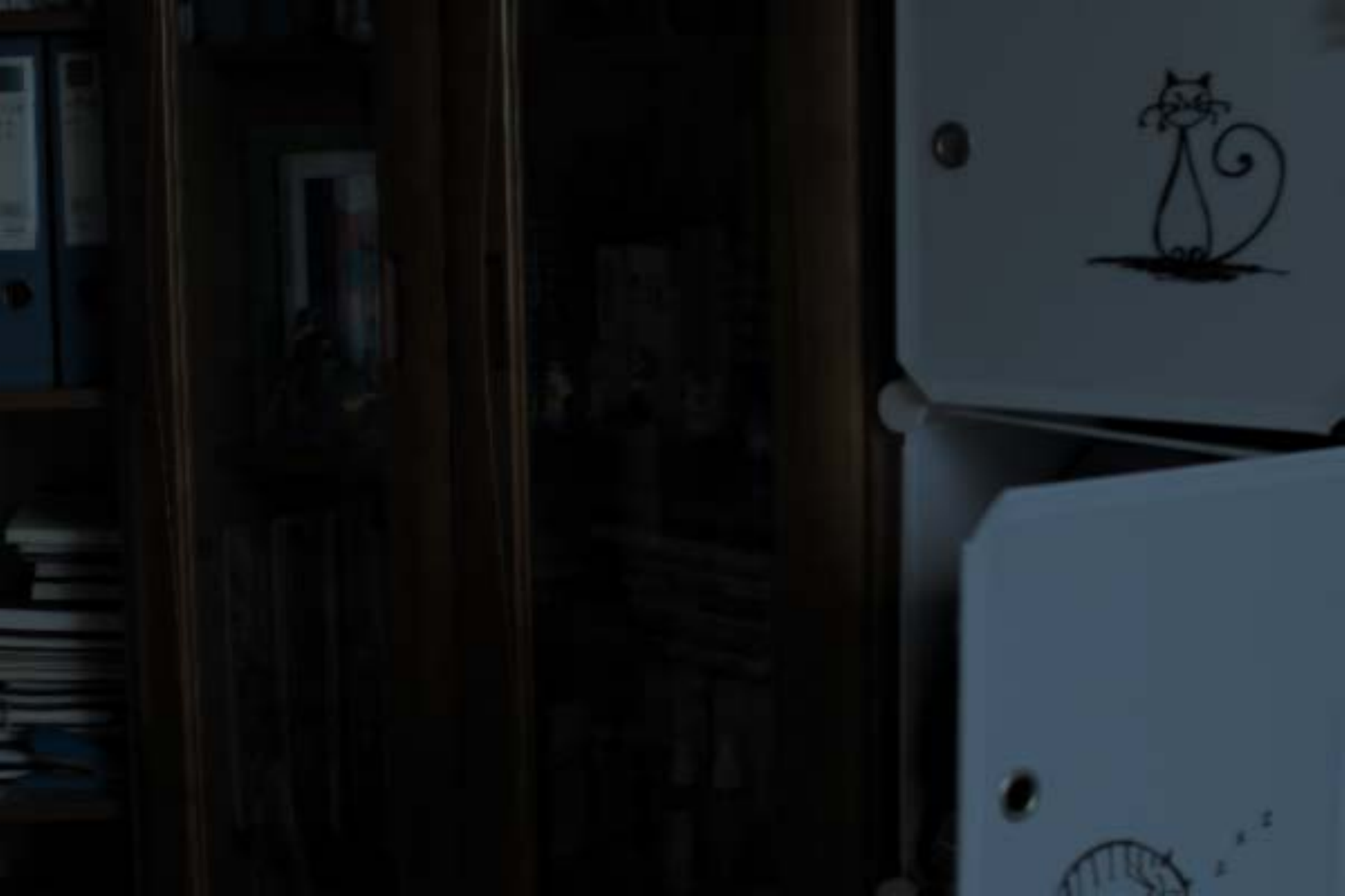}}
	\subfigure{
		\includegraphics[width=0.16\linewidth]{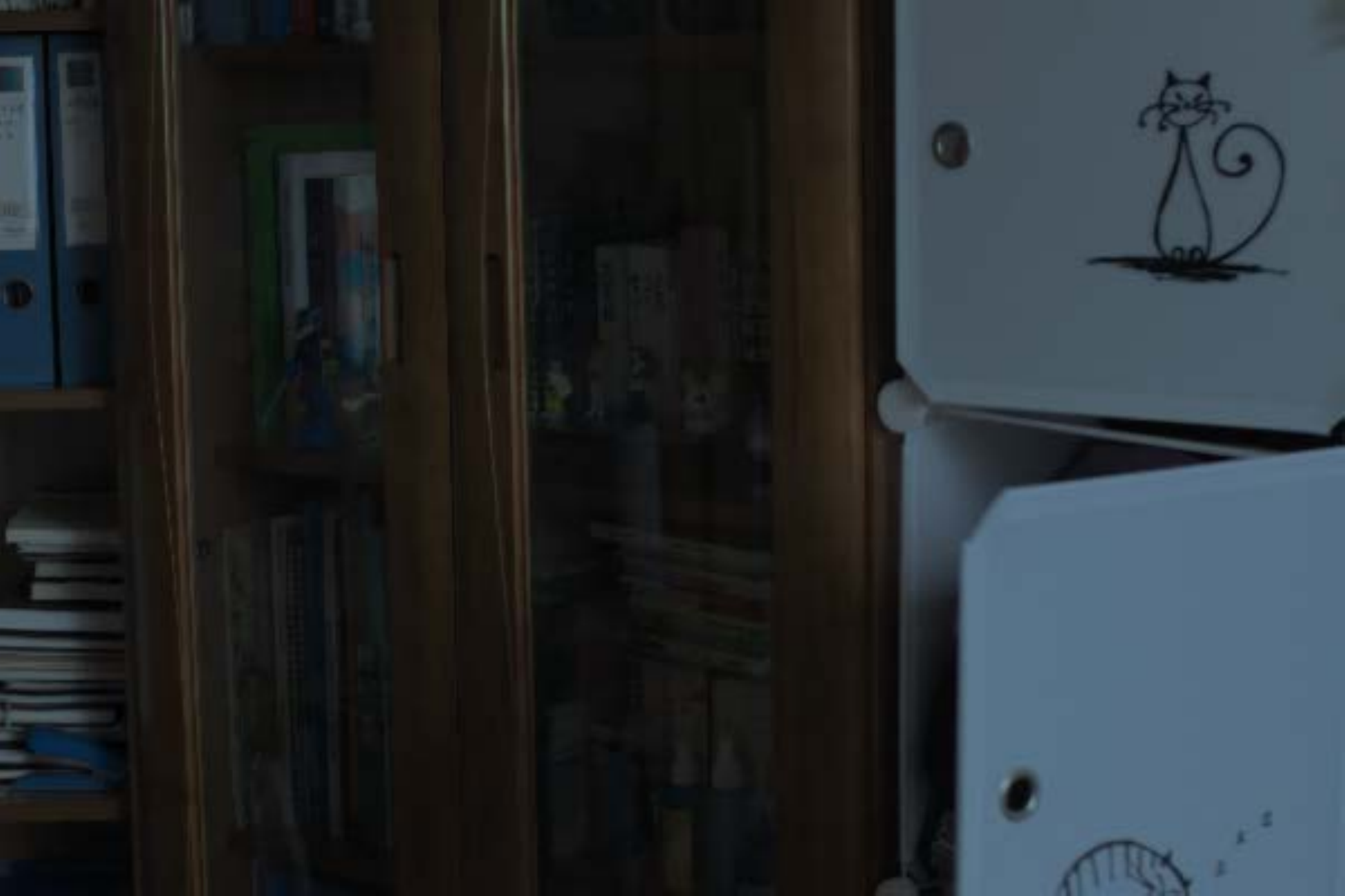}}
	\subfigure{
		\includegraphics[width=0.16\linewidth]{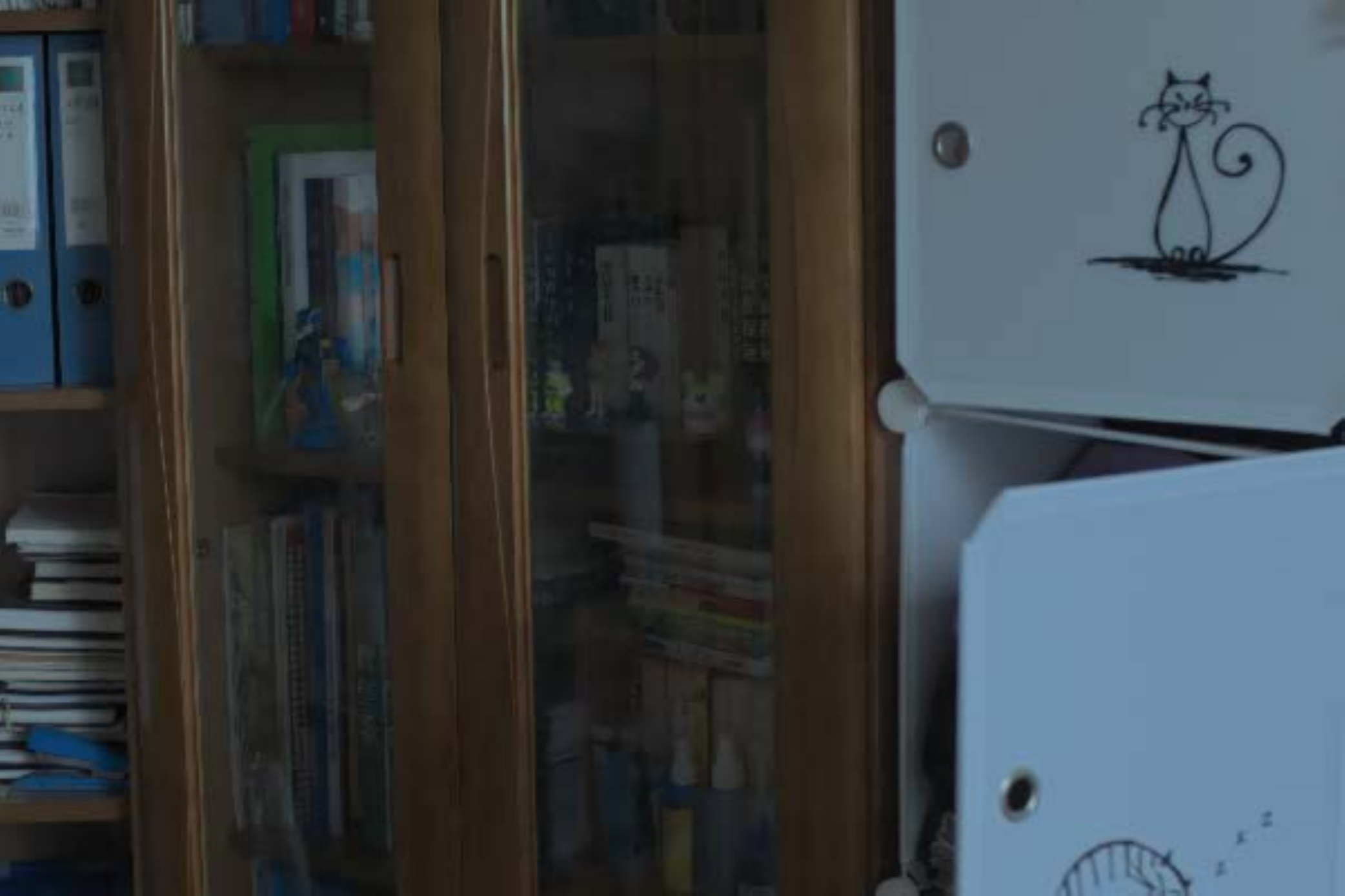}}
	\subfigure{
		\includegraphics[width=0.16\linewidth]{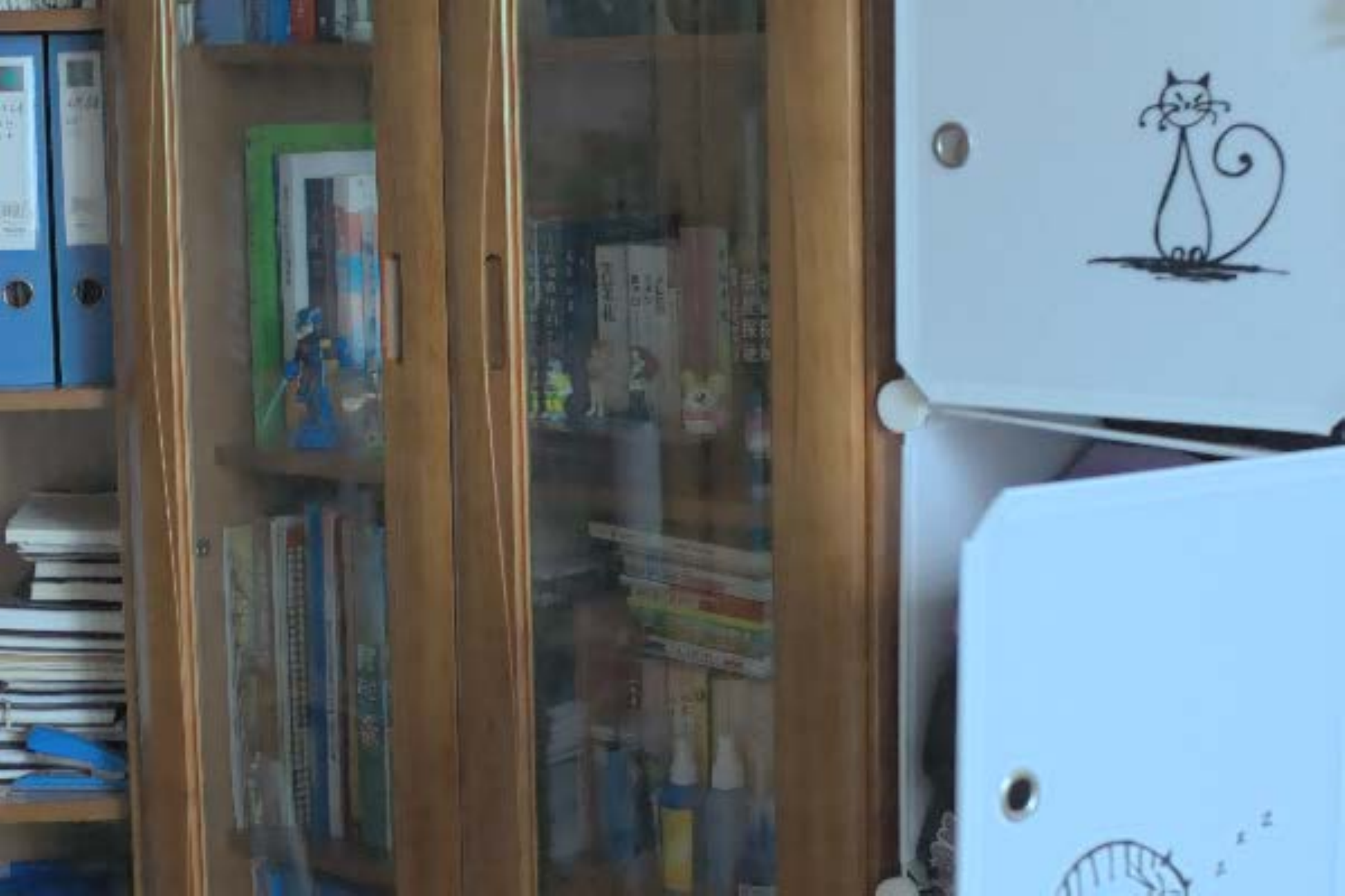}}
	\subfigure{
		\includegraphics[width=0.16\linewidth]{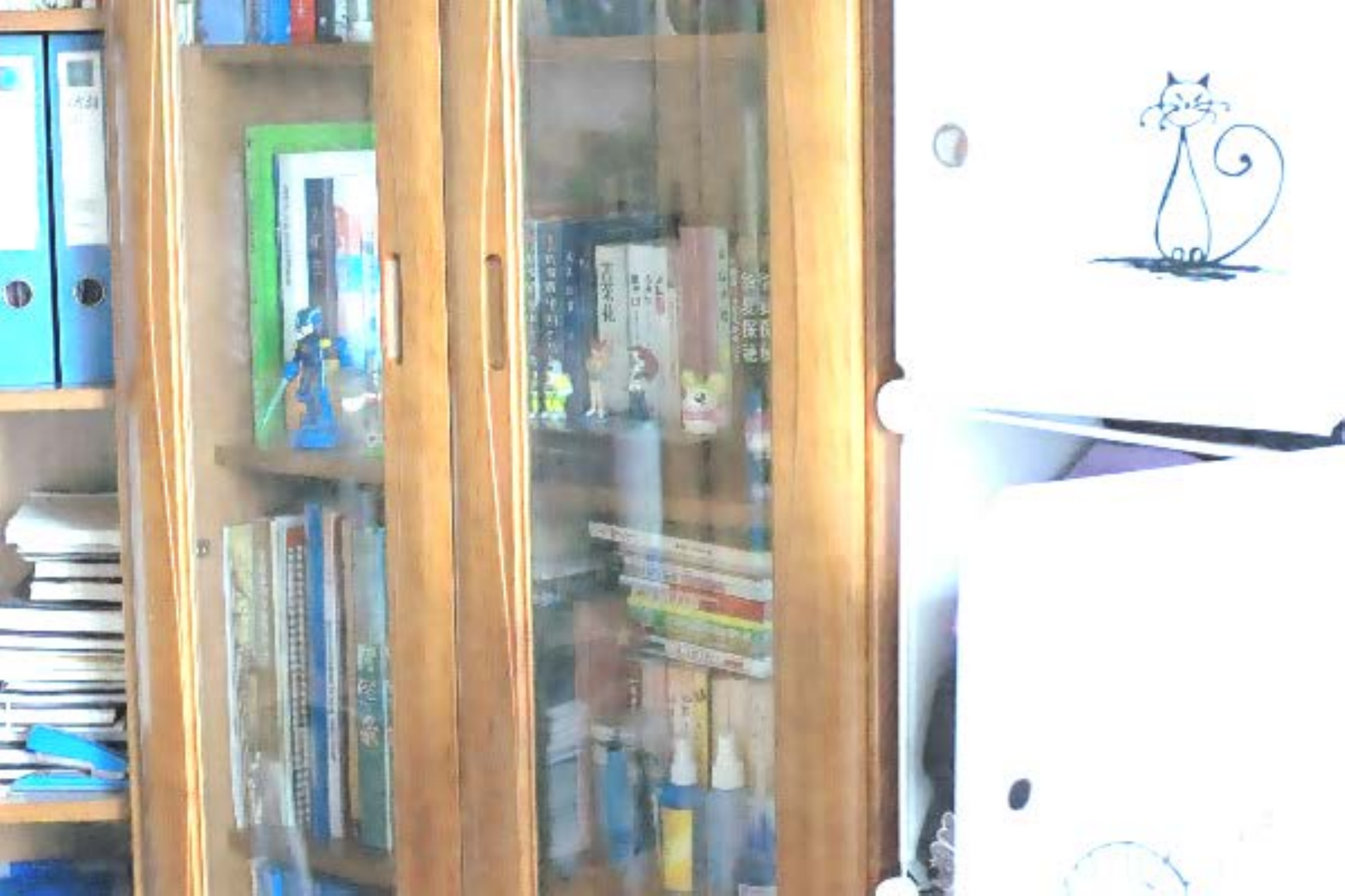}}\\
	\subfigure{
		\includegraphics[width=0.16\linewidth]{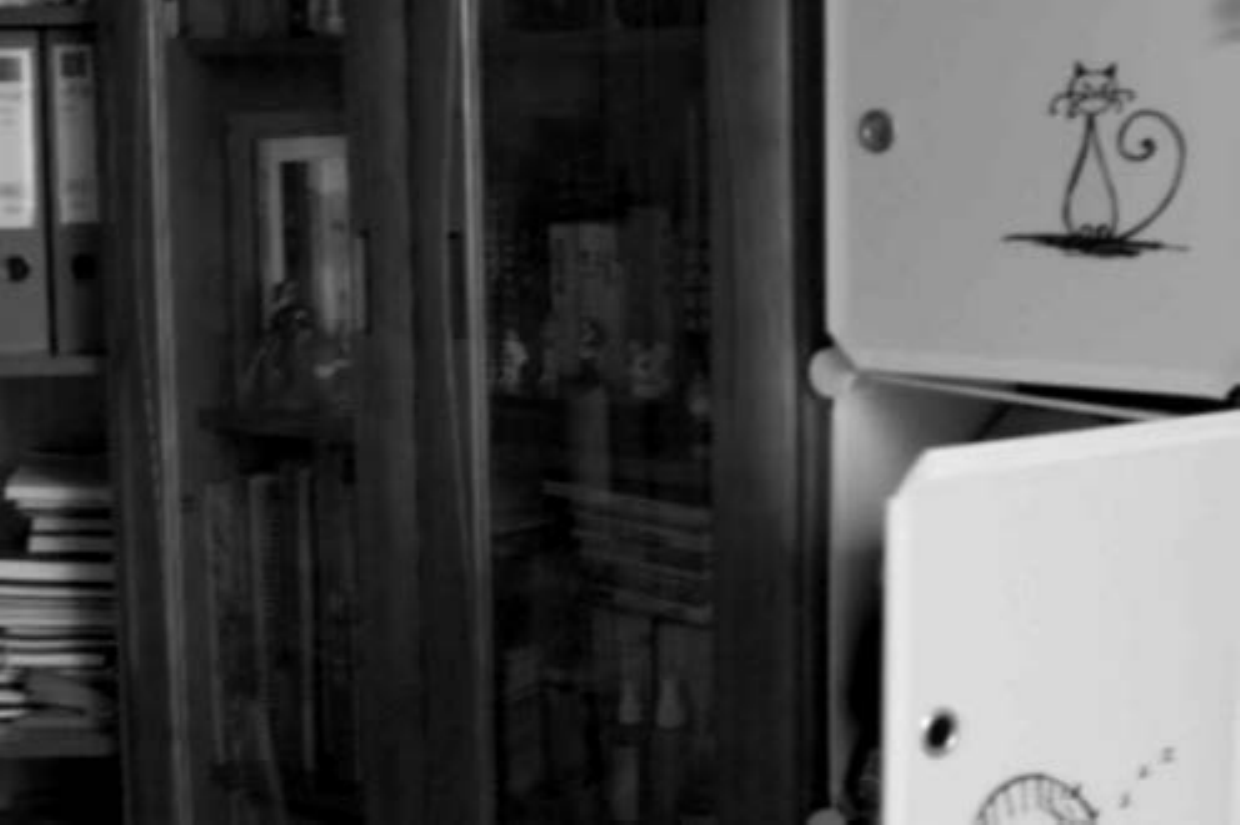}}
	\subfigure{
		\includegraphics[width=0.16\linewidth]{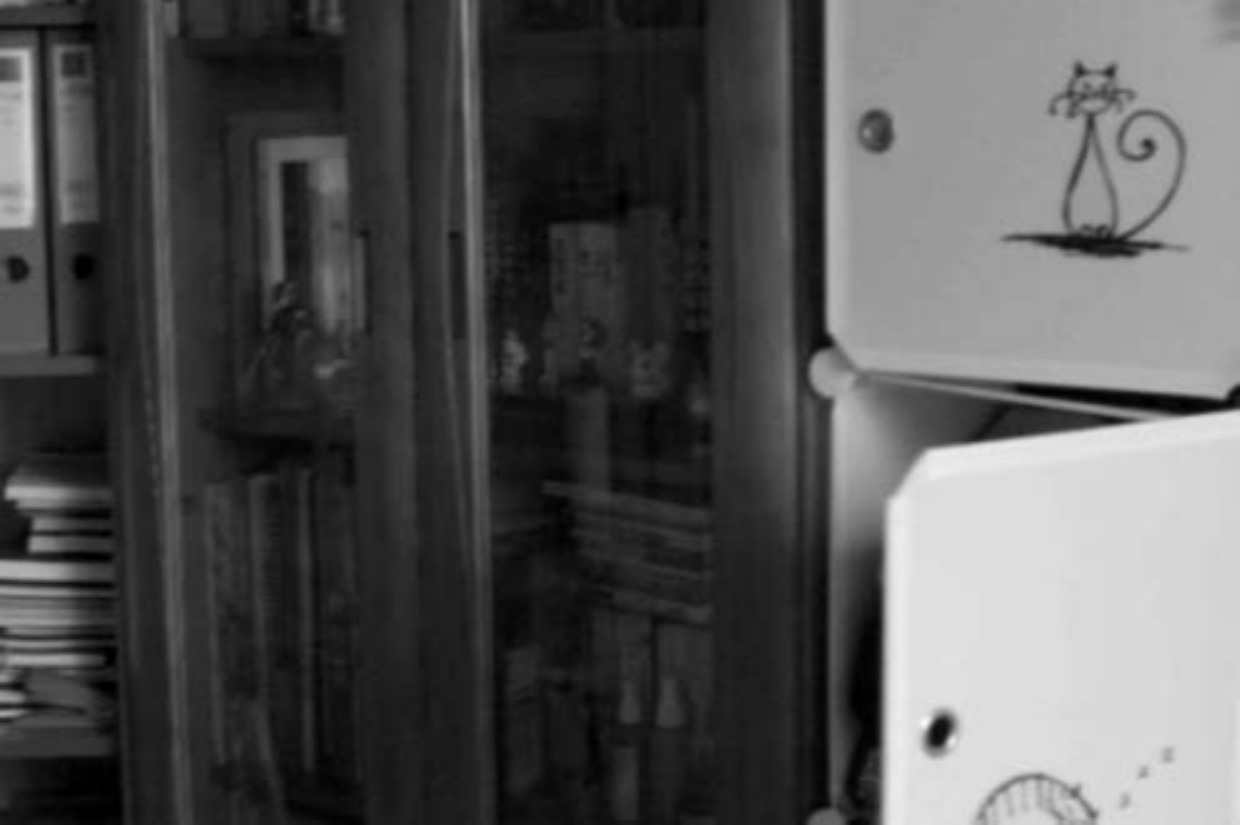}}
	\subfigure{
		\includegraphics[width=0.16\linewidth]{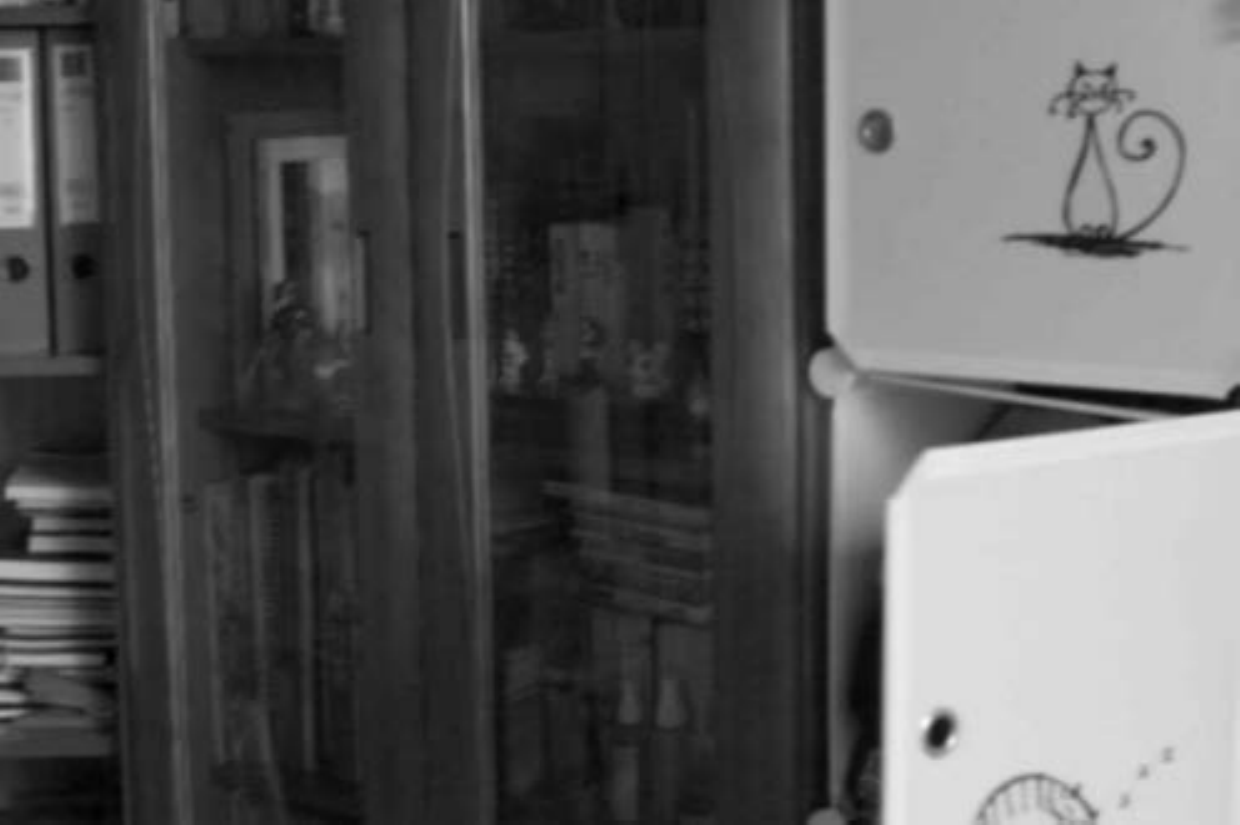}}
	\subfigure{
		\includegraphics[width=0.16\linewidth]{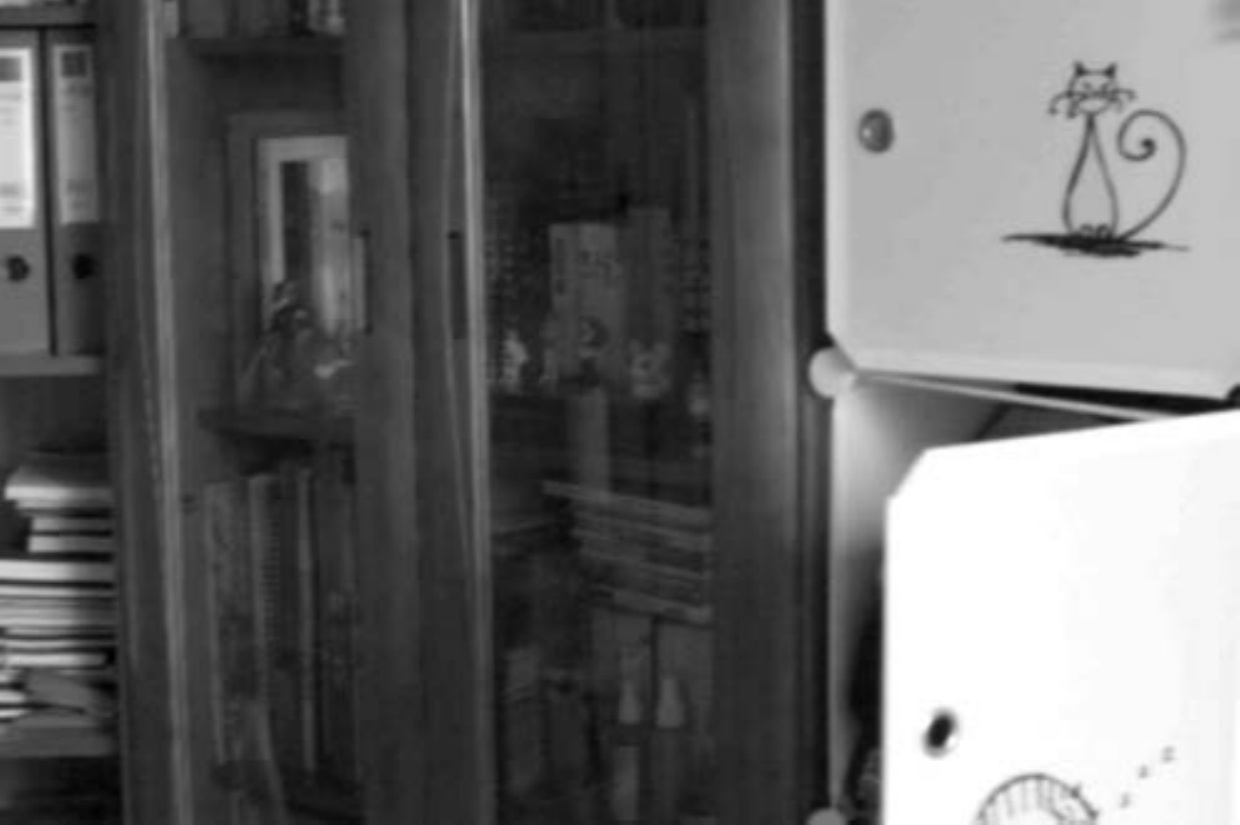}}
	\subfigure{
		\includegraphics[width=0.16\linewidth]{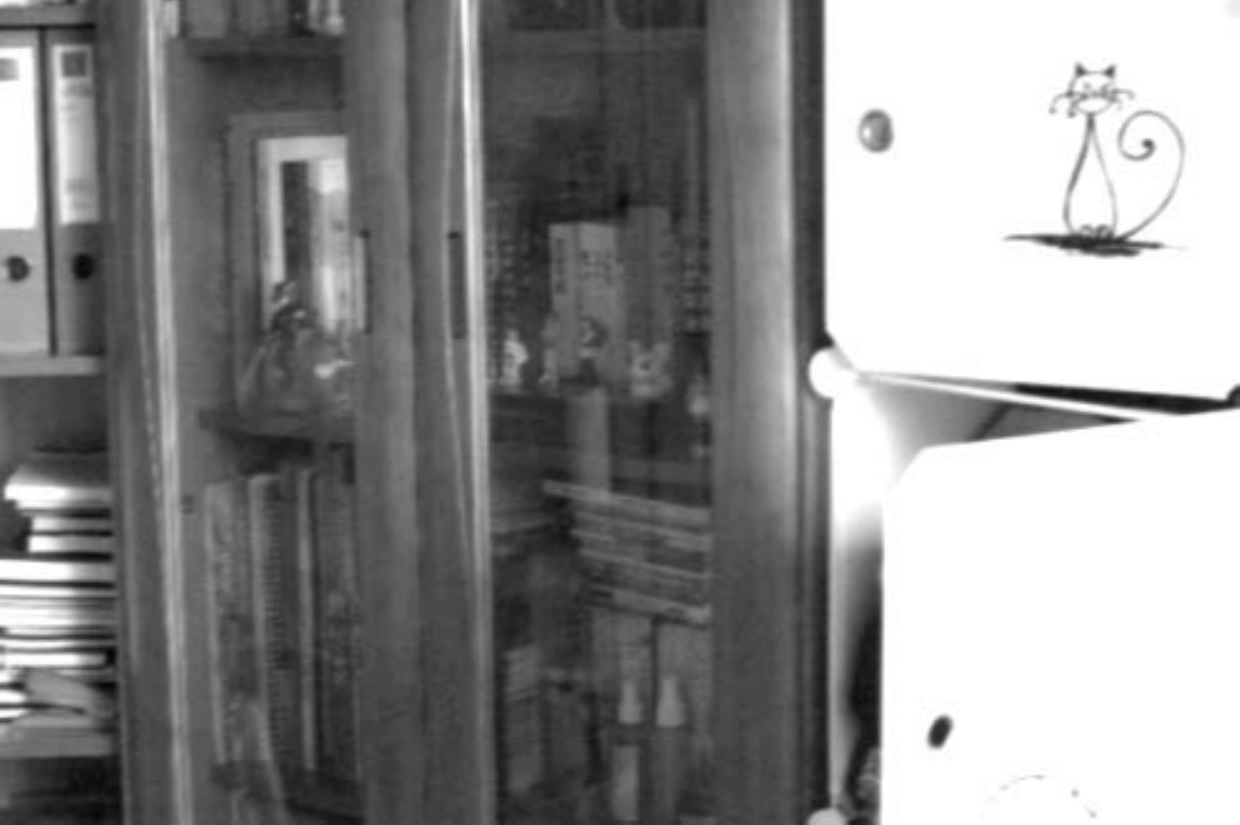}}
	\caption{Enhanced image (top row) and corresponding illumination layer (bottom row) with setting $\alpha=0,0.3,0.5,0.8,1.0$ from left to right. }
	\label{fig:alpha}
\end{figure*}

\subsection{Adjustment Networks}
As discussed before, %and illustrated in Fig. \ref{fig:lol_dec}, 
due to complicated degradations, the Retinex decomposition of a low-light image is not perfect, and thus adjustment is crucial for LIE. In this subsection, we present adjustment networks for both illumination and reflectance. The overall structures of the two networks are shown in Fig. \ref{fig:adj}.

\subsubsection{Illumination Adjustment with Global Brightness}\label{sec:l_adj}
Since the illumination layer is with only one channel, and contains less information about the image, compared with the reflectance layer, we can use relatively simpler network structures. Specifically, we stack two ResBlocks to construct the adjustment network for $\L$, denoted as \emph{L-AdjNet}. Besides, considering that the LIE problem does not have a unique and standard solution, and level of the enhancement may be decided by users, we also introduce a global brightness parameter $\alpha$, similar to that in \citep{zhang2019kindling, zhang2021beyond, fan2020integrating}. Different from concatenating this parameter to $\L$ as the input of the adjustment network considered in \citep{zhang2021beyond}, we multiply this parameter, with a small transformation block, to the skip connection of the second ResBlock, whose effect can be more intuitively explainable. Specifically, our L-AdjNet results in the following adjusting process:
\begin{equation}
	\L^{(s)}=\mathrm{ReLu}\!\left(\L^{(s-1)}\!+\!t_s(\alpha)h_s(\L^{(s-1)})\right)\!,~s=1,2,
\end{equation}
where $t_s(\cdot),~s=1,2$ denote transformations to $\alpha$, with that $t_1(\cdot)\equiv1$ and $t_2(\cdot)$ being a small network, and $h_s(\cdot),~s=1,2$ denote the convolution layers before added to the skip connection. Therefore, the whole adjusting process can be intuitively interpreted as a two-step projected gradient descent \citep{boyd2004convex} for $\L$ (note that the $\mathrm{ReLu}$ activation is intrinsically a projection operation to ensure $\L$ non-negative), with the second step size controlled by $\alpha$, and gradient computed by $-h_s(\cdot),~s=1,2$. We have tried to let $t_1(\cdot)$ also being a network, but the performance degenerated a little, which indicates that the first step adjustment should be sufficient enough. The influence of $\alpha$ is visually illustrated in Fig. \ref{fig:alpha}.

The next issue is that how to determine $\alpha$. In the training phase, since both the low-light and normal-light images are available, we calculate this parameter by $\alpha=\frac{1}{HW}\left\|\big(\tilde{\I}_h-\tilde{\I}_l\big)\oslash\tilde{\I}_h\right\|_1$, where $\tilde{\I}$ denotes the gray image converted from its color version $\I$, which can be interpreted as the mean relative differences of the light intensity between normal-light and low-light images. In the testing phase, if normal-light image corresponding to the input is available, this calculation is still applicable, as in our experiments on paired datasets. However, in practice, normal-light image is not available, therefore, users may vary this parameter and choose the one leading to the best visual result subjectively. For easing such manually parameter specification, we also propose a self-supervised strategy to fine-tune this parameter, as well as the whole adjustment network, detailed in Section \ref{sec:finetune}.

\begin{figure} 
	\centering
	\subfigure[]{
		\includegraphics[width=0.28\linewidth]{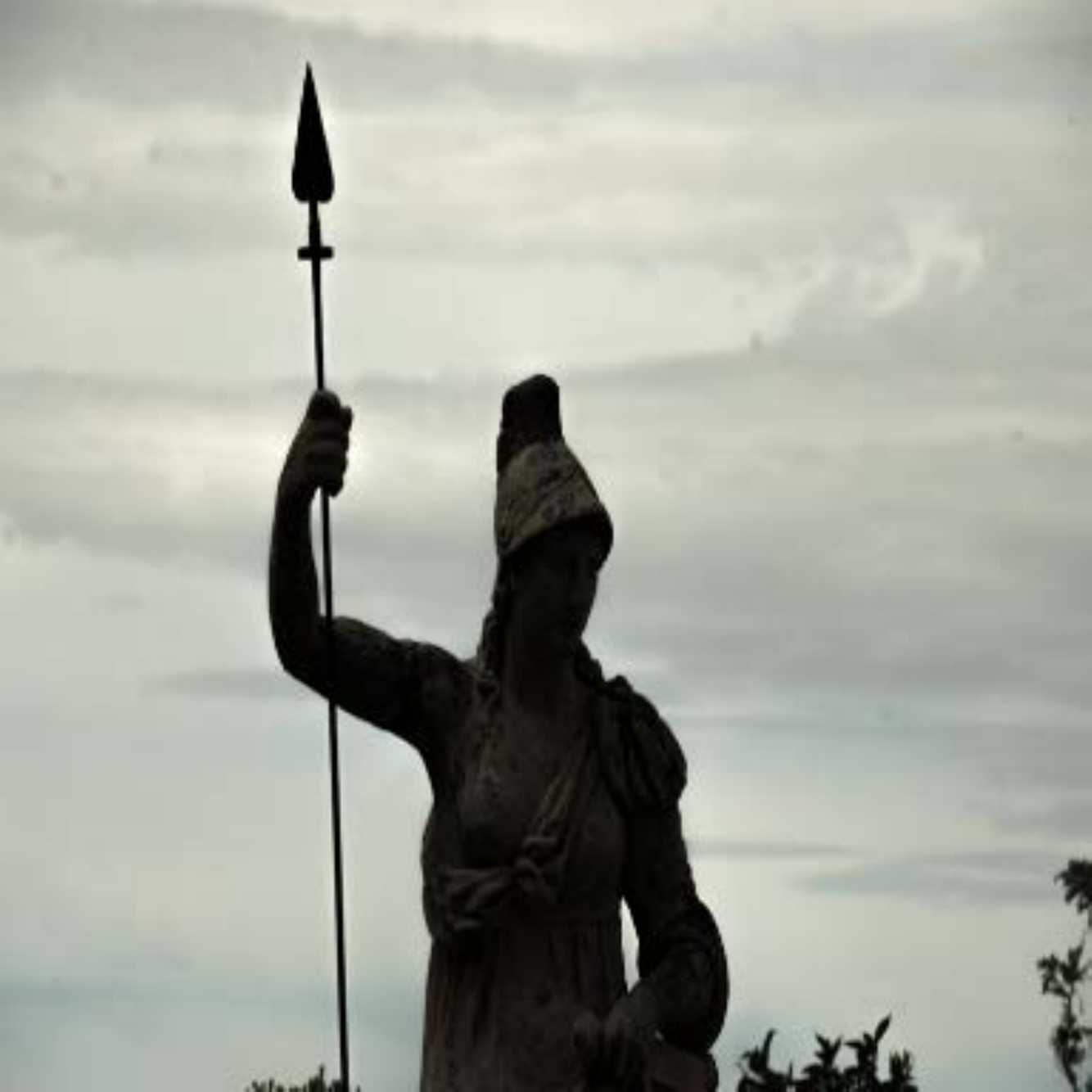}\label{lbs-ab1}}
	\subfigure[]{
		\includegraphics[width=0.28\linewidth]{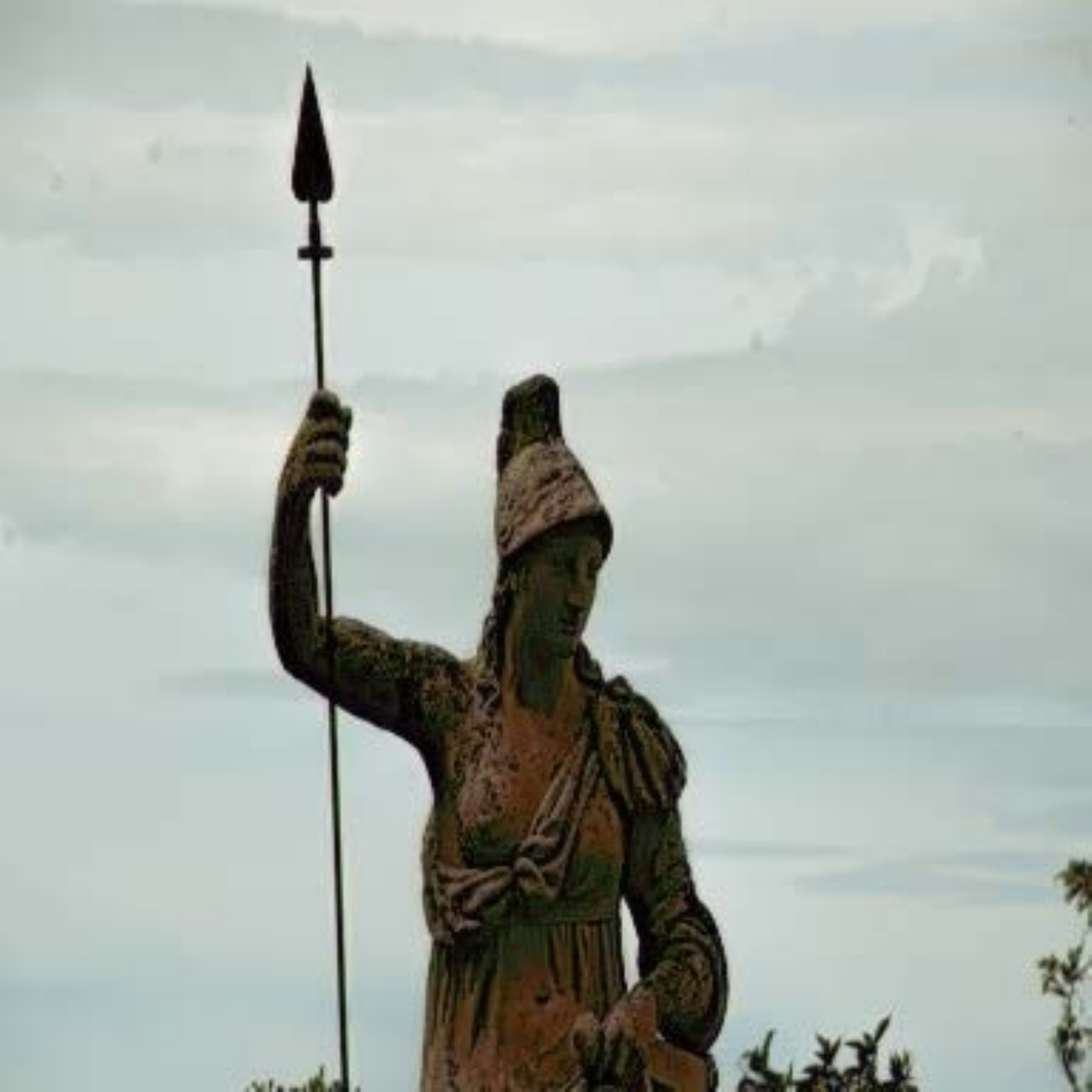}\label{lbs-ab2}}
	\subfigure[]{
		\includegraphics[width=0.28\linewidth]{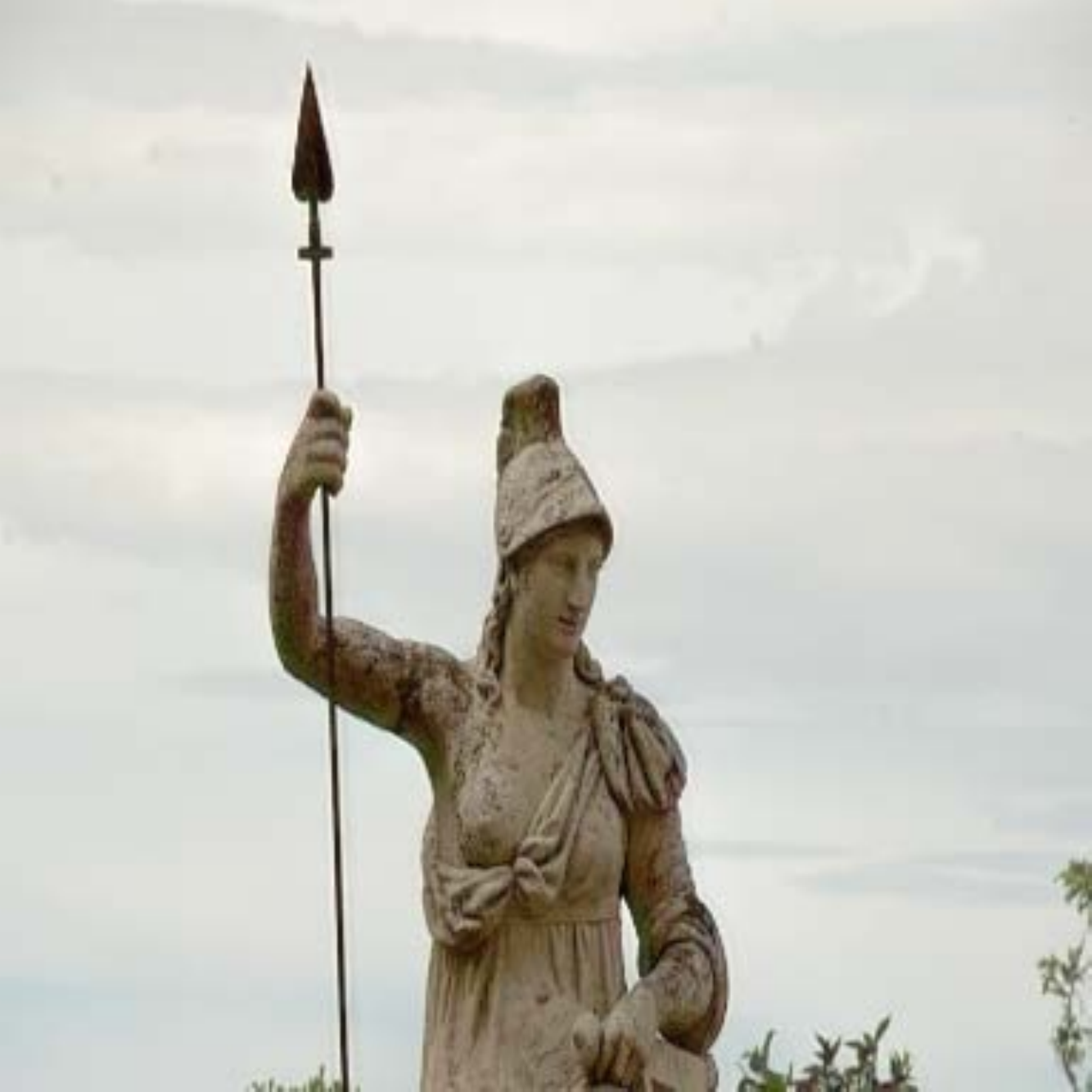}\label{lbs-ab3}}\\
	\subfigure[]{
		\includegraphics[width=0.28\linewidth]{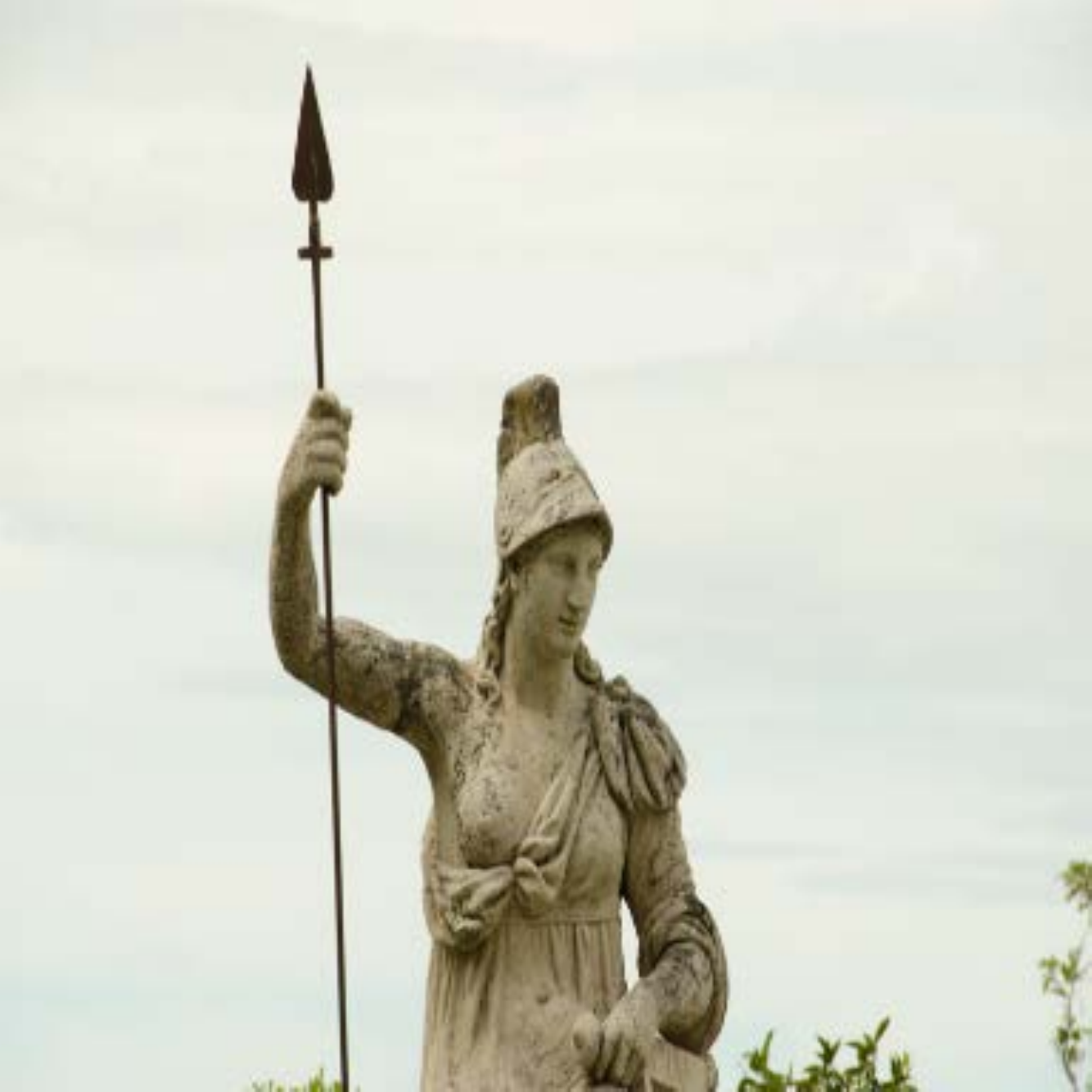}\label{lbs-ab4}}
	\subfigure[]{
		\includegraphics[width=0.28\linewidth]{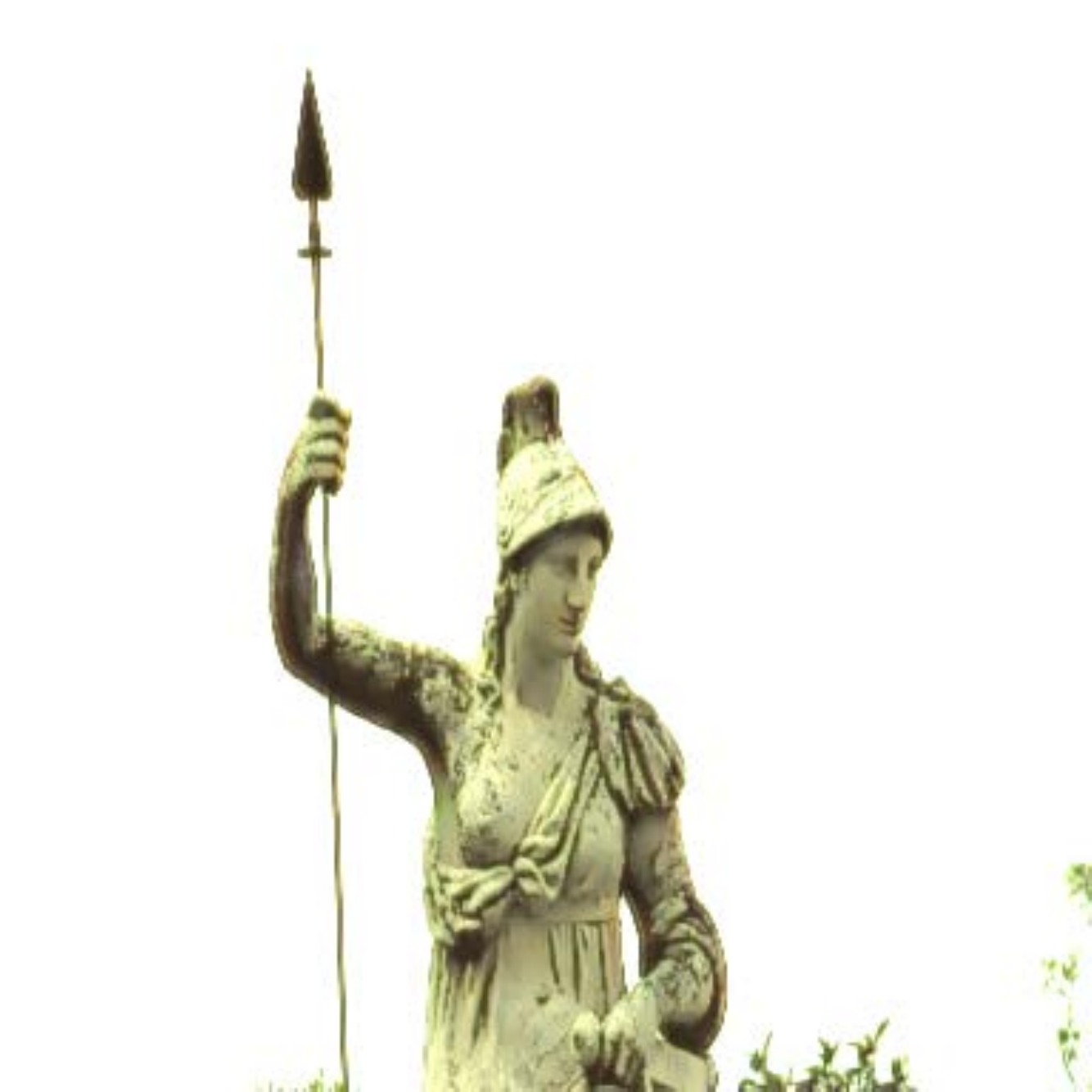}\label{lbs-ab5}}
	\subfigure[]{
		\includegraphics[width=0.28\linewidth]{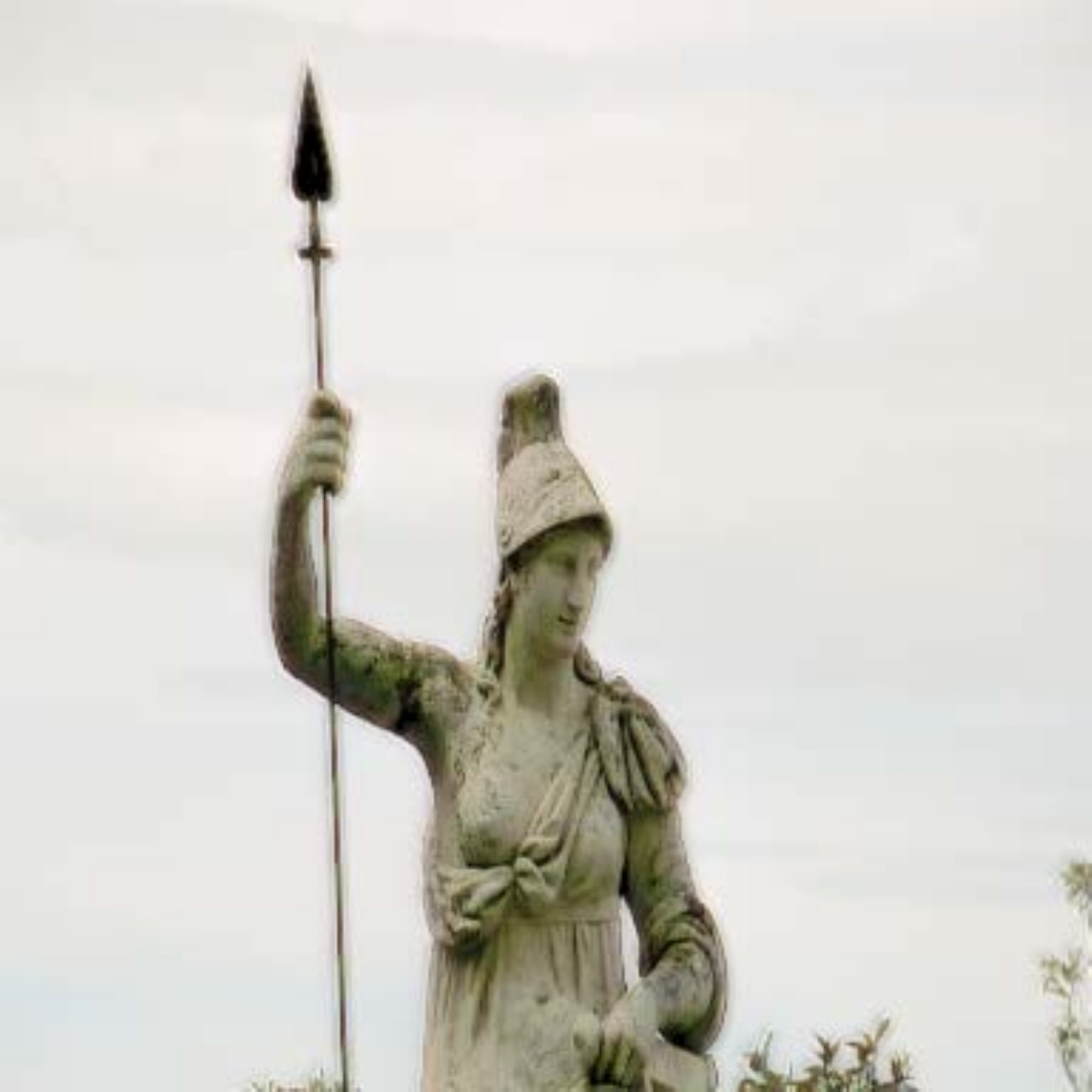}\label{lbs-ab6}}
	\caption{(a) is the input image, and (d) is the corresponding normal-light image. (b) and (c) are the enhanced results without and with the LBS module, respectively, and (e) and (f) are the corresponding reflectance layers.}
	\label{fig:ab-lbs} 
\end{figure}

\subsubsection{Reflectance Adjustment with Local Brightness Sensitivity}\label{sec:r_adj}
As discussed in \citep{zhang2021beyond}, the degradations involved in $\R$ are more complicated, and 
information in addition to $\R$ itself is important to guide adjusting $\R$. Our major observation is that, as shown in Fig. \ref{lbs-ab1}, different regions in the image have different sensitivities in brightness under the same environment, which tends to result in spatially variant degradations of $\R$. In contrast, the brightness of normal-light image, as shown in Fig. \ref{lbs-ab4}, is more spatially homogeneous. Therefore, the pixel-wise brightness difference between low-light and normal-light images can be a useful feature for adjusting $\R$. However, since the groundtruth normal-light image is unavailable at test time, we try to predict such information by the low-light image. Specifically, we design a simple local brightness sensitivity (LBS) module with a two-layer convolution structure to achieve this goal. During training, the target output of LBS module is defined as $(\tilde{\I}_h-\tilde{\I}_l)\oslash\tilde{\I}_h$, where $\tilde{\I}$ denotes the gray image converted from its color version $\I$. This target can be roughly regarded as the relative brightness difference between low-light and normal-light images. The extracted LBS feature is then concatenated with $\R$, as the input of a ResNet with 4 ResBlocks for final adjustment. As can be seen from Fig. \ref{fig:ab-lbs}, with this LBS feature extraction module, the obtained reflectance is with more details, which leads to better enhancement result, well addressing the non-uniform brightness issue.

\subsection{Loss Functions and Network Training}\label{sec:loss_train}
\subsubsection{Loss Function for Decomposition Network}
The training of decomposition network is not easy, since only image pairs of $\I_h$ and $\I_l$ are available, while neither of them have groundtruth reflectance and illumination. Therefore, following the idea from \citep{zhang2021beyond}, we resort to several properties by Retinex model: consistency in reflectance between low-light and normal-light images, smoothness in illumination and reconstruction by decomposition. Specifically, denoting $(\I_l,\I_h)$ as the training pair of low-light image and corresponding normal-light one, and $\left(\R_l^{(k)},\L_l^{(k)}\right)$ and $\left(\R_h^{(k)},\L_h^{(k)}\right)$ the Retinex decomposition for low-light and normal-light images, respectively, at the $k$th stage, there are three terms included in our loss function: 
\begin{itemize}
	\item $\mathcal{L}_R^{(k)}=\frac{1}{HW}\left\|\R_l^{(k)}-\R_h^{(k)}\right\|_F^2$, which measures the closeness between the reflectance layers obtained from low-light and normal-light images, respectively;
	\item $\mathcal{L}_L^{(k)}=\frac{1}{HW}\Big(\Big\|\frac{\nabla\L_l^{(k)}}{\max(\lvert\nabla\I_l\rvert,\epsilon)}\Big\|_1+\Big\|\frac{\nabla\L_h^{(k)}}{\max(\lvert\nabla\I_l\rvert,\epsilon)}\Big\|_1\Big)$\footnote{Here, $\nabla$ stands for the first order difference operator containing both horizontal and vertical directions, i.e., $\nabla\I=\left[\d_x\otimes\I,\d_y\otimes\I\right]$ using the notation in Eq. \eqref{model_used}.}, which is computed in the gradient domain in order to enforce smoothness to the illumination layers;
	\item $\mathcal{L}_{\textrm{rec}}^{(k)}=\frac{1}{HW}\Big(\left\|\I_l-\R_l^{(k)}\circ\L_l^{(k)}\right\|_F^2+\left\|\I_h-\R_h^{(k)}\circ\L_h^{(k)}\right\|_F^2\Big)$, which tries to to reconstruct the images by decomposed reflectance and illumination according to Retinex model.
\end{itemize}

Combining these terms together, the overall loss is then defined as
\begin{equation}
\begin{split}
\mathcal{L}_{\textrm{dec}}=&\gamma_{R}\sum\nolimits_{k=1}^{K}\mathcal{L}_R^{(k)}+\gamma_{L}\sum\nolimits_{k=1}^{K}\mathcal{L}_L^{(k)}\\
&+\gamma_{\textrm{rec}}\sum\nolimits_{k=1}^{K}\mathcal{L}_{\textrm{rec}}^{(k)},
\end{split}
\end{equation}
where $\gamma_{R}$, $\gamma_{L}$ and $\gamma_{\textrm{rec}}$ are the balance parameters, and set to $0.1$, $1$ and $1000$, respectively, in our experiments. Note that, since our estimation for reflectance and illumination is realized stage-wise in a progressive way, we can place loss at each stage instead of only penalize the final output, by virtue of the algorithm unrolling methodology.

\subsubsection{Loss Function for Adjustment Networks}
For the adjustment networks, there are also several components to be considered in loss function. Specifically, the adjustment results $\hat{\R}_{\textrm{adj}}$ and $\hat{\L}_{\textrm{adj}}$ for both reflectance and illumination should be supervised, as well as the final enhanced image. In addition, as mentioned in Section \ref{sec:r_adj}, the LBS module should also be supervised. We then discuss the loss term for each of them.

For $\hat{\R}_{\textrm{adj}}$ and $\hat{\L}_{\textrm{adj}}$, considering that ideally the Retinex decomposition to the normal-light image can be treated as groundtruth, we can input the normal-light image to the decomposition network and use the output, with stop-gradient operation, as the pseudo groundtruth to supervise $\hat{\R}_{\textrm{adj}}$ and $\hat{\L}_{\textrm{adj}}$. Specifically, denoting $\left(\R_h^{(K)},\L_h^{(K)}\right)$ the output of the decomposition network by inputting normal-light image $\I_h$, we define
\begin{equation}
\mathcal{L}_{R_{\textrm{adj}}}=1-\mathrm{SSIM}\left(\hat{\R}_{\textrm{adj}},\mathrm{StopGradient}\left(\R_h^{(K)}\right)\right),
\end{equation}
and
\begin{equation}
	\mathcal{L}_{L_{\textrm{adj}}}=\frac{1}{HW}\left\|\hat{\L}_{\textrm{adj}}-\mathrm{StopGradient}\left(\L_h^{(K)}\right)\right\|_F^2,
\end{equation}
respectively, where $\mathrm{StopGradient}(\cdot)$ denotes the stop-gradient operation, and $\mathrm{SSIM}(\cdot,\cdot)$ is structural similarity (SSIM) \citep{wang2004image} between two images. Similarly, for the LBS module, we also simply measure the closeness between output and target by
\begin{equation}
	\mathcal{L}_{\mathrm{lbs}}=\frac{1}{HW}\left\|\mathrm{LBS}(\I_l)-\big(\tilde{\I}_h-\tilde{\I}_l\big)\oslash\tilde{\I}_h\right\|_F^2.
\end{equation}

Since the groundtruth normal-light image $\I_h$ is available in the training phase, the final enhanced image $\hat{\I}_{\textrm{en}}=\hat{\R}_{\textrm{adj}}\circ\hat{\L}_{\textrm{adj}}$ can be easily supervised by minimizing the closeness between them. In addition to the commonly used mean squared error (MSE) loss, we also adopt the perceptual loss \citep{perceptual_loss1, perceptual_loss2} and color loss \citep{wang2019underexposed} (the second and third term below), and the overall loss is
\begin{equation}\label{eq:loss_en}
\begin{split}
	\mathcal{L}_{\textrm{en}}=&\frac{1}{HW}\left\|\hat{\I}_{\textrm{en}}\!-\!\I_h\right\|_F^2\\
	&+\frac{1}{HW}\left\|\phi_{\mathrm{vgg}}(\hat{\I}_{\textrm{en}})-\phi_{\mathrm{vgg}}(\I_h)\right\|_F^2\\
	&+\frac{1}{HW}\sum\nolimits_p\angle\big((\hat{\I}_{\textrm{en}})_p,(\I_h)_p\big)
\end{split},
\end{equation}
where $\phi_{\mathrm{vgg}}(\cdot)$ is a pre-trained VGG-Net \citep{simonyan2014very}, $\angle(\cdot,\cdot)$ denotes the angle between two RGB vectors, and $p$ indicates the position of an image pixel. We will show in Section \ref{ablation} that each term in Eq. \eqref{eq:loss_en} facilitate to a promising performance.

Putting all of above loss terms together, the final loss function for adjustment networks is defined as
\begin{equation}
	\mathcal{L}_{\textrm{adj}}=\eta_{L}\mathcal{L}_{L_{\textrm{adj}}}+\eta_{R}\mathcal{L}_{R_{\textrm{adj}}}+\eta_{\textrm{lbs}}\mathcal{L}_{\textrm{lbs}}+\eta_{\textrm{en}}\mathcal{L}_{\textrm{en}},
\end{equation}
where the balance parameters $\eta_{L}$, $\eta_{R}$, $\eta_{\textrm{lbs}}$ and $\eta_{\textrm{en}}$ are set to $0.05$, $0.05$, $0.1$ and $20$, respectively.

\subsubsection{Training Strategy}\label{sec:train_strategy}
Our whole framework, including both decomposition and adjustment networks, is trained in an end-to-end fashion. However, considering that decomposition network and adjustment network play different roles in the LIE task, we use two optimizers for these two parts (detailed in Section \ref{sec:exp_implement}), respectively.

As for the training data, we use a simple yet effective data augmentation technique. Specifically, note that the normal-light image itself can be seen as a special case of light-light image, except that the light condition is normal. Therefore, we add image pairs in the form of $(\I_h,\I_h)$, in addition to $(\I_l,\I_h)$, to the training set. In this sense, the range of light conditions of low-light image in the dataset can be enlarged.

\subsection{Self-supervised Fine-tuning at Test Time}\label{sec:finetune}
As discussed in Section \ref{sec:l_adj}, there is a global brightness parameter $\alpha$ to be specified manually in applying our network to real low-light images. Therefore, we present a self-supervised fine-tuning strategy to avoid such manual specification at test time. It should be mentioned that this fine-tuning strategy can not only be used to tune $\alpha$, but indeed also be used to fine-tune a part of the whole network. To our knowledge, the test time training (fine-tuning) can be dated back to \citep{test-time-training}, and similar methodology has been used in image dehazing problem \citep{chen2021psd}.

\begin{figure*} 
	\centering
	\subfigure[Input]{
		\includegraphics[width=0.18\linewidth]{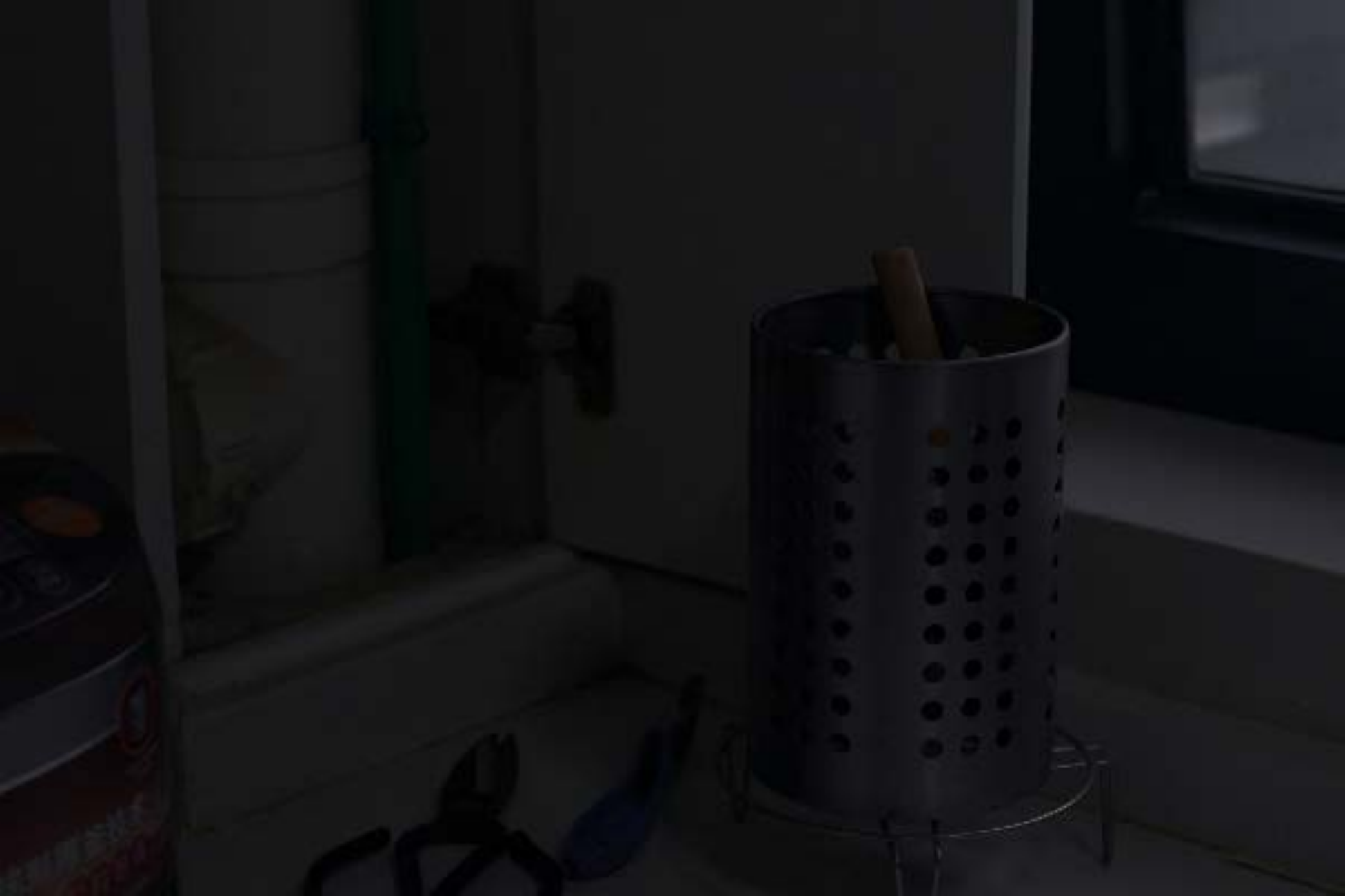}}
	\subfigure[Synthesized]{
		\includegraphics[width=0.18\linewidth]{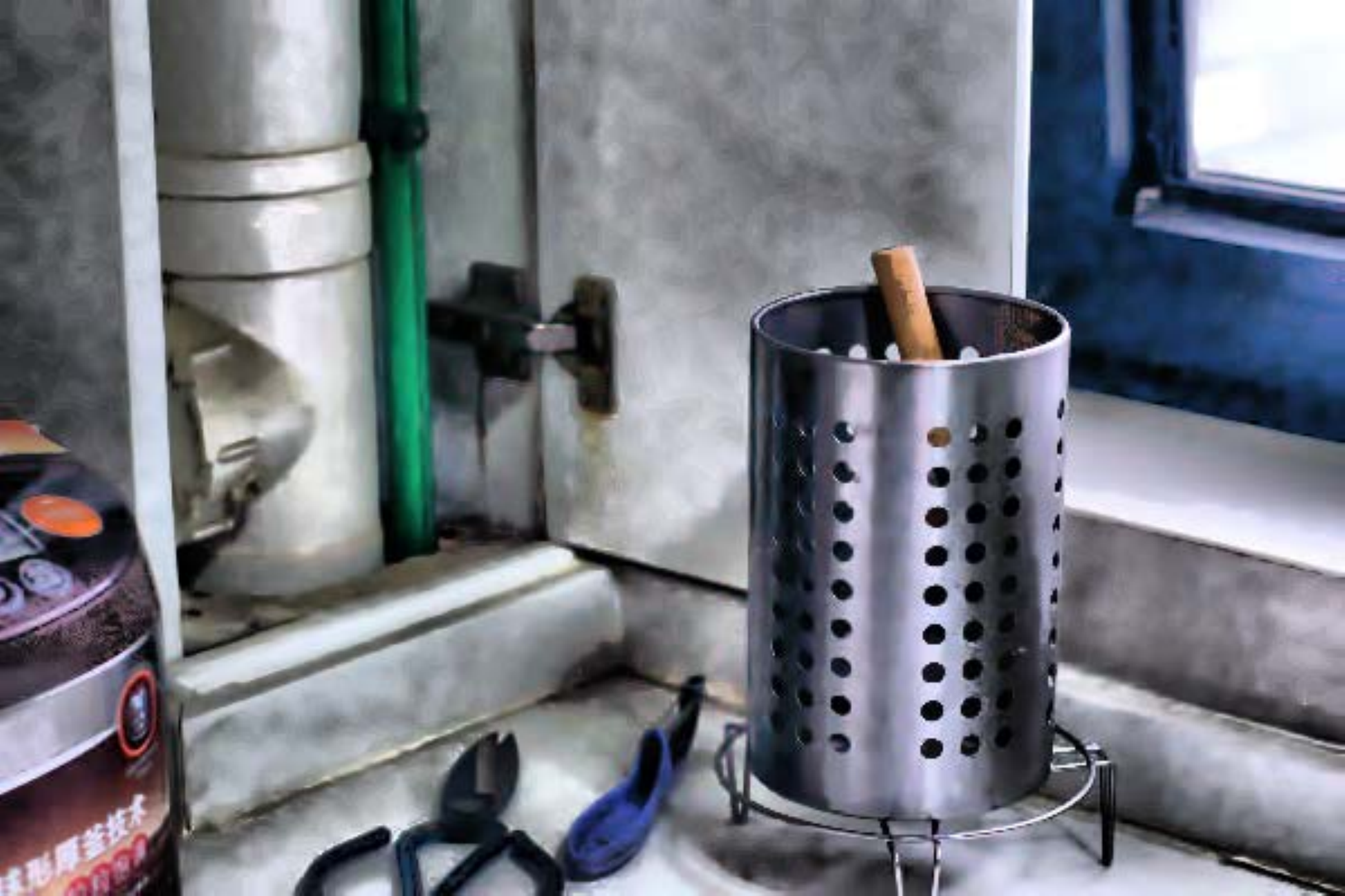}}
	\subfigure[Without fine-tuning]{
		\includegraphics[width=0.18\linewidth]{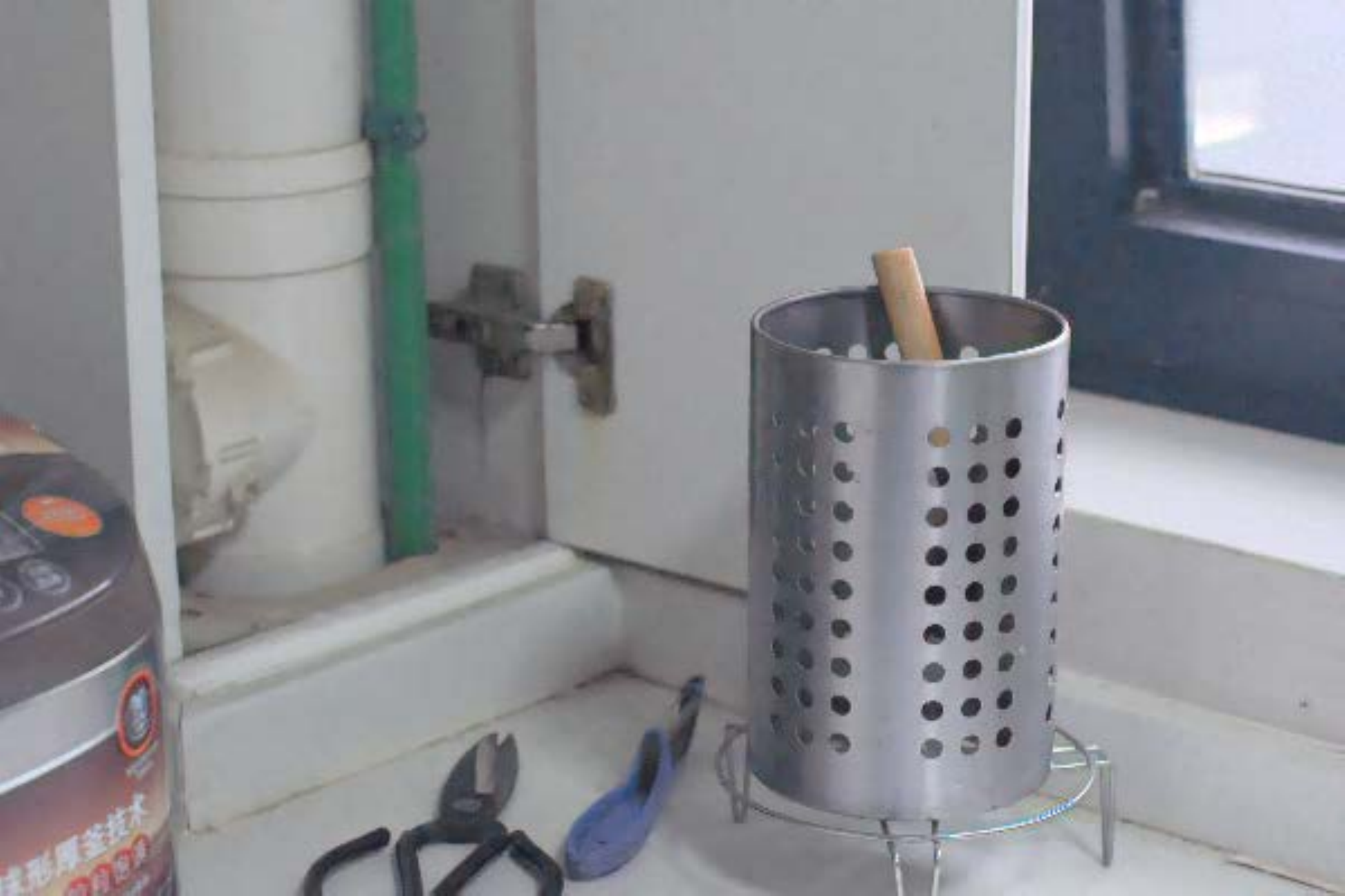}}
	\subfigure[With fine-tuning]{
		\includegraphics[width=0.18\linewidth]{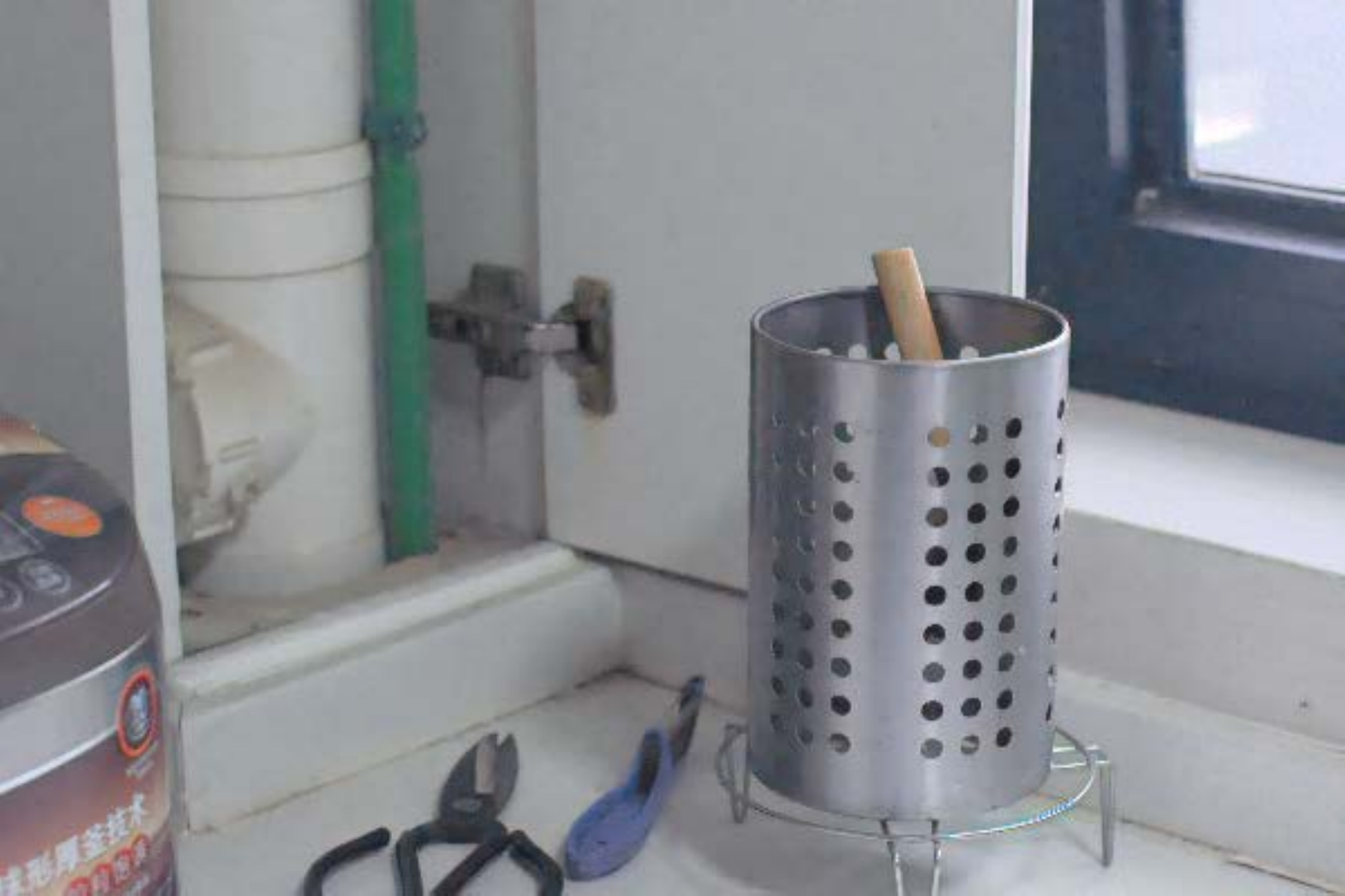}}
	\subfigure[Ground Truth]{
		\includegraphics[width=0.18\linewidth]{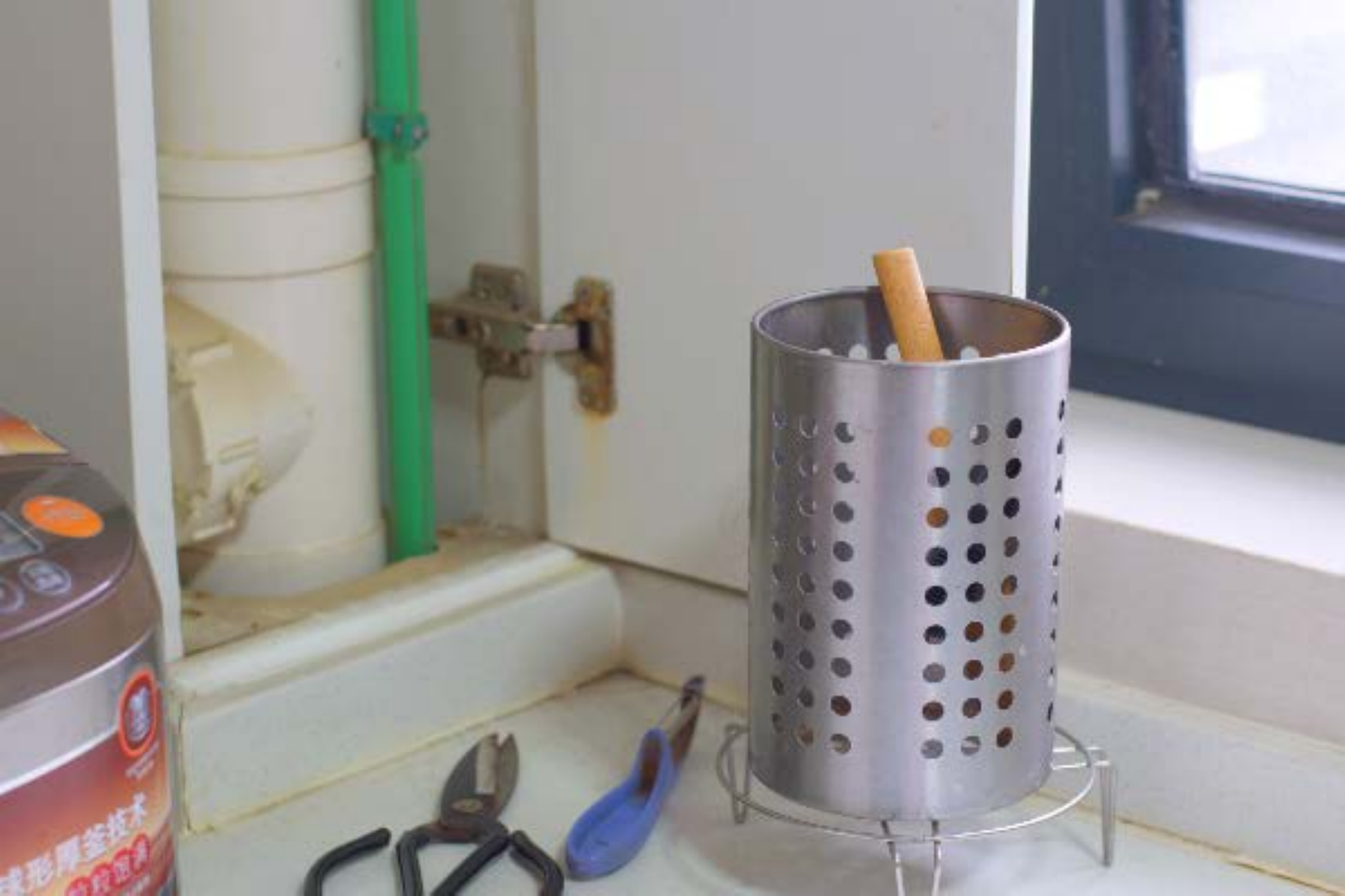}}
	\caption{Visual comparison of the enhanced images with and without the self-supervised fine-tuning strategy, together with the synthesized guiding image.}
	\label{fig:finetune} 
\end{figure*}

The core idea of our fine-tuning strategy is to first synthesize a pseudo normal-light image $\hat{\I}_{\textrm{syn}}$ from the low-light image itself, and then update the network by minimizing the following loss function:
\begin{equation}\label{loss_ft}
	\mathcal{L}_{\textrm{ft}}=\frac{1}{HW}\left\|\hat{\I}_{\textrm{ft}}-\hat{\I}_{\textrm{syn}}\right\|_F^2,
\end{equation}
where $\hat{\I}_{\textrm{ft}}$ denotes the enhanced image by the fine-tuned network. Here, the synthesized normal-light image needs not necessarily to be perfect, while it is enough to only provide some useful information as guidance for fine-tuning, such as brightness and contrast. Therefore, traditional image processing tools, that have been used in existing LIE methods, can be adopted to synthesize the normal-light image. Specifically, we first directly increase the global brightness of the low-light image, and then use CLAHE \citep{CLAHE} to enhance the local contrast, followed by BM3D \citep{dabov2007image} to suppress noise. Using the synthesized image and loss function, we can then fine-tune the adjustment networks, i.e., L-AdjNet and R-AdjNet, together with parameter $\alpha$, with about 30 iterations. The reason why we only fine-tune the adjustment networks instead of the whole architecture is two-fold. First, after training, the decomposition network is already powerful enough for the task we concerning. Second, due to the learning capacity of DNN, the network could overfit to the synthesized image, especially considering that the decomposition network has much more parameters.

Fig. \ref{fig:finetune} demonstrate the effectiveness of the proposed self-supervised fine-tuning strategy. As can be seen, though the quality of $\hat{\I}_{\mathrm{syn}}$ is not satisfactory, it provides useful information for fine-tuning, especially the color and contrast information. Therefore, the fine-tuned image looks slightly better than the direct output of the network.

It should also be mentioned that, though effective, the test time fine-tuning could be costly regarding the inference time, and thus there is a balance between performance and time that should be taken into consideration in practice.

\section{Experiments}\label{sec:exp}
In this section, we present experimental results to demonstrate the effectiveness of the proposed method. We first briefly introduce the experimental settings, including datasets, competing methods, performance metrics and some implementation details. Then we present both the quantitative and visual results of our method, in comparison with existing methods. In addition, ablation studies are also provided to more comprehensively analyze the proposed framework. In the following, our method is referred to as RAUNA, which is short for ``Retinex based Algorithm UNrolling and Adjustment'', and an additional subscript ``ft'' indicates the proposed fine-tuning strategy.

\subsection{Experimental Settings}

\subsubsection{Datasets}
We adopt four popular datasets for the LIE problem in our experiments, including LOL \citep{Chen2018Retinex}, MIT-Adobe FiveK \citep{fivek},  NPE \citep{wang2013naturalness} and DICM \citep{lee2013contrast}. Among these datasets, LOL and MIT-Adobe FiveK are composed of low/normal light image pairs, while the rest are with only low-light images. It should also be mentioned that, though both are with low/normal light image pairs, the characteristics of LOL and MIT-Adobe FiveK datasets are very different due to the collection ways. Specifically, image pairs in LOL dataset are collected by varying exposure time and ISO at the same scene, and thus the low-light images are with complex noise and contrast bias. In contrast, for MIT-Adobe FiveK dataset, only the low-light images are real-captured without noise, while the corresponding normal-light images are obtained by expert retouching. We use the default training/testing split for LOL dataset, while follow the split used in \citep{wang2019underexposed, zhang2021beyond} for MIT-Adobe FiveK. Summary of used datasets is provided in Table \ref{tabdata}.

\begin{table*}[t]
	\begin{center}
		\caption{Summary of used datasets in experiments.}
		\label{tabdata}
		\begin{tabular}{ l | l | l | l }
			\hline
			Dataset & Size (Train/Test) & Paired & Type\\
			\hline
			LOL \citep{Chen2018Retinex} & 485/15 & Yes & Real\\
			MIT-Adobe FiveK \citep{fivek} &4500/500 & Yes & Synthesized\\
			NPE \citep{wang2013naturalness} & 84 & No & Real\\
			%LIME \citep{guo2016lime} &10 & No & Real\\
			DICM \citep{lee2013contrast} &64& No & Real\\
			\hline 
		\end{tabular}
	\end{center}
\end{table*}

\subsubsection{Competing Methods}
For competing methods, we consider both traditional and deep-learning based methods for a comprehensive comparison. In specific, we adopt four representative traditional methods, including CLAHE \citep{CLAHE}, CLAHE with BM3D, LR3M \citep{LR3M} and NPE \citep{wang2013naturalness}. As for deep learning based methods, we adopt recently proposed state-of-the-art ones, including Zero-DCE++ \citep{Zero-DCE++}, KinD++ \citep{zhang2021beyond}, CSDNet \citep{ma2021learning}, DeepUPE \citep{wang2019underexposed}, MBLLEN \citep{Lv2018MBLLEN}, RetinexNet \citep{Chen2018Retinex}, RUAS \citep{liu2021retinex}, TBEFN \citep{lu2020tbefn}, SGM \citep{yang2021sparse}, DRBN \citep{yang2020fidelity} and KinD++ \citep{zhang2021beyond}. Among the deep learning based methods, CSDNet, RetinexNet, MBLLEN, TBEFN, SGM, KinD++ and RUAS are fully supervised, Zero-DCE++ is unsupervised, and DRBN is semi-supervised (using additional unpaired data).

\begin{table*}
	\begin{center}
		\caption{Quantitative comparison of all competing methods on LOL dataset. The best and second best results, with respect to each IQA metric, are highlighted in {\color{red}RED} and {\color{blue}BLUE}, respectively.}
		\label{tab:lol}
		\tiny
		\begin{tabular}{ l | c | c | c | c | c | c | c | c | c | c }
			\hline
			\multirow{2}{*}{Method} & \multicolumn{5}{c|}{Without GC} & \multicolumn{5}{c}{With GC} \\
			\cline{2-11}
			& SSIM $\uparrow$ & PSNR $\uparrow$ & NIQE $\downarrow$ & LOE$_{\textrm{ref}}$ $\downarrow$ & LOE $\downarrow$ & SSIM $\uparrow$ & PSNR $\uparrow$ & NIQE $\downarrow$ & LOE$_{\textrm{ref}}$ $\downarrow$ & LOE $\downarrow$\\
			\hline
			CLAHE \citep{CLAHE} & 0.494 & 14.36 & 9.032 & 808.6 & 787.1 & 0.772 & 17.77 & 7.789 &358.1 &285.9 \\
			CLAHE+BM3D&0.702 &15.65&5.731&767.7 &749.5 &0.836 &18.21 &6.024 &350.4& 284.7 \\
			%BIMEF\citep{ying2017bio} &0.706&13.977&7.5162&\textit{223.7}&133.2 &0.780&17.149&7.4738&226.5&145.9\\
			LR3M \citep{LR3M} & 0.440 & 10.31 & 7.540 & 290.9 & 332.1 & 0.707 & 15.77 & 6.628 & 298.6 & 348.4 \\
			NPE \citep{wang2013naturalness} &0.697&17.14&8.439&463.6&422.7&0.709&17.78&8.371&468.9&430.8\\
			\hline
			Zero-DCE++ \citep{Zero-DCE++}&0.718&15.42&7.896&257.3&195.1&0.768&18.08&7.793&260.4&210.4\\
			\hline
			CSDNet \citep{ma2021learning} & 0.870 & 20.75 & 4.023 & 373.9&341.0&0.873&22.55&4.018&374.0&339.5\\
			DeepUPE \citep{wang2019underexposed} & 0.549 & 12.80 & 7.524 & 369.6&297.3&0.772&18.43&7.863&359.0&305.4\\
			MBLLEN \citep{Lv2018MBLLEN}&0.788&17.91&3.584&362.2&383.7&0.789&18.29&{\color{blue}\textbf{3.544}}&369.3&394.2\\
			RetinexNet \citep{Chen2018Retinex}&0.648&16.98&8.878&462.0&486.2&0.646&16.96&8.799&466.0&490.1\\
			RUAS \citep{liu2021retinex}&0.666&16.50&6.340 & 220.0 & {\color{red}\textbf{125.6}}&0.742&17.03&6.258&215.5&133.4\\
			TBEFN \citep{lu2020tbefn}&0.831&17.43 & 3.437 & 325.7&342.8&0.836&17.74&{\color{red}\textbf{3.418}}&327.4&345.6\\
			SGM \citep{yang2021sparse}&0.782&16.90&4.613&428.8&411.2&0.771&16.46&4.627&429.1&410.2\\
			DRBN \citep{yang2020fidelity} & 0.853 & 18.67 & 5.110 &488.1 &501.2&0.853&19.13&5.449&487.9&501.9\\
			KinD++ \citep{zhang2021beyond}&0.863&19.82&5.111&254.2&207.9&-&-&-&-&-\\
			\hline
			RAUNA & {\color{blue}\textbf{0.874}} & {\color{blue}\textbf{22.82}} &4.213 & {\color{red}\textbf{189.2}} & {\color{blue}\textbf{126.5}}&-&-&-&-&-\\
			RAUNA$_{\textrm{ft}}$ & {\color{red}\textbf{0.887}}& {\color{red}\textbf{24.58}} & 4.120 & {\color{blue}\textbf{196.5}} & {\color{blue}\textbf{126.5}}&-&-&-&-&-\\
			\hline 
		\end{tabular}
	\end{center}
\end{table*}

\subsubsection{Evaluation Metrics} 
\begin{figure*}[t] 
	\centering
	\subfigure[Input]{
		\includegraphics[width=0.15\linewidth]{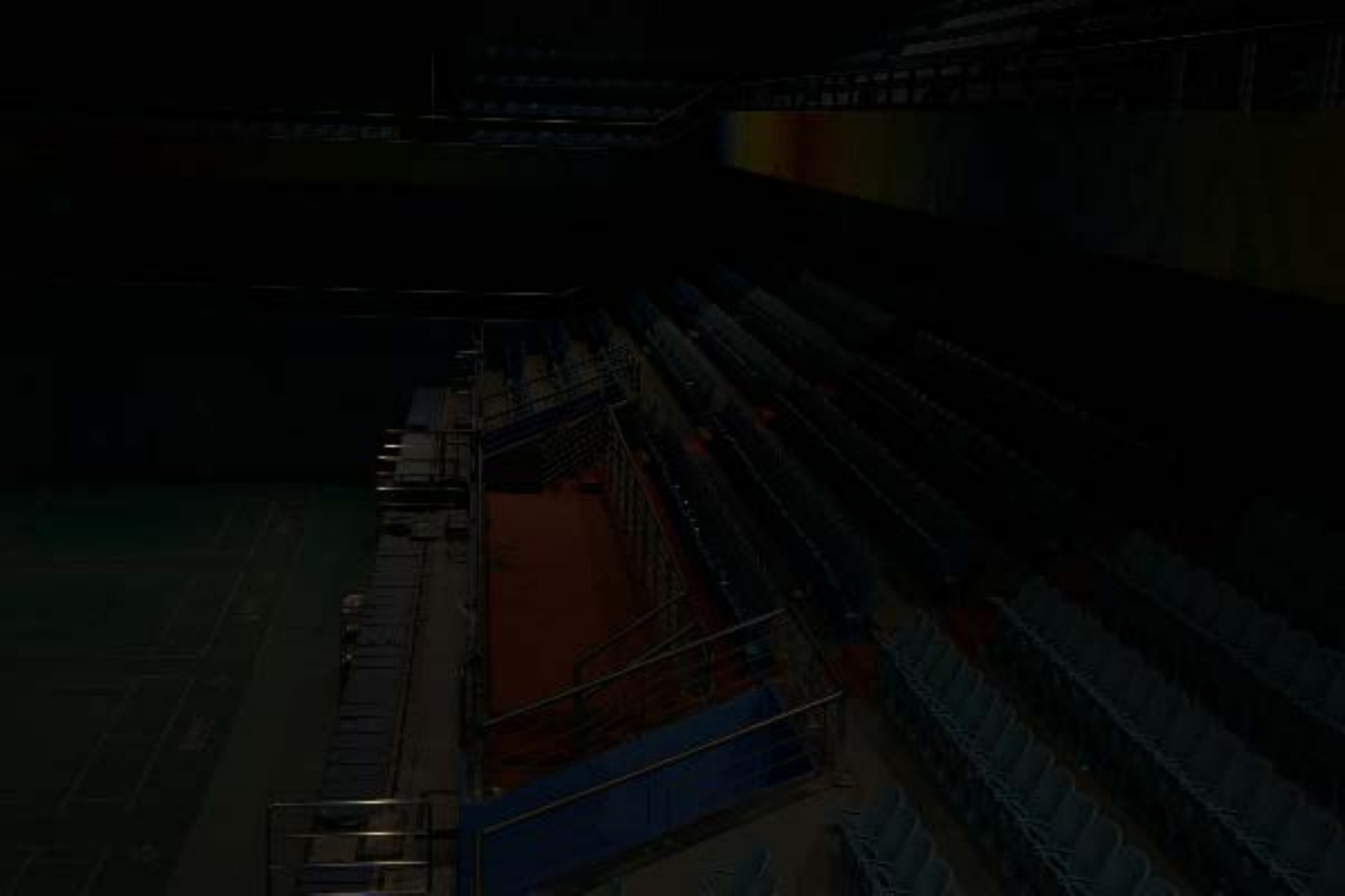}}
	\subfigure[CLAHE]{
		\includegraphics[width=0.15\linewidth]{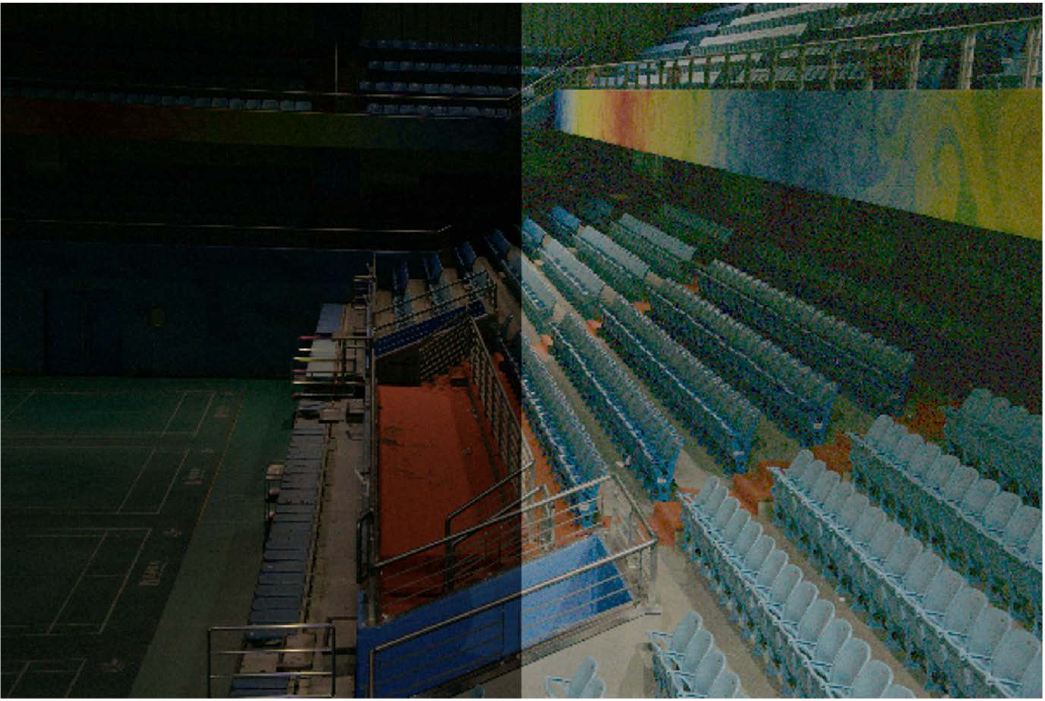}}
    \subfigure[CLAHE+BM3D]{
		\includegraphics[width=0.15\linewidth]{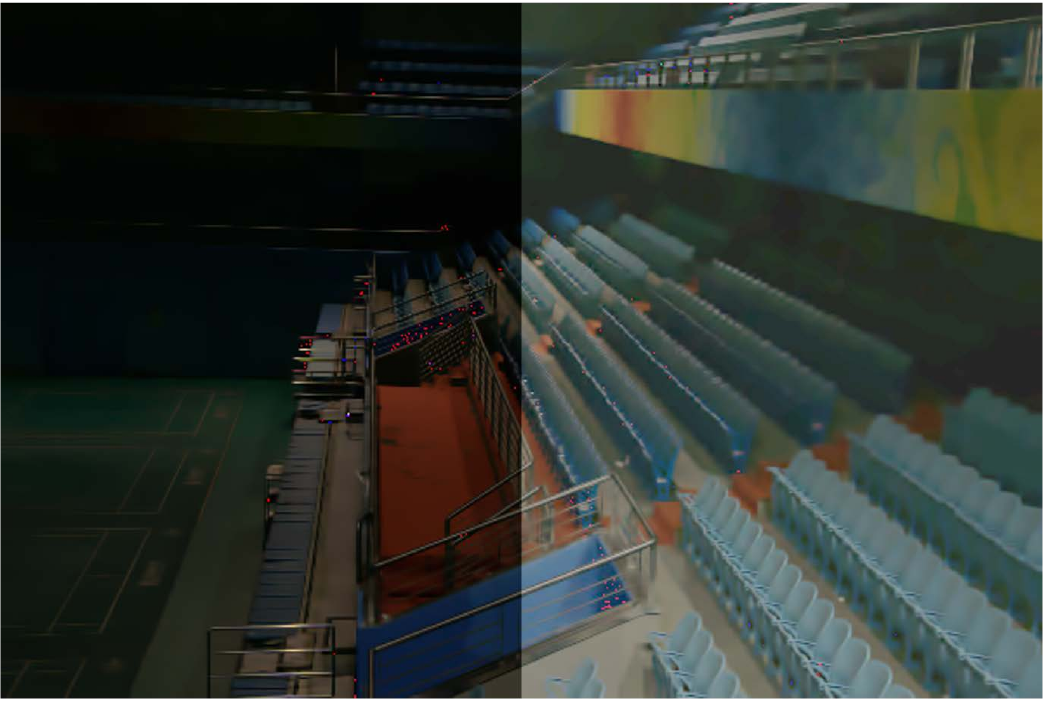}}
	\subfigure[LR3M]{
		\includegraphics[width=0.15\linewidth]{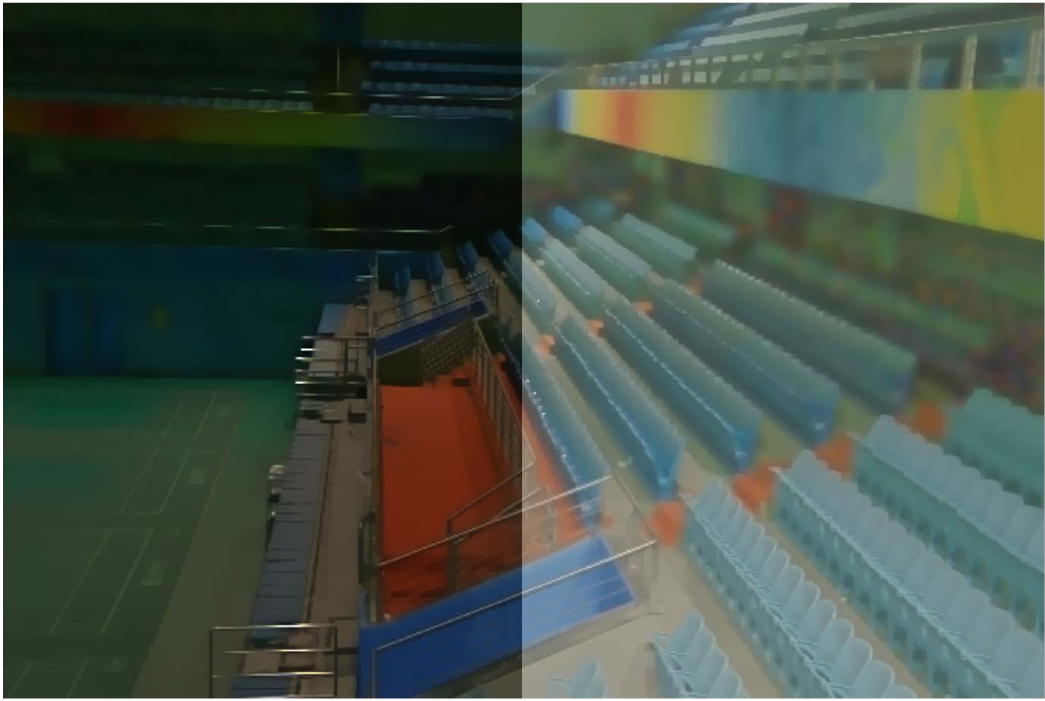}}
	\subfigure[NPE]{
		\includegraphics[width=0.15\linewidth]{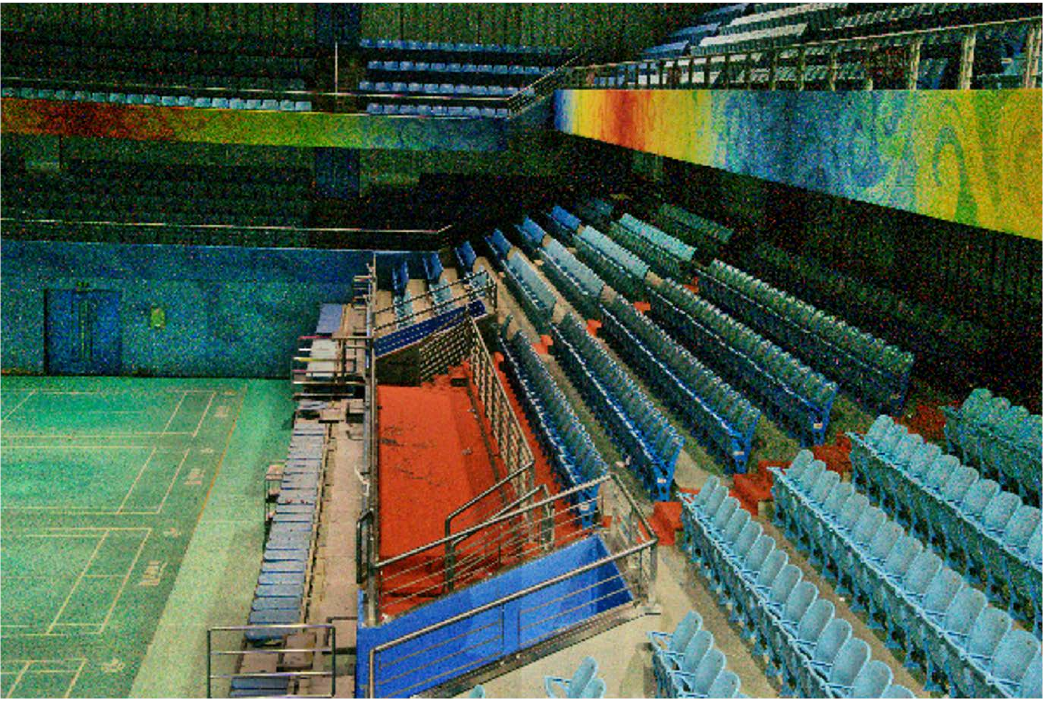}}
	\subfigure[Zero-DCE++]{
		\includegraphics[width=0.15\linewidth]{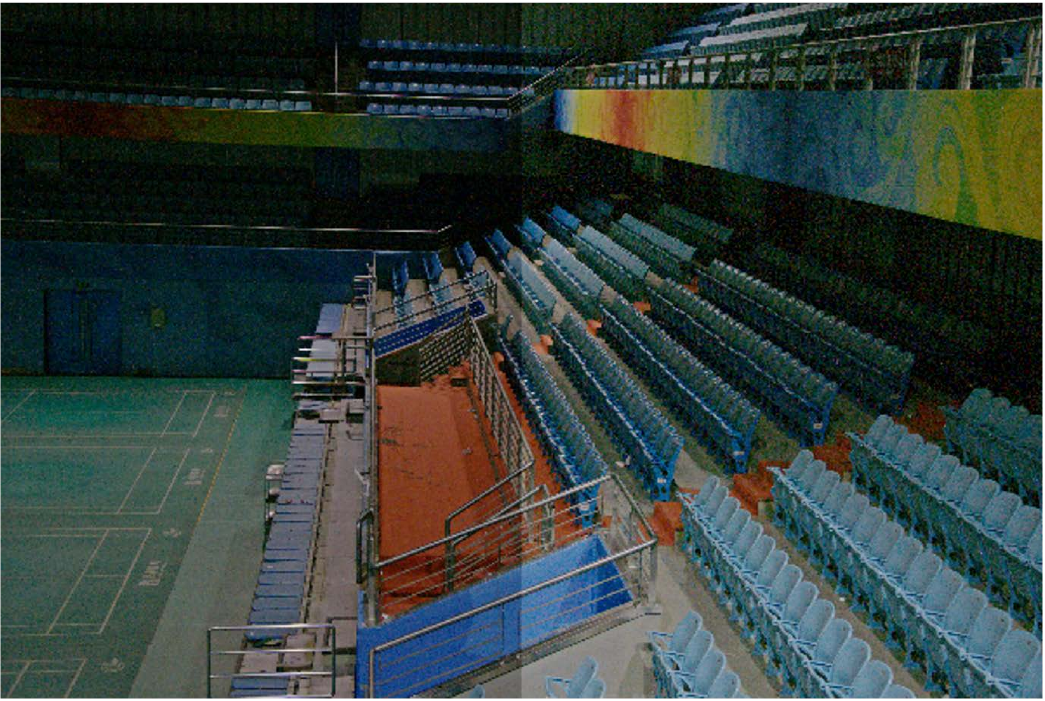}}\\
	\subfigure[CSDNet]{
		\includegraphics[width=0.15\linewidth]{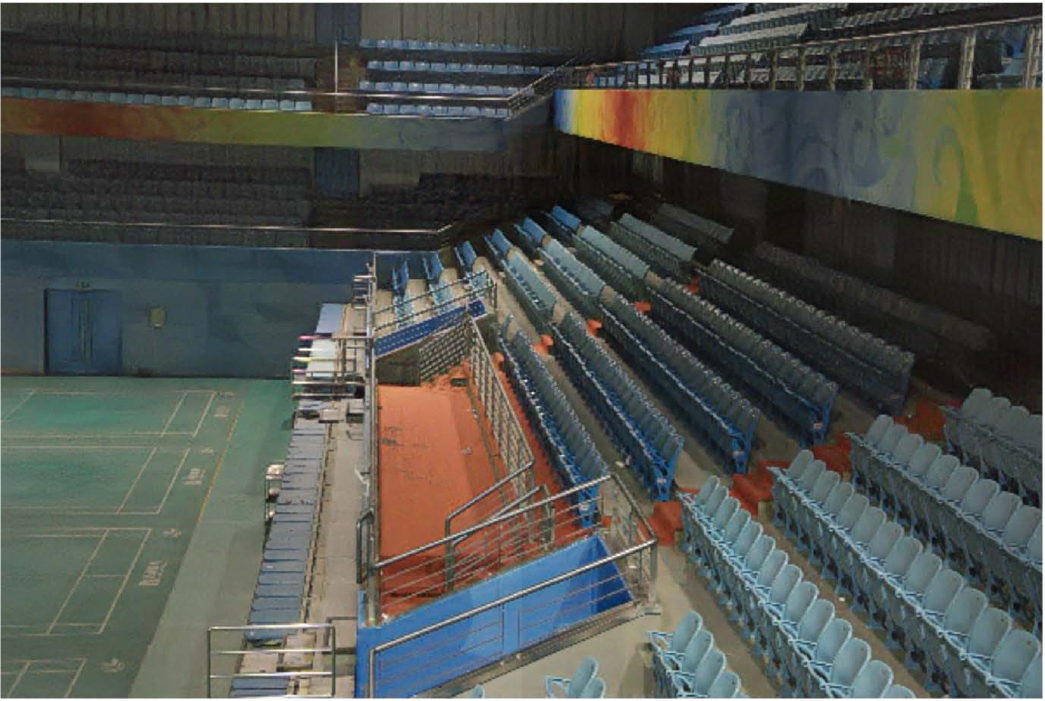}}
	\subfigure[DeepUPE]{
		\includegraphics[width=0.15\linewidth]{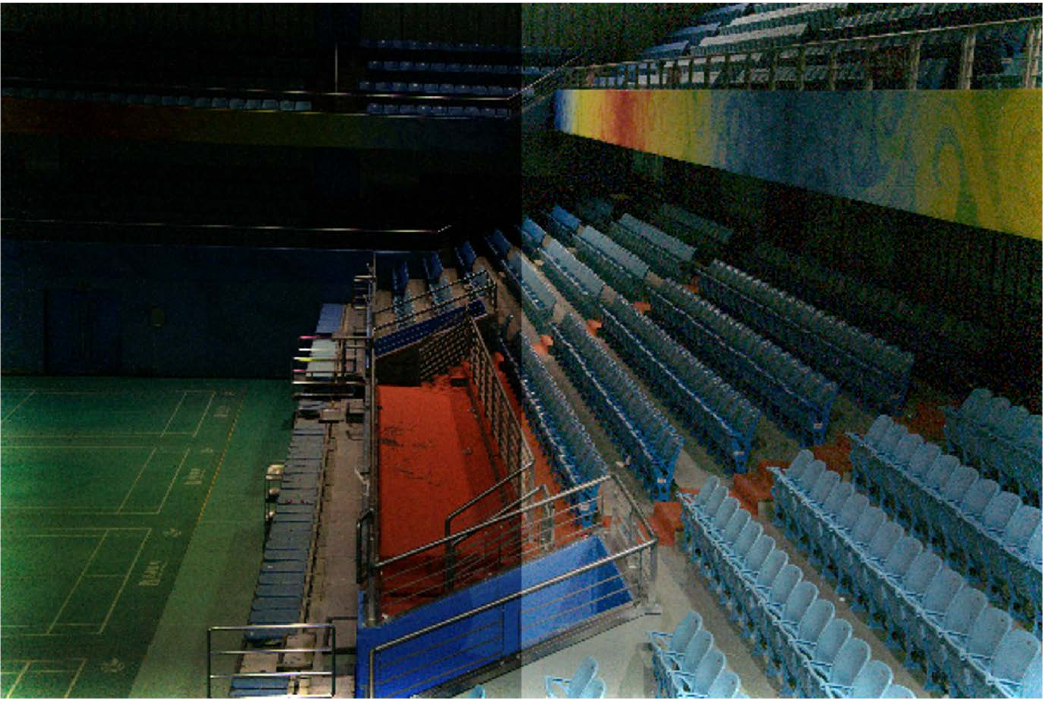}} 
	\subfigure[MBLLEN]{
		\includegraphics[width=0.15\linewidth]{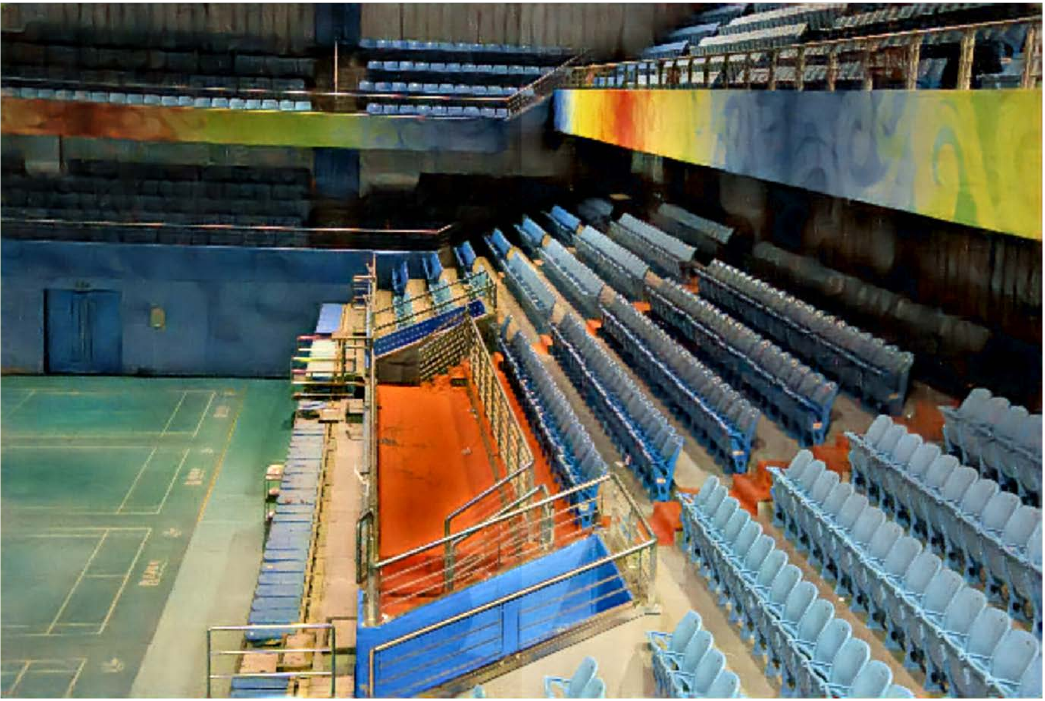}}
	\subfigure[RetinexNet]{
		\includegraphics[width=0.15\linewidth]{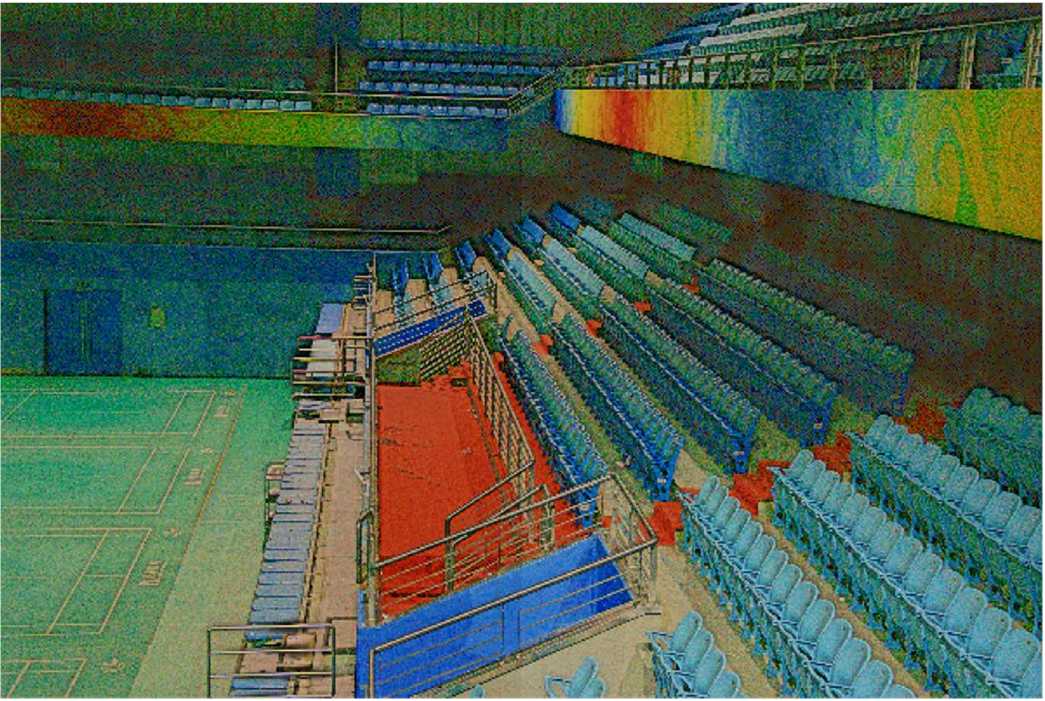}} 
	\subfigure[RUAS]{
		\includegraphics[width=0.15\linewidth]{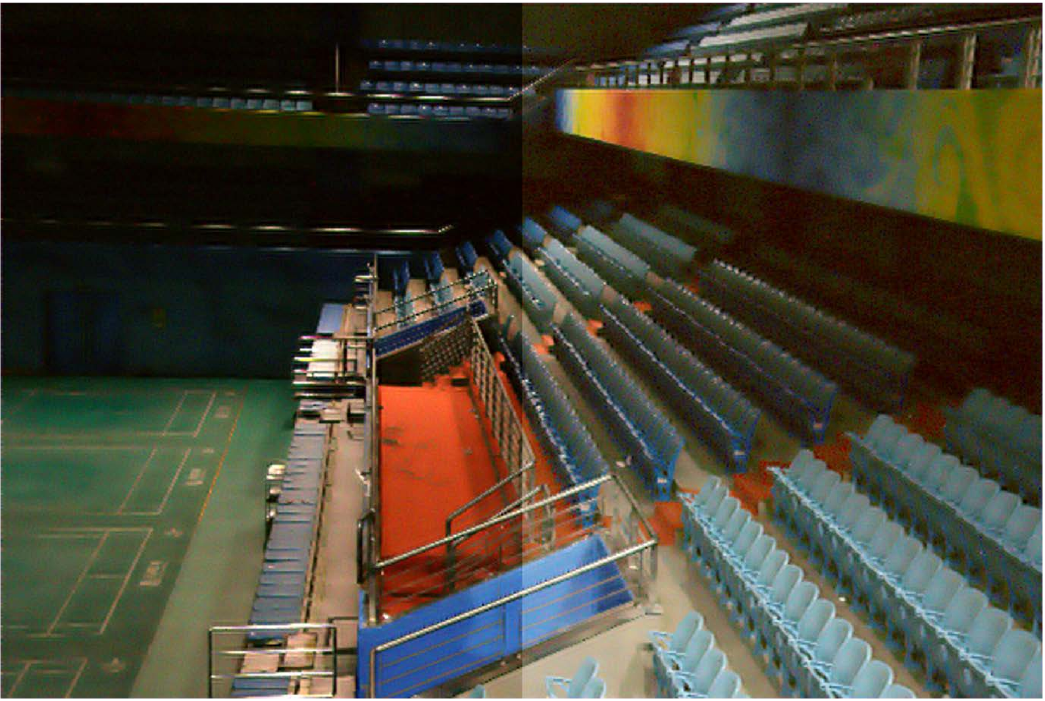}}
	\subfigure[TBEFN]{
		\includegraphics[width=0.15\linewidth]{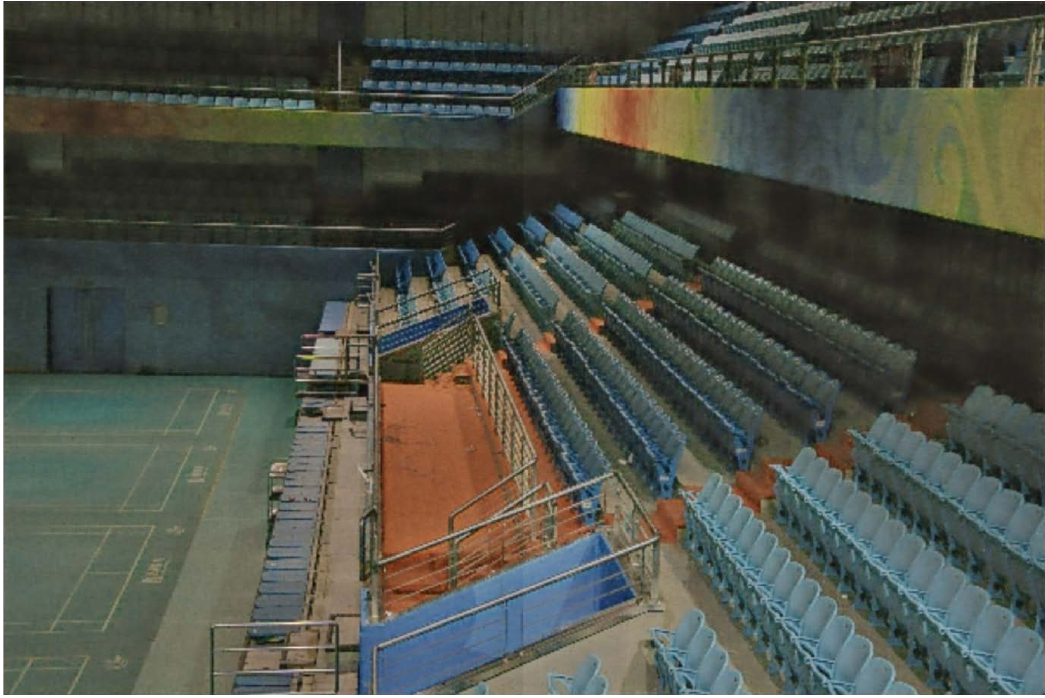}}\\
	\subfigure[SGM]{
		\includegraphics[width=0.15\linewidth]{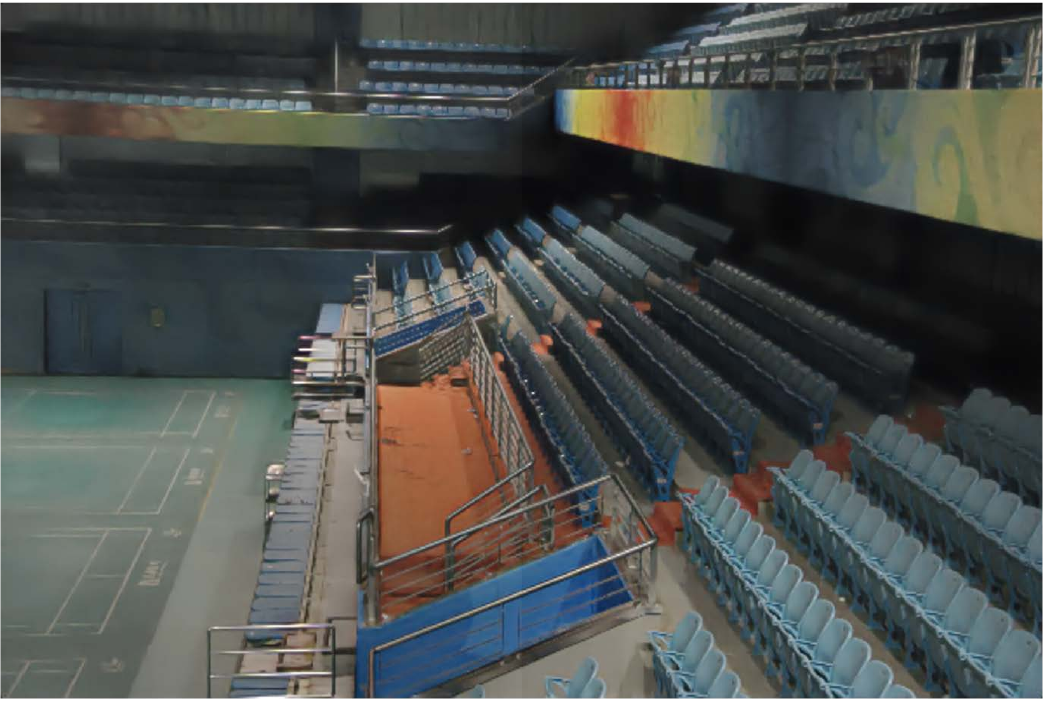}}
	\subfigure[DRBN]{
		\includegraphics[width=0.15\linewidth]{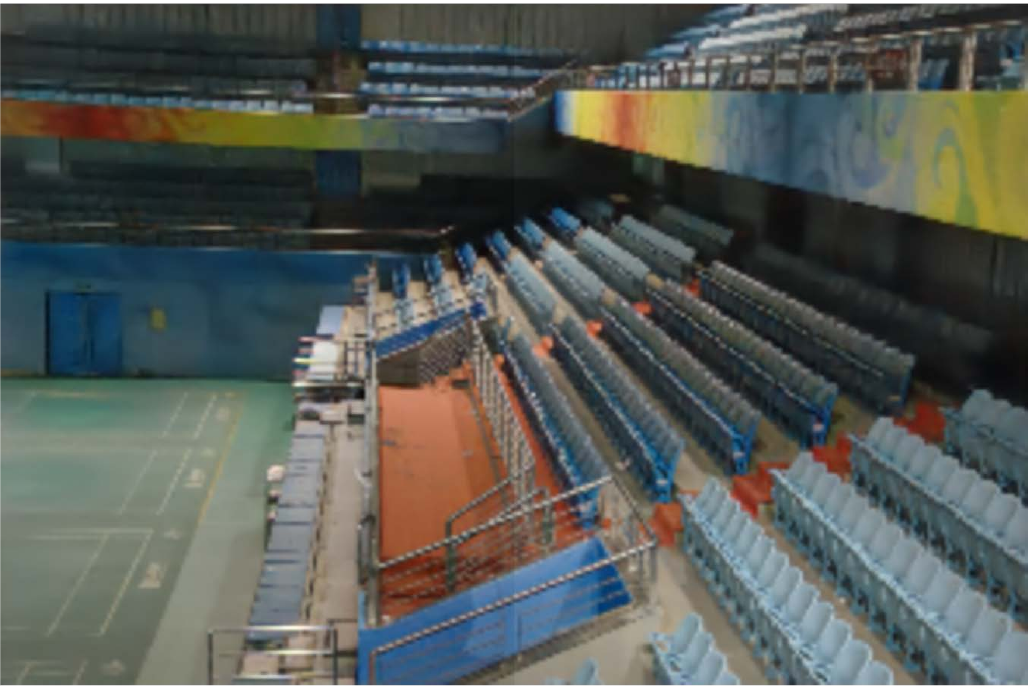}}
	\subfigure[KinD++]{
		\includegraphics[width=0.15\linewidth]{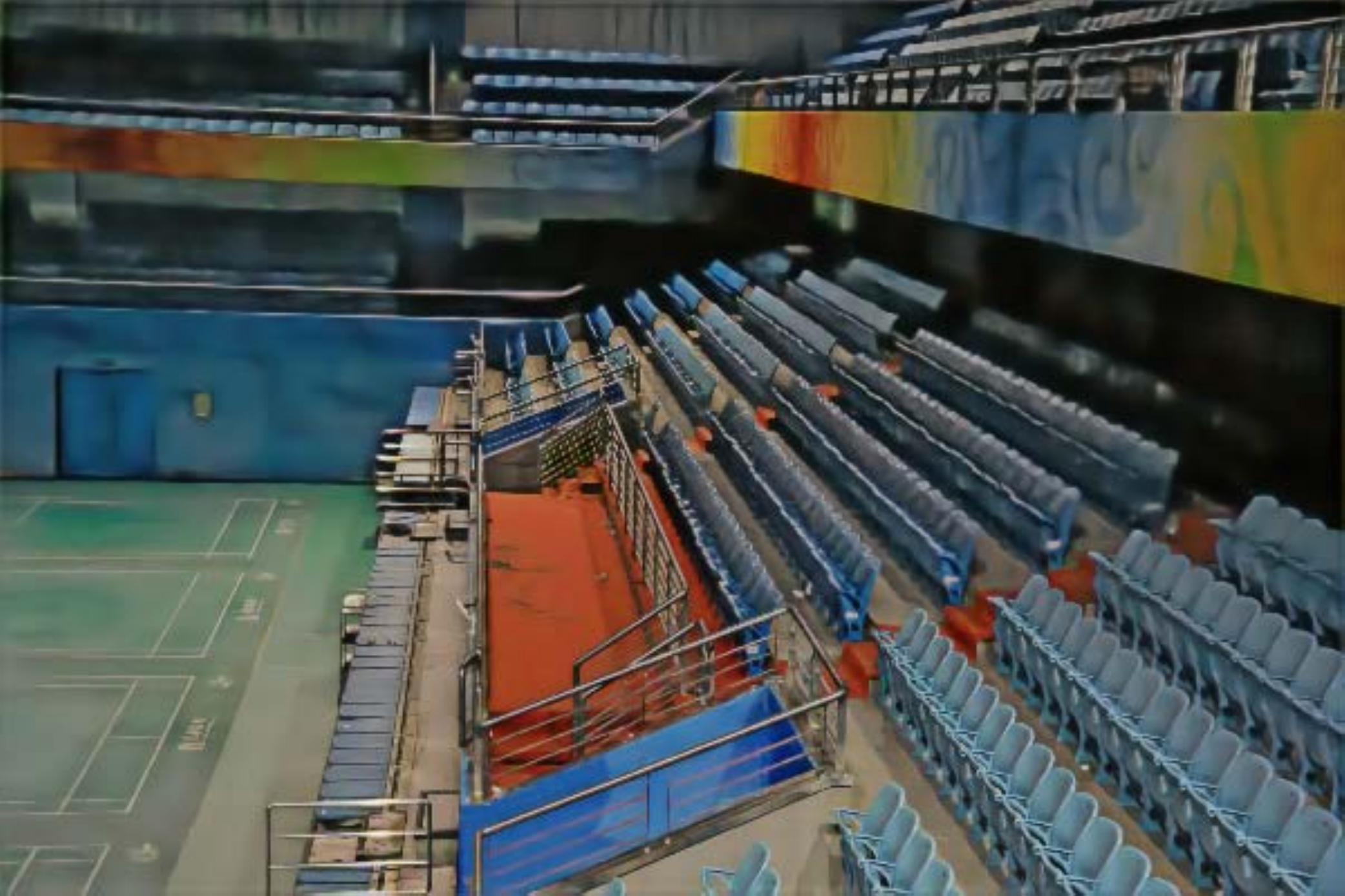}}
	\subfigure[RAUNA]{
		\includegraphics[width=0.15\linewidth]{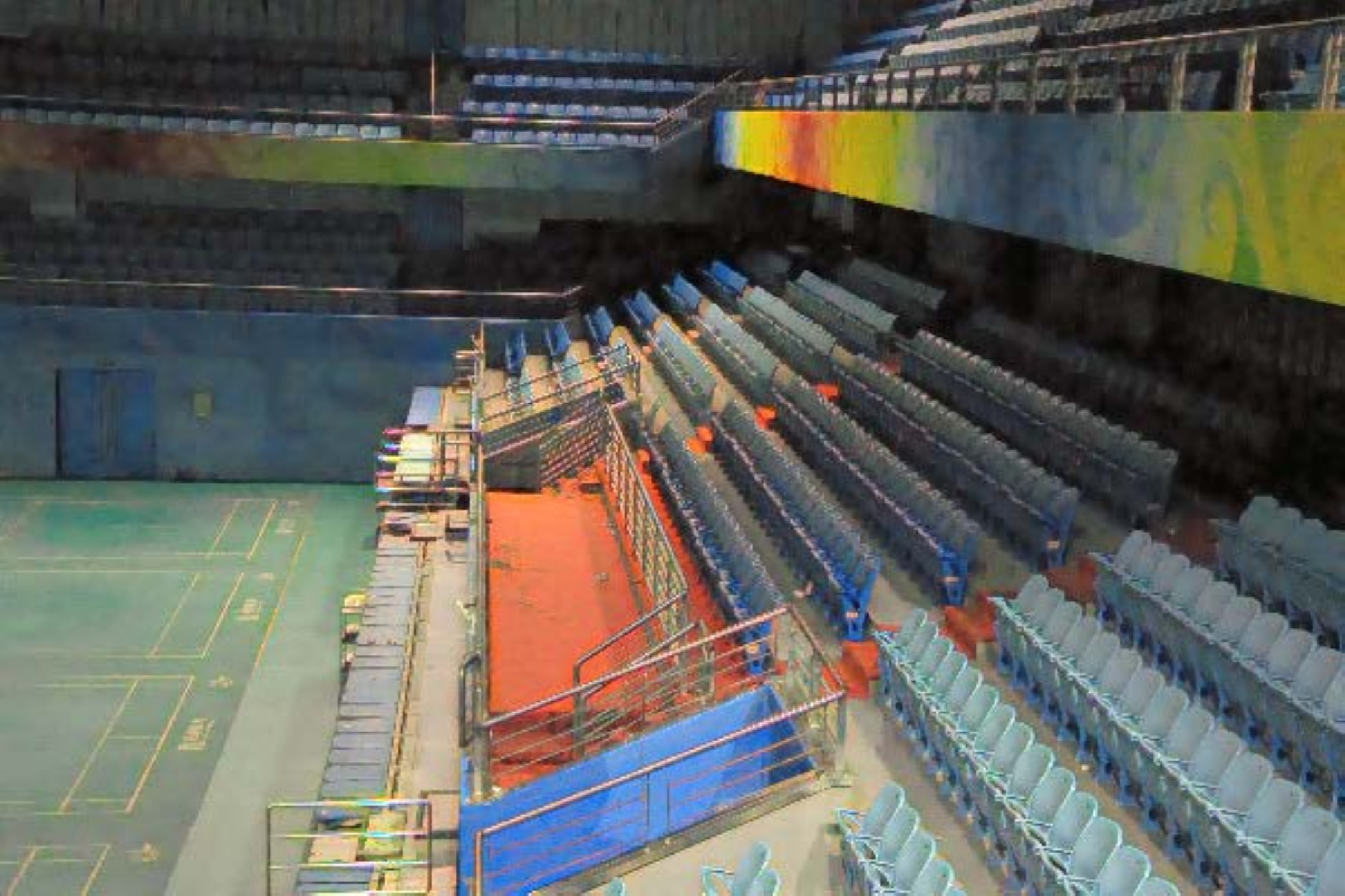}}
	\subfigure[RAUNA$_{\textrm{ft}}$]{
		\includegraphics[width=0.15\linewidth]{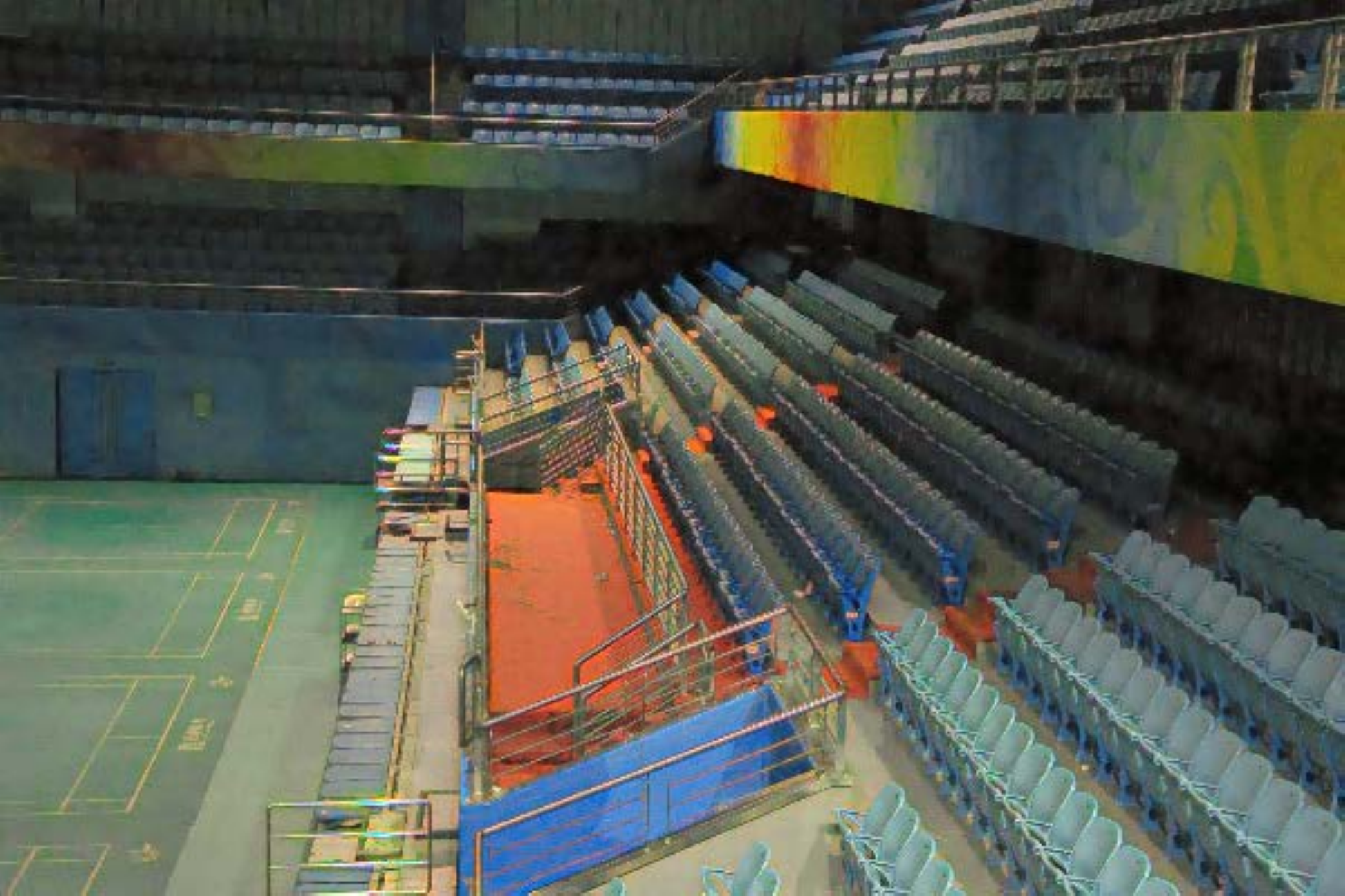}}
	\subfigure[Ground Truth]{
		\includegraphics[width=0.15\linewidth]{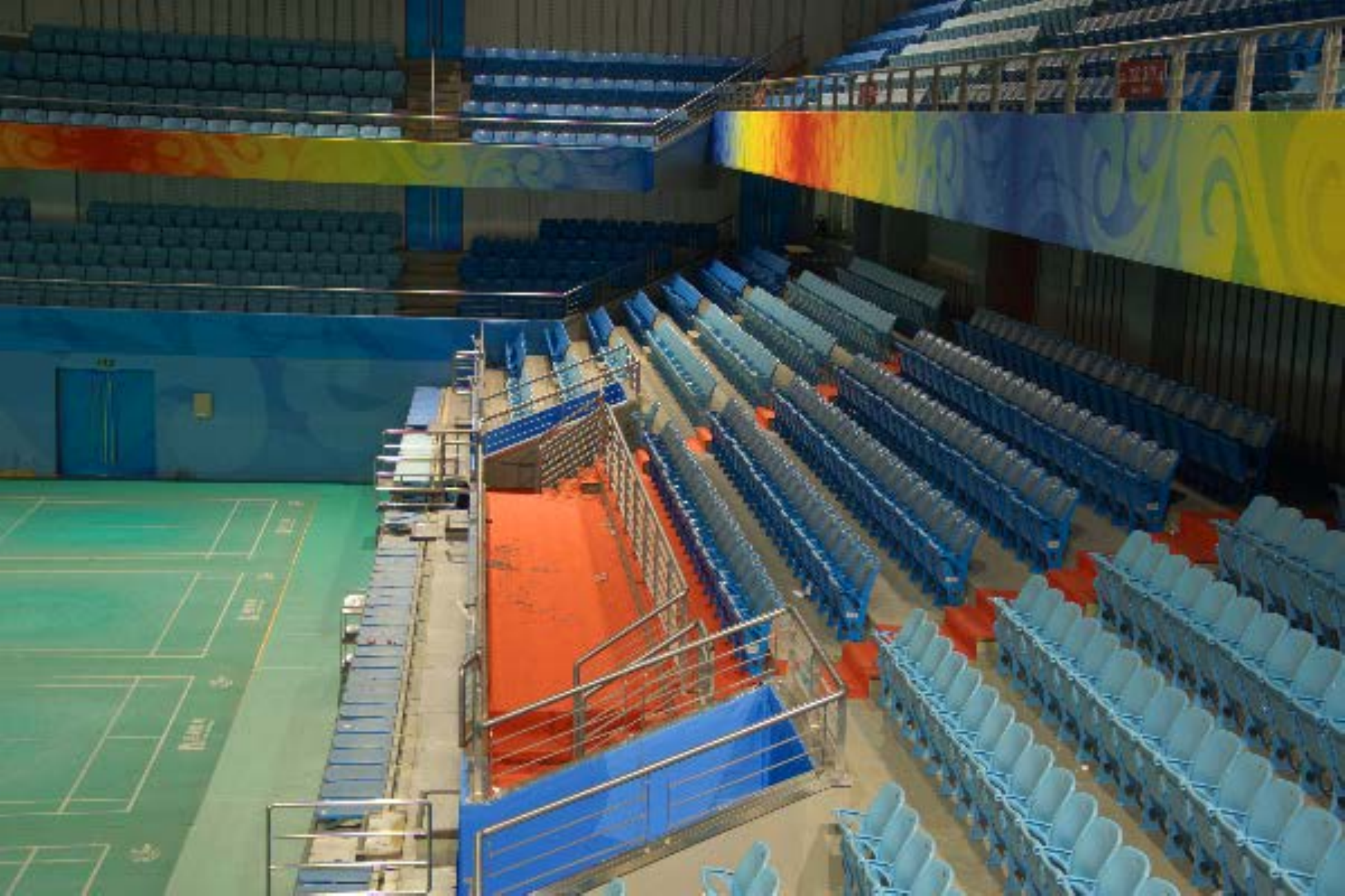}}
	\caption{Visual results of competing methods on  LOL dataset. For (b)-(n), the left and parts are from enhanced images without and with Gamma Correction, respectively.}
	\label{fig:lol1} 
\end{figure*}
We adopt in total five metrics, including PSNR, SSIM \citep{wang2004image}, NIQE \citep{niqe}, LOE \citep{wang2013naturalness} and LOE$_{\textrm{ref}}$ \citep{zhang2021beyond}, to comprehensively assess the performance of all competing methods. Among them, PSNR and SSIM are two commonly used metrics for image quality assessment (IQA) with reference image, while NIQE is a general IQA measurement without reference. LOE is a metric specifically designed for evaluating the performance of LIE methods. However, it was original defined in a blind way, that the reference image to compute this metric is chosen as the input low-light image \citep{wang2013naturalness}, which is more or less problematic as pointed in \citep{guo2016lime}. Therefore, LOE can be reasonable modified to LOE$_{\textrm{ref}}$ by replacing the low-light image by its corresponding normal-light one as the reference \citep{zhang2021beyond}, when groundtruth is available. Another thing should be mentioned that all the metrics in our experiments are evaluated with RGB channels.

\subsubsection{Implementation Details}\label{sec:exp_implement}
All experiments are conducted on a computer with Intel i7 3.40GHz (CPU), NVIDIA RTX3080 (GPU) and Ubuntu 18.04 LTS (OS). Our method is implemented using PyTorch 1.7.0, while other methods are implemented according to their codes. We initialize our model parameters following \citep{he2016deep}, and ADAM \citep{adam} is used for optimization. The initial learning rates are set to $1\times 10^{-5}$ and $1\times 10^{-3}$ for decomposition and adjustment networks, respectively, and then is divided by 10 at the 2nd and 3rd epochs for decomposition network, while at the 60th epoch for adjustment network. The model is trained with 70 epochs in total.
%The learning rate of decomposition networks is divided by 10 at $2_{th}$ and $3_{th}$ epochs, and then fixed. The learning rate of decomposition networks is divided by 10 at the $60_{th}$ epoch and then fixed. They are trained for 70 epochs totally. For fine-tuning, the learning rate for the adjustable parameter $alpha$ is $5\times 10^{-2}$, and that of adjustment networks is set to $1\times 10^{-3}$. We train the fine-tuning for 30 epochs.
Since LOL and MIT-Abobe FiveK datasets contain both training and testing sets, we can train and test our model on corresponding ones. For NPE and DICM datasets, since no groundtruth is provided, we directly test using our model trained on LOL. During training, we use cropped image patches in each training batch. Specifically, for LOL dataset, 8 pairs of patches with size 64$\times$64 are used in each batch, and for MIT-Adobe FiveK dataset, 16 pairs of patches with size 48$\times$48 are used. Since LOL dataset is more challenging than MIT-Adobe FiveK dataset, we use a 17-stage decomposition network for LOL, while 13-stage for MIT-Adobe FiveK.
%{\color{red}For LOL dataset, we set batch and patch size as 8 and 64, respectively, and for MIT dataset, we set batch and patch size as 16 and 48, respectively. 
%Different numbers of the stage in decomposition networks are adopted for LOL and MIT datasets. We use 17 stages for decomposition of LOL dataset and 13 stages for decomposition of MIT dataset.

\begin{table*}

	\begin{center}
		\caption{Quantitative comparison of all competing methods on MIT-Adobe FiveK dataset. The best and second best results are highlighted in {\color{red}RED} and {\color{blue}BLUE}, respectively.}
		\label{tab:5K}
		\tiny
		\begin{tabular}{ l | c | c | c | c | c | c | c | c | c | c }
			\hline
			\multirow{2}{*}{Method} & \multicolumn{5}{c|}{Without GC} & \multicolumn{5}{c}{With GC} \\
			\cline{2-11}
			& SSIM $\uparrow$ & PSNR $\uparrow$ & NIQE $\downarrow$ & LOE$_{\textrm{ref}}$ $\downarrow$ & LOE $\downarrow$ & SSIM $\uparrow$ & PSNR $\uparrow$ & NIQE $\downarrow$ & LOE$_{\textrm{ref}}$ $\downarrow$ & LOE $\downarrow$\\
			\hline
			CLAHE \citep{CLAHE} & 0.543 & 14.06 & 4.391 & 965.1 & 966.8 & 0.847&19.00&3.731&360.8&347.2\\
			CLAHE+BM3D&0.620&14.06&5.627&958.8&960.5 & 0.798&18.84&5.186&360.2&346.5\\
			%BIMEF\citep{ying2017bio} &0.798&18.069&3.8318&145.8&130.3&0.806&18.021&3.8373&146.5&129.0\\
			LR3M \citep{LR3M} & 0.708 & 15.99 & 5.275 & 273.0  & 283.8  & 0.737 & 18.19 & 5.258 & 272.0  & 283.6 \\
			NPE\citep{wang2013naturalness} &0.795&17.50&3.735&454.1&473.4&0.804&18.10&3.743&452.1&470.4\\
			\hline
			Zero-DCE++\citep{Zero-DCE++}&0.766&15.65&\emph{3.694}&311.0&370.6&0.810&17.49&3.708&309.2&326.9\\
			\hline
			CSDNet\citep{ma2021learning} &  0.866 & 19.97 & 3.794 & 298.0&316.4 & {\color{blue}\textbf{0.904}} & 22.90&3.815&297.6&311.6 \\
			DeepUPE\citep{wang2019underexposed} & 0.872 & 19.71 & 3.817 & 801.5&372.8&0.879 & 20.60 & 3.833 & 370.3&372.4\\
			MBLLEN\citep{Lv2018MBLLEN}&0.815&19.61&3.851&{\color{red}\textbf{74.5}} &{\color{red}\textbf{67.4}} &0.837 &20.81&3.740&77.5&84.1 \\
			RetinexNet\citep{Chen2018Retinex}&0.687&12.59&4.475&1333.0&1369.3&0.699&13.87&4.586&1330.7&1366.3\\
			RUAS\citep{liu2021retinex}&0.836&18.33&3.927&164.9&154.9&0.854&19.58&3.969&168.5&154.2\\
			TBEFN\citep{lu2020tbefn}&0.752&14.77&3.799&492.7&514.8&0.792&16.81&3.762&490.2&510.8\\
			SGM\citep{yang2021sparse}&0.773&16.29& 3.724 & 614.3&679.2 &0.785&16.87&3.725&614.0&632.3 \\
			DRBN \citep{yang2020fidelity} &0.674 & 16.21 & 5.438 & 708.7 & 711.6  & 0.775 & 18.27 & 5.444 & 708.4 & 711.8  \\
			KinD++\citep{zhang2021beyond}&0.850&22.64&4.151&143.3&142.5&-&-&-&-&-\\
			\hline
			RAUNA &0.901 & {\color{blue}\textbf{23.52}} & {\color{blue}\textbf{3.691}} & 97.8& 98.1&-&-&-&-&-\\
			RAUNA$_{\textrm{ft}}$& {\color{red}\textbf{0.910}} & {\color{red}\textbf{25.43}} & {\color{red}\textbf{3.677}} & {\color{blue}\textbf{92.0}} & {\color{blue}\textbf{95.8}}& -&-&-&-&-\\
			\hline 
		\end{tabular}
	\end{center}
\end{table*}

\begin{figure*}[t]
	\centering
	\subfigure[Input]{
		\includegraphics[width=0.15\linewidth]{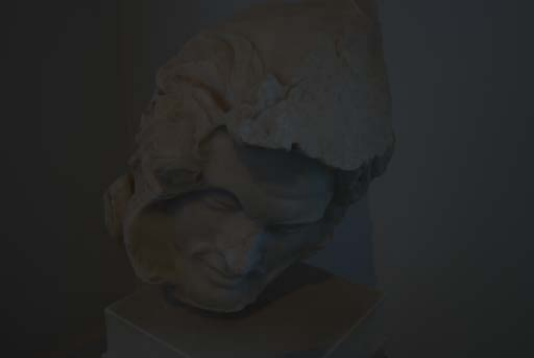}}
	\subfigure[CLAHE]{
		\includegraphics[width=0.15\linewidth]{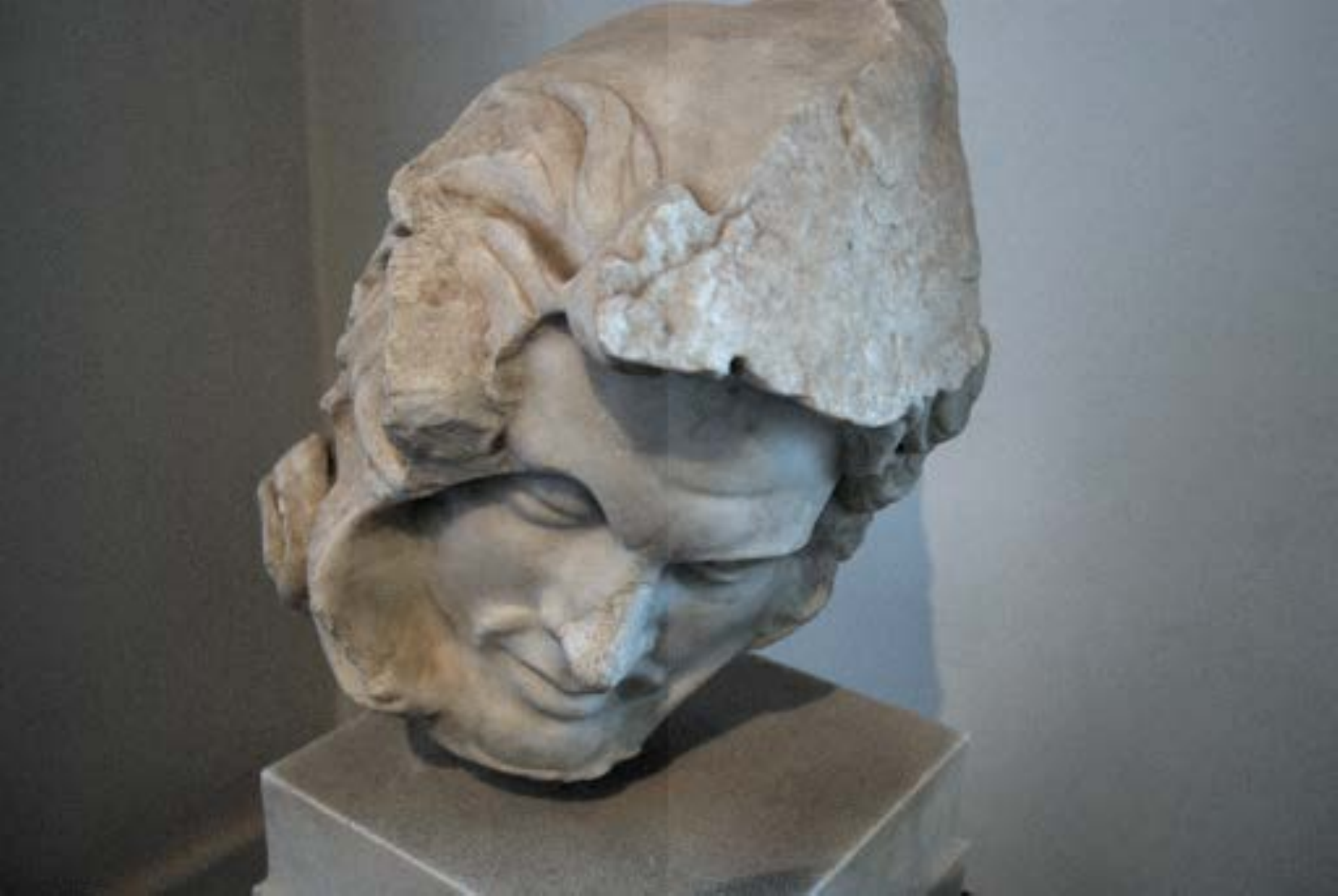}}
	\subfigure[CLAHE+BM3D]{
		\includegraphics[width=0.15\linewidth]{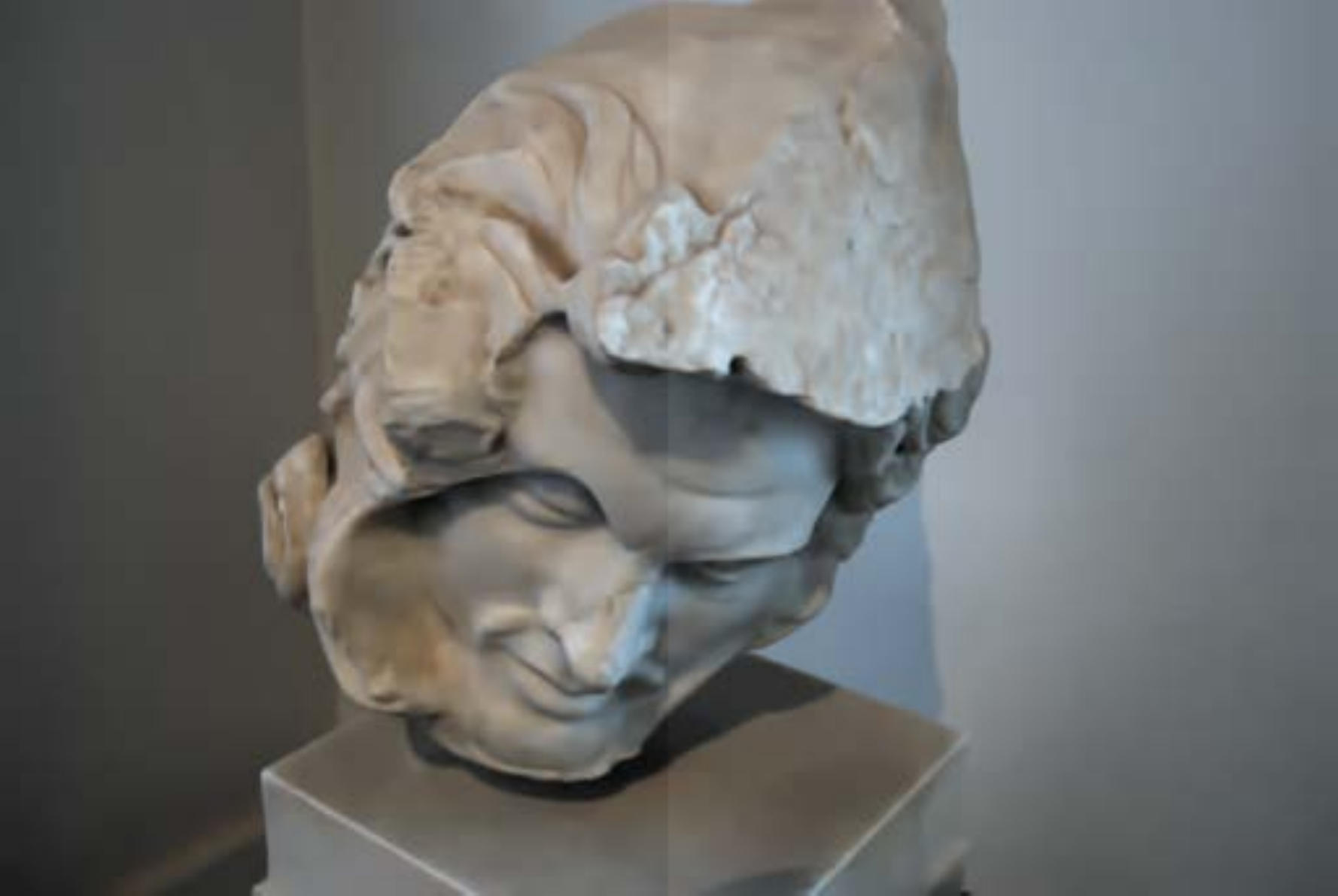}}
	\subfigure[LR3M]{
		\includegraphics[width=0.15\linewidth]{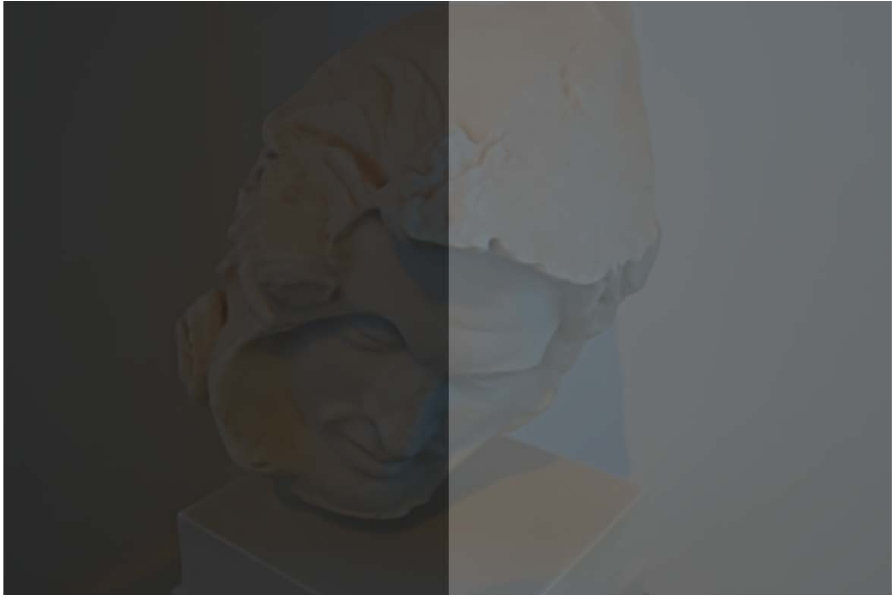}}
	\subfigure[NPE]{
		\includegraphics[width=0.15\linewidth]{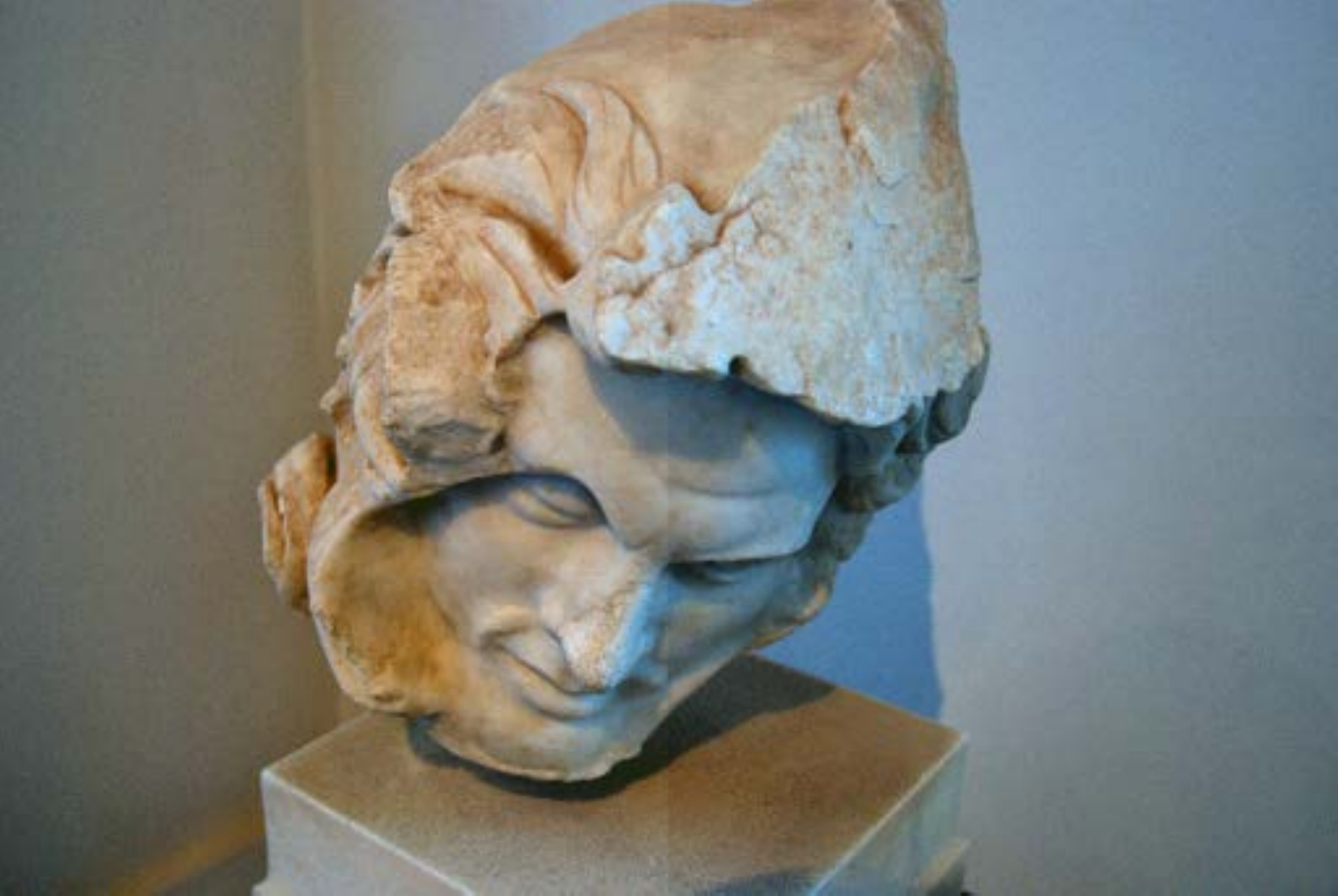}}
	\subfigure[Zero-DCE++]{
		\includegraphics[width=0.15\linewidth]{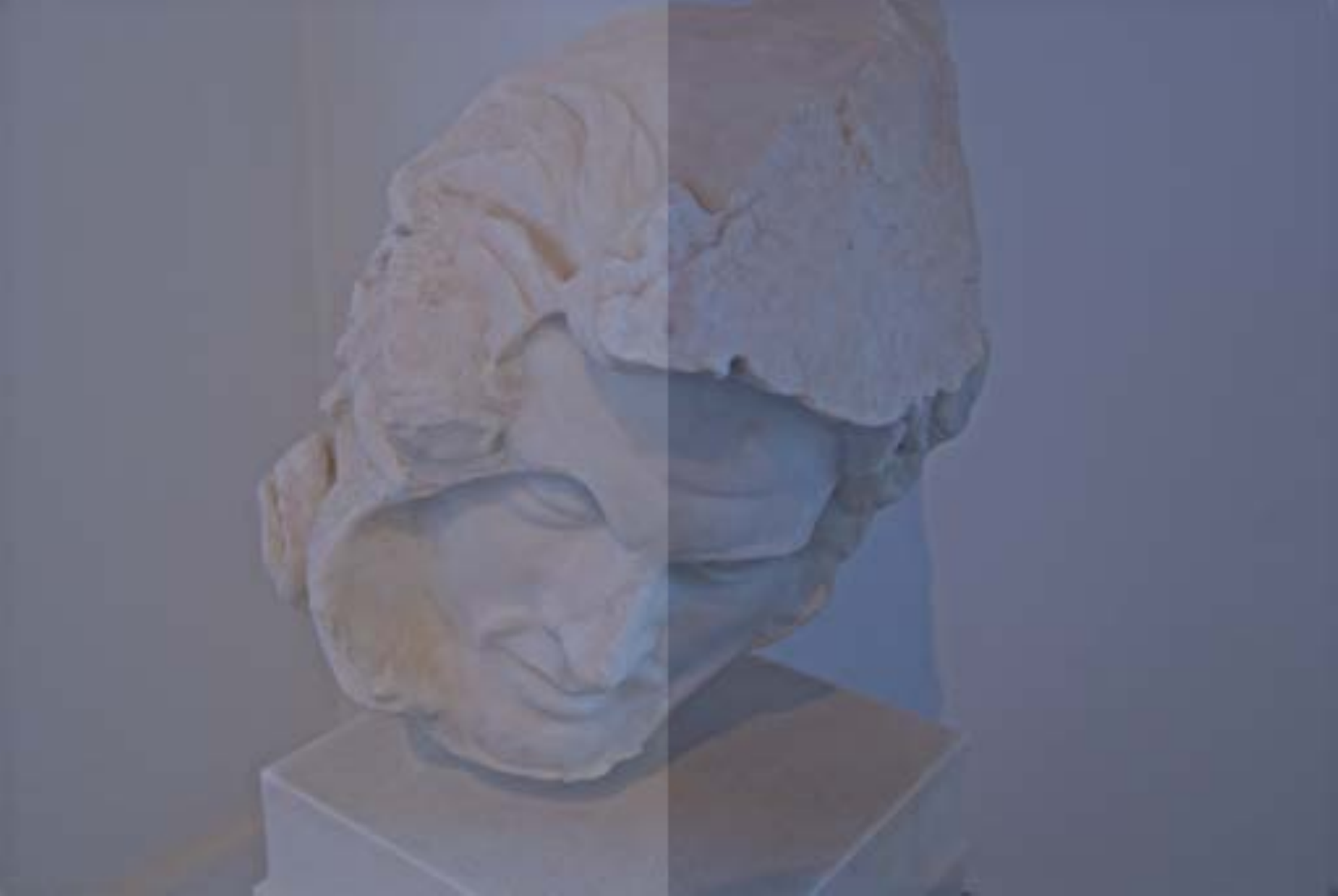}}\\
	\subfigure[CSDNet]{
		\includegraphics[width=0.15\linewidth]{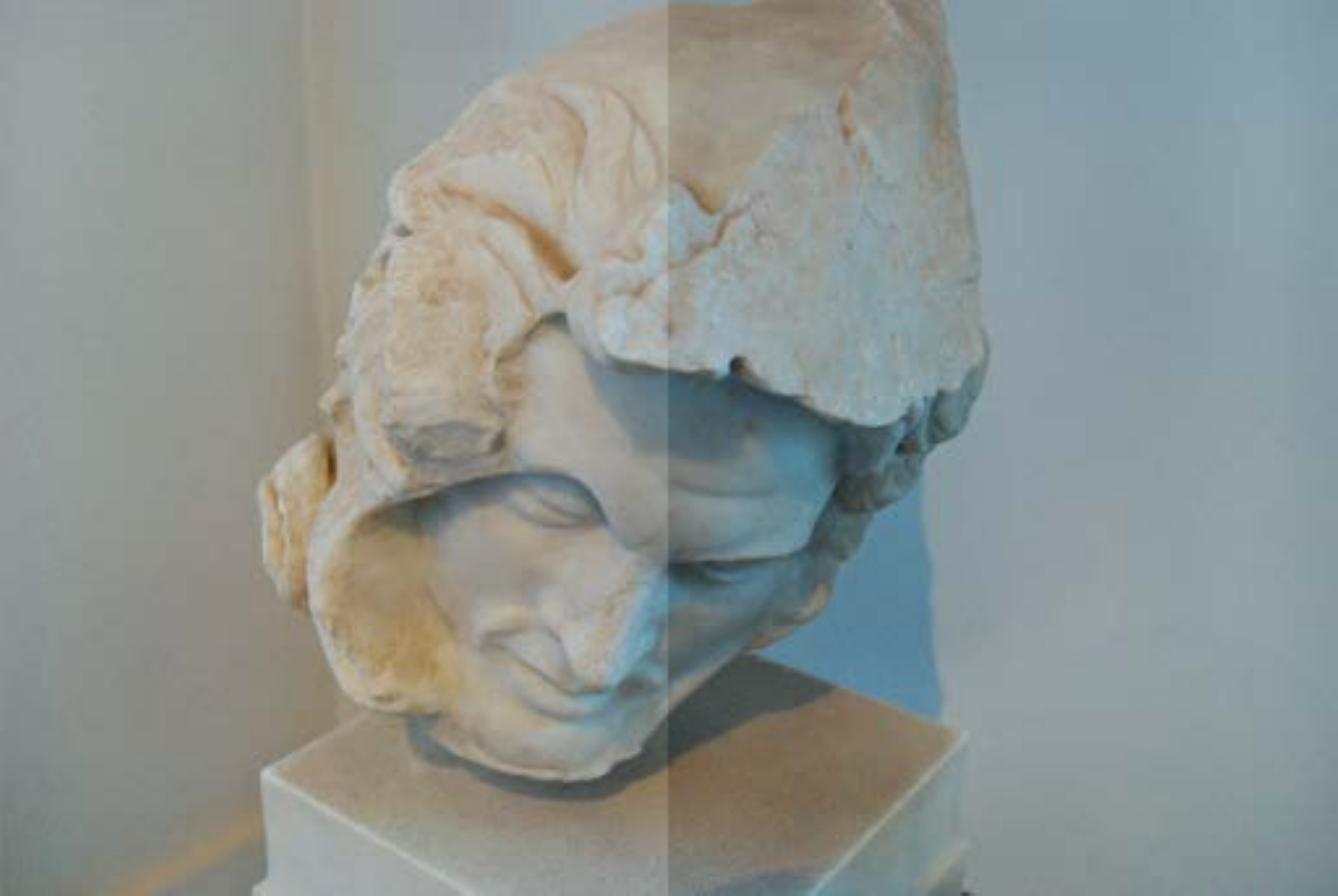}}
	\subfigure[DeepUPE]{
		\includegraphics[width=0.15\linewidth]{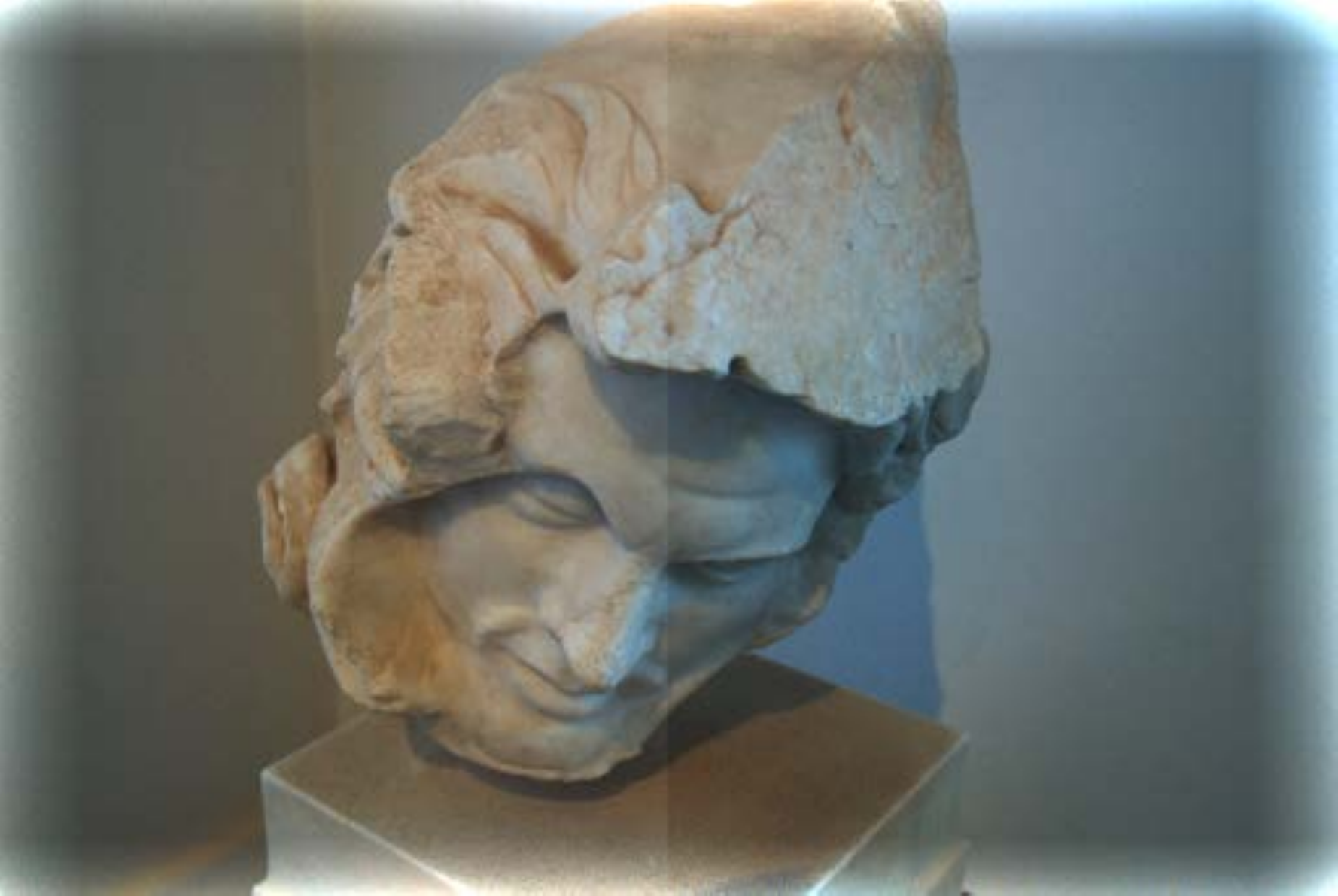}}
	\subfigure[MBLLEN]{
		\includegraphics[width=0.15\linewidth]{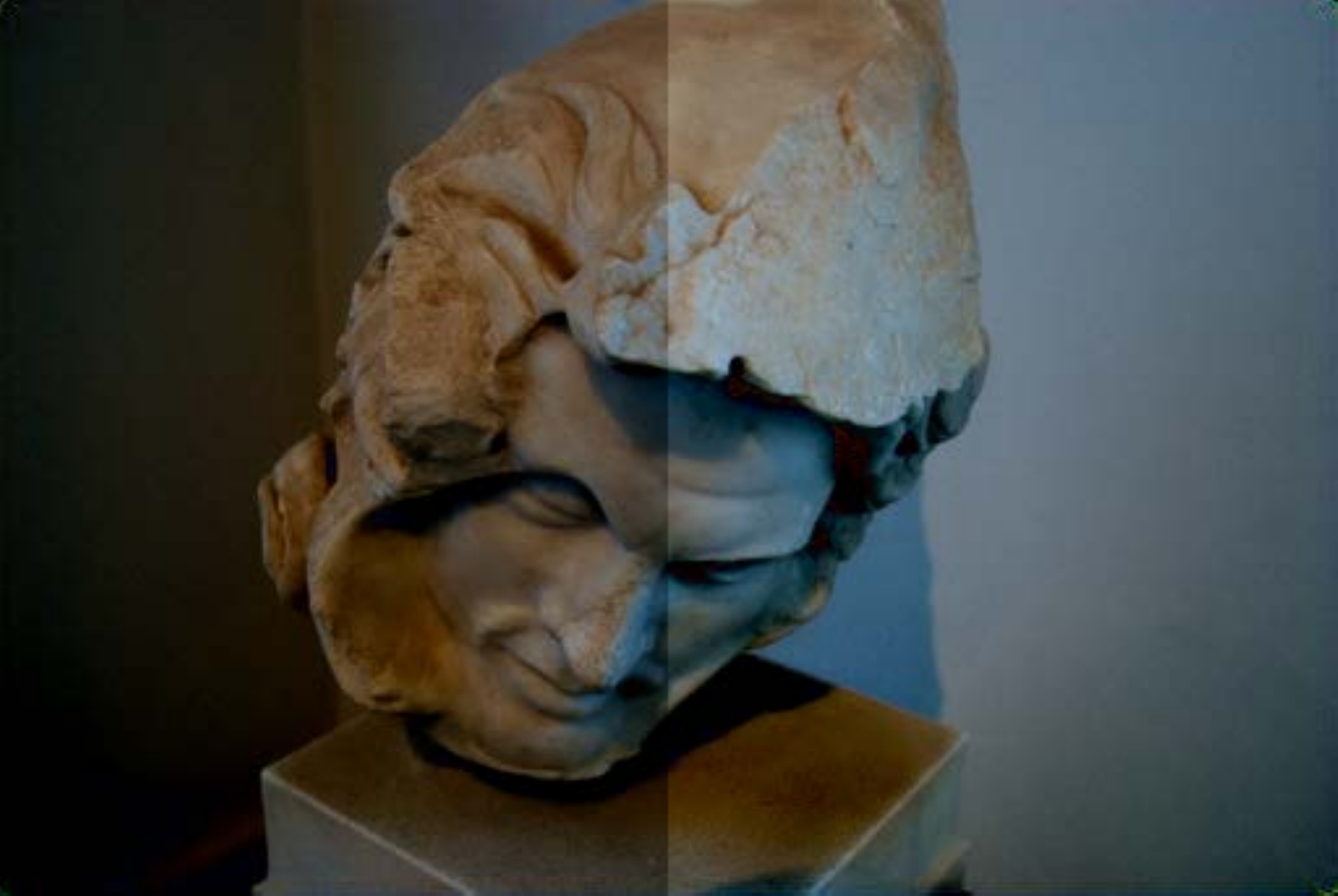}}
	\subfigure[RetinexNet]{
		\includegraphics[width=0.15\linewidth]{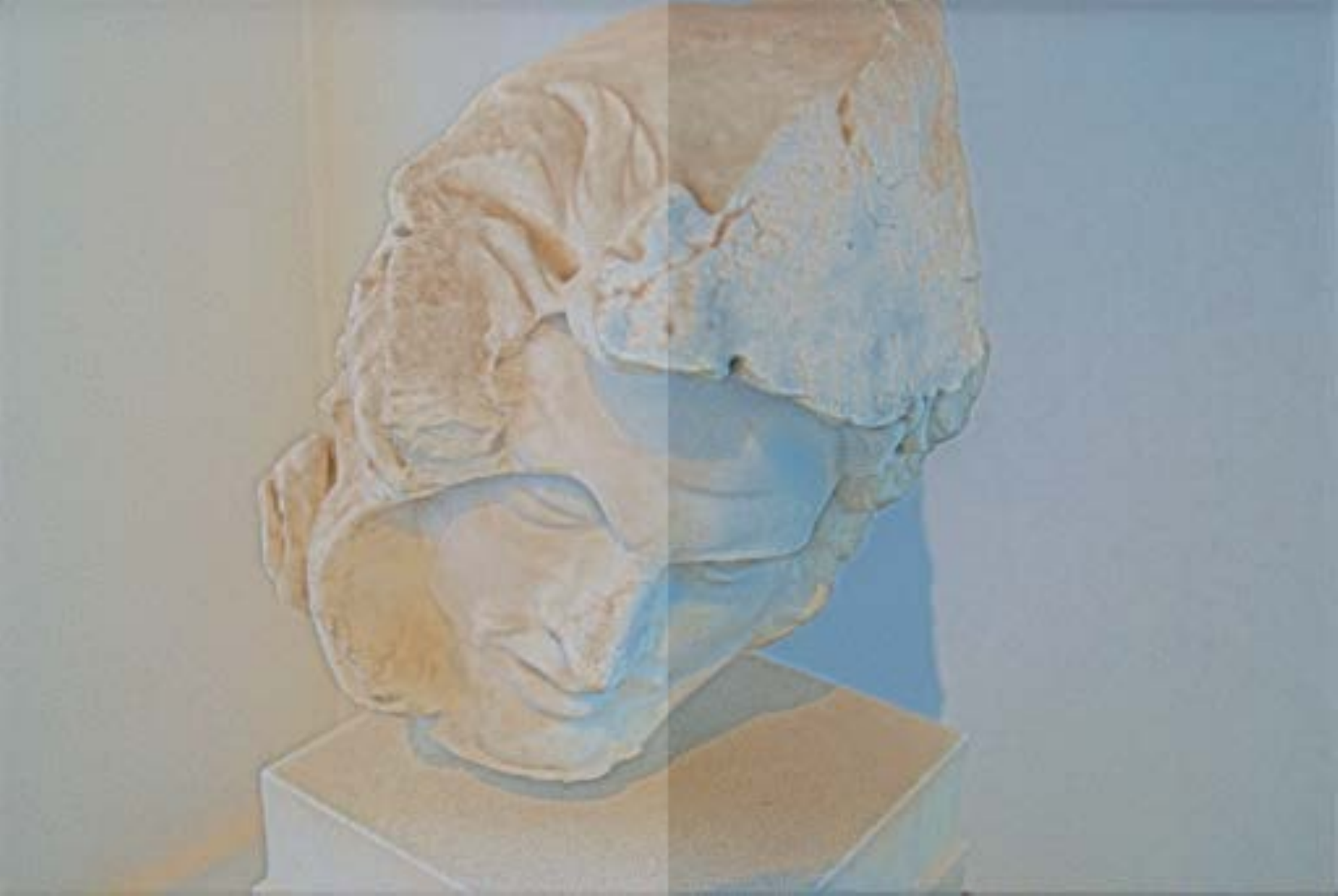}} 
	\subfigure[RUAS]{
		\includegraphics[width=0.15\linewidth]{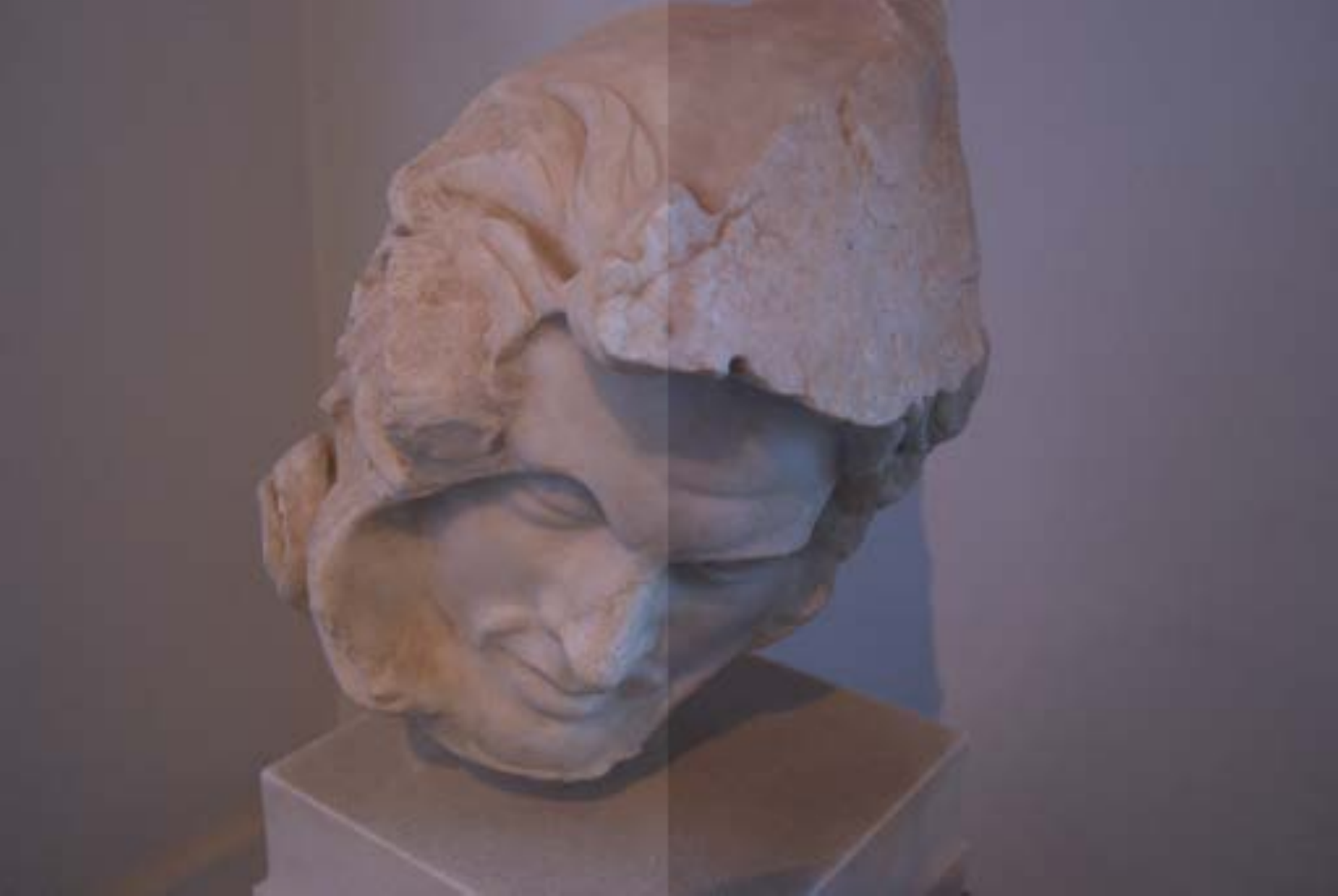}} 
	\subfigure[TBEFN]{
		\includegraphics[width=0.15\linewidth]{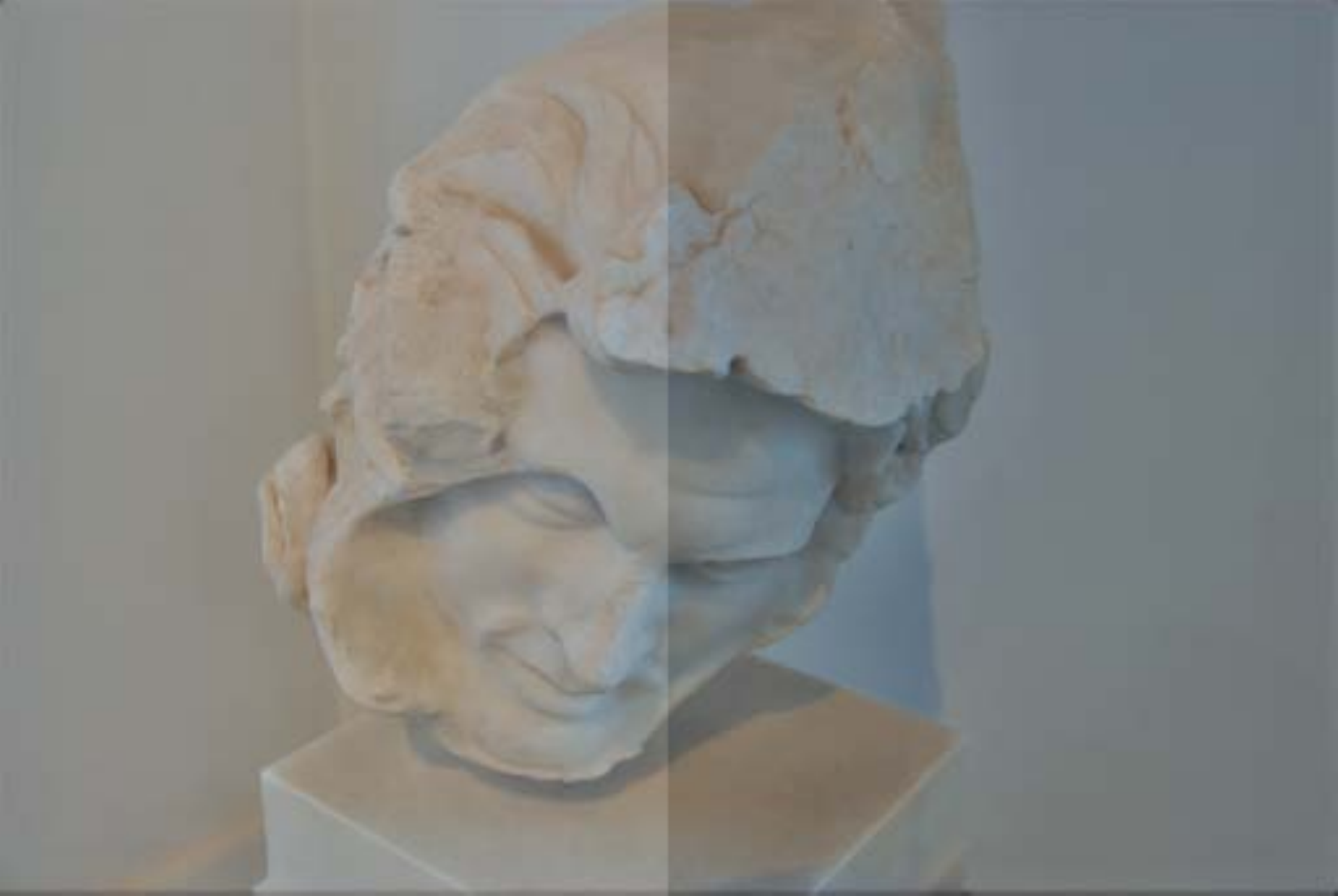}}\\
	\subfigure[SGM]{
		\includegraphics[width=0.15\linewidth]{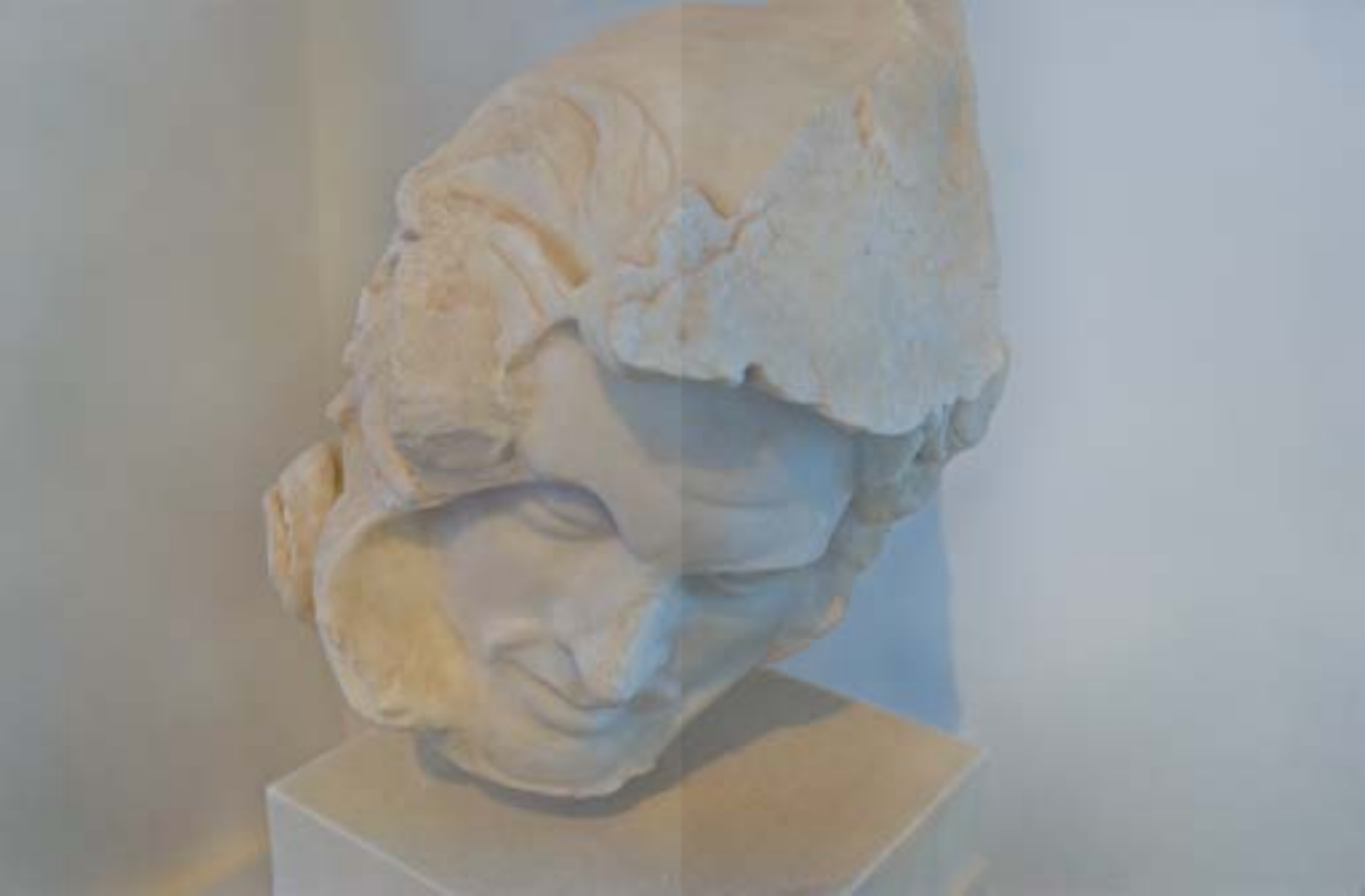}}
	\subfigure[DRBN]{
		\includegraphics[width=0.15\linewidth]{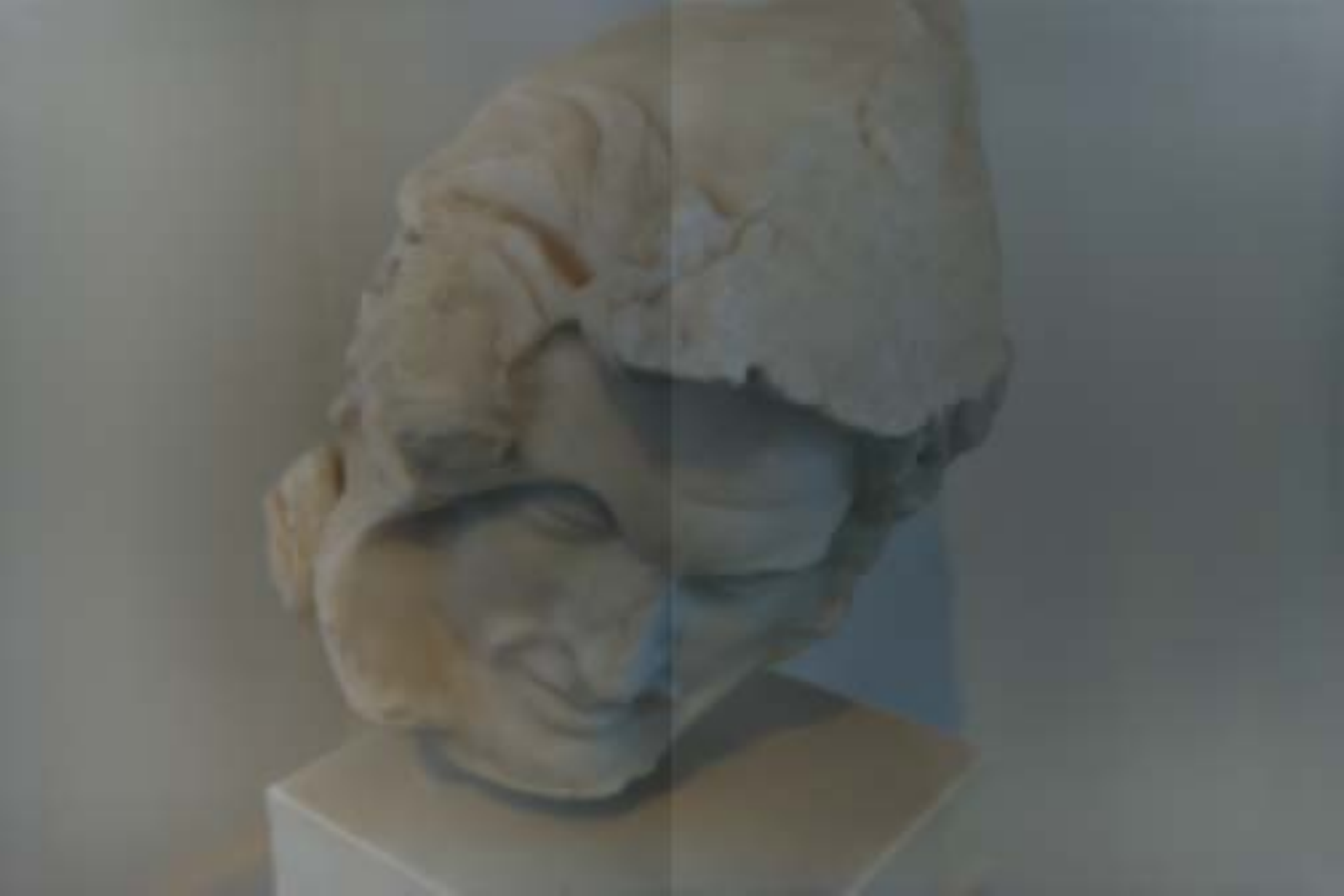}}
	\subfigure[KinD++]{
		\includegraphics[width=0.15\linewidth]{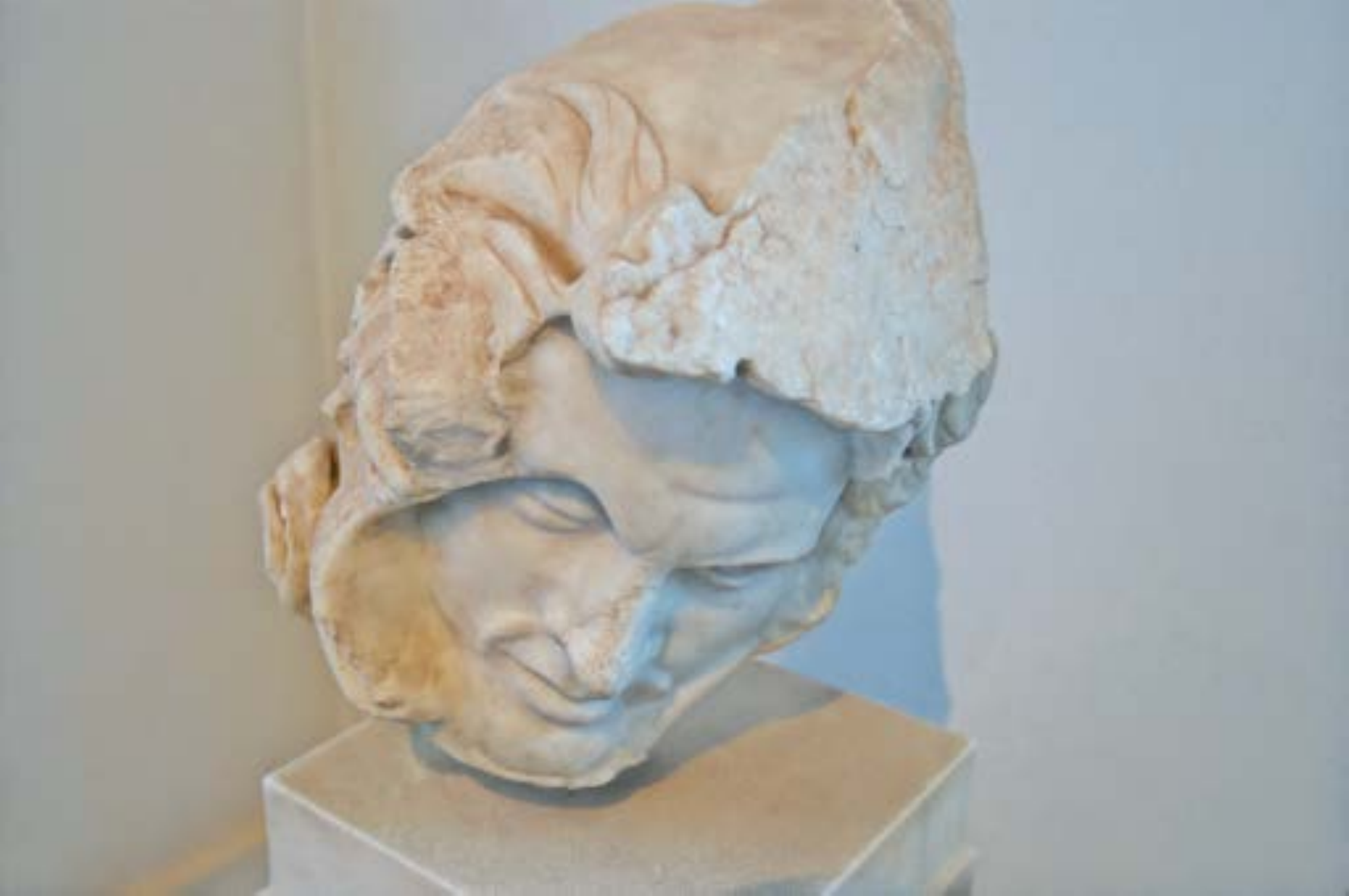}}
	\subfigure[RAUNA]{
		\includegraphics[width=0.15\linewidth]{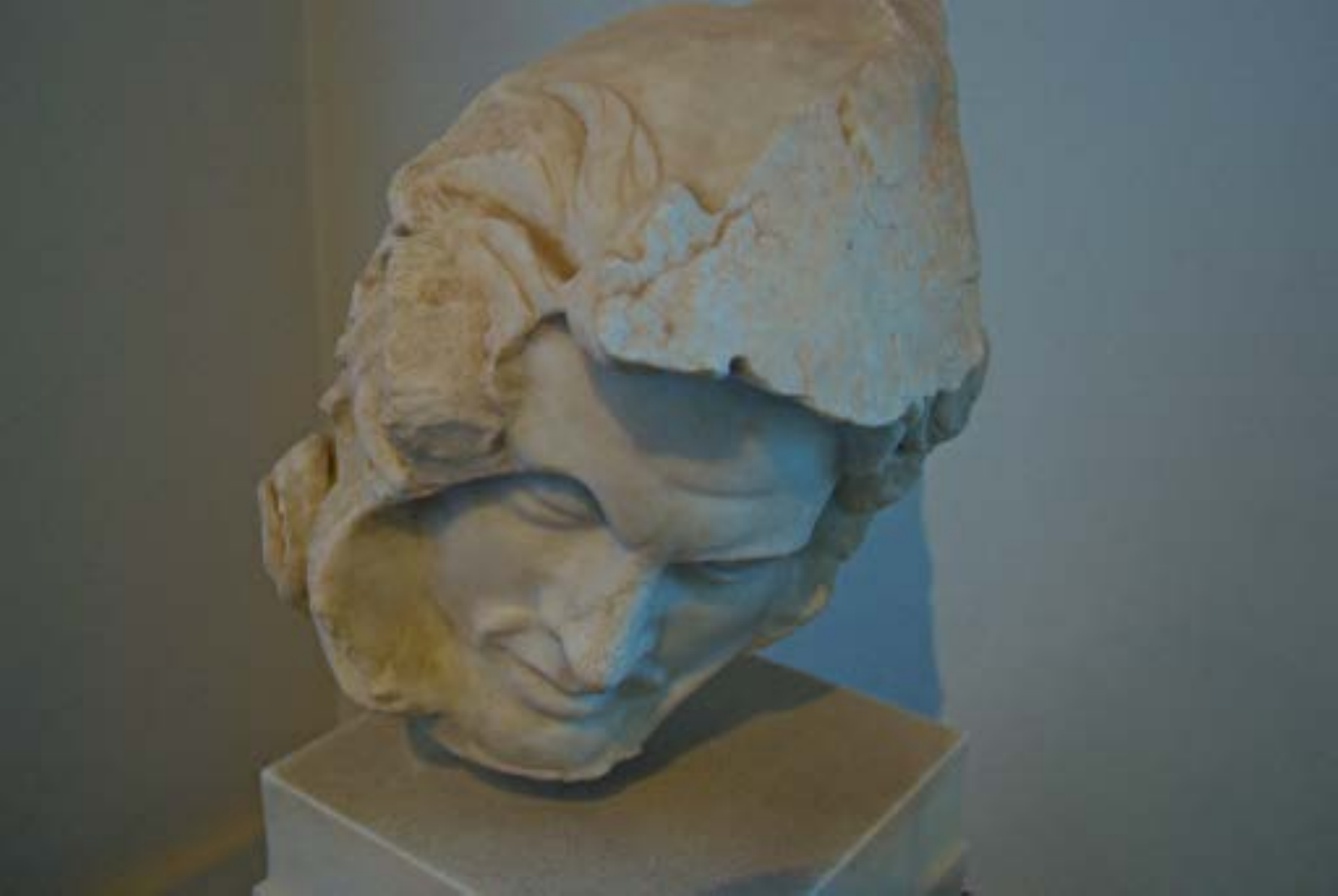}}
	\subfigure[RAUNA$_{\textrm{ft}}$]{
		\includegraphics[width=0.15\linewidth]{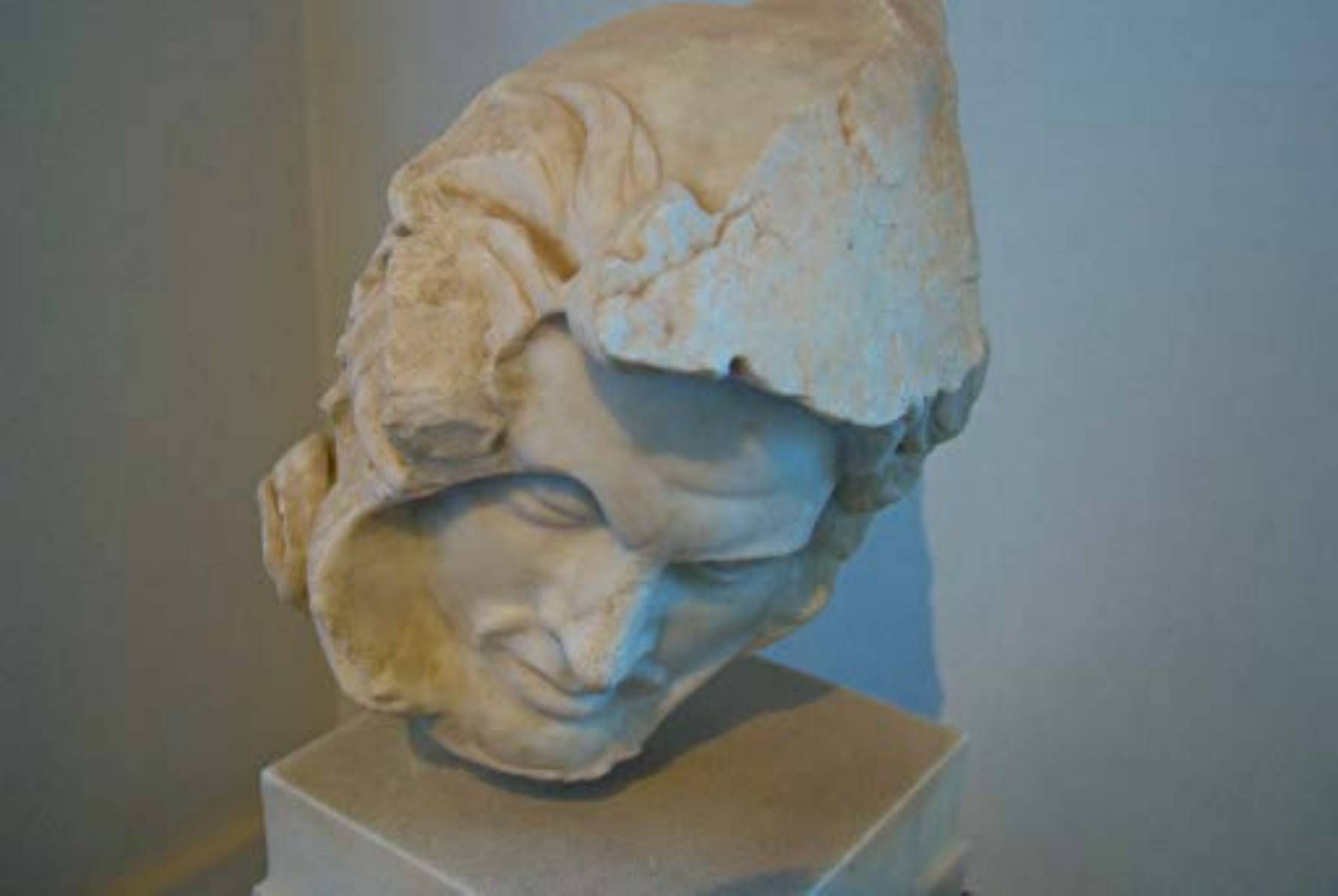}}
	\label{lol15}
	\subfigure[Ground Truth]{
		\includegraphics[width=0.15\linewidth]{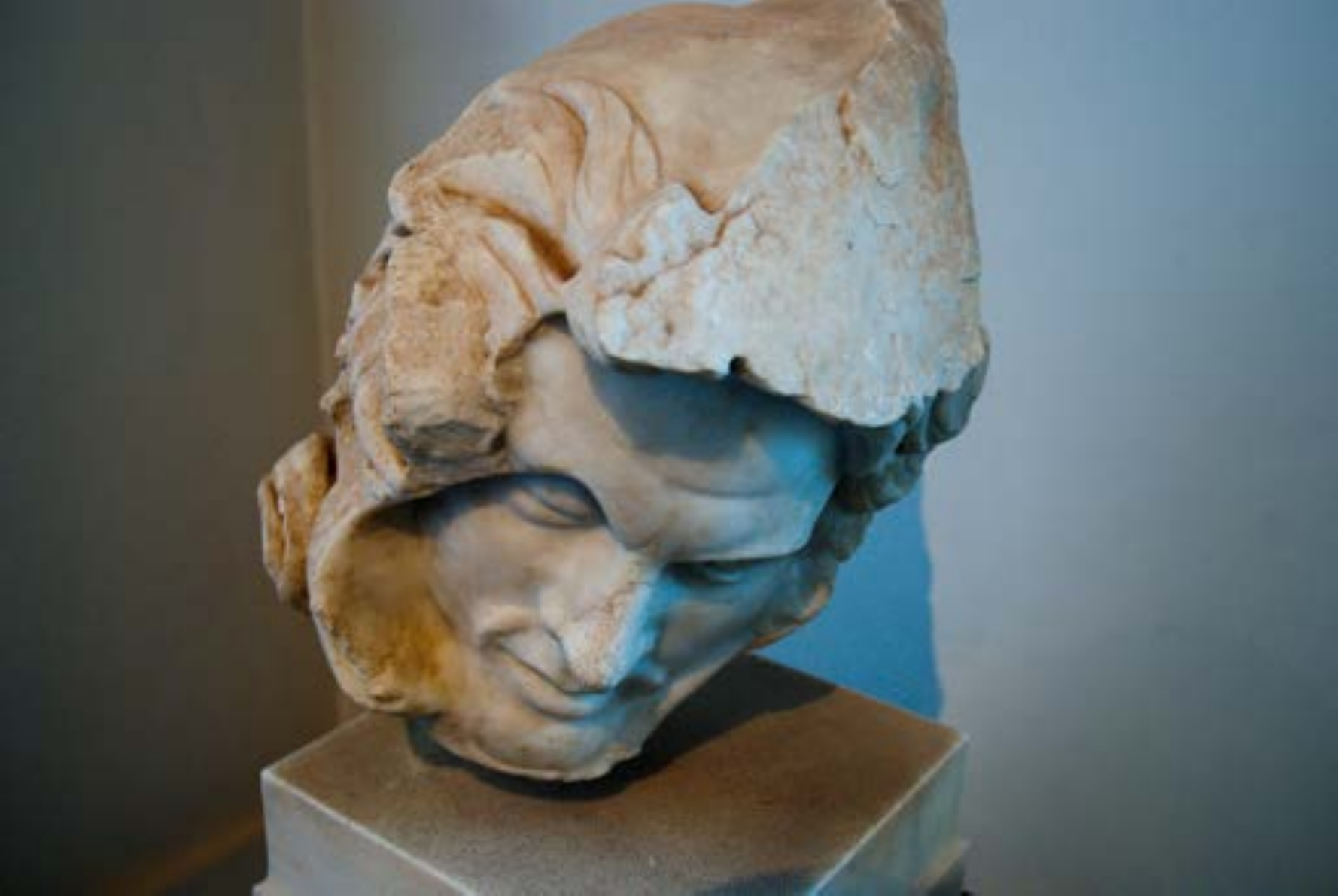}}
	\caption{Visual results of competing methods on  MIT-Adobe FiveK dataset. For (b)-(n), the left and parts are from enhanced images without and with Gamma Correction, respectively.}
	\label{fig:mit1} 
\end{figure*}

\subsection{Comparison with Existing Methods}
\subsubsection{Results on LOL Dataset}
The quantitative and visual results of all competing methods on LOL dataset are shown in Table \ref{tab:lol} and Fig. \ref{fig:lol1}. %(more visual results are provided in supplementary material). 
%The subscript ``ref'' for KinD++ and our method means that the global brightness parameter is determined by the reference normal-light image, and ``ft'' for our method indicates the self-supervised fine-tuning strategy introduced in Section \ref{sec:finetune}. 
Following \citep{yang2020fidelity}, we also report the results after GC to emphasize detail fidelity of the enhanced images in evaluation, except for KinD++ and our methods since the adjustment networks within their frameworks can automatically adjust the global illumination, with reference to groundtruth normal-light image or pseudo one.

It can be seen from Table \ref{tab:lol}, our methods achieve the best performance under all metrics that computed with reference to groundtruth normal-light images, including both the general IQA metrics, i.e., PSNR and SSIM, and that specifically for LIE problem, i.e., LOE$_{\textrm{ref}}$. Besides, for blind IQA metrics, our methods are also very competitive. The better performance of the proposed methods can be more directly observed from Fig. \ref{fig:lol1}. Specifically, most of the competing methods do not perform very well on this challenging image, that their results could still contain unexpected degradations, such as low brightness, color bias and noise, while the results by CSDNet, KinD++ and our methods are relatively more acceptable. However, the result by CSDNet seems more noisy and with less contrast, and KinD++ over-smooths some regions of the image. In comparison with other methods, our methods can produce relatively more balanced results regarding brightness, contrast, color bias and noise.

\begin{table*}[t]
	\begin{center}
		\caption{Quantitative comparison of all competing methods on NPE and DICM Datasets. The best and second best results are highlighted in {\color{red}RED} and {\color{blue}BLUE}, respectively.}
		\label{tab:un}
		\begin{tabular}{ l | c | c | c | c  }
			\hline
			& \multicolumn{2}{c|}{DICM} & \multicolumn{2}{c}{NPE}  \\
			\hline
			Method & NIQE $\downarrow$ & LOE $\downarrow$ & NIQE $\downarrow$ & LOE $\downarrow$\\
			\hline
			CLAHE \citep{CLAHE} & 3.659 & 237.1&3.093&337.2\\
			CLAHE+BM3D &5.481&241.0&4.140&338.6\\
			%BIMEF\citep{ying2017bio} & 3.529 & 190.9 & 3.832 & 244.8 \\
			LR3M \citep{LR3M} & 4.963  & 283.4  & 4.377  & 295.3 \\
			NPE \citep{wang2013naturalness} & 3.669 & 320.6 & {\color{red}\textbf{3.030}} & 366.6\\
			\hline
			Zero-DCE++ \citep{Zero-DCE++} & 3.589 & 189.1 & 3.211 & 227.9\\
			\hline
			CSDNet \citep{ma2021learning}& 3.697 & 312.3 & 3.287 & 344.1  \\
			DeepUPE \citep{wang2019underexposed} & 3.687 & 245.2 & 3.440 & 299.4\\
			MBLLEN \citep{Lv2018MBLLEN}& 3.841 & 266.0 & 4.165 & 228.2 \\
			RetinexNet \citep{Chen2018Retinex}& 5.003 & 504.9 & 4.151 & 1478.5\\
			RUAS \citep{liu2021retinex}& 4.055 & 213.8 & 4.918 & 361.5\\
			TBEFN \citep{lu2020tbefn} & {\color{red}\textbf{2.903}} & 351.4 & {\color{blue}\textbf{3.062}} & 376.4\\
			SGM \citep{yang2021sparse}&{\color{blue}\textbf{3.083}} & 411.8 & 3.079 & 510.9 \\
			DRBN \citep{yang2020fidelity} & 4.739  & 642.3  & 4.583  & 660.1 \\
			KinD++ \citep{zhang2021beyond} & 3.173 & 371.0 & 3.406 & 455.9 \\
			\hline
			RAUNA & 3.155 & {\color{blue}\textbf{188.7}} & 3.153 & {\color{blue}\textbf{227.2}} \\
			RAUNA$_{\textrm{ft}}$ & 3.124 & {\color{red}\textbf{188.4}} & 3.238 & {\color{red}\textbf{224.9}}\\
			\hline 
		\end{tabular}
	\end{center}
\end{table*}

\begin{figure*}[t]
	\centering
	\subfigure[Input]{
		\includegraphics[width=0.15\linewidth]{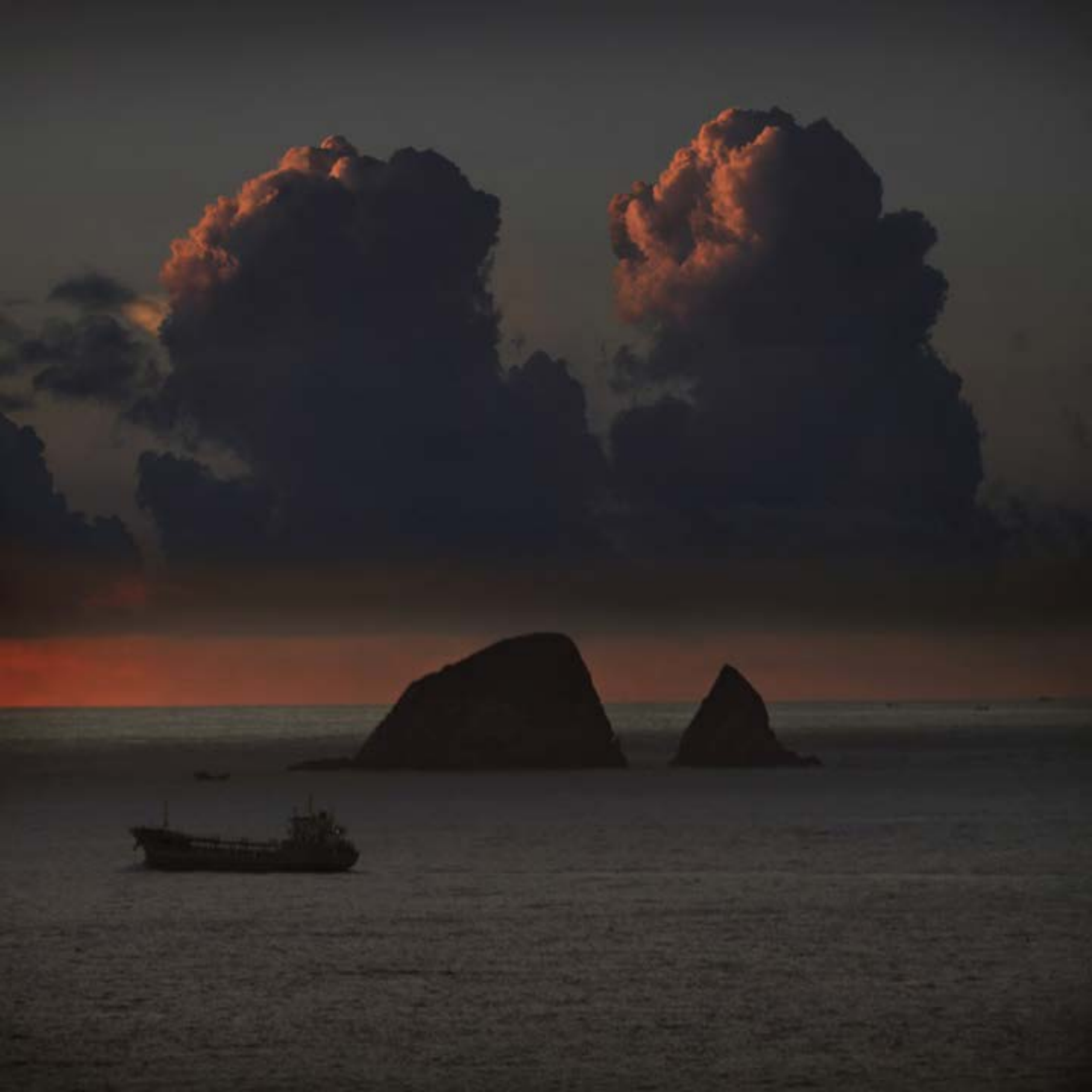}}
	\subfigure[CLAHE]{
		\includegraphics[width=0.15\linewidth]{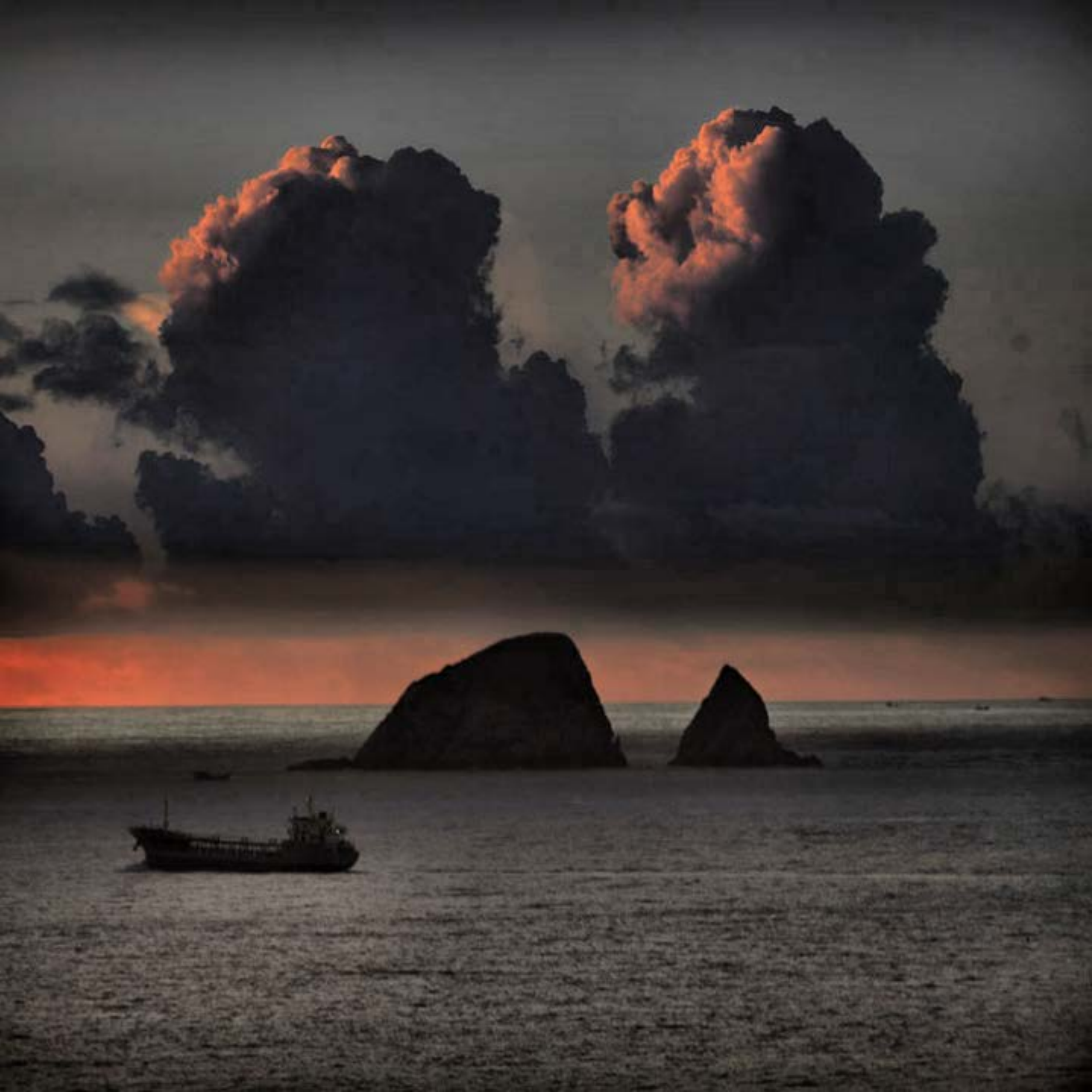}}
	\subfigure[CLAHE+BM3D]{
		\includegraphics[width=0.15\linewidth]{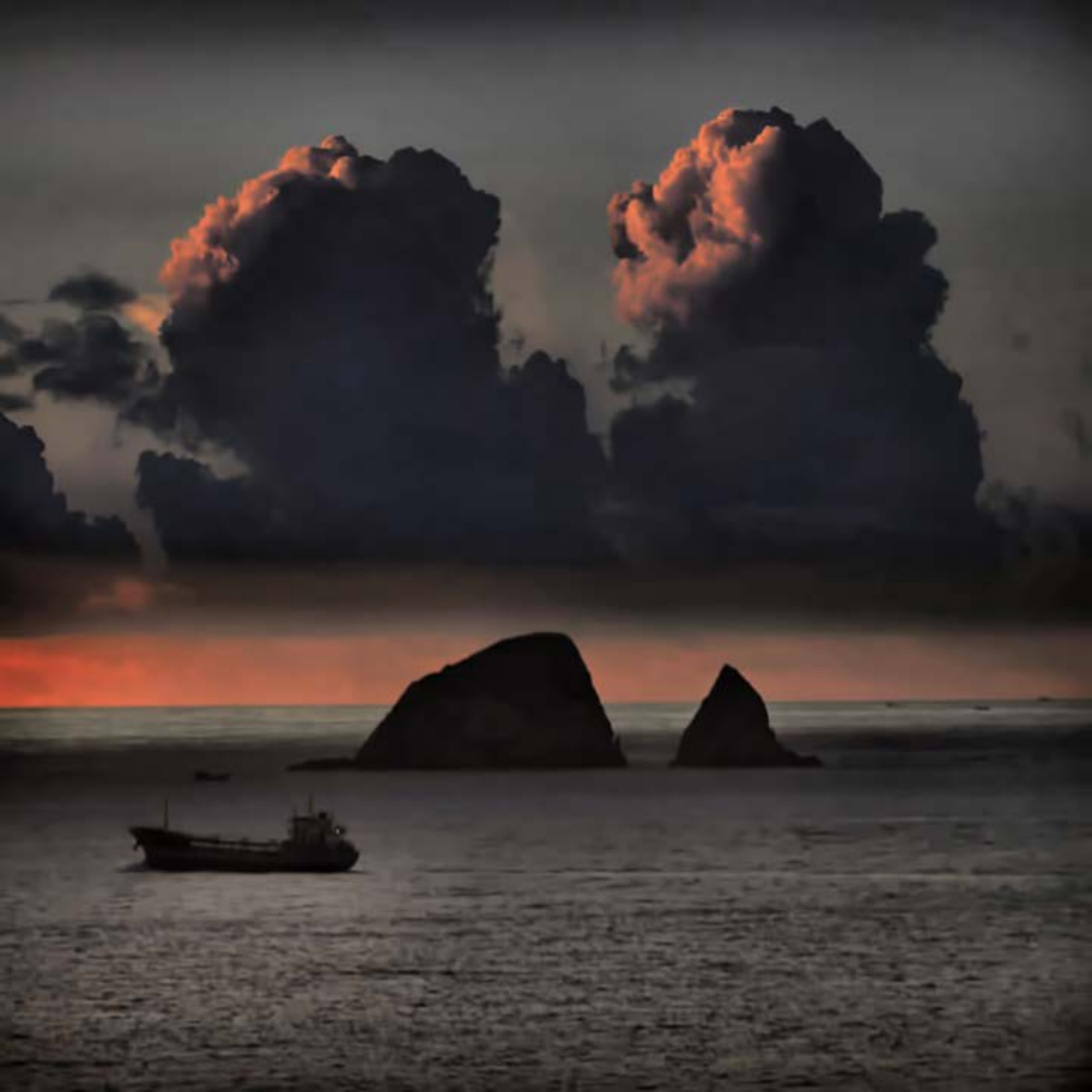}}
	\subfigure[NPE]{
		\includegraphics[width=0.15\linewidth]{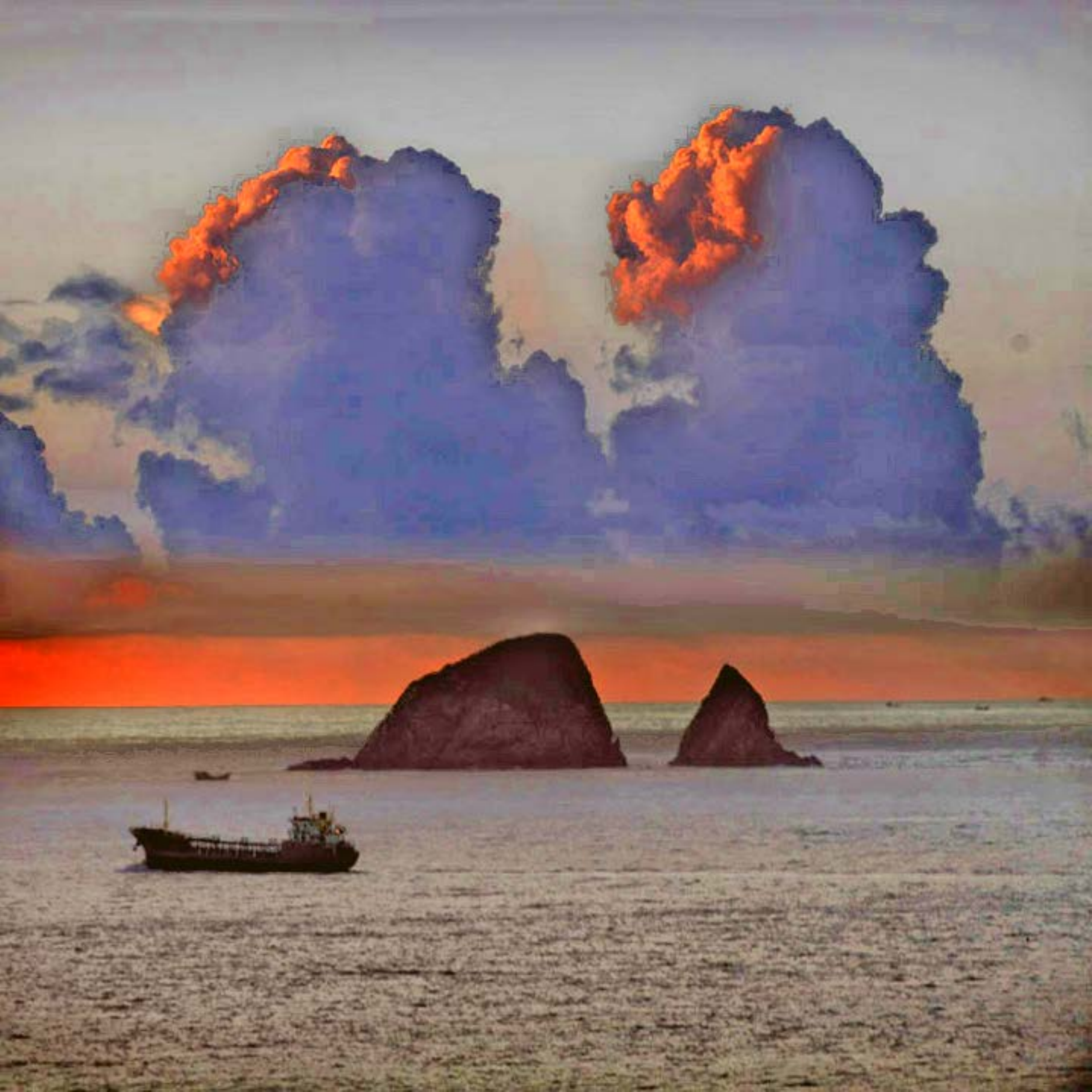}}
	\subfigure[LR3M]{
		\includegraphics[width=0.15\linewidth]{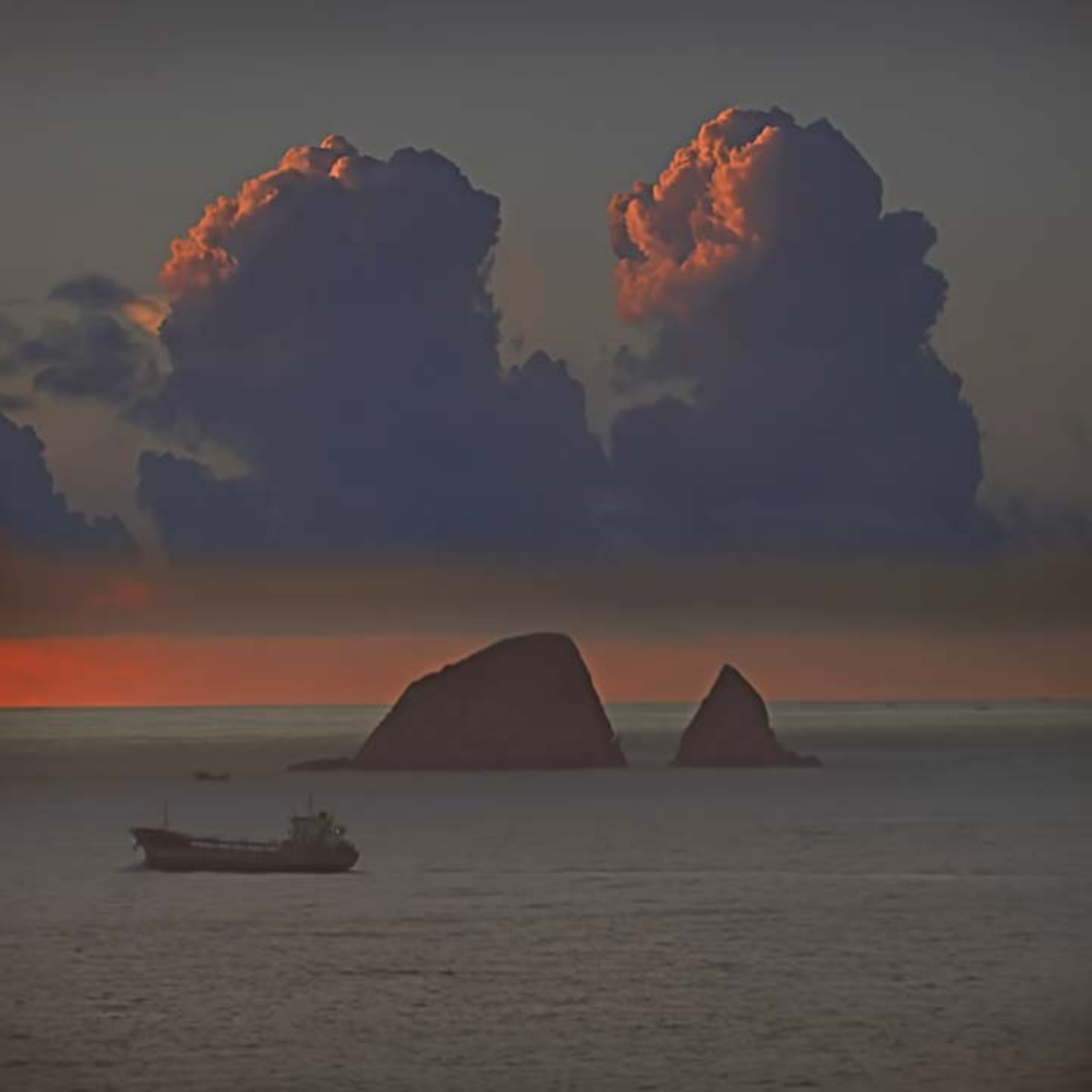}}
	\subfigure[Zero-DCE++]{
		\includegraphics[width=0.15\linewidth]{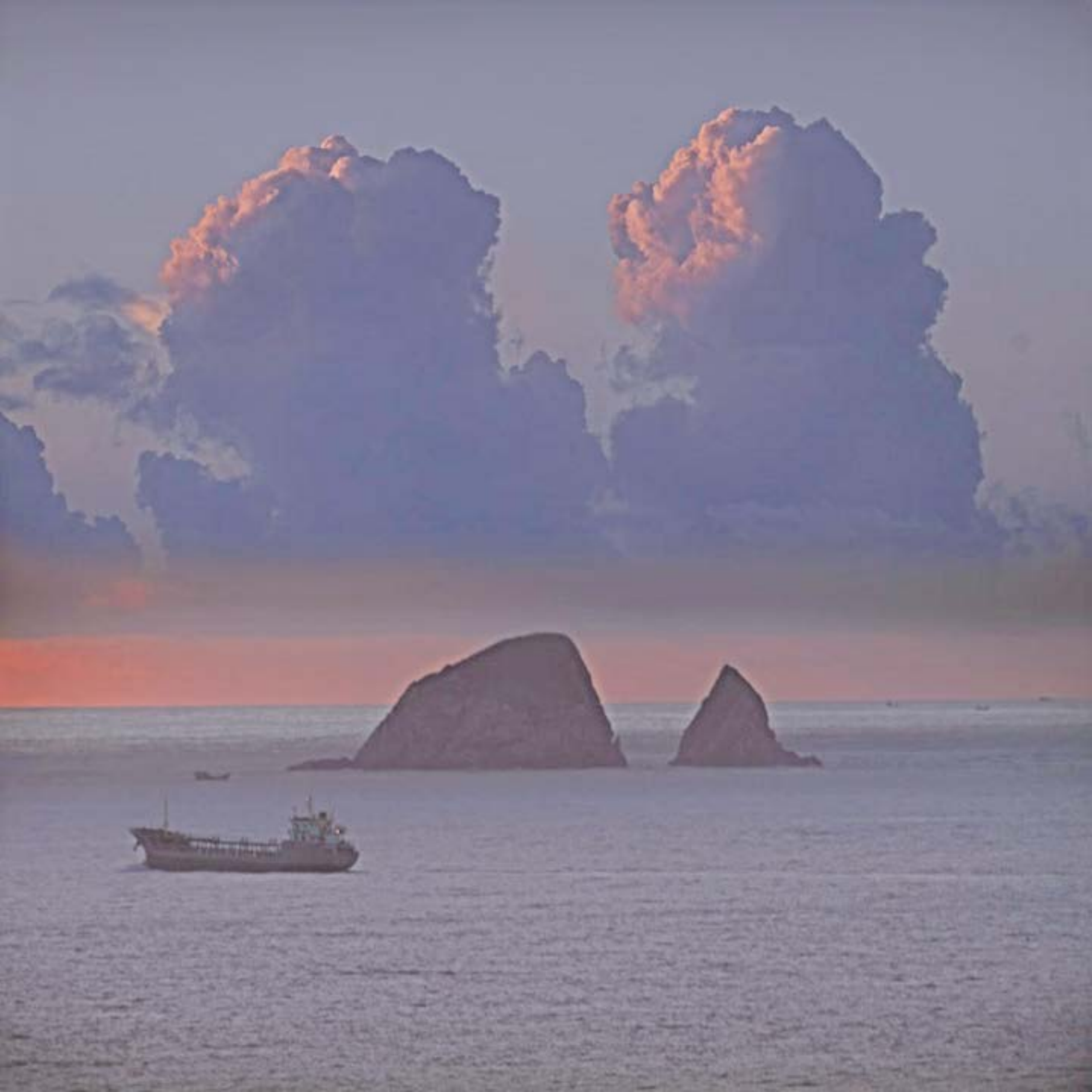}}\\
	\subfigure[CSDNet]{
		\includegraphics[width=0.15\linewidth]{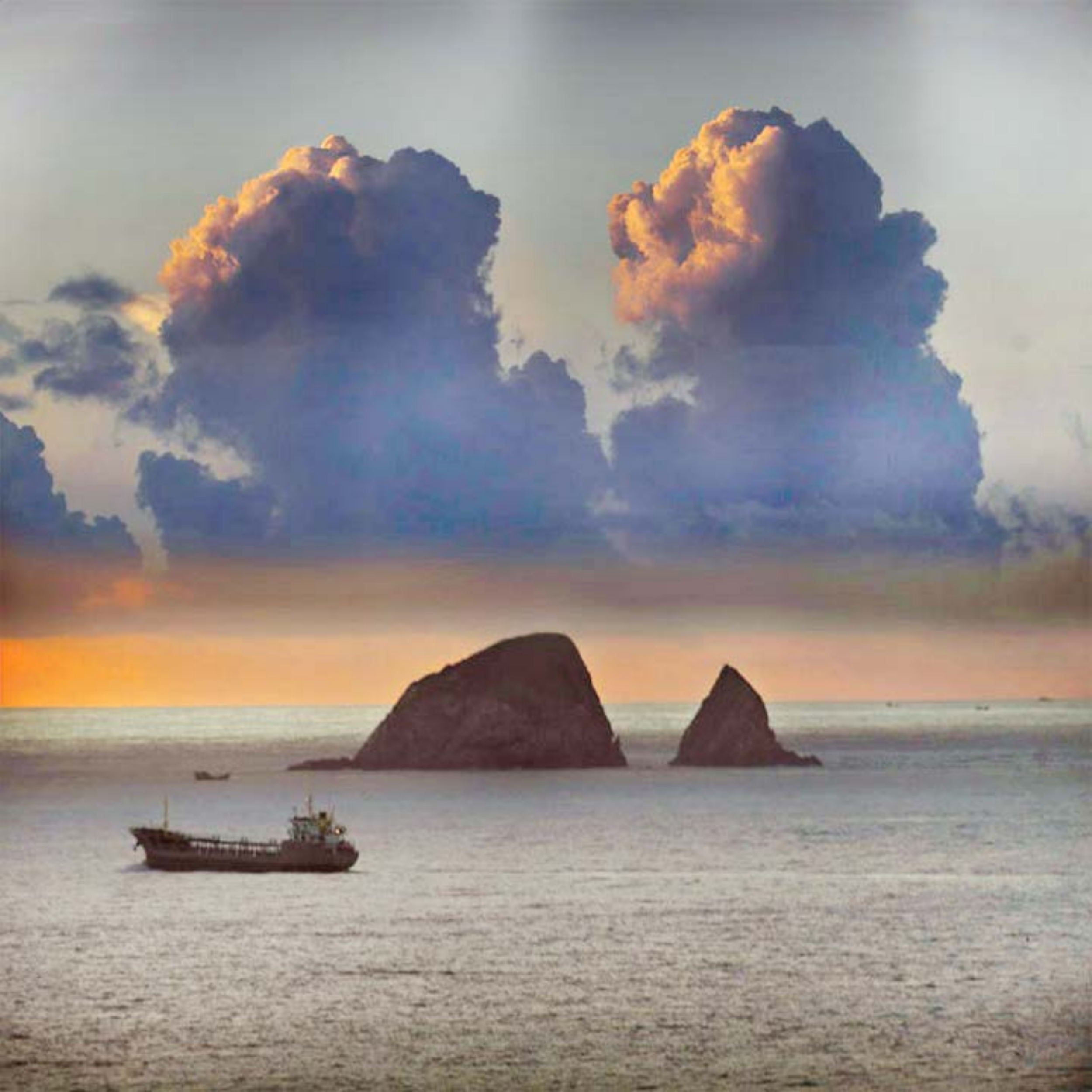}}
	\subfigure[DeepUPE]{
		\includegraphics[width=0.15\linewidth]{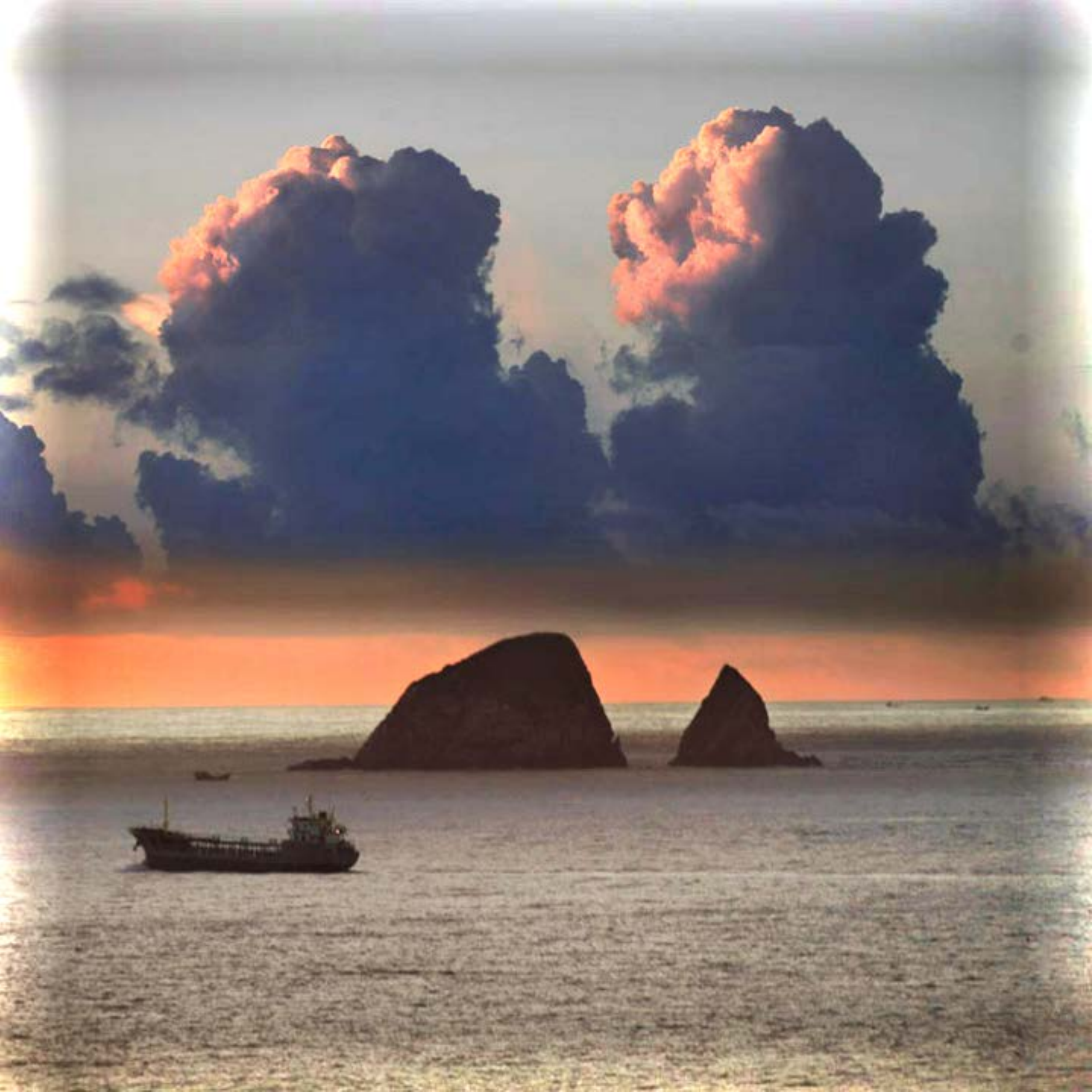}}
	\subfigure[MBLLEN]{
		\includegraphics[width=0.15\linewidth]{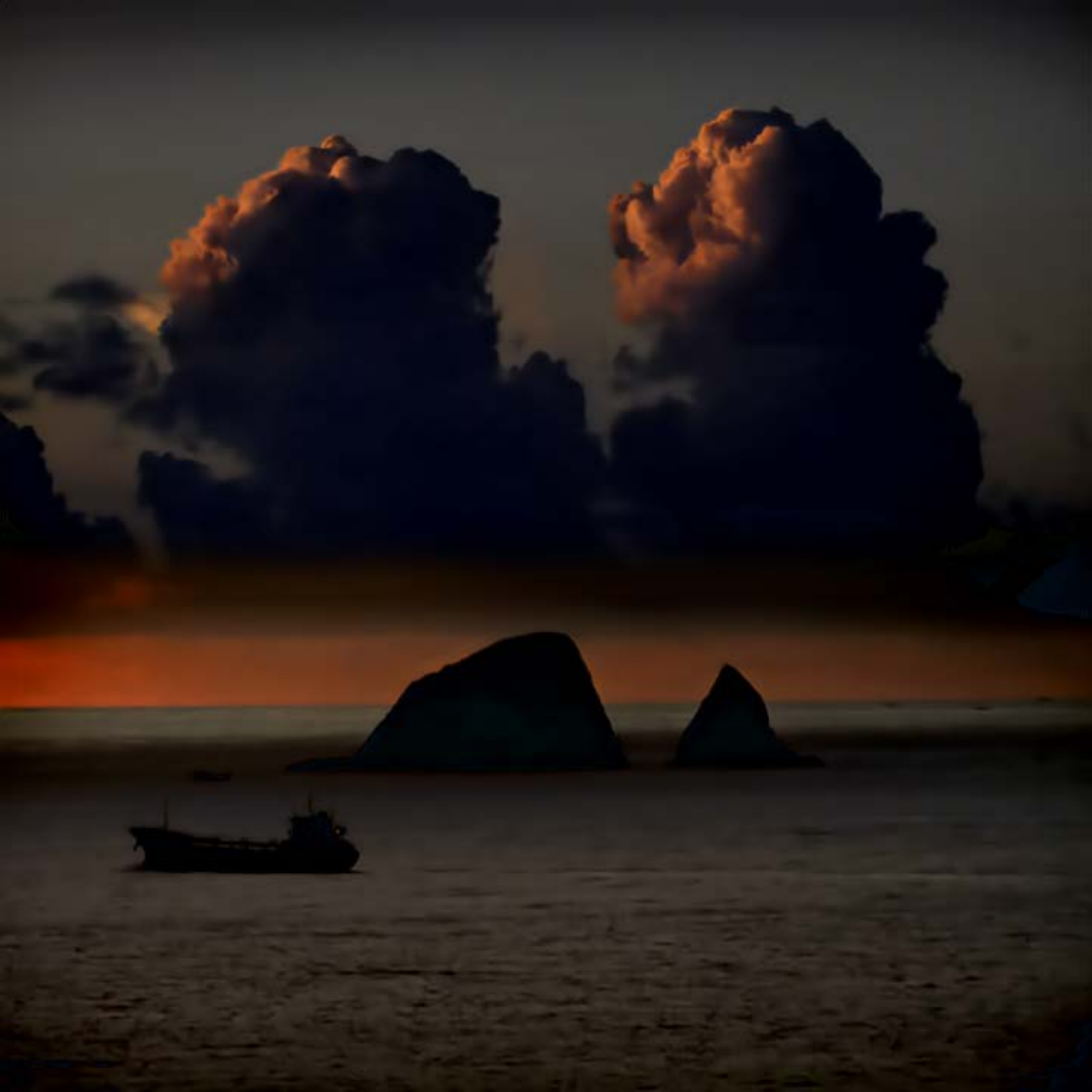}}
	\subfigure[RetinexNet]{
		\includegraphics[width=0.15\linewidth]{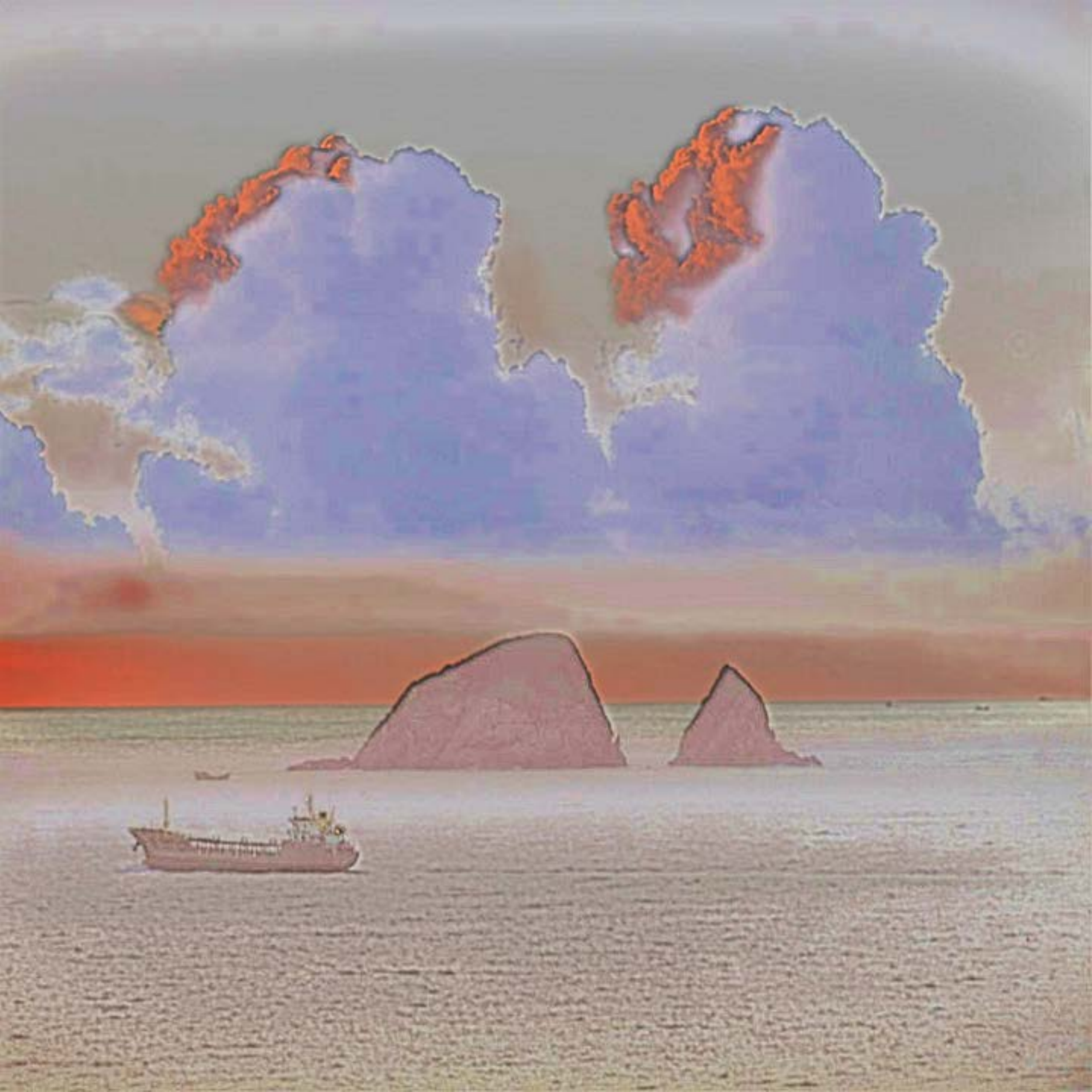}}
	\subfigure[RUAS]{
		\includegraphics[width=0.15\linewidth]{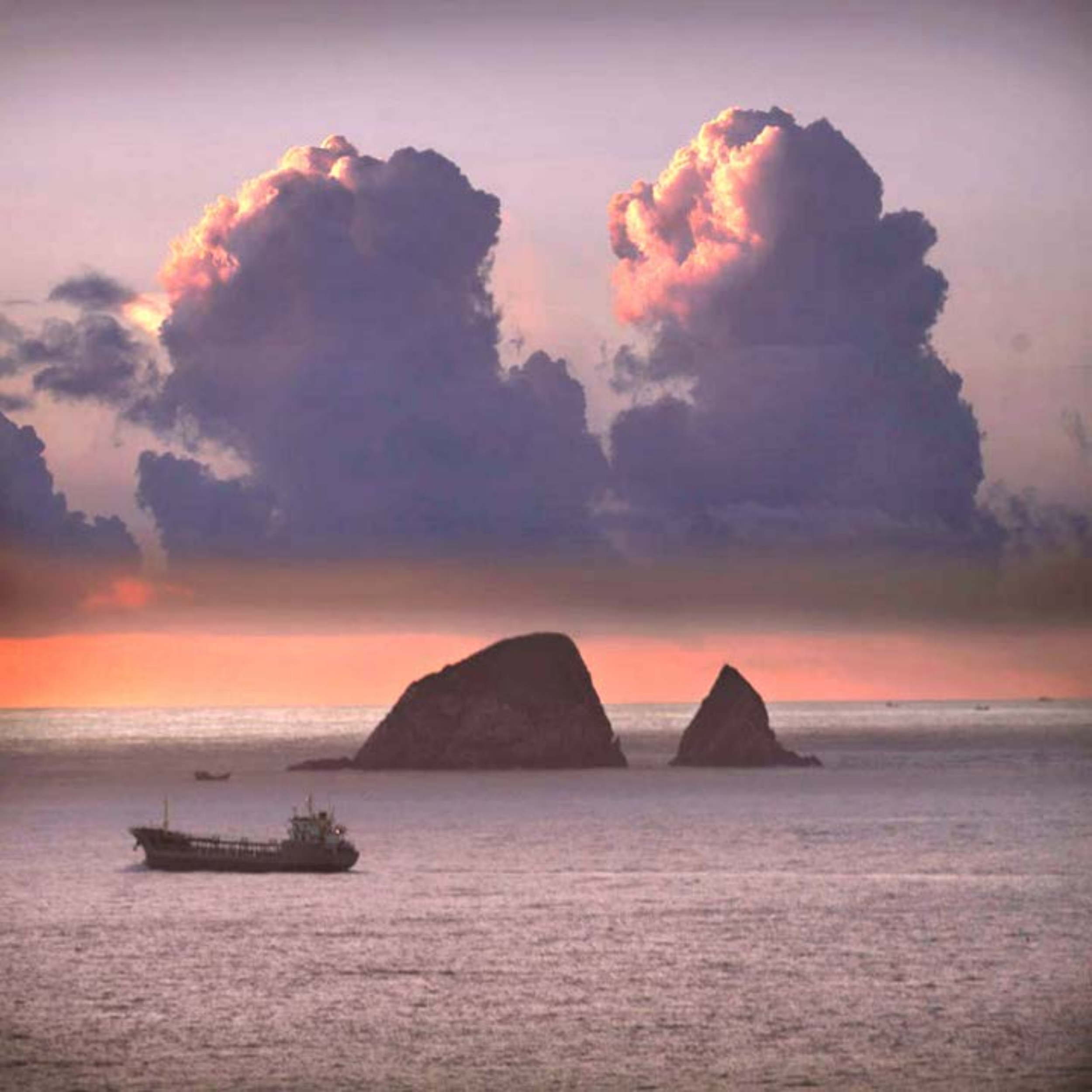}}
	\subfigure[TBEFN]{
		\includegraphics[width=0.15\linewidth]{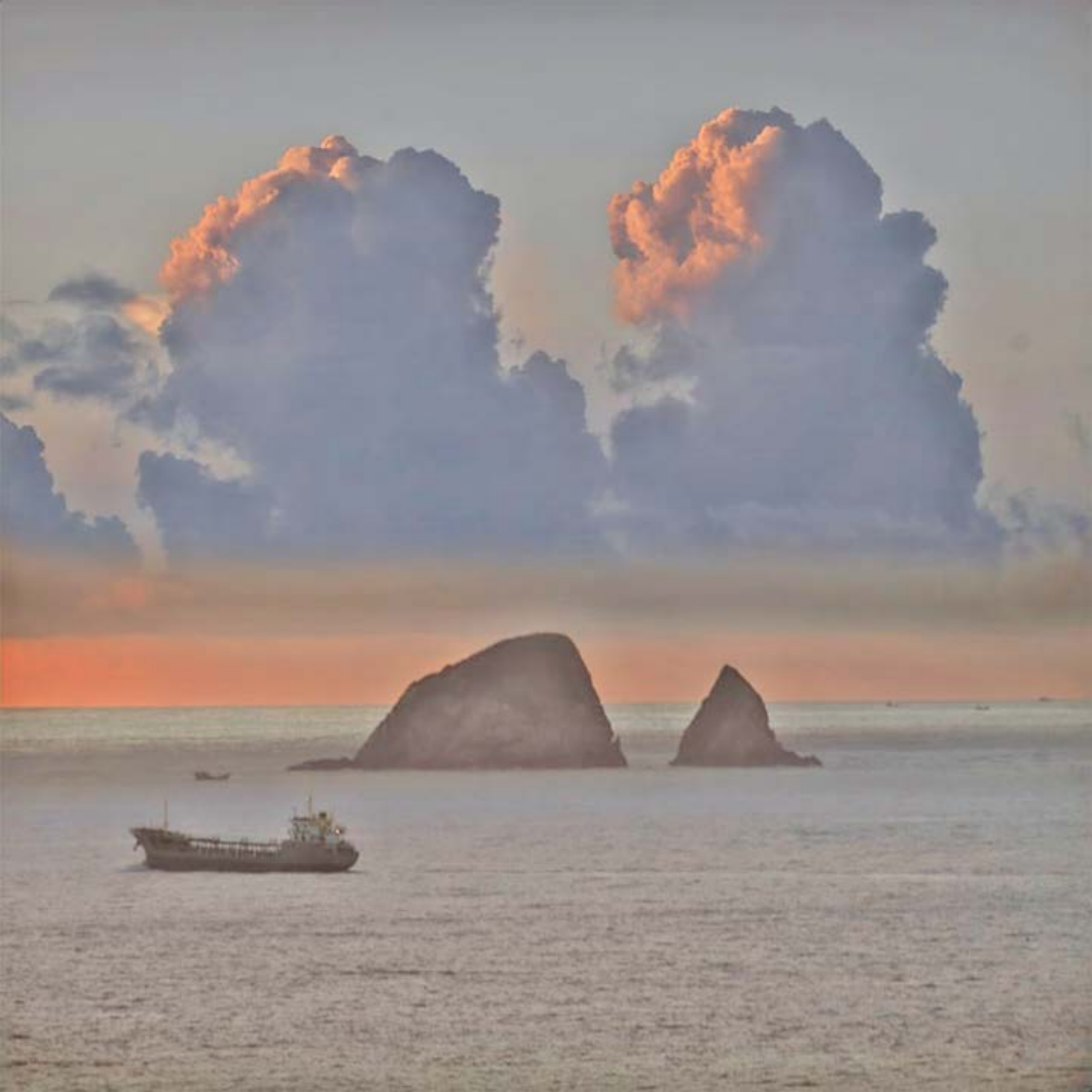}}\\
	\subfigure[SGM]{
		\includegraphics[width=0.15\linewidth]{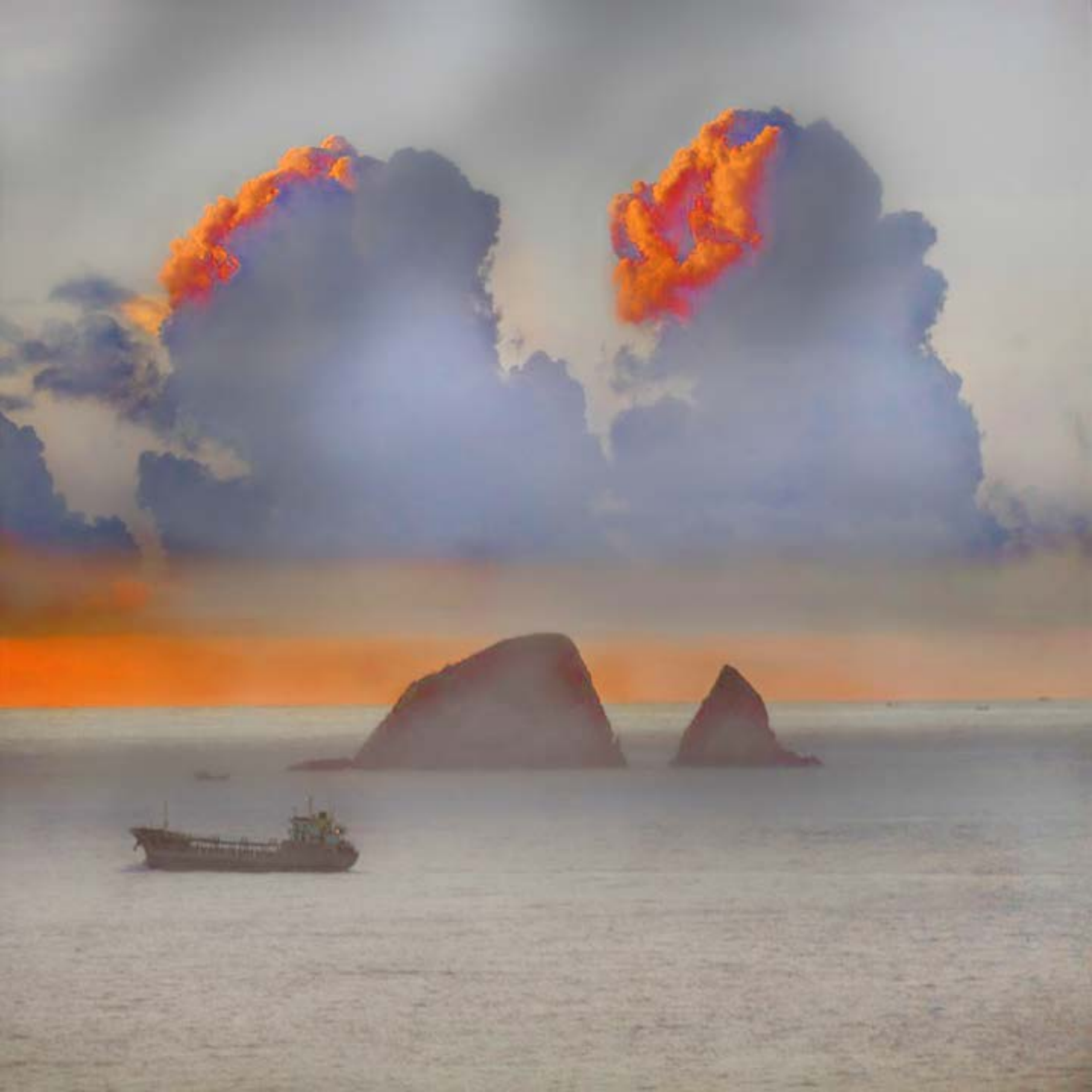}}
	\subfigure[KinD++]{
		\includegraphics[width=0.15\linewidth]{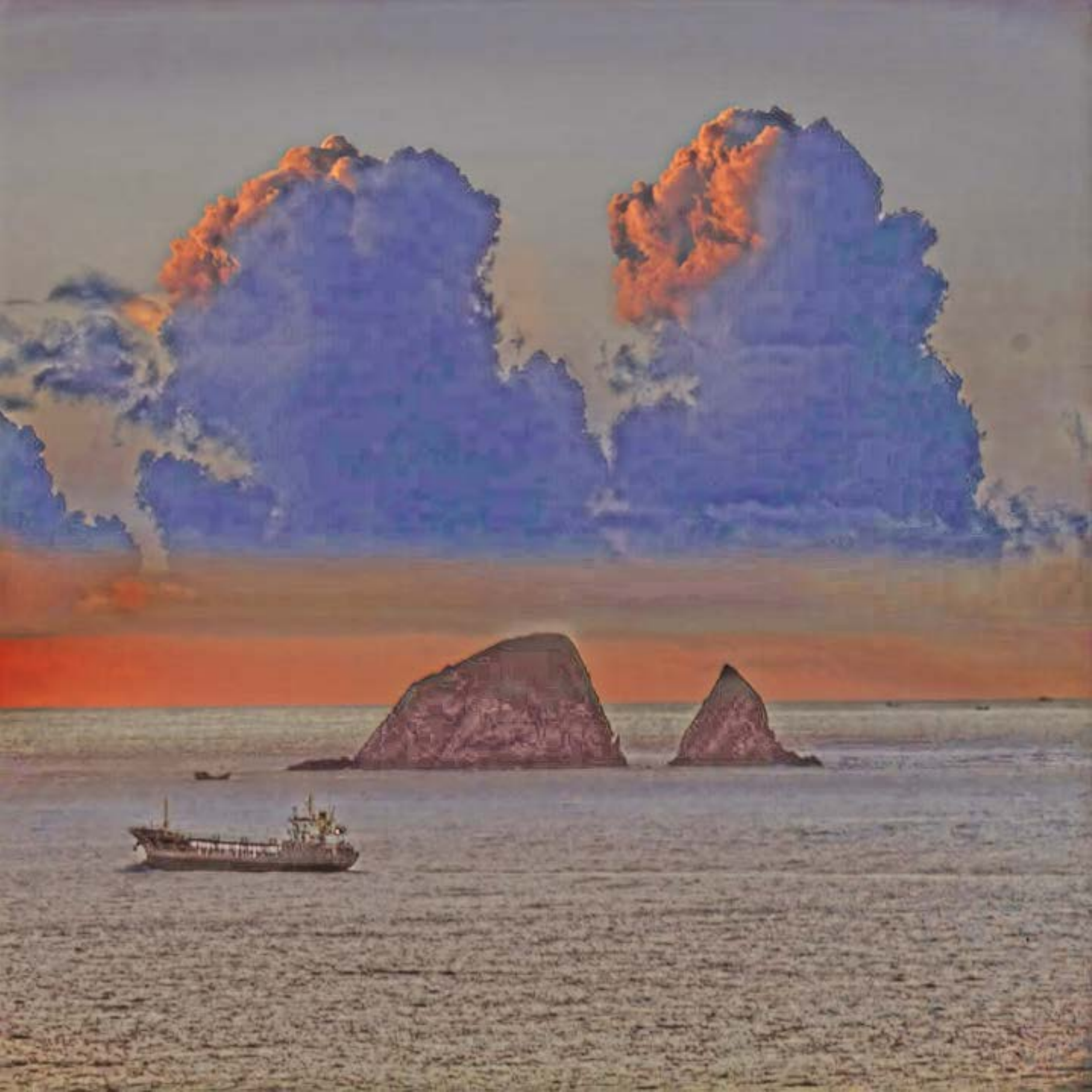}}
	\subfigure[DRBN]{
		\includegraphics[width=0.15\linewidth]{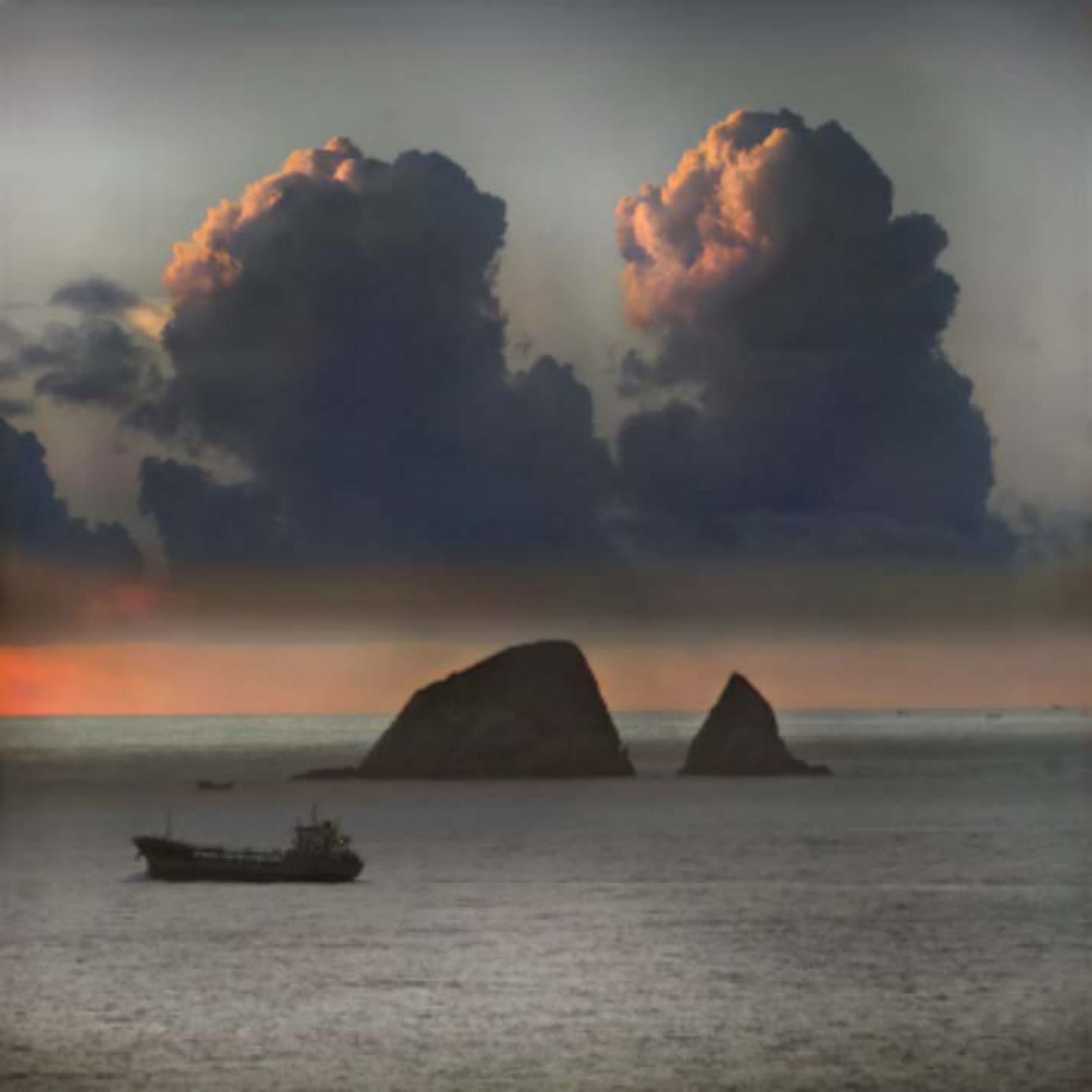}}
	\subfigure[RAUNA]{
		\includegraphics[width=0.15\linewidth]{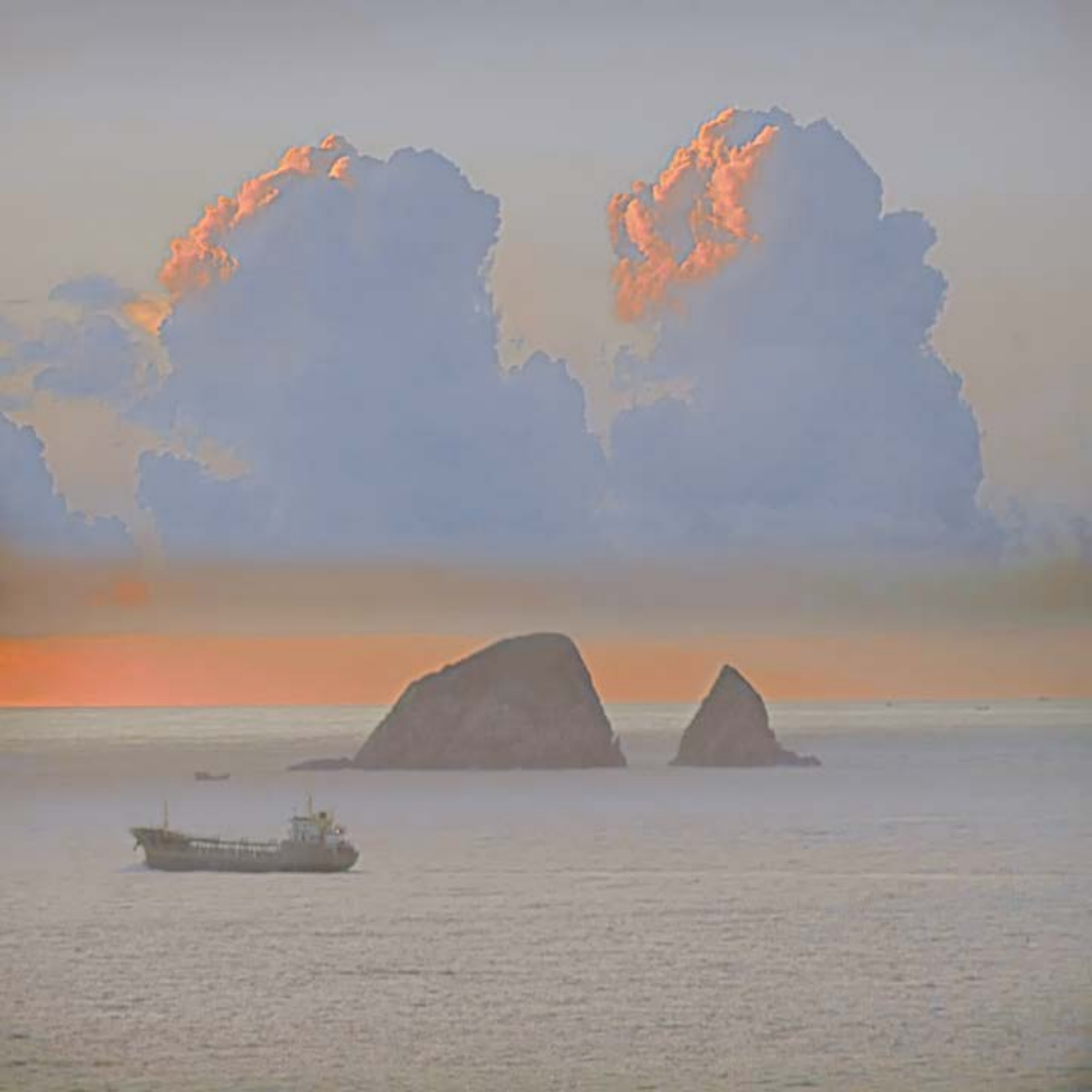}}
	\subfigure[RAUNA$_{\textrm{ft}}$]{
		\includegraphics[width=0.15\linewidth]{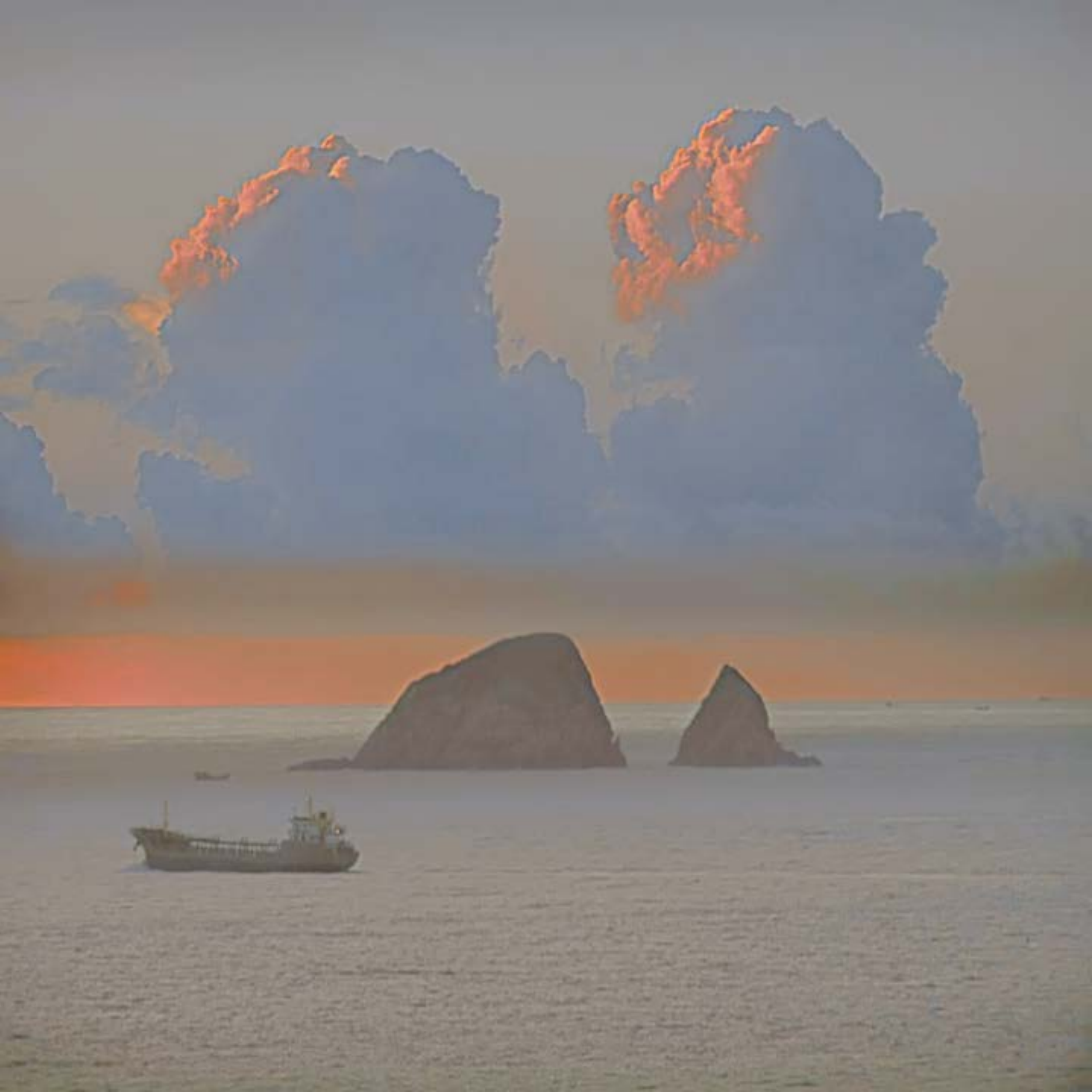}}
	\caption{Visual results of competing methods on NPE dataset.}
	\label{fig:npe} 
\end{figure*}

\begin{figure*}[t]
	\centering
	\subfigure[Input]{
		\includegraphics[width=0.15\linewidth]{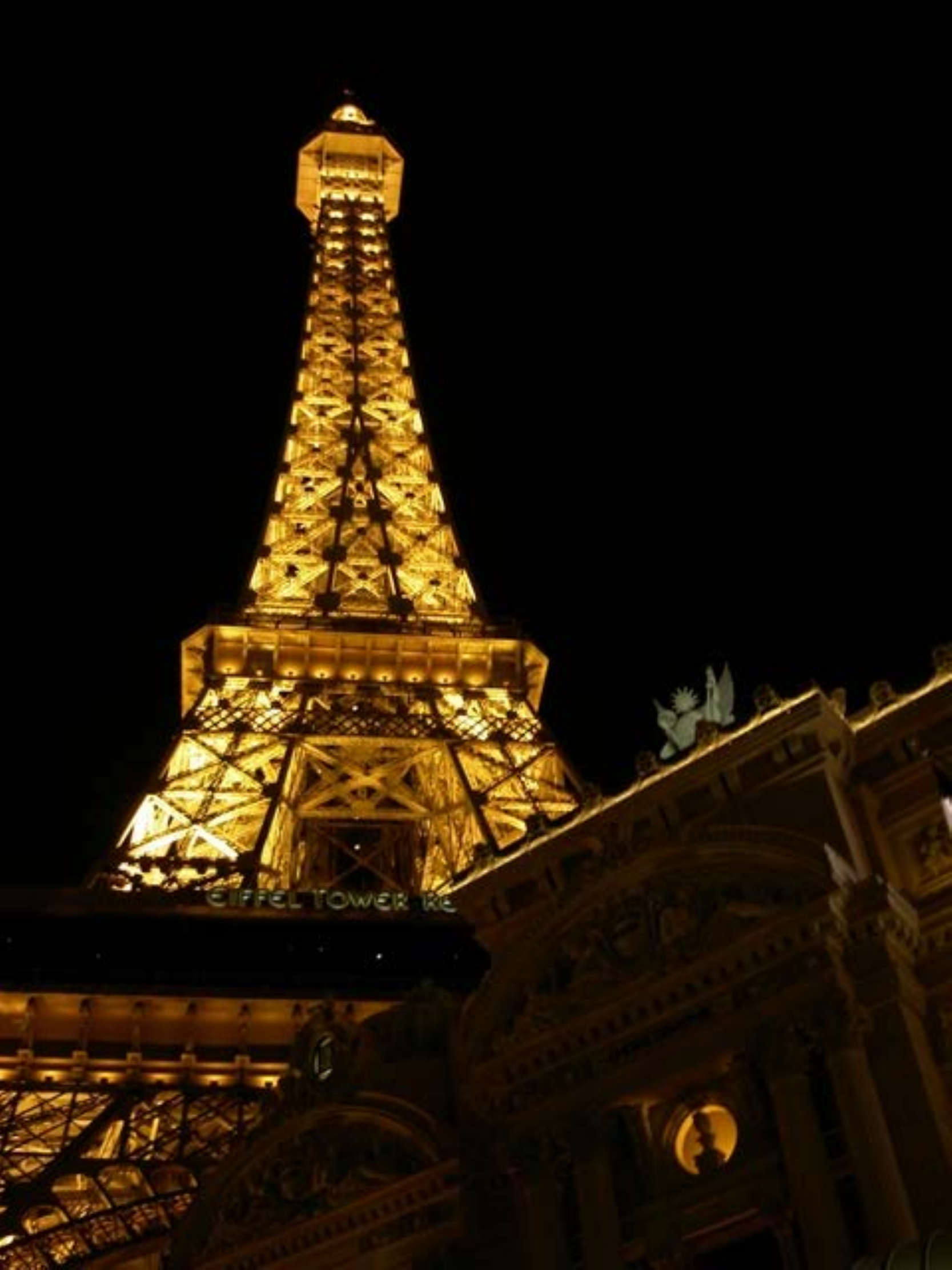}}
	\subfigure[CLAHE]{
		\includegraphics[width=0.15\linewidth]{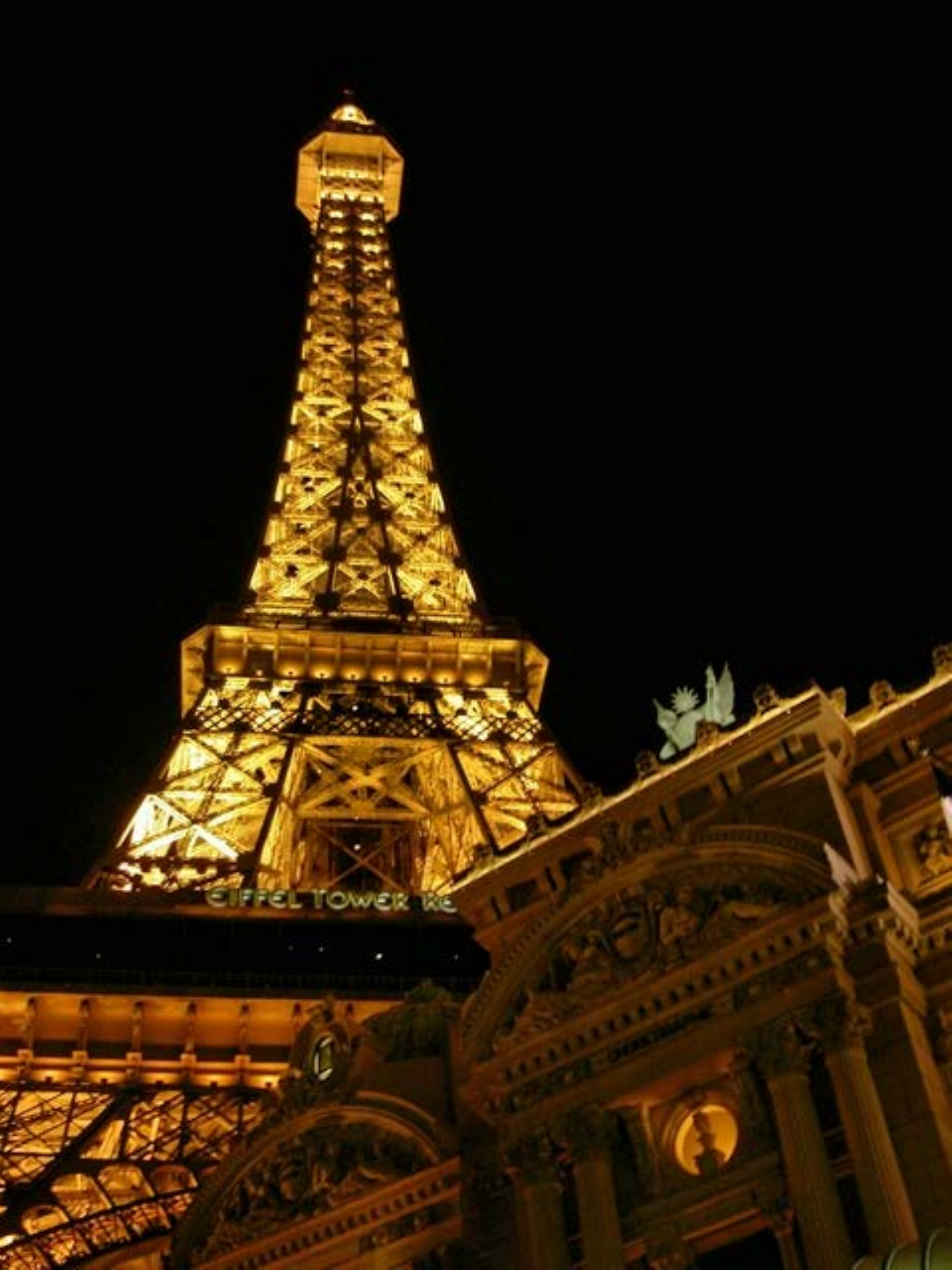}}
	\subfigure[CLAHE+BM3D]{
		\includegraphics[width=0.15\linewidth]{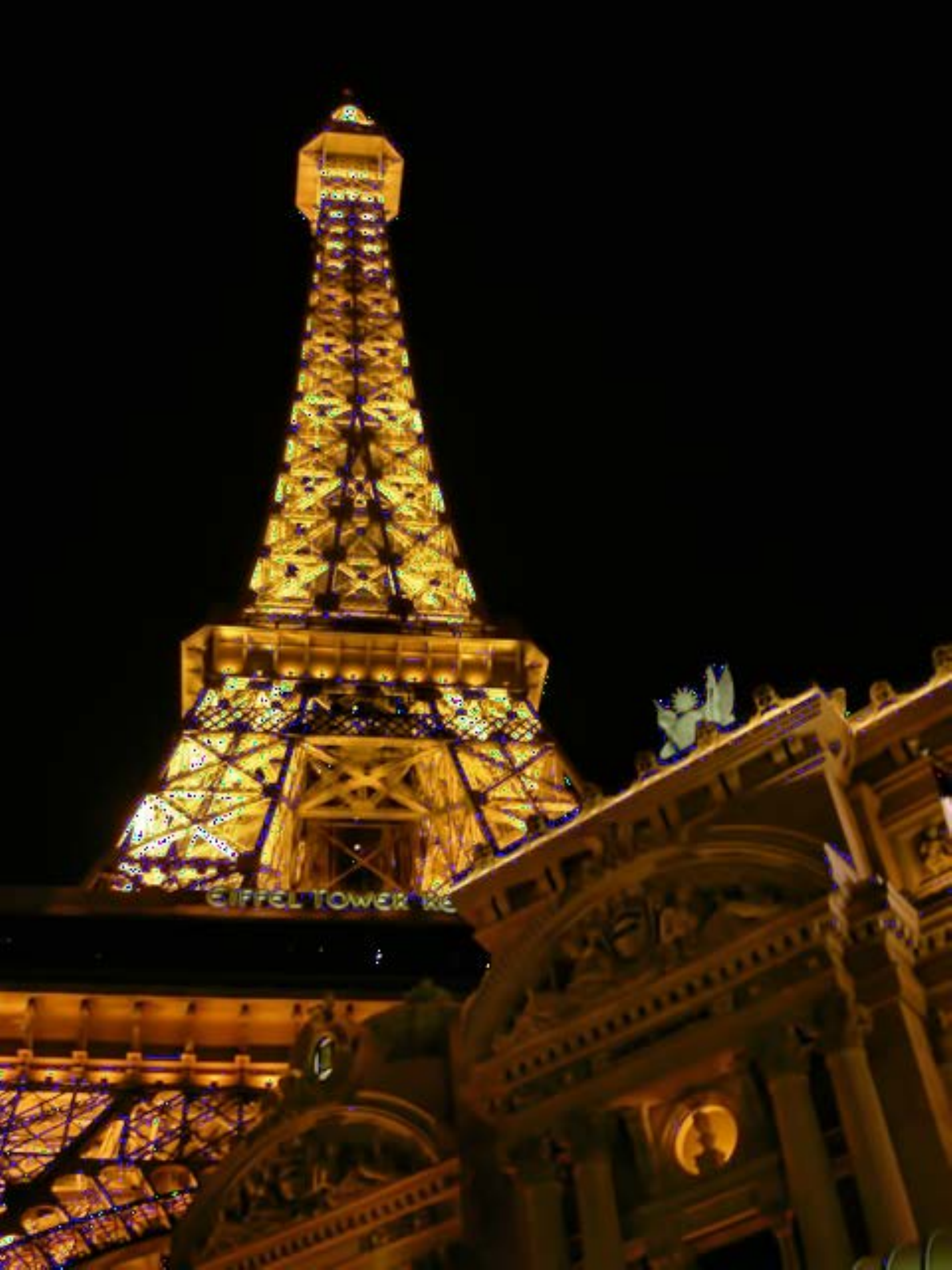}}
	\subfigure[NPE]{
		\includegraphics[width=0.15\linewidth]{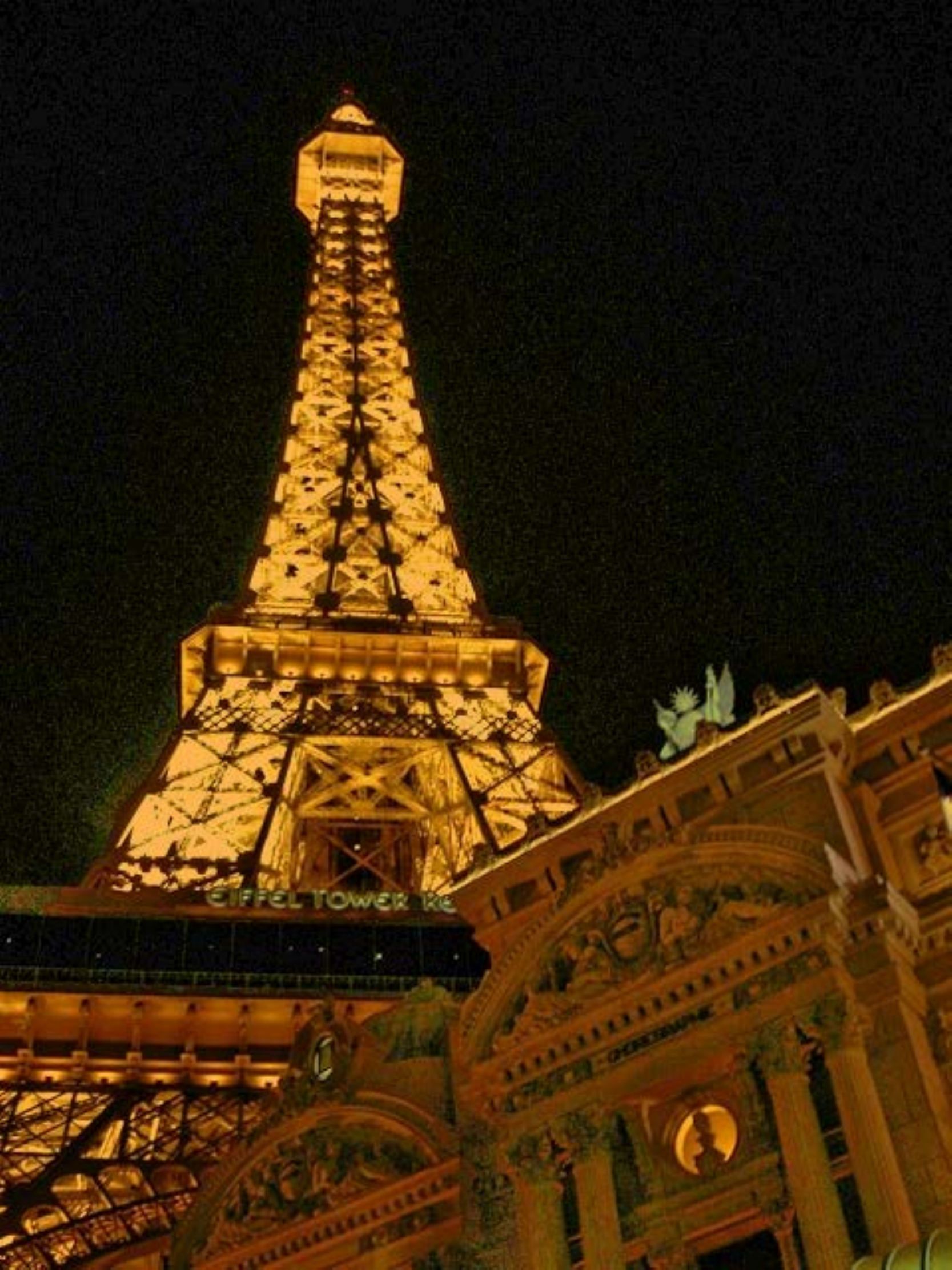}}
	\subfigure[LR3M]{
		\includegraphics[width=0.15\linewidth]{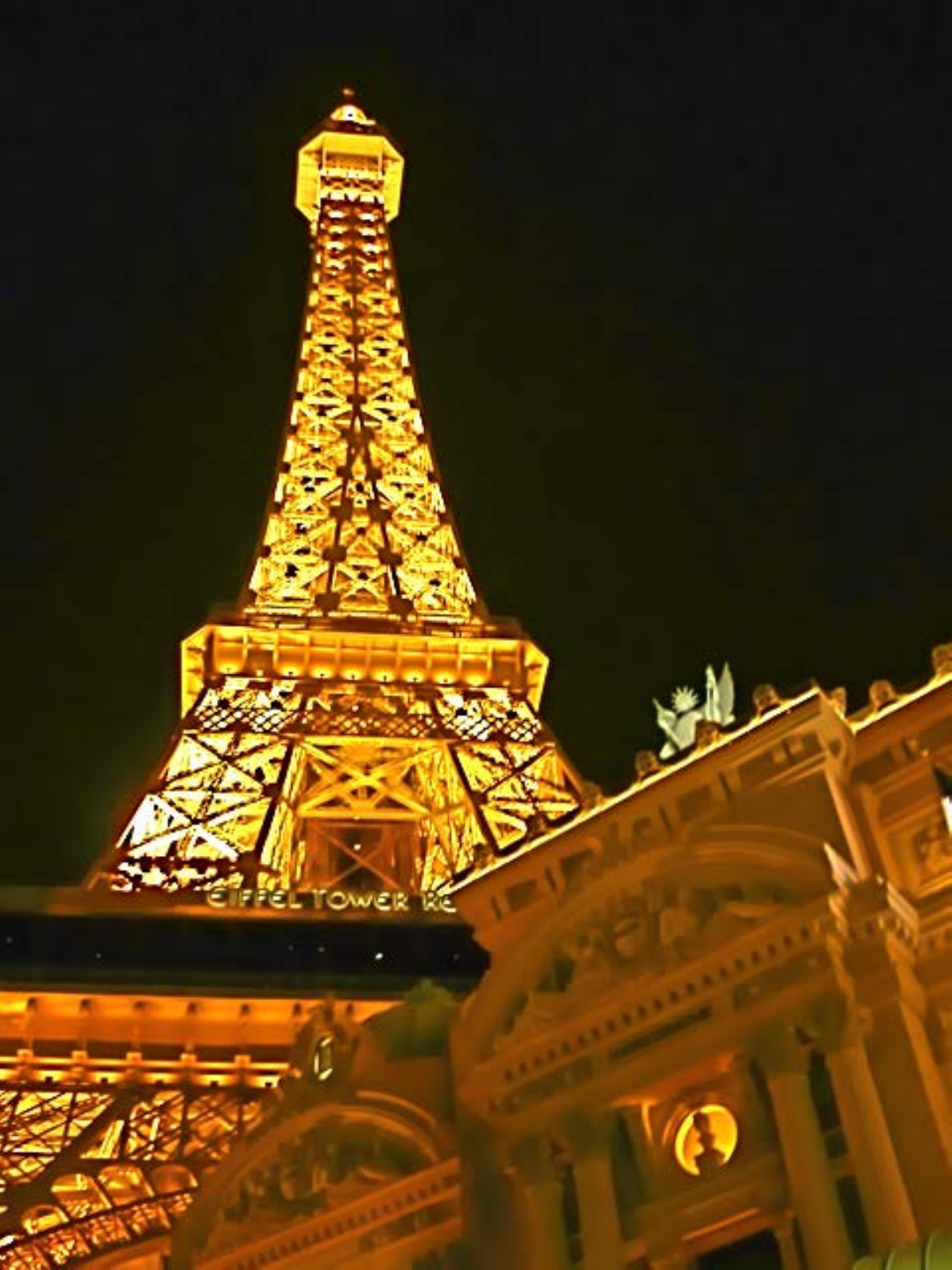}}
	\subfigure[Zero-DCE++]{
		\includegraphics[width=0.15\linewidth]{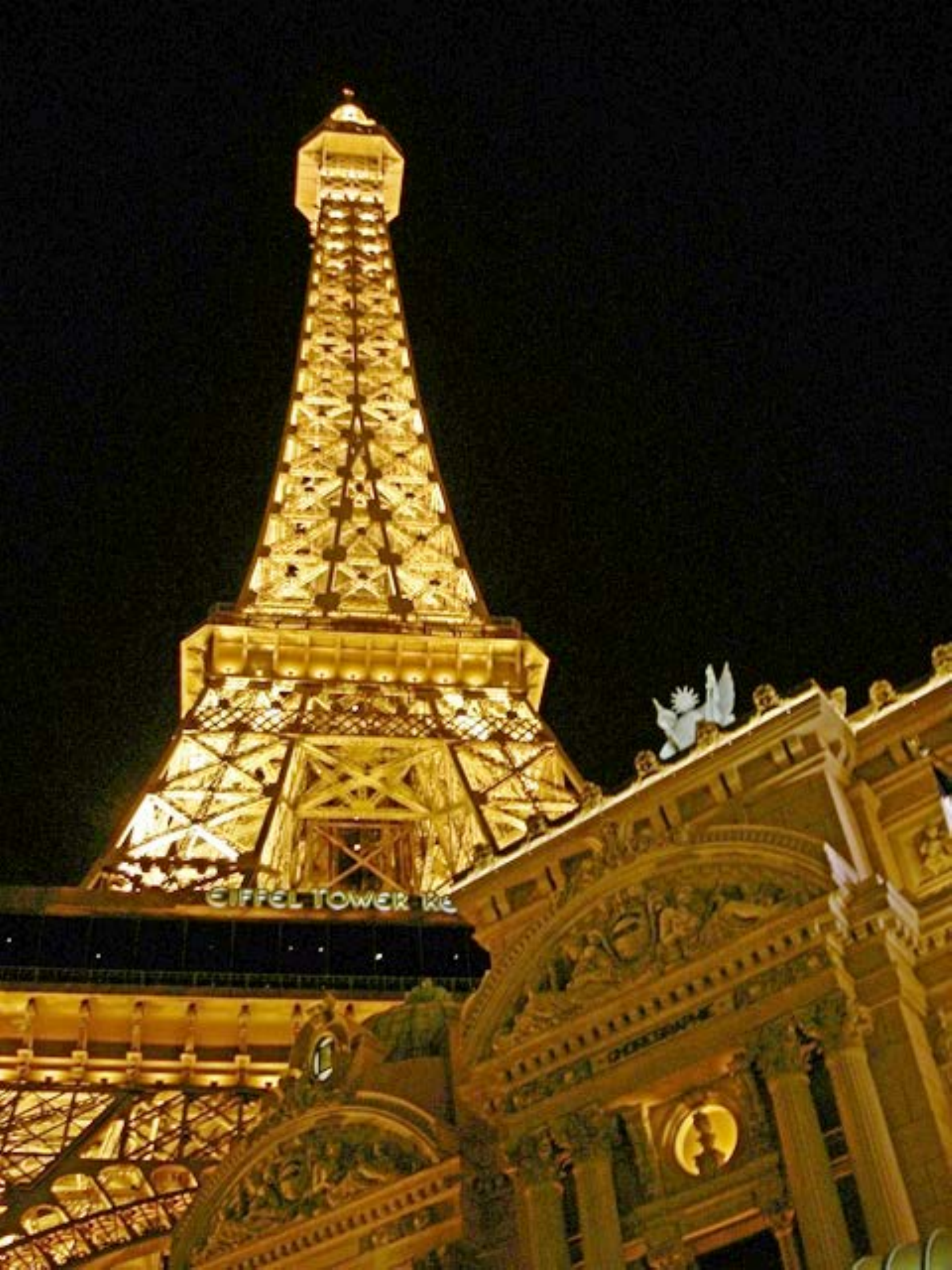}}\\
	\subfigure[CSDNet]{
		\includegraphics[width=0.15\linewidth]{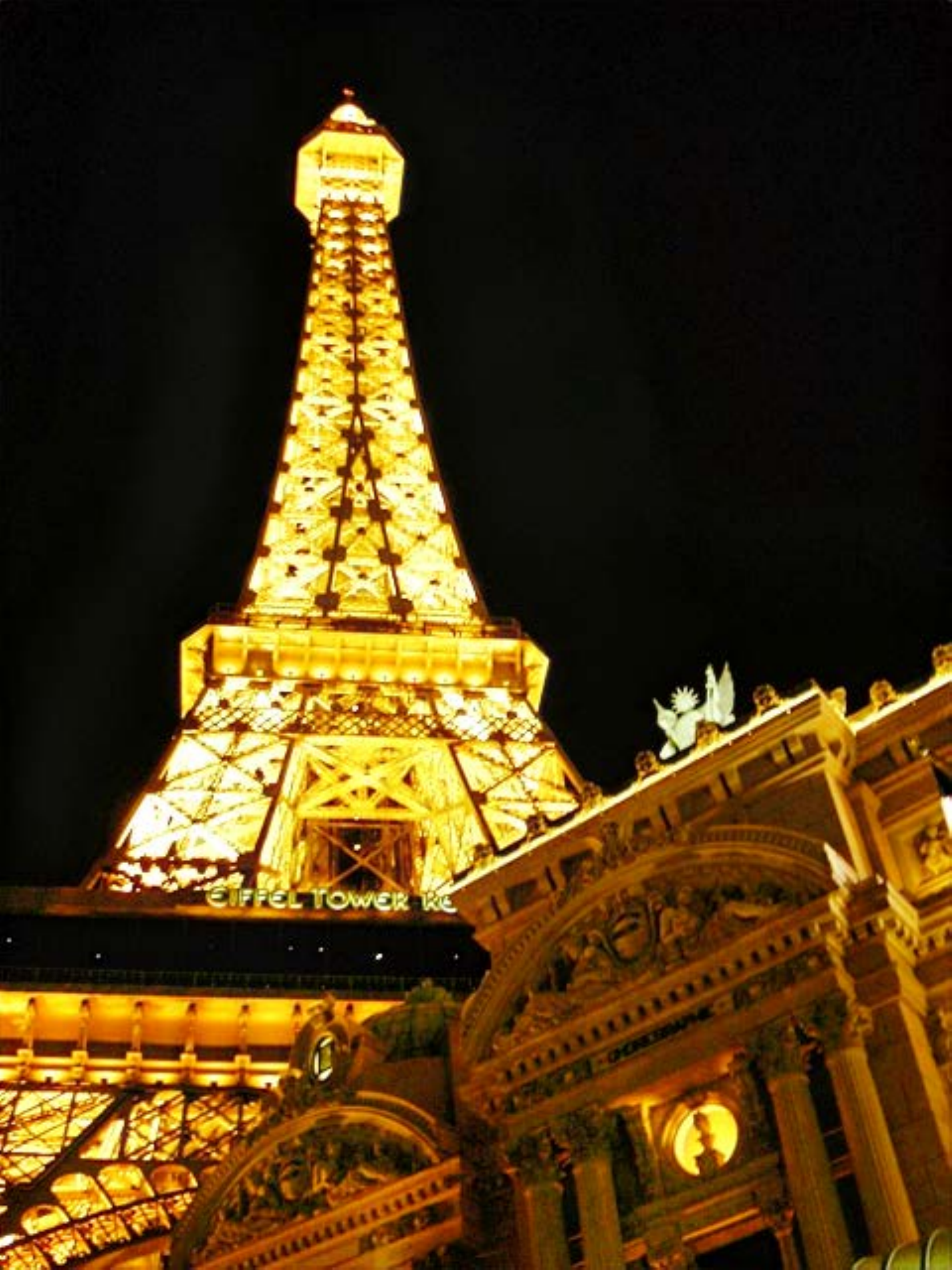}}
	\subfigure[DeepUPE]{
		\includegraphics[width=0.15\linewidth]{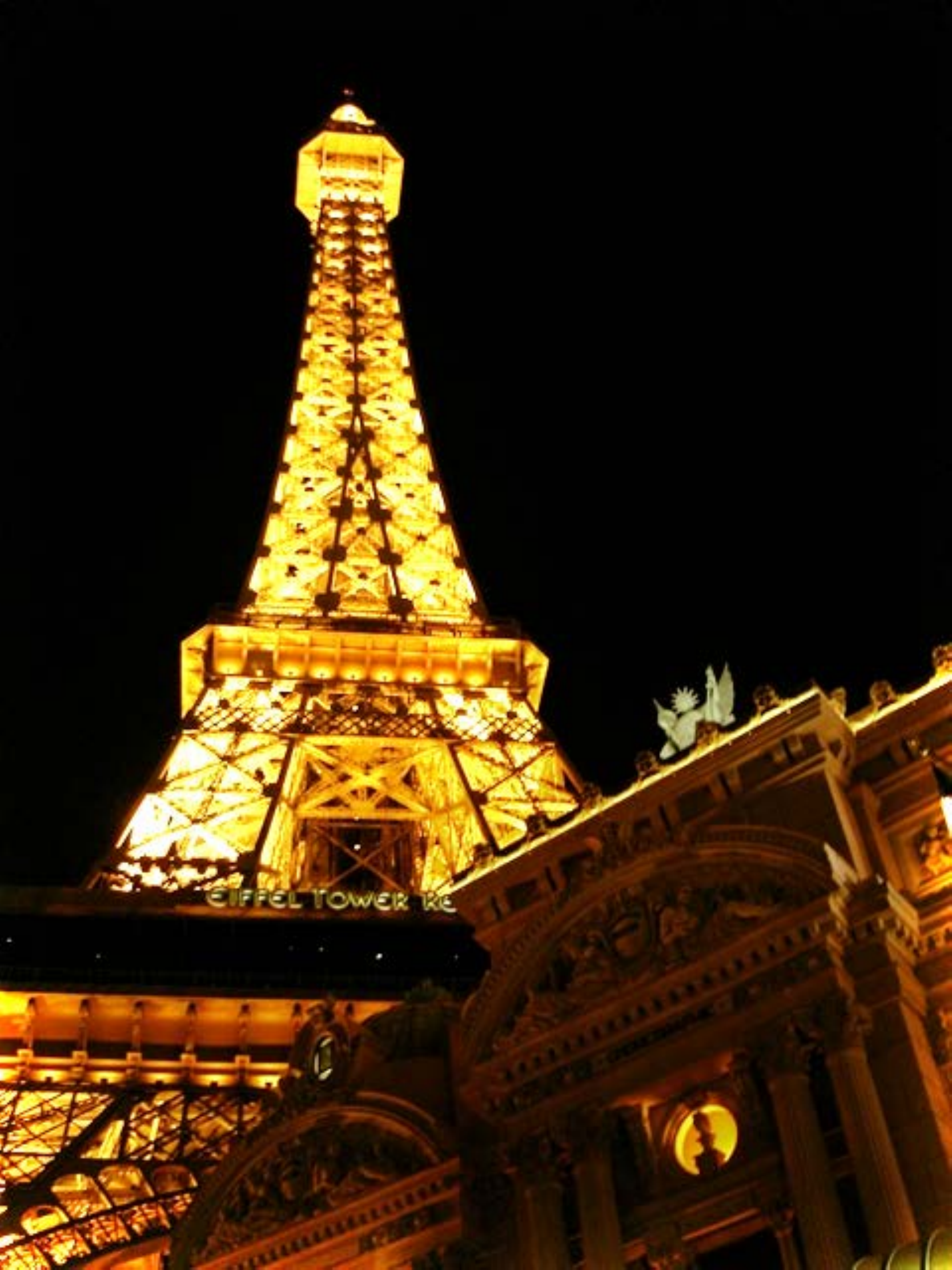}} 
	\subfigure[MBLLEN]{
		\includegraphics[width=0.15\linewidth]{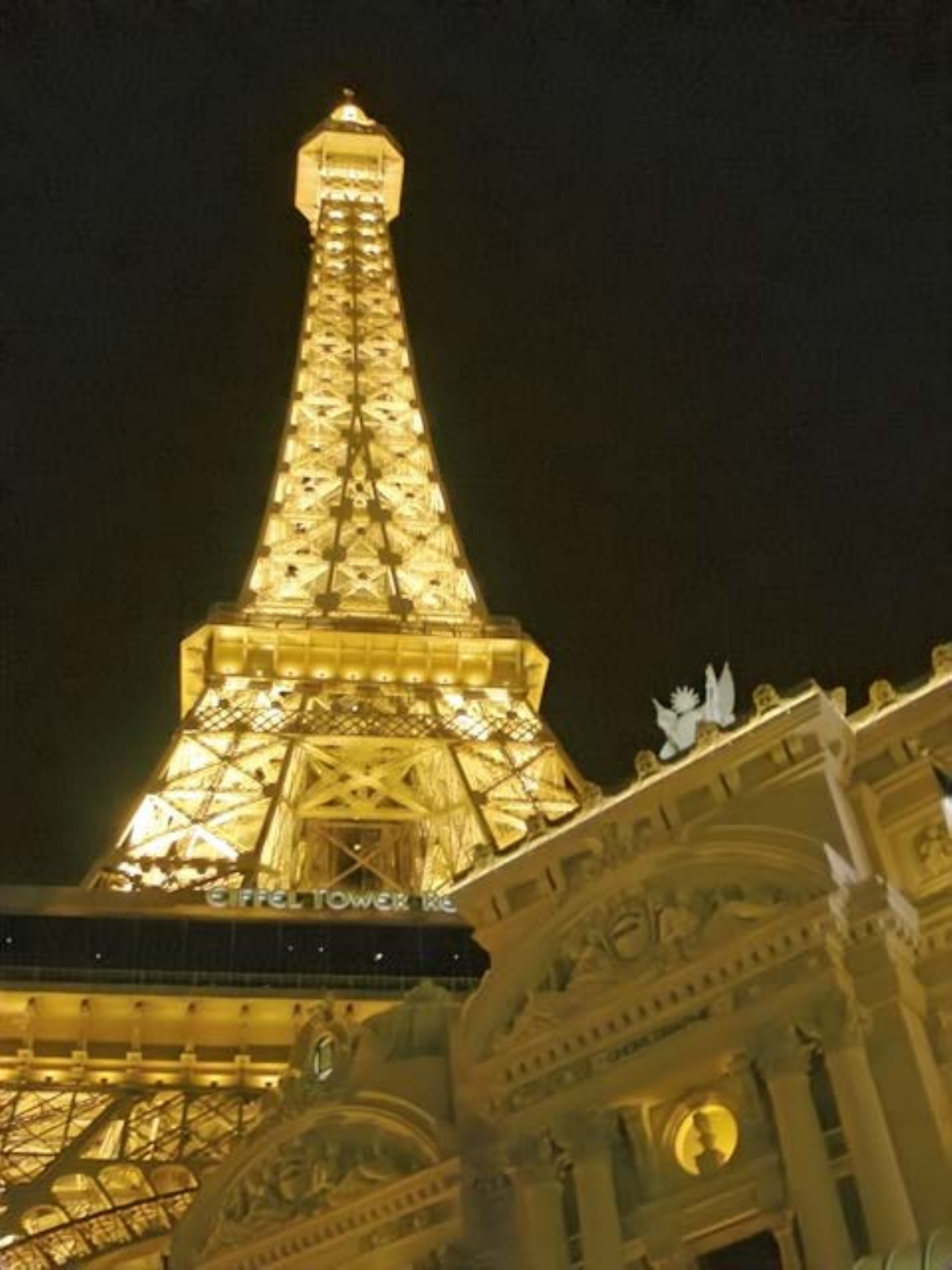}}
	\subfigure[RetinexNet]{
		\includegraphics[width=0.15\linewidth]{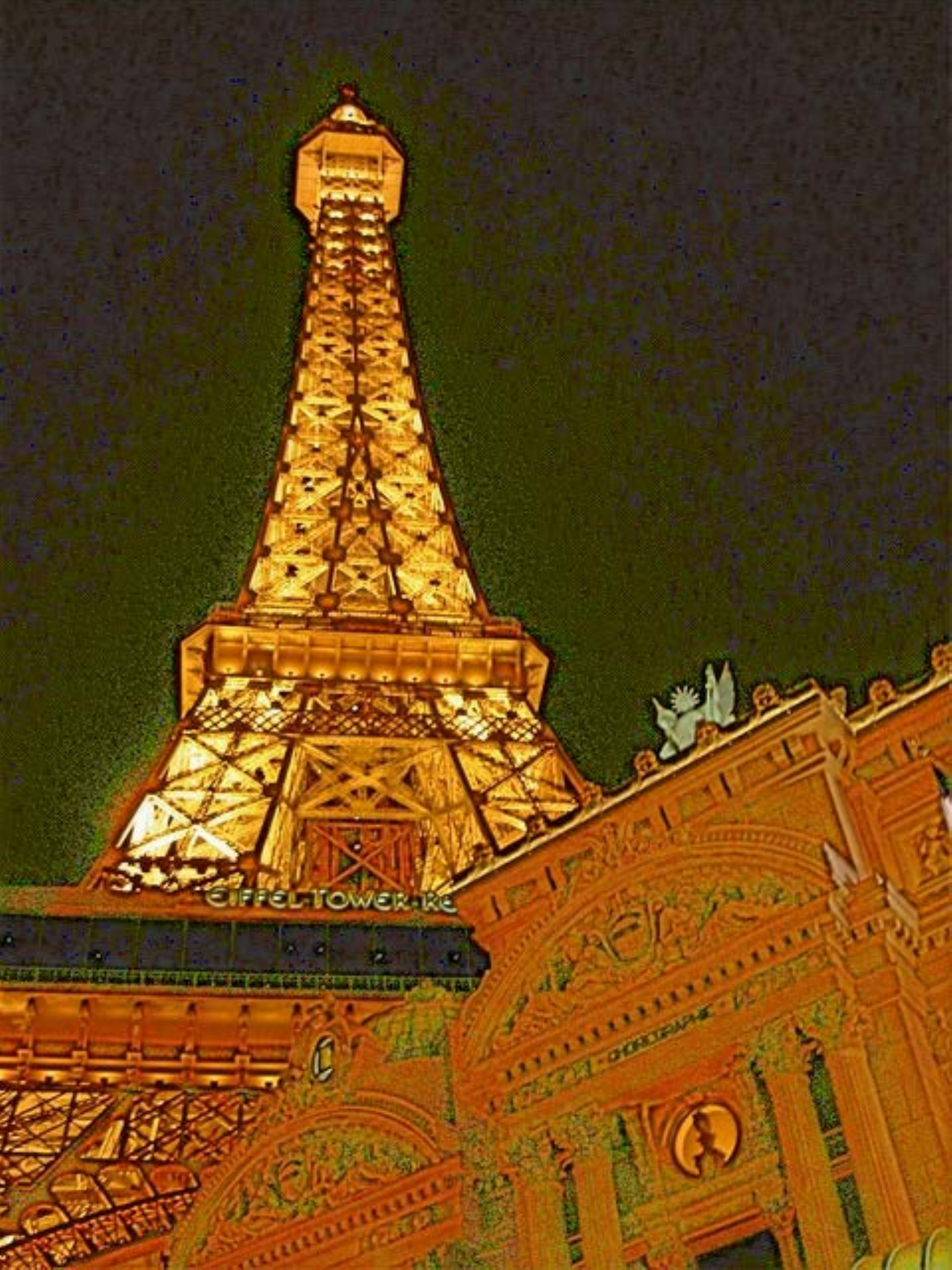}} 
	\subfigure[RUAS]{
		\includegraphics[width=0.15\linewidth]{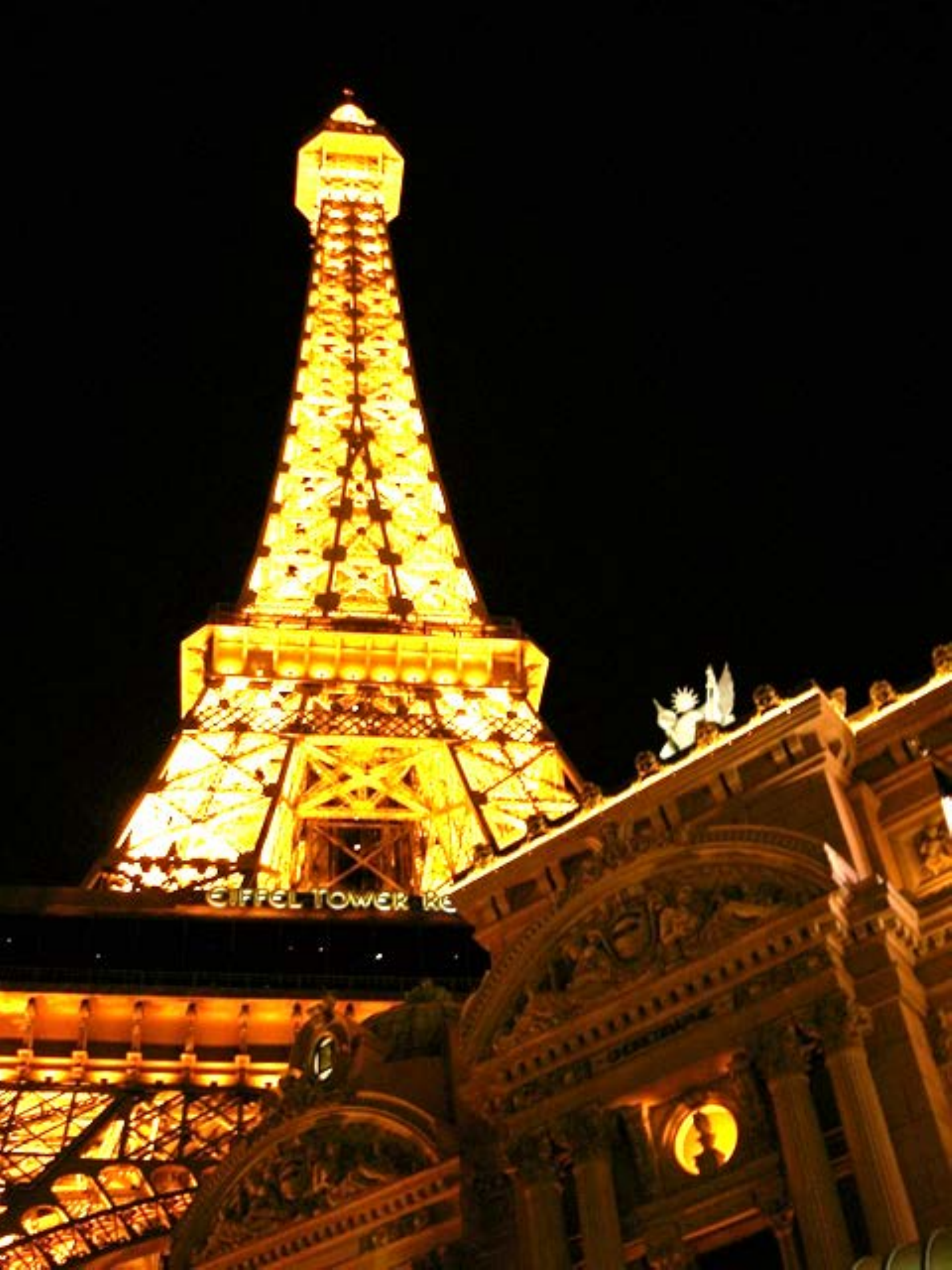}}
	\subfigure[TBEFN]{
		\includegraphics[width=0.15\linewidth]{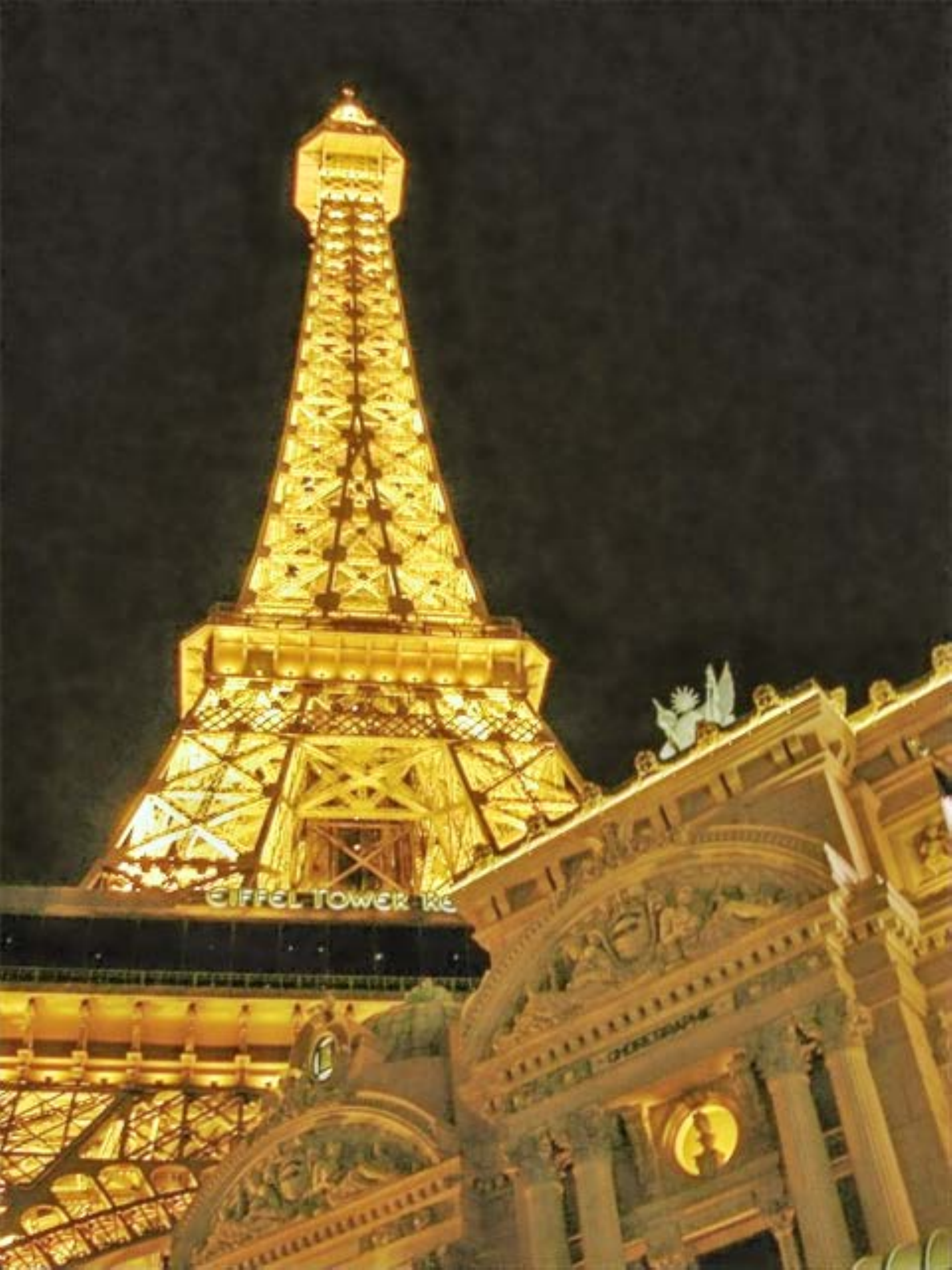}}\\
	\subfigure[SGM]{
		\includegraphics[width=0.15\linewidth]{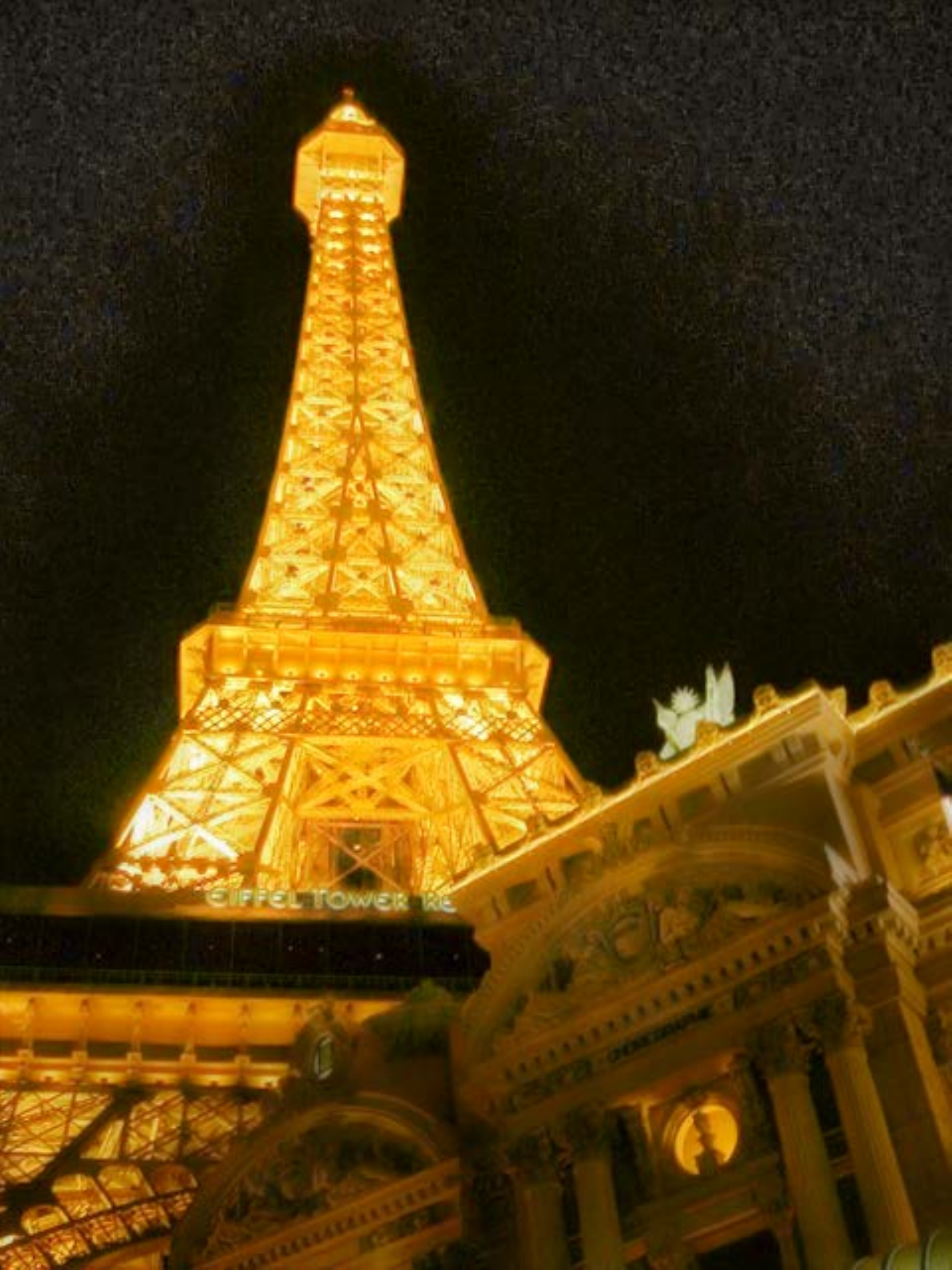}}
	\subfigure[KinD++]{
		\includegraphics[width=0.15\linewidth]{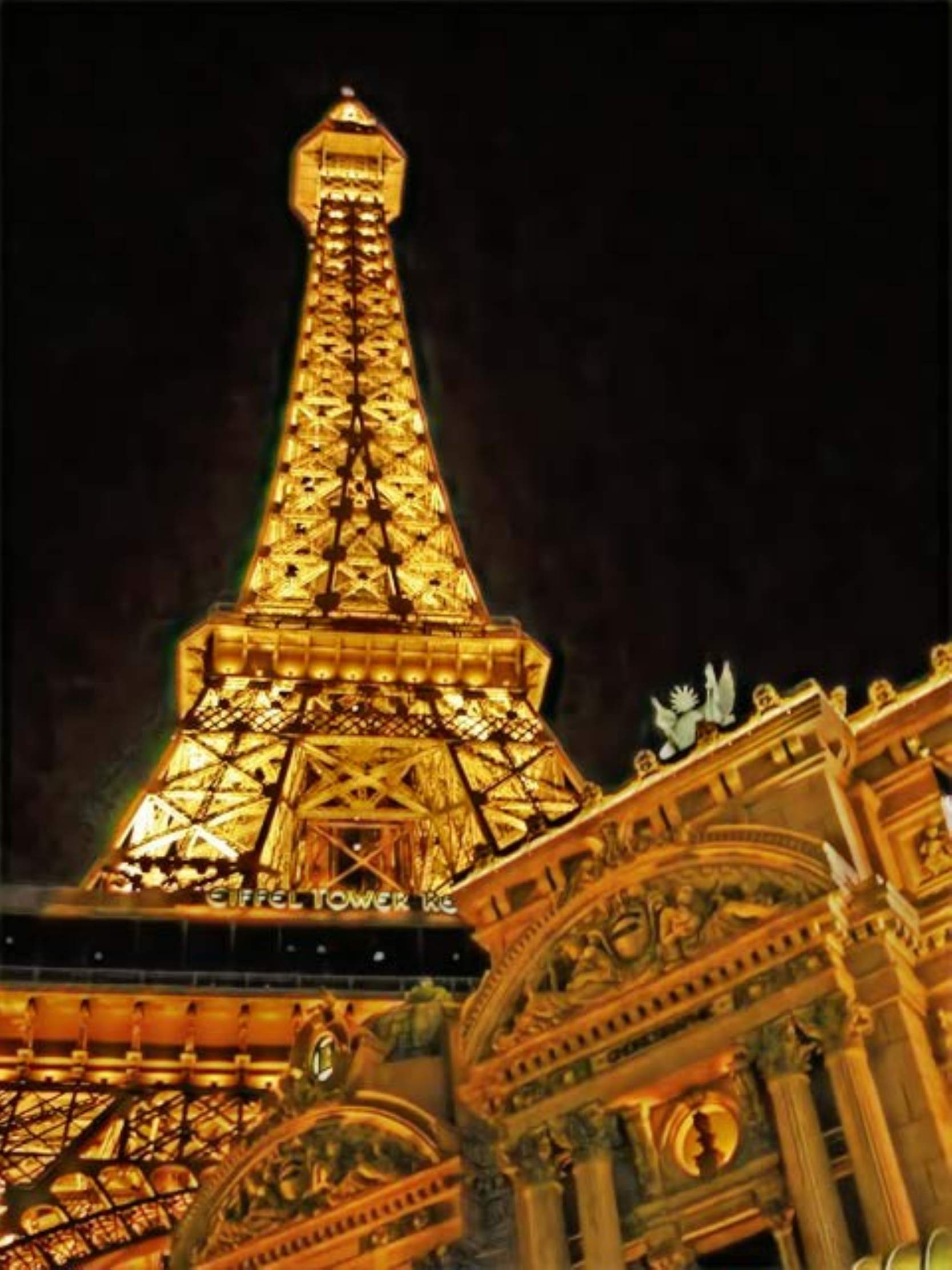}}
	\subfigure[DRBN]{
		\includegraphics[width=0.15\linewidth]{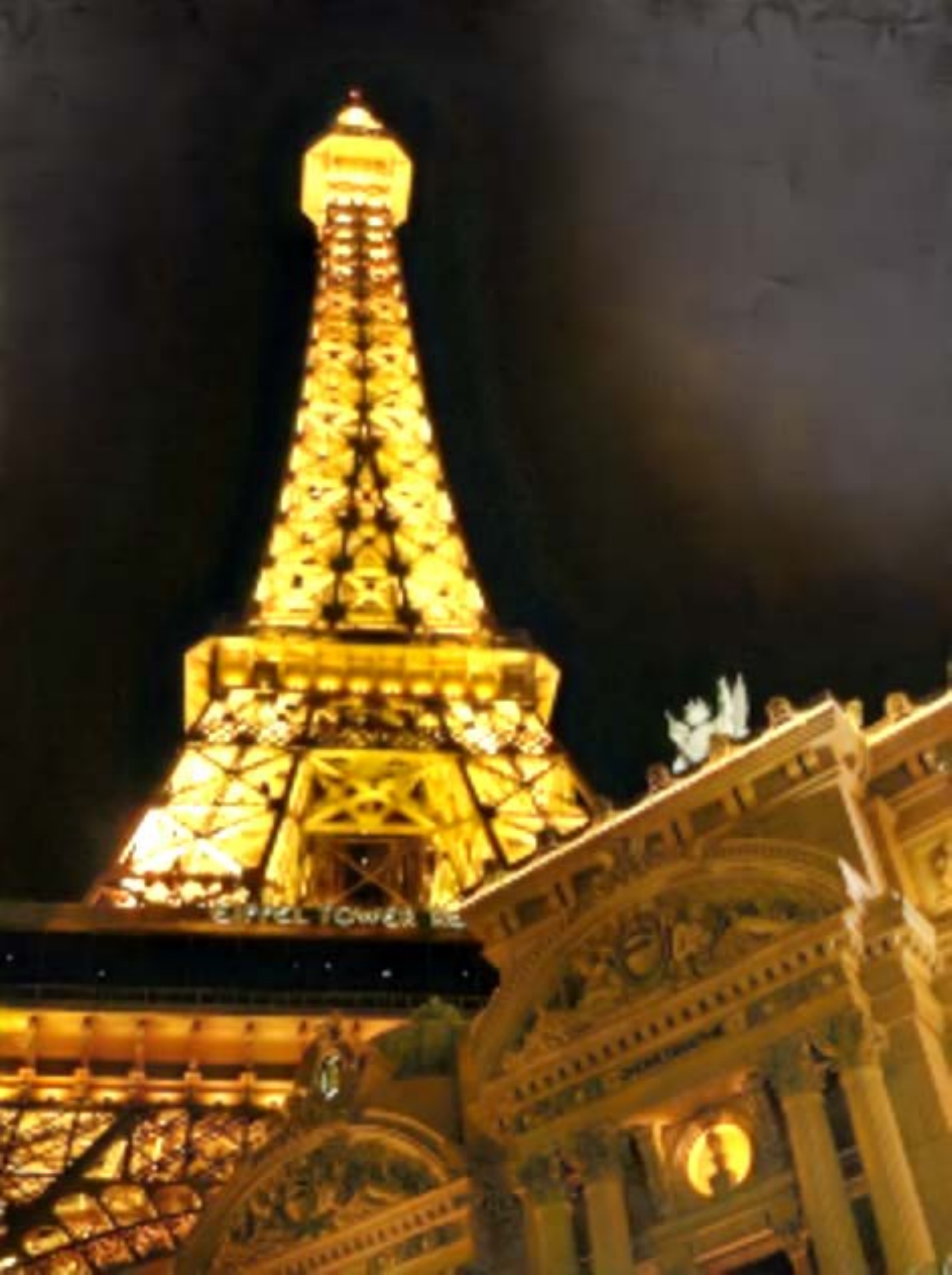}}
	\subfigure[RAUNA]{
		\includegraphics[width=0.15\linewidth]{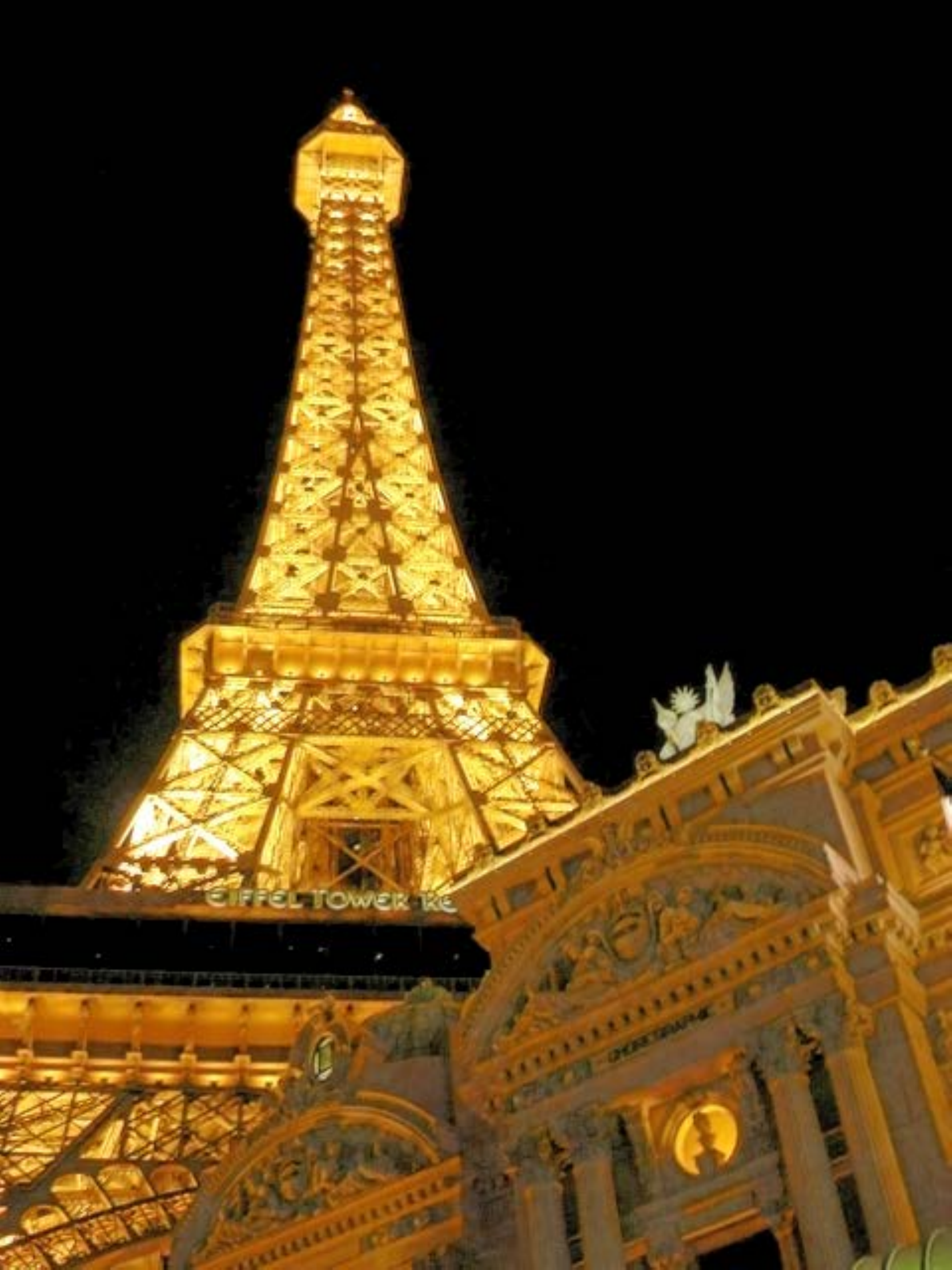}}
	\subfigure[RAUNA$_{\textrm{ft}}$]{
		\includegraphics[width=0.15\linewidth]{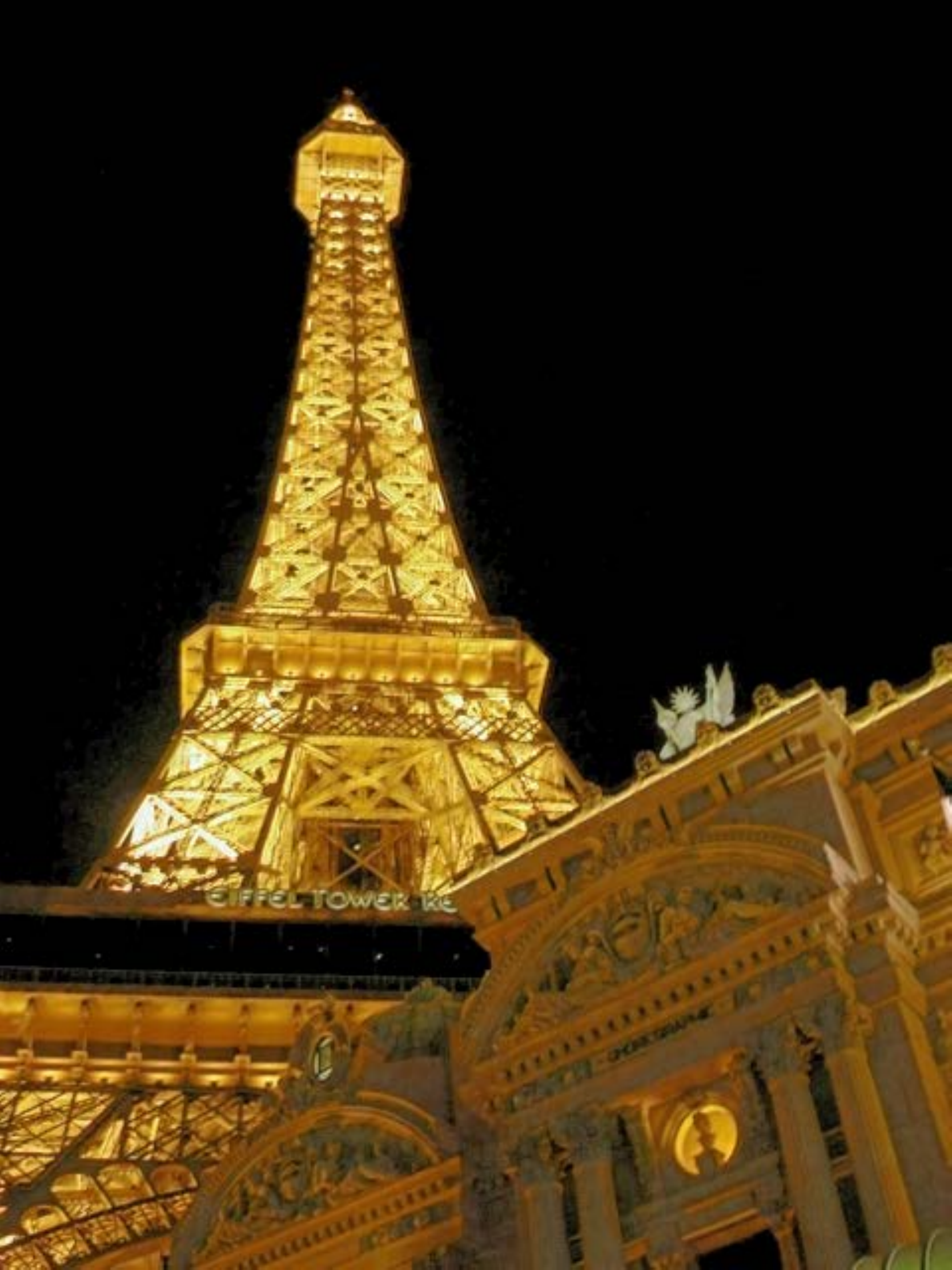}}
	\caption{Visual results of competing methods on DICM dataset.}
	\label{fig:dicm} 
\end{figure*}

\subsubsection{Results on MIT-Abobe FiveK Dataset}
As mentioned before, images in MIT-Abobe FiveK dataset are collected and synthesized without noise, and thus it is relatively easier for a method to perform well. Similar as before, we report the both results with and without GC, except for KinD++ and our methods. The quantitative and visual results of all competing methods are shown in Table \ref{tab:5K} and Fig. \ref{fig:mit1}.
%(more visual results are provided in supplementary material).

As can be seen from Table \ref{tab:5K}, the proposed methods again achieve the best performance in most of the adopted IQA metrics, while perform the second best in the LOE metrics. However, as show in Fig. \ref{fig:mit1}, the visual result of MBLLEN, that achieves the best performance regarding LOEs, is still not very satisfactory, especially for the brightness. Also, the visual results of almost all methods shown in Fig. \ref{fig:mit1} are better than those on LOL dataset, which is due to the fact that the dataset is less noisy as mentioned before. These results further substantiate the effectiveness of our LIE framework.

\subsubsection{Results on Unpaired Datasets}
Since there are no groundtruth normal-light images available on the unpaired NPE and DICM datasets, we adopt the blind IQA metrics, i.e., NIQE and LOE, and summarize the results in Table \ref{tab:un}. As can be seen from the table, our methods have the best performance with respect to the LIE-oriented LOE metric, showing its promising ability in this task. As for the NIQE metric, which is for general purpose IQA, though not the best, our methods still perform very competitively.

The better performance of our methods can be more directly observed from the visual results shown in Figs. \ref{fig:npe} and \ref{fig:dicm}. Specifically, our methods can not only better adjust the global brightness, but also successfully deal with the non-homogeneous light conditions. For example, the input image in Fig. \ref{fig:dicm} has a bright region in the tower, while a dark region in the building, and most of the competing methods either fail to brighten the darker region, or over-enhance the brighter region. In comparison, our methods can properly tune the brightness of the darker region, while avoid overexposure in the brighter region. Besides, the results of our methods are with less color bias, compared with another two competitive methods, Zero-DCE++ and KinD++.

\subsubsection{Model and Inference Complexity}
\begin{figure}[!t]
	\centering
	\includegraphics[width=0.49\textwidth]{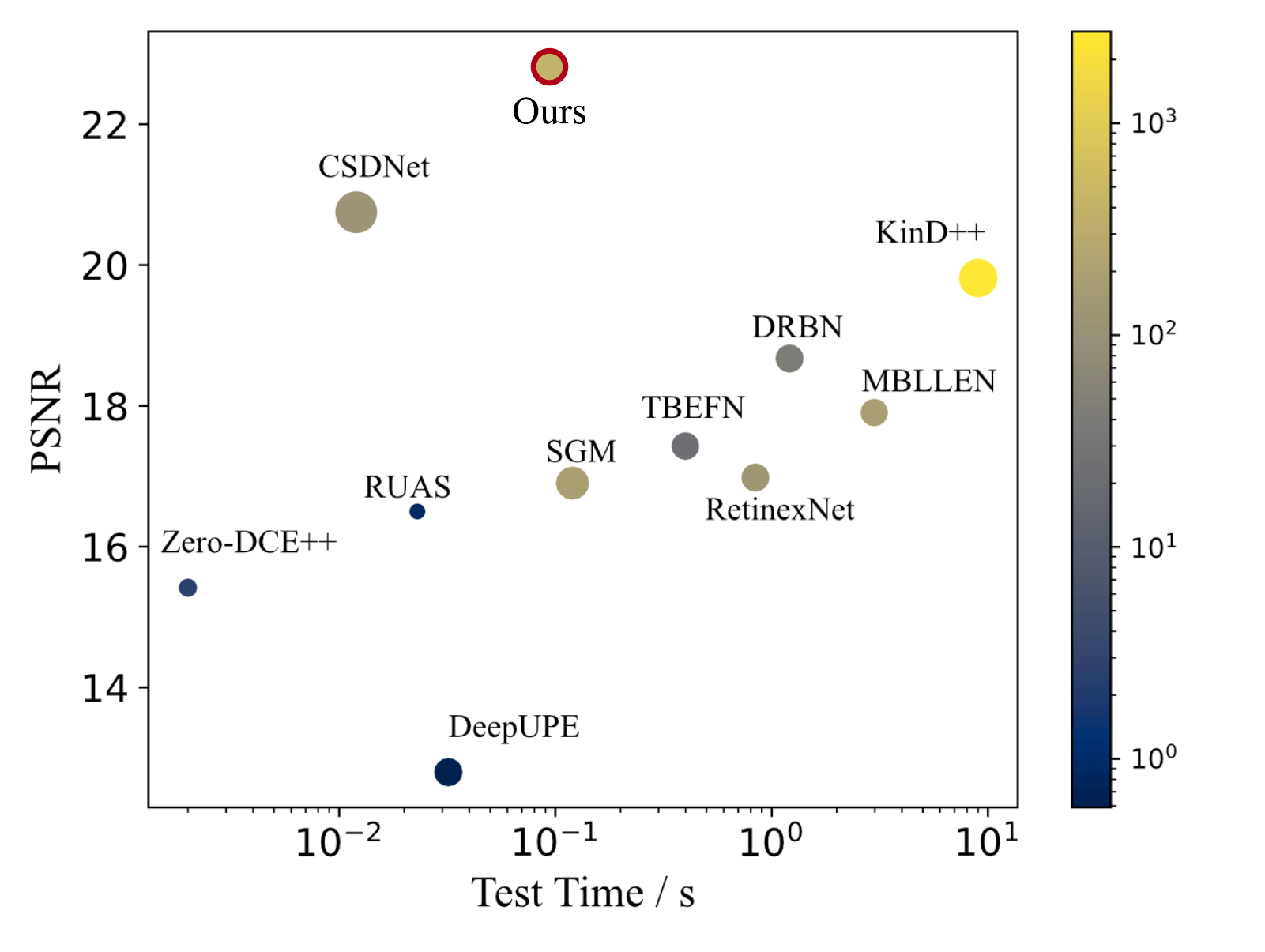}
	\caption{Visualization of model and inference complexities of deep learning based methods. The horizontal axis refers to number of parameters. The vertical axis refers to PSNR value. The size of the circles refers to the test time, that the smaller circle indicates a shorter time. The color refers to value of FLOPs. }
	\label{fig:m_com} 
\end{figure}

For deep learning methods, there is always a trade-off between performance and complexity (model and inference), which is important for practical applications. Therefore, we also compare the model and inference complexities of deep learning based methods, including ours. Specifically, we first summarize the statistics regarding complexity on LOL dataset in Table \ref{tab:m_comp}, and these statistics show that, though not the best, both the model and inference complexities of our method are not too high. Then we plot in Fig. \ref{fig:m_com} the performance with respect to PSNR on LOL dataset, versus the statistics from Table \ref{tab:m_comp}. It can be seen that our method performs the best with relatively low inference time, and the model size is also not too large, which indicates that our method achieves a promising trade-off between performance and complexity.

\subsection{Ablation Study}\label{ablation}
In this Section, we conduct experiments to verify the necessity of each component in our framework, including network architectures, loss functions and data augmentation. The overall results on LOL dataset are summarized in Table \ref{tab:im} and visualized in Fig. \ref{fig:ab}. The details are provided in the following.

\begin{table*}[!t]
	\centering
	\caption{Comparison of model and inference complexities of deep learning based LIE methods.}\label{tab:m_comp}
	\begin{tabular}{l|r|r|r|l}
		\hline
		Method & Test Time/s & Size/M & FLOPs/G & Platform  \\ \hline	
		Zero-DCE++ \citep{Zero-DCE++} & 0.002  & 0.011  & 2.535  & PyTorch  \\	
		CSDNet \citep{ma2021learning} & 0.012 & 17.272  & 121.888  & PyTorch  \\
		DeepUPE \citep{wang2019underexposed} & 0.032 & 0.594  & 0.059  & PyTorch  \\ 
		MBLLEN \citep{Lv2018MBLLEN} & 2.981  & 0.450  & 192.923  & TensorFlow  \\ 
		RetinexNet \citep{Chen2018Retinex} & 0.841  & 0.555  & 135.997  & TensorFlow  \\ 
		RUAS \citep{liu2021retinex} & 0.023  & 0.003  & 0.870  & PyTorch  \\ 
		TBEFN \citep{lu2020tbefn} & 0.399  & 0.145  & 24.112  & TensorFlow  \\ 
		SGM \citep{yang2021sparse} & 0.120  & 2.308  & 193.103  & PyTorch  \\ 		 
		DRBN \citep{yang2020fidelity} & 1.210  & 0.577  & 42.865  & PyTorch  \\ 
		KinD++ \citep{zhang2021beyond} & 9.013  & 8.275  & 2710.247  & TensorFlow  \\ 
		RAUNA & 0.095  & 1.850  & 413.303  & PyTorch \\ \hline
	\end{tabular}
\end{table*}

\subsubsection{Ablation on Model Components}\label{sec:ab-model}
\begin{figure*}[t]
	\centering
	\subfigure[w/o Unrolling]{
		\includegraphics[width=0.15\linewidth]{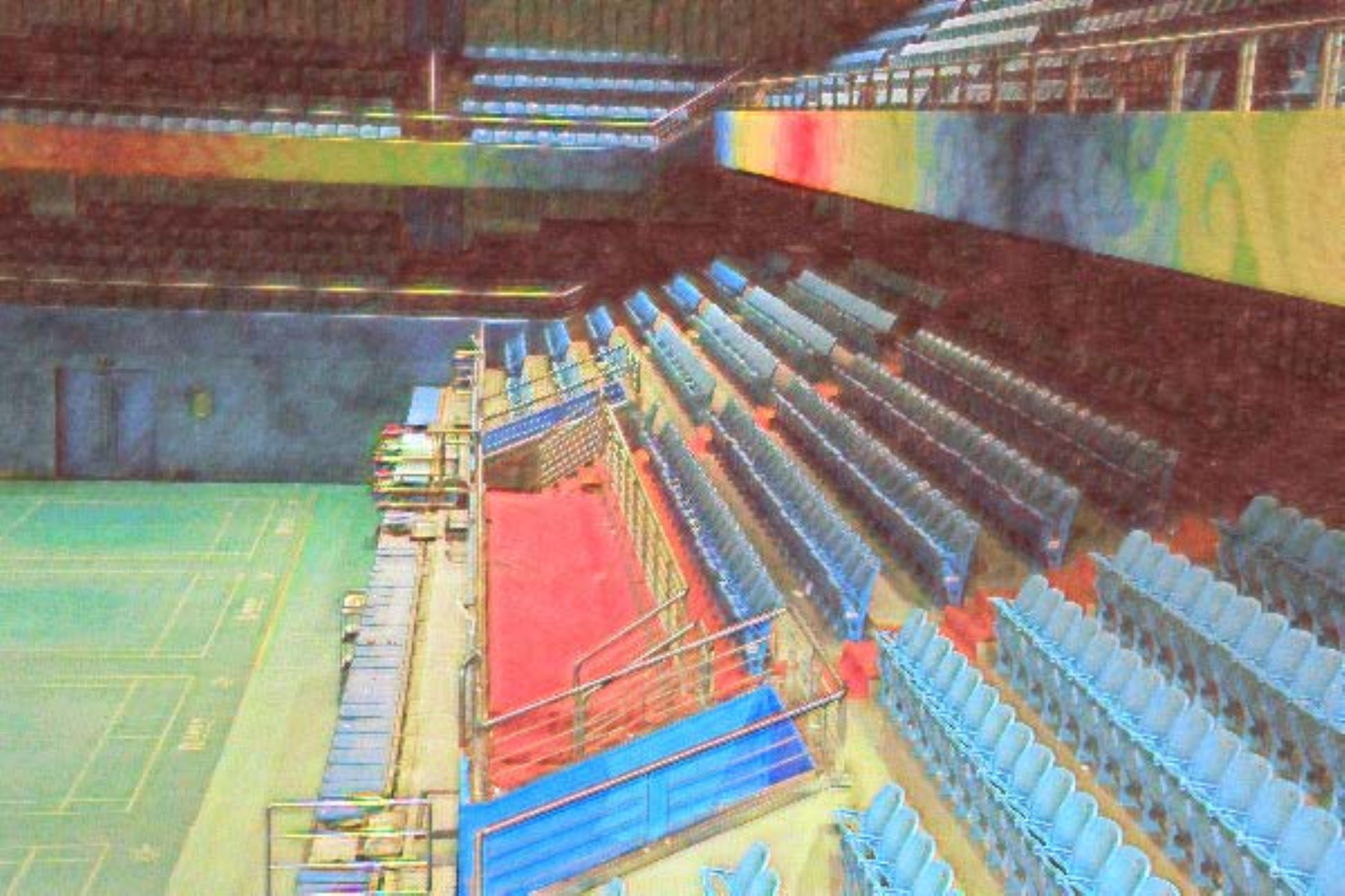}\label{ab-unroll}}
	\subfigure[w/o Exp. Prior]{
		\includegraphics[width=0.15\linewidth]{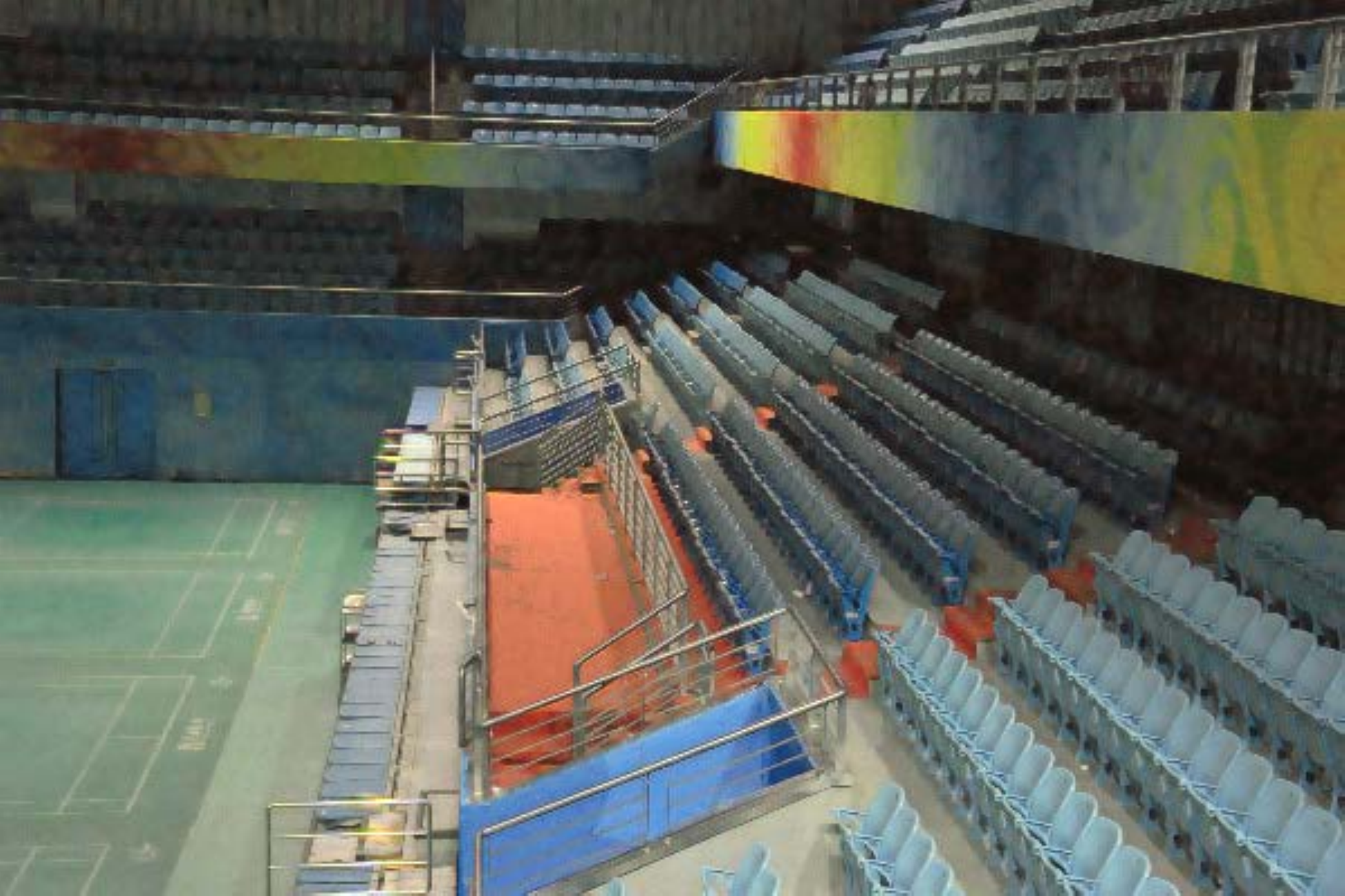}\label{ab-prior}}  
	\subfigure[MSE]{
		\includegraphics[width=0.15\linewidth]{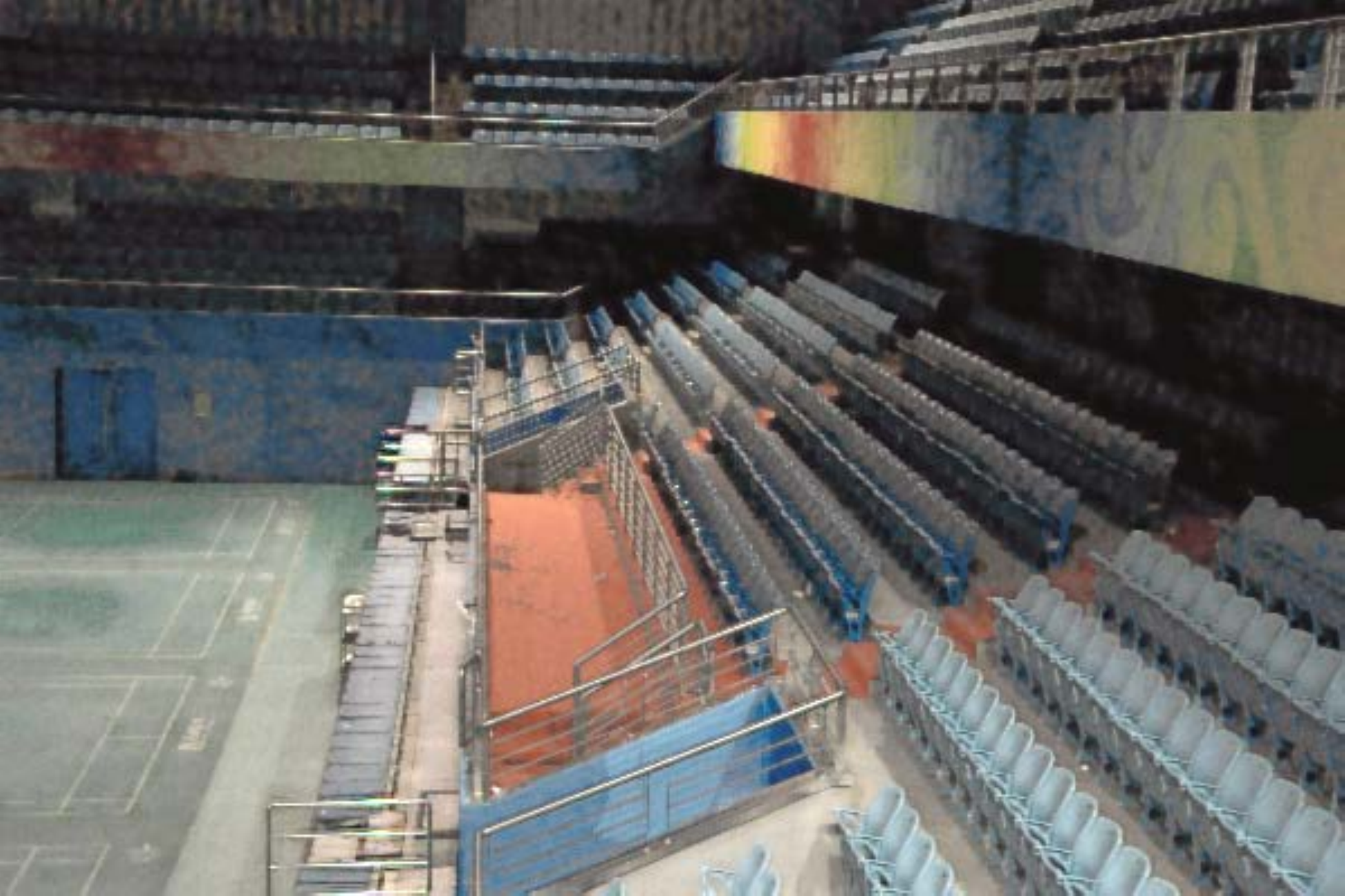}\label{ab-loss1}}   
	\subfigure[MSE+Col]{
		\includegraphics[width=0.15\linewidth]{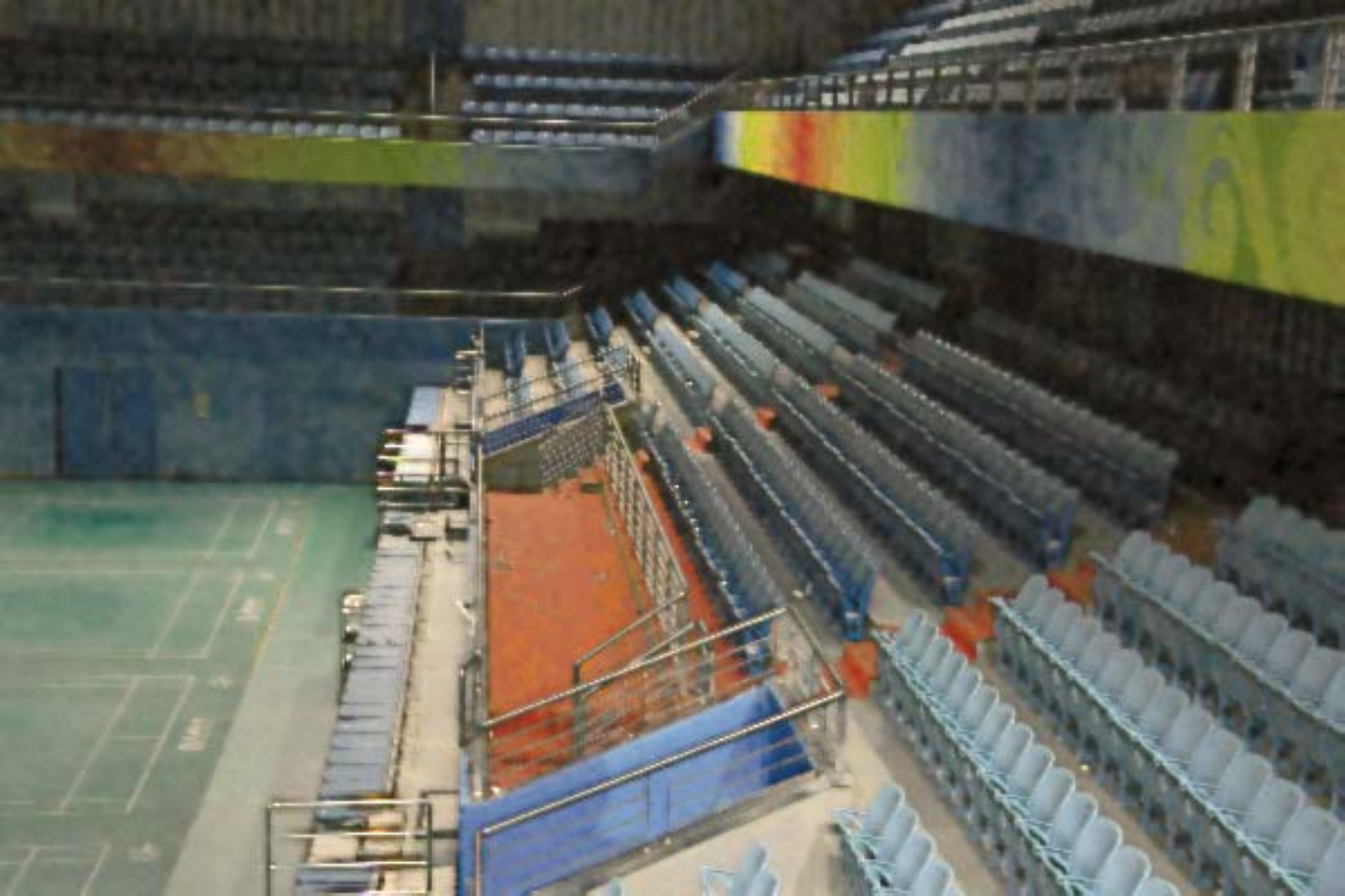}\label{ab-loss2}} 
	\subfigure[w/o Augment.]{
		\includegraphics[width=0.15\linewidth]{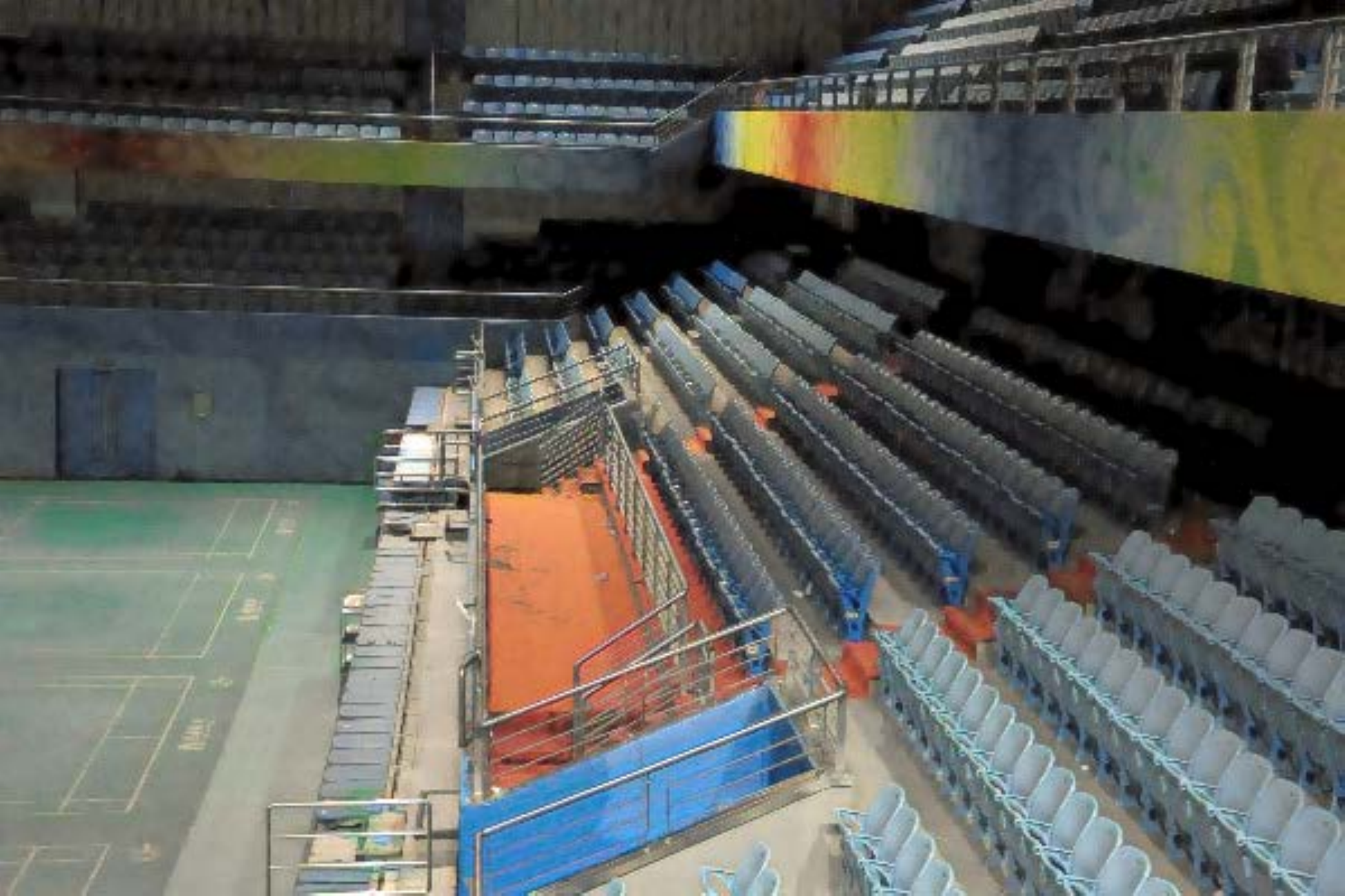}\label{ab-augment}}
	\subfigure[Ours]{
		\includegraphics[width=0.15\linewidth]{LOL1/778_ours.pdf}\label{ab-ours}}
	\caption{Visual comparison of ablation experiments. }
	\label{fig:ab} 
\end{figure*}

We first demonstrate the necessity of the algorithm unrolling design for the decomposition network. Specifically, since the main building blocks in our decomposition network is ResBlocks in R-DecNet and convolution layers in L-DecNet, we construct a network by keeping such structures while ignoring other operations led by algorithm unrolling. It is easy to see that this newly constructed network is with the similar complexity as that of the original decomposition network. We then use this network in our framework to see its performance. It can be seen from Table \ref{tab:im} that the performance significantly drops, which demonstrates the importance of algorithm unrolling. Visual results in Fig. \ref{fig:ab} also clearly substantiate this observation. We then diagnose the effect of the number of stages in the decomposition network, and the results are shown in Table \ref{tab:ab-stage}. As can be seen, with relatively more stages, the performance can be improved. However, if too many stages were used, the performance could drop, which could be due to overfitting issue.

\begin{figure*} 
    \centering
         \subfigure[Input]{
        \includegraphics[width=0.15\linewidth]{LOL1/778_low.pdf}}
     \subfigure[0.001]{
        \includegraphics[width=0.15\linewidth]{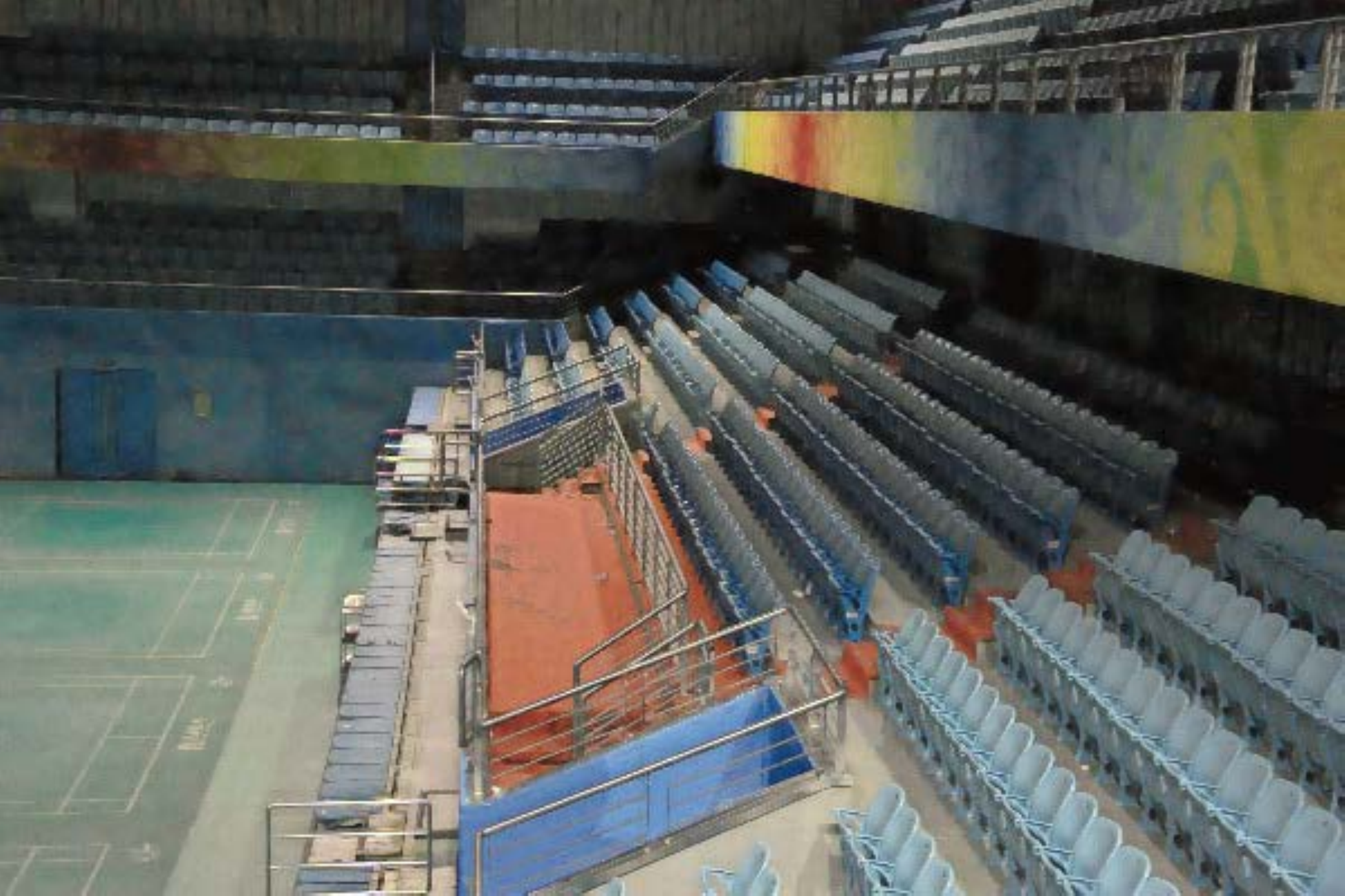}}
	  \subfigure[0.01]{
        \includegraphics[width=0.15\linewidth]{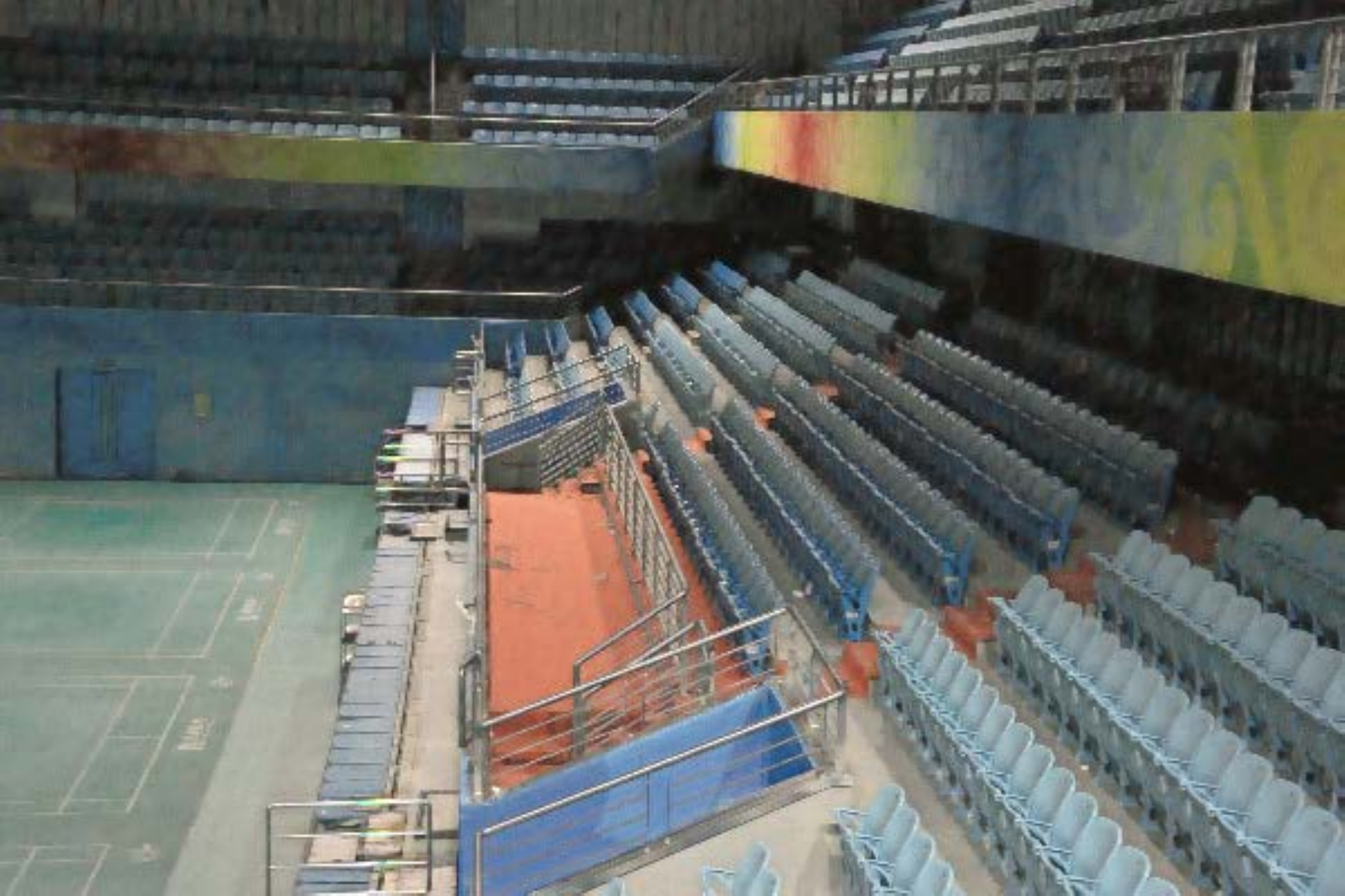}}
	  \subfigure[0.1]{
        \includegraphics[width=0.15\linewidth]{LOL1/778_ours.pdf}}
    	 \subfigure[1]{
       \includegraphics[width=0.15\linewidth]{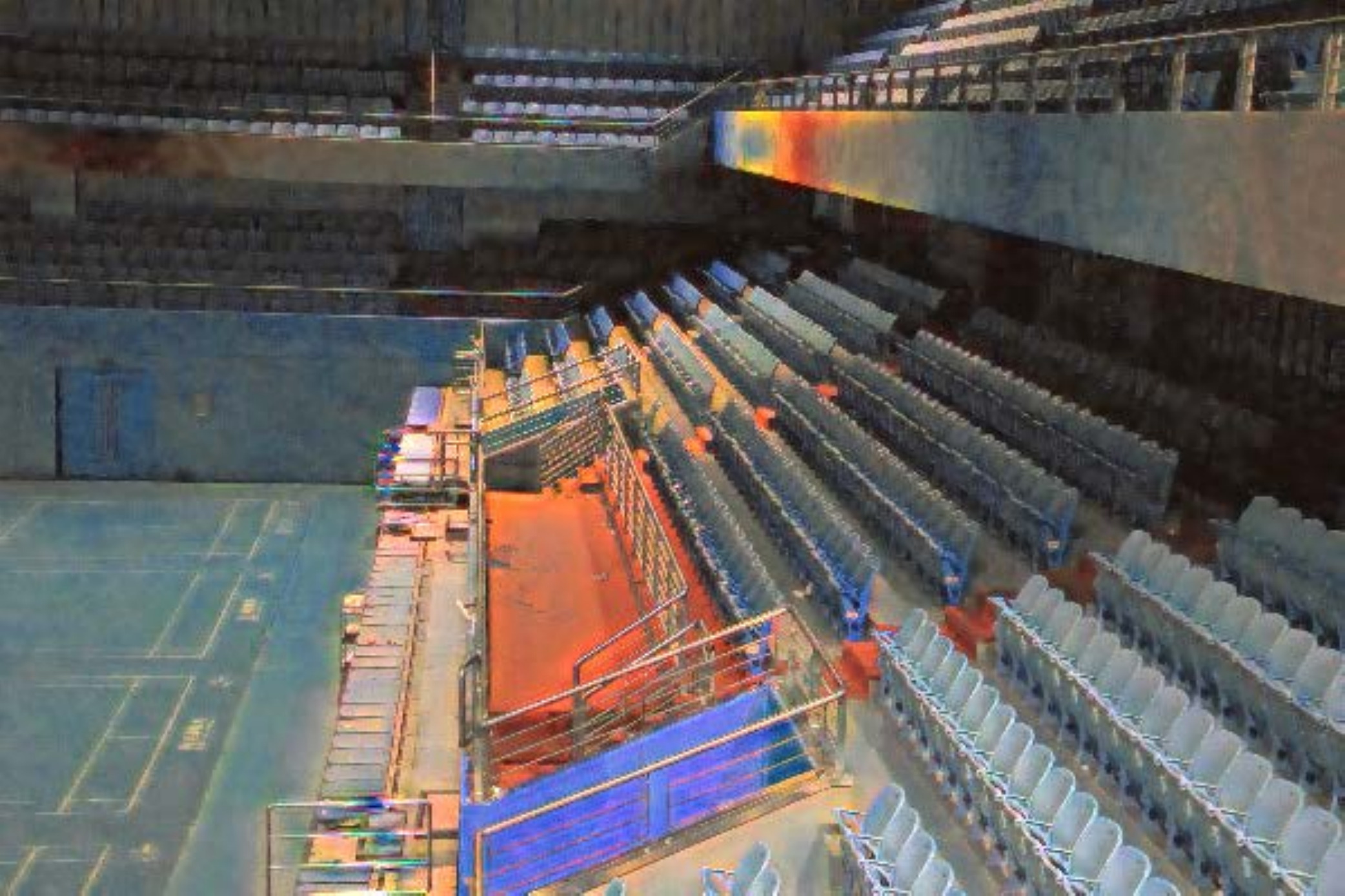}}
         \subfigure[Ground Truth]{
        \includegraphics[width=0.15\linewidth]{LOL1/778_gt.pdf}}
	  \caption{Visual comparison of using different values of weight $\gamma$ for the structure-revealing prior in (\ref{model_used}).}
	  \label{fig:gam} 
\end{figure*}

Then we justify the role of the explicit  structure-revealing prior, i.e., the last term in Eq. \eqref{model_used}, in the whole network, as using this term is one of the major differences between our method and existing algorithm unrolling inspired methods for LIE. Specifically, we can remove all the operations related to this prior, and see the performance of the resulted network. It is seen from Table \ref{tab:im} that without this explicit prior, the performance also drops a lot, which indicates that it provides an effective hint for the LIE problem, especially in contrast as can be observed from Fig. \ref{fig:ab}. Since there is a tuning parameter $\gamma$ in this prior term, we also visualize its influence to the enhancement result in Fig. \ref{fig:gam}. As can be seen from the figure, the saturation is approximately in proportion to $\gamma$. In our experiments, we set $\gamma=0.1$ for promising performance.

Next we quantify the effect of the LBS module, as we have already visually demonstrated its effectiveness in Fig. \ref{fig:ab-lbs}. The result in Table \ref{tab:im} shows that this LBS module, though very simple, is very important to the final performance. 

\begin{table}[t]
	\begin{center}
		\caption{Quantitative results of ablation experiments.}
		\label{tab:im}
		\begin{tabular}{ c | l | c | c   }
			\hline
			\multicolumn{2}{c|}{Setting} & SSIM $\uparrow$ & PSNR $\uparrow$   \\ \hline
			\multirow{4}*{Model} & w/o Unrolling & 0.771  & 19.37   \\ 
			~ & w/o Explicit Prior & 0.870  & 22.19    \\
			%			~ & w/ ResNet & 0.822  & 21.409 & 4.5551  & 268.8  & 246.6   \\ 
			~ & w/o LBS & 0.767  & 19.91   \\ \hline
			\multirow{2}*{Loss} & MSE & 0.839  & 22.02   \\
			~ & MSE+Color & 0.848  & 21.94   \\ 
			\hline
			Data & w/o Augmentation & 0.860  & 22.39   \\
			\hline
			\multicolumn{2}{c|}{Ours} & \textbf{0.874}  & \textbf{22.82}  \\
			\hline 
		\end{tabular}
	\end{center}
\end{table}

\subsubsection{Ablation on Loss Function}
In Section \ref{sec:loss_train}, we have introduced several loss terms in order to train the network. For most modules, we only place one loss term each, while the exception is the final reconstruction loss $\mathcal{L}_{\textrm{en}}$ containing three terms, i.e., MSE, perceptual and color losses. Therefore, we conduct ablation experiments by removing one or two terms, and the results on LOL dataset are shown in Table \ref{tab:im}. As can be seen, all of the three terms can benefit to the final performance. Although there seems to be a trade-off in using color loss, regarding IQA metrics, the visual results in Fig. \ref{fig:ab} show that the color bias can be alleviated by introducing color loss in addition to pure MSE loss.

\subsubsection{Ablation on Data Augmentation}
In Section \ref{sec:train_strategy}, we have discussed a simple data augmentation strategy for training. Here, we also conduct experiment to show its effectiveness. Specifically, we train our model on LOL dataset without data augmentation, and record the result in Table \ref{tab:im}. As can be seen from the table, the proposed data augmentation strategy can facilitate to the overall performance, especially for the SSIM metric.

%After ensuring the indispensability of unrolling algorithm, we present experiments to choose appropriate number of the iteration, and we call it the number of stages in our model. Because our model consists of DecNet with unfolding and AdjNet, the values of metrics and the number of stages are not simple linear relationship. Combining the various evaluation metrics, we choose 17 as the iterative number for best performance.

\begin{table}
	\centering
	\caption{Quantitative results on LOL dataset, with respect to the number of stages within the decomposition network.}\label{tab:ab-stage}
	\footnotesize
	\begin{tabular}{l|c|c|c|c|c|c}
		\hline
		Stage & 5 & 10 & 12 & 15 & 17 & 20  \\
		\hline
		SSIM $\uparrow$  & 0.850 & 0.851 & 0.852 & 0.851 & \textbf{0.874} & 0.855\\
		\hline
		PSNR $\uparrow$ & 20.53 & 21.15 & 21.65 & 21.69 & \textbf{22.82} & 20.57\\
		\hline
	\end{tabular}
	\normalsize
\end{table}

\subsection{Failure Cases}
\begin{figure} 
	\centering
	\subfigure{
		\includegraphics[width=45pt,height=45pt]{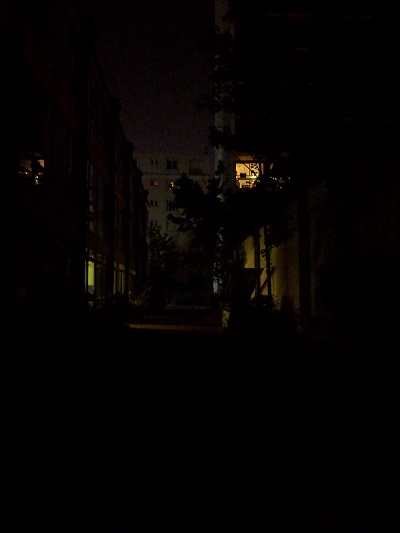}}
	\subfigure{
		\includegraphics[width=45pt,height=45pt]{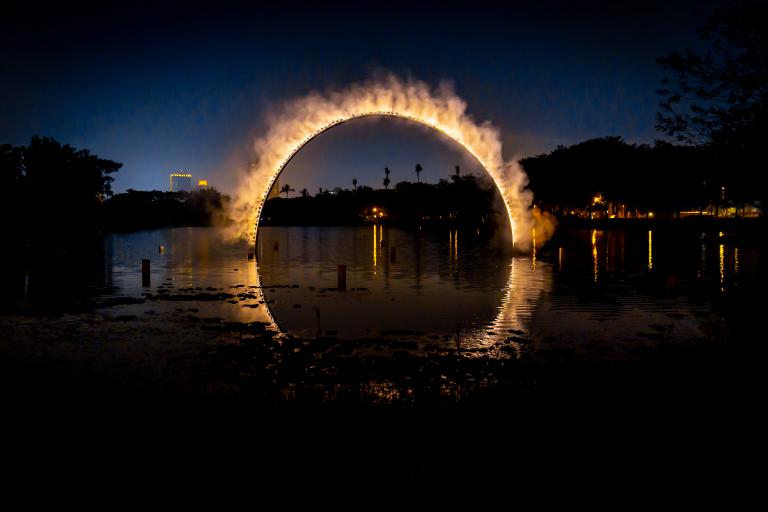}}
	\subfigure{
		\includegraphics[width=45pt,height=45pt]{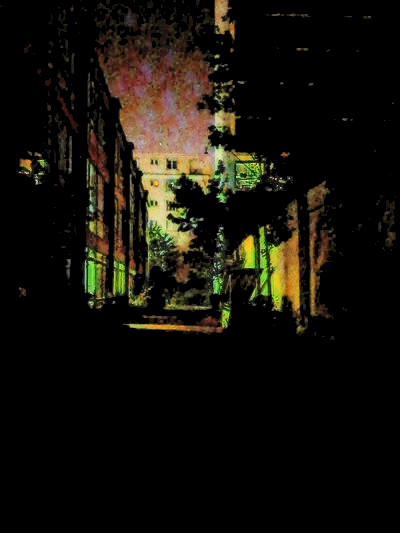}}
	\subfigure{
		\includegraphics[width=45pt,height=45pt]{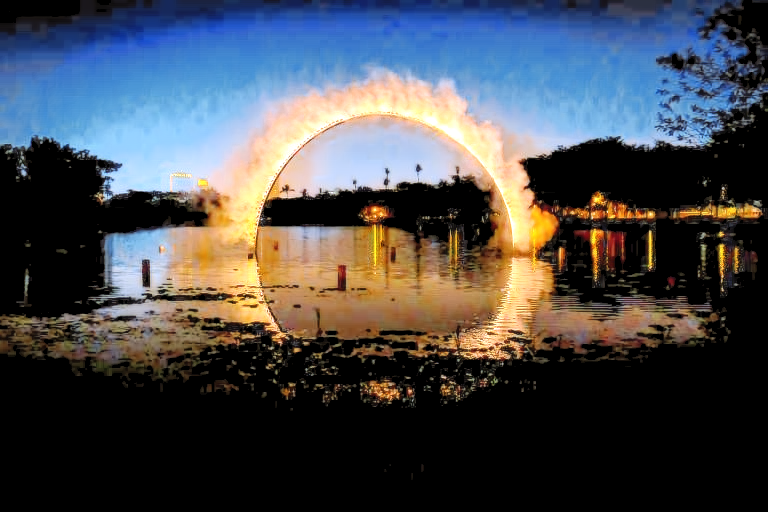}}	
	\caption{Two typical failure examples based on our model. The first and second columns are the input low-light images, and the third and fourth column are the corresponding enhanced images by our method.}
	\label{fig:lim} 
\end{figure}

Although achieved promising performance in the experiments, our method still has its limitations, especially under some real challenging scenarios. Here we show two typical failure cases of our method in Fig. \ref{fig:lim}, for which the enhanced images are still unsatisfactory. The first image in Fig. \ref{fig:lim} is with extremely low light intensity, and a lot of detail information in the dark area is lost. Consequently, our method fails to lighten the dark area of the image. In comparison, the second image in the figure is with extremely non-homogeneous light conditions in different regions and our method produces a distorted result. 

\section{Conclusions}\label{sec:conclusion}
In this work, we have proposed a new deep learning framework, following a decomposition-adjustment pipeline based on Retinex theory, called RAUNA, for the LIE problem. Specifically, for decomposition, we have designed a Retinex decomposition network inspired by unrolling an algorithm for solving the optimization based Retinex decomposition, with both explicit and implicit priors; for adjustment, we have presented simple yet effective structures by considering both global brightness and local brightness sensitivity. To avoid manually parameter tuning, a self-supervised fine-tuning strategy has also been adopted. Experiments on a series of typical datasets have demonstrated the effectiveness of the proposed RAUNA method, as compared with existing ones. In the future, we will devote to addressing more complex degradations for images captured in real low-light conditions, e.g., the failure cases shown in Fig. \ref{fig:lim}.

\begin{appendices}
\section{Calculations of the Descent Directions in Algorithm 1}
\subsection{Calculations of $d_{\L}$}
As summarized in Eq. (9) of the main text,
\begin{equation}
	d_{\L}=\left(\nabla^2f_1(\L)\right)^{-1}\nabla f_1(\L),
\end{equation}
where 
\begin{equation}
	f_1(\L)=\frac{1}{2}\Vert\I-\R\circ \L\Vert_{F}^{2}.
\end{equation}
Reshaping matrix to vector, the above equations can be equivalently rewritten as
\begin{equation}\label{eq:direction_dl_vec}
	d_{\l}=\left(\nabla^2f_1(\l)\right)^{-1}\nabla f_1(\l),
\end{equation}
and
\begin{equation}
	f_1(\l)=\frac{1}{2}\Vert\i-\r\circ \l\Vert_{2}^{2},
\end{equation}
respectively, where $\i,\l,\r$ are the vector forms of $\I,\L,\R$. Then the gradient of $f_1(\l)$ can be calculated as
\begin{equation}\label{eq:grad_f1}
	\nabla f_1(\l)=\r\circ\left(\r\circ\l-\i\right),
\end{equation}
and the Hessian matrix can be further calculated as
\begin{equation}\label{eq:hessian_f1}
	\nabla^2 f_1(\l)=\mathrm{diag}\left(\r\circ\r\right),
\end{equation}
where $\mathrm{diag}(\x)$ denotes the diagonal matrix with the diagonal elements being vector $\x$. Substituting $\nabla f_1(\l)$ and $\nabla^2 f_1(\l)$ by \eqref{eq:grad_f1} and \eqref{eq:hessian_f1} in \eqref{eq:direction_dl_vec}, we can get
\begin{equation}
	d_{\l}=\left(\mathrm{diag}\left(\r\circ\r\right)\right)^{-1}\r\circ\left(\r\circ\l-\i\right).
\end{equation}
Since $\mathrm{diag}\left(\r\circ\r\right)$ is a diagonal matrix, then the above equation can be further written as
\begin{equation}
	d_{\l}=\r\circ\left(\r\circ\l-\i\right)\oslash\left(\r\circ\r\right),
\end{equation}
which can finally be reshaped back to matrix as
\begin{equation}
	d_{\L}=(\R\circ (\R\circ\L-\I))\oslash(\R\circ \R),
\end{equation}
by noting that all the involved operations are performed element-wise.

\subsection{Calculations of $d_{\R}$}
According to Eq. (12) of the main text,
\begin{equation}
	d_{\R}=\left(\nabla^2f_2(\R)\right)^{-1}\nabla f_2(\R),
\end{equation}
where
\begin{equation}
	f_2(\R)={\frac{1}{2}}\Vert \I-\R\circ \L\Vert_F^2+\frac{\gamma}{4}\!\sum_{i=x,y}\!\Vert\d_i\otimes\R-\G_i\Vert_F^2.
\end{equation}
Similar as that for calculating $d_{\L}$, we can equivalently rewrite the above equations with respect to vectors:
\begin{equation}
	d_{\r}=\left(\nabla^2f_2(\r)\right)^{-1}\nabla f_2(\r),
\end{equation}
\begin{equation}
f_2(\r)={\frac{1}{2}}\Vert\i-\r\circ\l\Vert_2^2+\frac{\gamma}{4}\sum_{i=x,y}\left\Vert{\D_i\r-\g_{i}}\right\Vert_{2}^{2},
\end{equation}
where $\D_i\r$ is the matrix-vector representation of the convolution $\d_{i}\otimes \R$ defined in Eq. (6) of the main text, with $\D_i$ being the difference operator matrix with respect to $\d_i$, and $\g_i$ is the vectorization of $\G_i$, also defined in Eq. (6) of the main text. Similar as before, the gradient and Hessian of $f_2(\r)$ with respect to $\r$ can be calculated as
\begin{equation}
\nabla f_2(\r)=\l\circ\left(\r\circ\l-\i\right)+\frac{\gamma}{2}\sum_{i=x,y}\D_i^{\mathsf{T}}\left(\D_i\r-\g_i\right),
\end{equation}
and
\begin{equation}
\nabla^2 f_2(\r)=\mathrm{diag}\left(\l\circ\l\right)+\frac{\gamma}{2}\sum_{i=x,y}\D_i^{\mathsf{T}}\D_i,
\end{equation}
respectively. Since $\nabla^2 f_2(\r)$ now is not a diagonal matrix, the calculating of $d_{\R}$ is problematic due to the involved matrix inversion. Therefore, we resort to a diagonal approximation to $\nabla^2 f_2(\r)$, such that the matrix inversion can be done elementwise as that for calculating $\nabla^2 f_1(\l)$. Specifically, considering the eigenvalues of ${\D_i}^{\mathsf{T}}\D_i$ are upper-bounded by $4$, by virtue of the property of difference operator matrix, we have
\begin{equation}
	\mathrm{diag}\left(\l\circ\l\right)+4\gamma\F\succeq\nabla^2 f_2(\r),
\end{equation}
where $\F$ is the identity matrix, and $\A\succeq\B$ means that $\A-\B$ is a positive semi-definite matrix. Consequently, we can approximate the Hessian $\nabla^2 f_2(\r)$ by
\begin{equation}
	\nabla^2 f_2(\r)\approx\mathrm{diag}\left(\l\circ\l\right)+4\gamma\F,
\end{equation}
which then results in a quasi-Newton approximation to the exactly Newton descent direction $d_{r}$:
\begin{equation}
\begin{split}
d_{r}\approx&\Big(\l\circ\left(\r\circ\l-\i\right)+\frac{\gamma}{2}\sum_{i=x,y}\D_i^{\mathsf{T}}\left(\D_i\r-\g_i\right)\Big)\\
&\oslash\left(\mathrm{diag}\left(\l\circ\l\right)+4\gamma\F\right).
\end{split}
\end{equation}
Rewriting $d_{r}$ with matrix form, we can finally get
\begin{equation}
\begin{split}
d_{\R}\approx&\Big(\left(\R\circ\L-\I\right)\circ\L+\frac{\gamma}{2}\sum_{i=x,y}\d_i\otimes^{\mathsf{T}}\left(\d_i\otimes\R-\G_i\right)\Big)\\
&\oslash\!\left(\L\circ\L+4\gamma\E\right).
\end{split}
\end{equation}

\end{appendices}

\bibliography{LIE_unfolding_IJCV}

\end{document}